\ifdefined\pdfsuppressptexinfo
\fi

\ifdefined\pdfinfoomitdate
\fi

\ifdefined\pdftrailerid
  \pdftrailerid{}
\fi

\documentclass{article}
\usepackage{iclr2027_conference,times}

\usepackage{amsmath,amsfonts,bm}

\def\eqref#1{equation~\ref{#1}}

\def\1{\bm{1}}

\DeclareMathAlphabet{\mathsfit}{\encodingdefault}{\sfdefault}{m}{sl}
\SetMathAlphabet{\mathsfit}{bold}{\encodingdefault}{\sfdefault}{bx}{n}

\usepackage[utf8]{inputenc}
\usepackage[T1]{fontenc}
\usepackage[english]{babel}
\usepackage{microtype}

\usepackage{amsmath}
\usepackage{amssymb}
\usepackage{amsfonts}
\usepackage{mathtools}
\usepackage{amsthm}
\usepackage{eucal}
\usepackage{bm}
\usepackage{nicefrac}

\usepackage{graphicx}
\usepackage{tikz}
\usepackage{subfigure}
\usepackage{wrapfig}
\usepackage{setspace}

\usepackage{booktabs}
\usepackage{colortbl}
\usepackage{multirow}
\usepackage{tabularx}
\usepackage{enumitem}

\usepackage{algpseudocode}

\usepackage{xcolor}
\usepackage{url}
\usepackage{xspace}
\usepackage{pifont}
\usepackage{stackengine}
\usepackage{etoc}
\usepackage{blindtext}

\usepackage{hyperref}
\hypersetup{
    breaklinks=true,
    colorlinks=true,
    citecolor=blue,
    linkcolor=blue,
    urlcolor=blue,
    hypertexnames=false
}

\newcolumntype{C}[1]{>{\centering\arraybackslash}p{#1}}
\newcolumntype{L}[1]{>{\raggedright\arraybackslash}p{#1}}

\DeclareRobustCommand{\cmark}{\ding{51}}

\newlength{\mylen}
\graphicspath{{figures/}}

\title{Dynamical Parameters: An Interpretability Framework for Time-Series Foundation Models}

\author{
  Kang Yang \hspace{2em}
  Gaofeng Dong \hspace{2em}
  Liying Han \hspace{2em}
  Mani Srivastava\thanks{Mani Srivastava holds concurrent appointments as a Professor of ECE and CS (joint) at the University of California, Los Angeles.} \\
  Department of Electrical and Computer Engineering\\
  University of California, Los Angeles\\
  \texttt{\{kyang73,gfdong,liying98\}@g.ucla.edu \quad mbs@ucla.edu}
}

\iclrfinalcopy

\begin{document}

\maketitle

\lhead{}

\addtocontents{toc}{\protect\setcounter{tocdepth}{-1}}

\begin{abstract}
This work studies a central gap in interpreting time-series foundation models~(TSFMs): a dynamical property may be accessible in a hidden state even when the forecast fails to respond correctly as that property changes.
We formalize these properties as \emph{Dynamical Parameters}, including trend slope, oscillation frequency, and autoregressive dependence.
We compare their \emph{representation accessibility}, measured by recovery from hidden states, with their \emph{forecast response}, measured by agreement with the expected forecast change.
Across nine frozen~TSFMs and thirteen laws, 42 of 63 model--parameter cells achieve accessibility above~\(0.95\), whereas their median reference-aligned response relative to the conditional reference is only~\(0.46\).
To explain this gap, causal geometry compares the hidden-state change required to produce the reference response with the change induced by the parameter intervention.
Directly modifying the hidden state recovers the reference response, but the parameter intervention often moves the state in a different direction.
These results show that accessible parameter information need not be expressed in forecasts when input changes miss the required hidden-state direction.
\end{abstract}

\vspace{\mylen}
\section{Introduction}
\label{sec:intro}
\vspace{\mylen}

Classical forecasting connects an observed history to its predicted future through an explicit pipeline.
A mathematical model specifies a governing law and its associated \emph{Dynamical Parameters}, such as trend slope, oscillation frequency, and autoregressive dependence.
These parameters are estimated from the history and determine how the fitted model extrapolates the future~\citep{kalman1960,harvey1989,durbin2012,ljung1999system}.
Time-series foundation models~(TSFMs), by contrast, produce general-purpose forecasts without explicitly specifying a governing law or its parameters~\citep{ansari2024chronos,das2024timesfm,woo2024moirai,shi2025timemoe,ekambaram2024ttm}.
Inspired by classical forecasting, we interpret TSFMs through two properties.
\emph{Representation accessibility} measures whether the governing law and its parameters can be recovered from hidden representations.
\emph{Forecast response} measures whether changing a parameter produces the corresponding change in the forecast.
These properties relate what a~TSFM represents to what its forecast expresses without assuming that it follows the classical pipeline internally.

Recent interpretability studies examine hidden representations through probing and forecast behavior through activation steering, concept erasure, and feature ablation~\citep{goswami2024moment,wilinski2025tsfm,sanyal2025time2time,pandey2025semantics,pagani2026frequency,dissectingchronos2026,eidos2026}.
Generator-level interventions further measure forecast sensitivity to controlled changes in temporal patterns~\citep{jander2026causal}.
However, successful probing establishes that a parameter is accessible, not that its changes are appropriately reflected in the forecast.
Likewise, altering a forecast through intervention establishes influence, not agreement with the direction, magnitude, and temporal shape implied by the changed dynamics.
Existing evidence therefore does not determine whether accessible Dynamical Parameters correspond to quantitatively correct forecast responses.

Determining whether accessible parameters are reflected in forecasts requires references for both parameter recovery and forecast change.
Because a finite, noisy history may not fully identify a parameter, we compare probe performance with a law- and nuisance-informed conditional reference.
To isolate the forecast consequence of a parameter change, we construct matched histories that differ in one designated parameter but share the remaining parameters, initial conditions, and stochastic innovations.
The predictable difference between their futures defines the corresponding reference response.
Reference-aligned gain and full-vector fidelity then measure whether the forecast change matches this target in direction, magnitude, and temporal shape.
These measurements account for finite-history uncertainty and distinguish response fidelity from mere forecast sensitivity.

When a parameter is accessible but its matched-input forecast response is weak, the gap can arise for two reasons.
The selected hidden state may not support the reference response, or the parameter intervention may not induce the hidden-state change required to produce it.
We develop \emph{causal geometry} to distinguish these possibilities.
The forecast Jacobian yields a small hidden-state edit toward the reference response, which we term a \emph{local causal direction}.
A nonlinear intervention then verifies its effect.
Comparing this required edit with the hidden-state change induced by the matched parameter intervention distinguishes limited response capacity from a mismatch between the required and induced changes.
Shared subspaces summarize structure across the effective edits, while a readout predicts their input-specific \emph{causal coordinates} from the original hidden state.

Across nine frozen~TSFMs, including Chronos-2, TimesFM~3, and Moirai~\citep{ansari2025chronos2,jain2026timesfm3,woo2024moirai}, we study ten single-parameter and three multi-parameter laws spanning linear trend, oscillation, and dual tone.
Three principal findings emerge:

\begin{itemize}[
    label=\textbullet,
    leftmargin=1em,
    topsep=-2pt,
    itemsep=1pt
]
    \item \textbf{Accessible Parameters Are Not Consistently Expressed in Forecasts.}
    Across 63 model--parameter cells, 42 achieve accessibility above~\(0.95\), while the median conditional-reference-normalized aligned gain is~\(0.46\).
    Three models reverse the expected trend-slope response, and full-vector metrics reveal further errors in magnitude and temporal shape.

    \item \textbf{Hidden-State Edits Recover Responses Missed by Input Changes.}
    Calibrated hidden-state edits recover the reference responses with low full-vector error, but input changes induce hidden-state changes that align weakly with these edits.
     Coordinates predicted from the original hidden states increase median aligned gain from~\(0.23\) to~\(0.58\).

    \item \textbf{Compact Geometry Persists Under Shuffled Targets.}
    Low-rank subspaces summarize effective edit directions, and shuffled targets often yield comparable compactness.
    The shared causal-subspace structure therefore extends beyond the correct parameter--response pairing.
\end{itemize}

These results separate parameter accessibility from forecast expression and link the gap to misalignment between required and input-induced hidden-state changes.
Shuffled targets show comparable compactness, indicating low-rank editability beyond the correct parameter--response pairing.

\vspace{\mylen}
\section{Related Work}
\label{sec:related_work}
\vspace{\mylen}

\textbf{Time-Series Foundation Models~(TSFMs).}
TSFMs are pretrained on large-scale time-series corpora and demonstrate strong zero-shot forecasting across diverse domains~\citep{garza2023timegpt,rasul2023lagllama,ansari2024chronos,das2024timesfm}.
This work studies whether \textit{Dynamical Parameters} are accessible in their hidden states and expressed through their forecast responses.

\textbf{Interpretability of TSFMs.}
Probing and representation analyses recover temporal properties and continuous parameters, including trend, frequency, periodicity, phase, and amplitude, from hidden representations~\citep{goswami2024moment,wilinski2025tsfm,pandey2025semantics,pagani2026frequency,chemeris2026aionoscope}.
Complementing this evidence about accessible information, steering, activation transplantation, sparse feature discovery, and component ablation show that internal features can alter forecasts~\citep{wilinski2025tsfm,sanyal2025time2time,dissectingchronos2026,oublal2026timesae,bao2026redundancies}.
At the level of the generating process, \citep{jander2026causal} measure scalar forecast responses to controlled parameter changes in two~TSFMs across six temporal patterns.
These results leave open how parameter accessibility relates to the direction, magnitude, and full-horizon shape of the forecast response when that parameter changes.
We characterize this relationship against conditional references across nine~TSFMs and thirteen laws.
To examine discrepancies between accessibility and response, causal geometry compares natural hidden changes with local edits derived by Jacobian inversion and tests input-specific causal coordinates.

\textbf{State-Space Models.}
Classical state-space inference estimates latent states and dynamical parameters from observed histories for forecasting~\citep{kalman1960,harvey1989,hamilton1994time,durbin2012,ljung1999system}.
Koopman forecasters impose spectral or stability constraints on their learned latent transitions, while Transformers trained on dynamical systems learn in-context estimation and filtering~\citep{forootani2025deepkoopformer,forootani2026learnable,garg2022what,akyurek2023what,vonoswald2023transformers,bai2023transformers,goel2024can,du2023can,xie2022explanation,muller2022transformers}.
We compare parameter accessibility and controlled forecast response in frozen general-purpose~TSFMs, relating them through hidden interventions.

\vspace{\mylen}
\section{Method}
\label{sec:method}
\vspace{\mylen}

Given an observed history~\(x_{1:T}\), a pretrained~TSFM produces hidden representations and a forecast.
We characterize two properties of \emph{Dynamical Parameters}:

\begin{itemize}[
    label=\textbullet,
    leftmargin=1em,
    topsep=-2pt,
    itemsep=1pt
]
    \item \textbf{Representation Accessibility~(\S\ref{sec:representation}).}
    This property measures whether the governing law and its parameter values are linearly accessible from the selected hidden representations.

    \item \textbf{Forecast Response~(\S\ref{sec:response}).}
    This property measures whether changing a parameter produces the reference forecast change in direction, magnitude, and temporal shape.
\end{itemize}

We relate these properties through \textbf{causal geometry~(\S\ref{sec:causal_geometry})}, which compares the hidden-state change required to produce the reference response with the change induced by the parameter intervention.

\vspace{\mylen}
\subsection{Controlled State-Space Framework}
\label{sec:framework}
\vspace{\mylen}

\textbf{Dynamical Laws and Parameters.}
Each time series is defined by a dynamical law~\(m\in\left\{1,\ldots,M\right\}\) and a parameter vector that shapes its temporal evolution
\[
z_m
=
\left(
z_{m,1},
\ldots,
z_{m,K_m}
\right)
\in
\mathcal{Z}_m.
\]
The law~\(m\) specifies the form of the dynamics, while~\(z_m\) contains the quantitative properties governing the evolution of the series.
Each component~\(z_{m,j}\) is a~\emph{Dynamical Parameter}.
Every law instantiates a common state-space form in which~\(z_m\) controls the transition dynamics, input, or readout.

Given~\(m\), \(z_m\), and stochastic variation~\(\varepsilon\), the controlled generator produces a history and future:
\begin{equation}
\label{eq:generator}
\left(
x_{1:T},
y_{T+1:T+H}
\right)
=
G_m
\left(
z_m,
\varepsilon
\right).
\end{equation}
The~TSFM observes only the history and produces a forecast and layer-wise hidden representations:
\[
\hat{y}_{T+1:T+H}
=
F_\theta
\left(
x_{1:T}
\right),
\qquad
h^\ell
=
F_\theta^\ell
\left(
x_{1:T}
\right).
\]
The generator exposes the law and parameters that the~TSFM can recover only from the history.

\textbf{Parameter Interventions.}
To isolate the effect of parameter~\(z_{m,j}\), two parameter vectors are constructed that differ only in that component:
\[
z_{m,j}^{B}
=
\mathcal{I}_{m,j}
\left(
z_{m,j}^{A},
\delta
\right),
\qquad
z_{m,j'}^{B}
=
z_{m,j'}^{A}
\quad
\text{for }j'\neq j.
\]
Both sequences are generated using the same stochastic realization~\(\varepsilon\):
\begin{equation}
\label{eq:counterfactual_pair}
\left(
x^{A},
y^{A}
\right)
=
G_m
\left(
z_m^{A},
\varepsilon
\right),
\qquad
\left(
x^{B},
y^{B}
\right)
=
G_m
\left(
z_m^{B},
\varepsilon
\right).
\end{equation}
Here, \(x^{A}\) and~\(x^{B}\) are the matched histories, while~\(y^{A}\) and~\(y^{B}\) are their futures.
Sharing~\(\varepsilon\) removes stochastic variation unrelated to the parameter intervention.
The resulting reference response is
\begin{equation}
\label{eq:reference_response}
\Delta y_{m,j}^{\mathrm{ref}}
=
y^{B}
-
y^{A}.
\end{equation}
The known law~\(m\), Dynamical Parameter vector~\(z_m\), and reference response~\(\Delta y^{\mathrm{ref}}_{m,j}\) provide targets for analyzing representation accessibility, forecast response, and causal geometry.

\vspace{\mylen}
\subsection{Representation Accessibility}
\label{sec:representation}
\vspace{\mylen}

Representation accessibility measures how accurately the dynamical law~\(m\) and its parameters~\(z_m\) can be recovered from hidden representations using linear probes.
The analysis has three steps.

\textbf{(I) Probing Laws and Parameters.}
At each layer~\(\ell\), a linear classifier predicts the dynamical law from~\(h^\ell\), while a linear ridge probe predicts each parameter within its corresponding law:
\[
\hat{m}^{\ell}
=
\operatorname*{argmax}_{m'}
\left(
W_{\ell}h^\ell+b_{\ell}
\right)_{m'},
\qquad
\hat{z}_{m,j}^{\ell}
=
w_{\ell,m,j}^{\top}h^\ell+b_{\ell,m,j}.
\]
\textbf{(II) Dynamical Law Decodability.}
For each law~\(m\in\left\{1,\ldots,M\right\}\), let~\(A_m^\star\) denote an upper reference for identifying~\(m\).
For balanced~\(M\)-way classification, \(A^0=1/M\) is the uniform-guessing accuracy.
The layer-wise accuracy and normalized law decodability are
\begin{equation}
\label{eq:law_decodability}
A_m^\ell
=
\Pr
\left(
\hat{m}^\ell=m
\mid
m
\right),
\qquad
D_m
\left(
\ell
\right)
=
\frac{
A_m^\ell-A^0
}{
A_m^\star-A^0
}.
\end{equation}
When a reliable Bayes ceiling is unavailable, \(A_m^\star=1\) provides a conservative upper reference.
A value of~\(D_m\left(\ell\right)\) near~\(1\) indicates that the hidden representation approaches this reference, while a value near~\(0\) indicates chance-level identification.

\textbf{(III) Parameter Accessibility.}
For each Dynamical Parameter~\(z_{m,j}\), we compare the probe correlation~\(r_{m,j}^{\ell}\) with a conditional posterior-mean reference:
\begin{equation}
\label{eq:state_decodability}
r_{m,j}^{\ell}
=
\operatorname{corr}
\left(
\hat{z}_{m,j}^{\ell},
z_{m,j}
\right),
\quad
r_{m,j}^{\star}
=
\operatorname{corr}
\left(
\mathbb{E}
\left[
z_{m,j}
\mid
x_{1:T},m,\eta
\right],
z_{m,j}
\right),
\quad
D_{m,j}(\ell)
=
\frac{r_{m,j}^{\ell}}{r_{m,j}^{\star}}.
\end{equation}
Here~\(\eta\) collects the scale, offset, noise, phase, and non-designated parameters supplied to the reference observer but not to the~TSFM.
A value of~\(D_{m,j}(\ell)\) near~\(0\) indicates no linear association, while a value near~\(1\) means the probe approaches this law- and nuisance-informed reference.

\vspace{\mylen}
\subsection{Forecast Response}
\label{sec:response}
\vspace{\mylen}

Forecast response measures whether changing a Dynamical Parameter changes the~TSFM forecast in the direction, magnitude, and temporal shape implied by the law.
The analysis has three steps.

\textbf{(I) Matched Forecast Response.}
For the matched pair in~\autoref{eq:counterfactual_pair}, the~TSFM forecast response is
\[
\Delta\hat{y}_{m,j}
=
F_\theta
\left(
x^{B}
\right)
-
F_\theta
\left(
x^{A}
\right).
\]
\textbf{(II) Response Agreement.}
To compare the forecast and reference responses while allowing a constant level offset, define the centered horizon response
\begin{equation}
\label{eq:center}
\mathcal{C}
\left(
a
\right)
=
a
-
\frac{1}{H}
\left(
\mathbf{1}^{\top}a
\right)
\mathbf{1},
\qquad
a\in\mathbb{R}^{H}.
\end{equation}
For~\(N\) matched pairs, the per-pair response coefficient and its mean are
\begin{equation}
\label{eq:response_coefficient}
u_{i,m,j}
=
\frac{
\left\langle
\mathcal{C}
\left(
\Delta\hat{y}_{i,m,j}
\right),
\mathcal{C}
\left(
\Delta y_{i,m,j}^{\mathrm{ref}}
\right)
\right\rangle
}{
\left\|
\mathcal{C}
\left(
\Delta y_{i,m,j}^{\mathrm{ref}}
\right)
\right\|_2^2
},
\qquad
U_{m,j}
=
\frac{1}{N}
\sum_{i=1}^{N}
u_{i,m,j}.
\end{equation}
Here, \(u_{i,m,j}\) is the least-squares slope of the forecast response on the reference response, with a free level offset.
The mean~\(U_{m,j}\) near~\(1\) indicates a reference-aligned component of the expected size, near~\(0\) indicates little aligned response, and a negative value indicates an opposing component.

\textsc{Full-Vector Fidelity and Aggregation.}
The response coefficient~\(U_{m,j}\) measures only the component aligned with the reference and therefore does not establish fidelity of the full forecast-change trajectory.
For each matched pair~\(i\), let~\(q_i\) denote the realized forecast change and~\(r_i\) the corresponding reference change.
Define~\(\widetilde q_i=\mathcal C(q_i)\), \(\widetilde r_i=\mathcal C(r_i)\), \(E=\sum_i\|\widetilde r_i\|_2^2\), and the pairwise aligned gain~\(g_i=\langle\widetilde q_i,\widetilde r_i\rangle/\|\widetilde r_i\|_2^2\).
We report four pooled metrics:
{\small
\begin{equation}
\label{eq:full_vector_metrics}
g=\frac{\sum_i\langle\widetilde q_i,\widetilde r_i\rangle}{E},
\quad
s=\frac{\sum_i\langle\widetilde q_i,\widetilde r_i\rangle}{\sqrt{\left(\sum_i\|\widetilde q_i\|_2^2\right)E}},
\quad
e=\sqrt{\frac{\sum_i\|\widetilde q_i-\widetilde r_i\|_2^2}{E}},
\quad
e_\perp=\sqrt{\frac{\sum_i\|\widetilde q_i-g_i\widetilde r_i\|_2^2}{E}}.
\end{equation}
}
Here, \(g\) measures the reference-aligned component, \(s\) measures directional agreement, \(e\) measures full centered-trajectory error, and \(e_\perp\) measures the component orthogonal to each pair's reference.
These quantities satisfy the pooled decomposition
\(e^2=\sum_i\|\widetilde r_i\|_2^2(g_i-1)^2/E+e_\perp^2\);
thus, total error reflects both gain mismatch and orthogonal trajectory error.
For the natural response, \(g_i=u_{i,m,j}\), so \(U_{m,j}=N^{-1}\sum_i g_i\) weights pairs equally, whereas pooled~\(g\) weights each pair by its reference energy.
We report both because a few large reference changes can dominate the pooled metric.

\textbf{(III) Conditional-Reference-Normalized Response.}
Finite histories and stochastic variation can prevent even an optimal history-based forecaster from reproducing the realized reference response.
We therefore define the law- and nuisance-informed conditional forecast
\[
F_m^\star(x_{1:T};\eta)
=
\mathbb{E}\!\left[
y_{T+1:T+H}\mid x_{1:T},m,\eta
\right],
\qquad
\Delta y_{m,j}^\star
=
F_m^\star(x^B;\eta)-F_m^\star(x^A;\eta),
\]
where~\(\eta\) collects the documented nuisance variables supplied to the conditional forecast.
Substituting~\(\Delta y_{i,m,j}^\star\) for~\(\Delta\hat y_{i,m,j}\) in~\autoref{eq:response_coefficient} gives the conditional-reference coefficient~\(U_{m,j}^\star\).
Across intervention magnitudes~\(\mathcal M_{m,j}\), the normalized forecast response is
\begin{equation}
\label{eq:normalized_response}
\bar U_{m,j}
=
\frac{\sum_{\delta\in\mathcal M_{m,j}}U_{m,j}(\delta)}
{\sum_{\delta\in\mathcal M_{m,j}}U_{m,j}^\star(\delta)}.
\end{equation}
A value of~\(\bar U_{m,j}\) near~\(1\) means the model matches the conditional reference's aggregate aligned component across magnitudes; near~\(0\) means little alignment, and negative values mean an opposing response.
Because this signed normalization conditions on~\(m\) and~\(\eta\), it is a comparison, not a history-only upper bound: \(\bar U_{m,j}\) can exceed one and does not establish full-vector fidelity.

\vspace{\mylen}
\subsection{Causal Geometry}
\label{sec:causal_geometry}
\vspace{\mylen}

Causal geometry characterizes how hidden-state interventions affect the frozen downstream computation and whether their effective directions share structure across inputs.
Here, ``causal'' refers only to the effect of modifying the hidden state at a specified layer; it neither identifies causal variables in the generating process nor establishes that the model naturally uses the resulting directions.
For a fixed law~\(m\), parameter~\(z_{m,j}\), and intervention magnitude~\(\delta\), we suppress dependence on~\(m\), \(j\), and~\(\delta\) below.

\textbf{(I) Local Causal Directions.}
For matched pair~\(i\) at layer~\(\ell\), let~\(h_i^\ell=F_\theta^\ell(x_i^A)\) denote the representation of~\(x_i^A\), and let~\(F_\theta^{>\ell}\) denote the remaining computation to the forecast.
The intervention modifies the selected hidden-state positions while leaving the downstream computation fixed.
Linearizing the forecast around~\(h_i^\ell\), regularized Jacobian inversion yields
{\small
\begin{equation}
\label{eq:local_direction}
J_i^\ell
=
\left.
\frac{
\partial F_\theta^{>\ell}
\left(
h
\right)
}{
\partial h
}
\right|_{h=h_i^\ell},
\qquad
v_i^\ell
=
\operatorname*{argmin}_{v}
\left\|
J_i^\ell v-\Delta y_i^{\mathrm{ref}}
\right\|_2^2
+
\lambda
\left\|
v
\right\|_2^2.
\end{equation}
}
Here, \(J_i^\ell\) maps hidden-state perturbations to first-order forecast changes, and \(v_i^\ell\) is the regularized, input-specific \emph{local causal direction}.
We evaluate this direction in the nonlinear~TSFM and summarize agreement across~\(N\) matched pairs.
For intervention strength~\(\alpha>0\), the realized response per unit intervention and pooled aligned gain are
{\small
\begin{equation}
\label{eq:pooled_response}
\Delta\hat{y}_i^\ell(v;\alpha)
=
\frac{
F_\theta^{>\ell}(h_i^\ell+\alpha v)-F_\theta^{>\ell}(h_i^\ell)
}{\alpha},
\qquad
U^{\mathrm{pool}}
=
\frac{
\sum_i
\left\langle
\mathcal C\!\left(\Delta\hat{y}_i^\ell(v_i;\alpha)\right),
\mathcal C\!\left(\Delta y_i^{\mathrm{ref}}\right)
\right\rangle
}{
\sum_i
\left\|
\mathcal C\!\left(\Delta y_i^{\mathrm{ref}}\right)
\right\|_2^2
}.
\end{equation}
}
The inversion targets the complete reference response.
The primary aligned-gain score removes constant level offsets using~\(\mathcal C\) in~\autoref{eq:center}, whereas raw and level diagnostics retain them.
A value of~\(U^{\mathrm{pool}}\) near~\(1\) indicates the correct pooled aligned component, near~\(0\) indicates little aligned response, and a negative value indicates an opposing component.
Setting~\(v_i=v_i^\ell\) defines~\(U_{\mathrm{local}}(\ell)\), which measures whether the Jacobian-derived directions remain effective in the nonlinear model.

\textbf{(II) Shared Causal Subspace.}
Although the local causal directions are input-specific, they may concentrate in a compact subspace shared across inputs.
The subspace is estimated from a discovery bank~\(\mathcal{D}_{\mathrm{disc}}\) and evaluated on a disjoint bank~\(\mathcal{D}_{\mathrm{eval}}\).
At layer~\(\ell\), singular-value decomposition gives
\begin{equation}
\label{eq:causal_space}
V_\ell
=
\left[
v_i^\ell
\right]_{i\in\mathcal{D}_{\mathrm{disc}}}
=
Q_\ell\Sigma_\ell R_\ell^\top,
\qquad
\mathcal{S}_{\ell,k}
=
\operatorname{span}
\left(
Q_{\ell,k}
\right),
\qquad
\tilde{v}_i^{\ell,k}
=
Q_{\ell,k}Q_{\ell,k}^{\top}v_i^\ell,
\end{equation}
where~\(Q_{\ell,k}\) contains the leading~\(k\) left singular vectors.
Substituting the projected directions~\(\tilde{v}_i^{\ell,k}\) into~\autoref{eq:pooled_response} on~\(\mathcal{D}_{\mathrm{eval}}\) defines~\(U_{\mathrm{projected}}\left(\ell,k\right)\).

For each valid layer, the smallest rank retaining~\(90\%\) of the local reference-aligned gain is
\[
k_{90}
\left(
\ell
\right)
=
\min
\left\{
k\in\mathcal{K}_{\ell}:
\frac{
U_{\mathrm{projected}}
\left(
\ell,k
\right)
}{
U_{\mathrm{local}}
\left(
\ell
\right)
}
\geq
0.9
\right\}.
\]
A small~\(k_{90}(\ell)\) indicates a low-rank shared causal subspace relative to the intervened state dimension.
If no evaluated rank reaches the threshold, \(k_{90}\left(\ell\right)\) is reported as a lower bound.
However, \(\tilde{v}_i^{\ell,k}\) still requires the reference-derived local direction~\(v_i^\ell\), so this analysis establishes subspace compactness but not whether its input-specific coordinates are accessible from the hidden representation.

\textbf{(III) Causal Coordinate Accessibility.}
The final step asks whether the input-specific coordinates of the local causal direction can be inferred from the unperturbed representation~\(h_i^\ell\).
The local direction is projected into the shared subspace, its coordinates are predicted from~\(h_i^\ell\), and the predicted direction is mapped back into the hidden space:
\begin{equation}
\label{eq:deployed_direction}
c_i^{\ell,k}
=
Q_{\ell,k}^{\top}v_i^\ell,
\qquad
\hat{c}_i^{\ell,k}
=
P_{\ell,k}h_i^\ell+b_{\ell,k}^{c},
\qquad
\hat{v}_i^{\ell,k}
=
Q_{\ell,k}\hat{c}_i^{\ell,k}.
\end{equation}
Here, \(P_{\ell,k}\) and \(b_{\ell,k}^{c}\) are the ridge-readout coefficients in the original hidden coordinates and incorporate discovery-bank standardization.
The readout alone is fitted on~\(\mathcal{D}_{\mathrm{disc}}\), while the~TSFM remains frozen.
On~\(\mathcal{D}_{\mathrm{eval}}\), the predicted direction~\(\hat v_i^{\ell,k}\) requires neither~\(J_i^\ell\) nor~\(\Delta y_i^{\mathrm{ref}}\).
Using these directions in~\autoref{eq:pooled_response} defines the deployed response score~\(U_{\mathrm{deployed}}(\ell,k)\).

A fixed-direction control uses the evaluation-bank mean direction, rescaled to match each input's oracle norm.
Its response score is~\(U_{\mathrm{fixed}}\left(\ell\right)\).
Because the control receives the oracle norm, comparison with~\(U_{\mathrm{deployed}}\left(\ell,k\right)\) isolates the benefit of input-specific coordinates from perturbation magnitude.
High projected response at low rank indicates a compact shared subspace, while high deployed response shows that its input-specific coordinates are accessible from unperturbed representations.

\vspace{\mylen}
\section{Evaluation}
\label{sec:eval}
\vspace{\mylen}

\textbf{Models.}
We evaluate Chronos-2~(\texttt{C2})~\citep{ansari2025chronos2}, TimesFM~2.5~(\texttt{F2})~\citep{das2024timesfm}, TimesFM~3~(\texttt{F3})~\citep{jain2026timesfm3}, Toto-Open-Base~1.0~(\texttt{TO})~\citep{cohen2025toto}, Moirai~(\texttt{MO})~\citep{woo2024moirai}, Sundial~(\texttt{SU})~\citep{liu2025sundial}, Time-MoE~(\texttt{ME})~\citep{shi2025timemoe}, TTM-R2~(\texttt{TT})~\citep{ekambaram2024ttm}, and Timer-XL~(\texttt{TX})~\citep{liu2025timerxl}.
We use \(H=16\) and \(T=256\) by default.
The panel spans diverse architectures, scales, contexts, and training objectives, as detailed in~\S\ref{app:models}.

\textbf{Laws.}
The analysis includes thirteen laws, each instantiating a common state-space form whose parameters determine the dynamics, input, or readout.
Ten are single-parameter laws: linear trend~(\texttt{TR}), oscillation~(\texttt{OS}), mean reversion~(\texttt{MR}), random walk with drift~(\texttt{RW}), exponential relaxation~(\texttt{RX}), saturating growth~(\texttt{SG}), level shift~(\texttt{LS}), damped oscillation~(\texttt{DO}), seasonal autoregression~(\texttt{SA}), and transient pulse~(\texttt{TP}).
Three are multi-parameter laws requiring joint identification: dual tone~(\texttt{DT}; low tone~\texttt{TL}, high tone~\texttt{TH}, and amplitude ratio~\texttt{TW}), linear state space~(\texttt{SS}; modal frequency~\texttt{SC} and eigenvalue modulus~\texttt{SR}), and forced diffusion~(\texttt{FD}; drive frequency~\texttt{WC} and diffusion time~\texttt{WD}), yielding seven parameters in total.
Generators, parameter ranges, and constraints appear in~\S\ref{app:generative_laws}.

\textbf{Evaluation Protocol.}
All TSFM weights remain frozen.
Probes are fitted on training datasets, with layers and penalties selected on validation datasets and evaluated once on independent test datasets.
Each standard forecast-response dataset contains~32{,}768 matched pairs per intervention magnitude.
Larger datasets are used for high-variance settings to improve stability across runs.
Causal geometry uses separate fitting, validation, and test datasets.
Implementation details for representation accessibility, forecast response, and causal geometry appear in~\S\ref{app:representation_impl}, \S\ref{app:response_impl}, and~\S\ref{app:geometry_impl}, respectively.

\vspace{\mylen}
\subsection{Dynamical Law Identification}
\label{sec:eval_law}
\vspace{\mylen}

\begin{wrapfigure}{r}{0.56\textwidth}
    \centering
    \includegraphics[width=\linewidth]{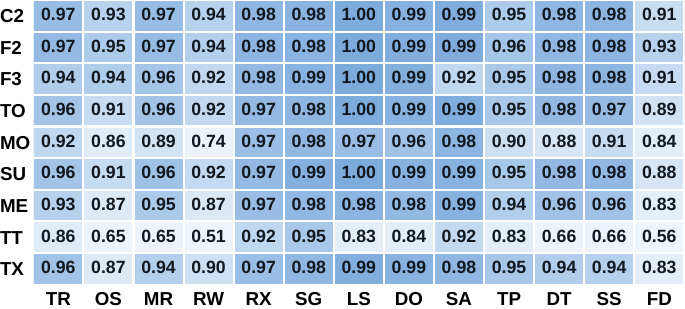}
    \caption{Thirteen-law decodability scores~\(D_m(\ell)\) across nine pretrained~TSFMs.}
    \label{fig:lawid}
\end{wrapfigure}

Dynamical-law identification tests whether TSFM representations distinguish the thirteen laws.
Balanced training, validation, and test banks treat each multi-parameter law as one class with jointly varying parameters.
A multinomial logistic probe is fitted at each layer, with layer and penalty selected on validation data.
For~\(D_m(\ell)\), we set chance accuracy to~\(A_0=1/13\) and the reference to~\(A_m^\star=1\), yielding normalized above-chance accuracy.
Results appear in~\autoref{fig:lawid}.

\textbf{Dynamical Laws Are Broadly Decodable.}
Every model distinguishes the thirteen laws far above chance.
Mean law decodability ranges from~\(0.757\) to~\(0.971\) across models, with a mean of~\(0.929\).
Thus, law identity remains broadly accessible even when parameters vary jointly, although this does not establish a unique learned mechanism.

\textbf{Difficulty Concentrates in Related Dynamics.}
Forced diffusion~(\texttt{FD}) has the lowest average decodability of~\(0.842\) and is the weakest law for seven models.
Random walk with drift~(\texttt{RW}) averages~\(0.851\) and is weakest for Moirai~(\texttt{MO}) and TTM-R2~(\texttt{TT}).
Other laws average at least~\(0.877\), indicating that errors concentrate in a small subset of dynamical families.

\textbf{Differences Are Model- and Law-Specific.}
Seven models average between~\(0.939\) and~\(0.971\), whereas Moirai~(\texttt{MO}) averages~\(0.908\) and TTM-R2~(\texttt{TT}) averages~\(0.757\).
TTM-R2~(\texttt{TT}) scores~\(0.51\) on random walk with drift~(\texttt{RW}) and~\(0.56\) on forced diffusion~(\texttt{FD}).
They are its weakest laws.
TimesFM~3~(\texttt{F3}) scores~\(0.92\) on seasonal autoregression~(\texttt{SA}), compared with~\(0.99\) for TimesFM~2.5~(\texttt{F2}), showing that aggregate scores can obscure law-specific differences.
Complete layer-wise profiles and confusion analyses for all models and thirteen laws appear in~\S\ref{app:law_results}.

\vspace{\mylen}
\subsection{Single-Parameter Laws}
\label{sec:eval_single}
\vspace{\mylen}

We study ten single-parameter laws.
Accessibility covers eight, excluding random walk with drift~(\texttt{RW}) and level shift~(\texttt{LS}) because their required references are unavailable.
Forecast response and causal geometry cover seven, additionally excluding exponential relaxation~(\texttt{RX}) because its response reference is unstable.
The common comparison spans seven laws and nine models, totaling 63 cells.

\vspace{\mylen}
\subsubsection{Representation Accessibility}
\label{sec:eval_representation}
\vspace{\mylen}

At each layer, a ridge probe predicts~\(z_{m,j}\), and its correlation relative to the law- and nuisance-informed reference~\(r_{m,j}^{\star}\) defines~\(D_{m,j}(\ell)\).
The layer and ridge penalty are selected on validation data and fixed for test evaluation.
The main comparison uses the forecast-response generative regime.

\begin{wrapfigure}{r}{0.45\textwidth}
    \centering
    \includegraphics[width=\linewidth]{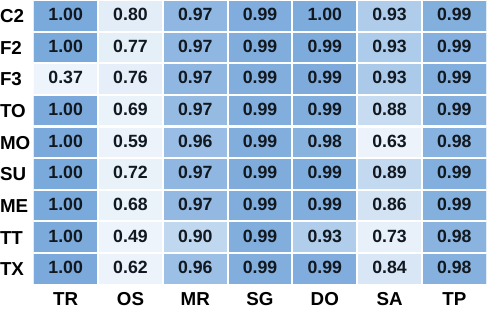}
    \caption{Representation accessibility~\(D_{m,j}\).
    }
    \label{fig:panel_D_laws}
\end{wrapfigure}

\textbf{Dynamical Parameters Are Broadly Accessible.}
In~\autoref{fig:panel_D_laws}, median accessibility across 63 common-regime model--law cells is~\(0.982\), and 42 cells exceed~\(0.95\).
Fifteen of 16 scores below~\(0.90\) occur for oscillation~(\texttt{OS}) or seasonal autoregression~(\texttt{SA}).
The exception is TimesFM~3~(\texttt{F3}) on linear trend~(\texttt{TR}), while the other five laws are highly accessible in nearly every model across diverse architectures and scales.

\textbf{Accessibility Reflects Both Preservation and Transformation.}
On the original eight-law probe banks, linear trend~(\texttt{TR}), saturating growth~(\texttt{SG}), and exponential relaxation~(\texttt{RX}) match near-ceiling raw-input controls, indicating preserved linear access.
Mean reversion~(\texttt{MR}), damped oscillation~(\texttt{DO}), and transient pulse~(\texttt{TP}) are highly accessible despite chance-level raw linear controls, showing TSFM processing makes them linearly accessible.

\textbf{Long-Range Parameters Differentiate Models.}
Common-regime accessibility ranges from~\(0.49\) to~\(0.80\) for oscillation~(\texttt{OS}) and from~\(0.63\) to~\(0.93\) for seasonal autoregression~(\texttt{SA}).
Layer profiles show early access to trend, whereas frequency has a near-zero raw-linear baseline and peaks beyond the network midpoint in seven models.
Exponential relaxation~(\texttt{RX}) ranges from~\(0.82\) for TimesFM~3~(\texttt{F3}) to~\(0.97\)--\(0.99\) for the other models.

Most scored parameters therefore approach the conditional reference.
Across 19 model--law cells, MLP probes improve median accessibility by~\(0.046\) for oscillation~(\texttt{OS}) and~\(0.029\) for seasonal autoregression~(\texttt{SA}), but do not recover the TimesFM~3~(\texttt{F3}) trend result.
Controls, per-law ranges, and layer-wise profiles for all models and laws appear in~\S\ref{app:representation_results}.

\vspace{\mylen}
\subsubsection{Forecast Response}
\label{sec:eval_response}
\label{sec:eval_resp_panel}
\vspace{\mylen}

Using the matched histories of~\S\ref{sec:framework}, we compare each TSFM forecast change with the reference change induced by the designated parameter.
For each law,~\(\bar U_{m,j}\) aggregates the evaluated intervention magnitudes and normalizes the reference-aligned response by the conditional reference.

\begin{wrapfigure}{r}{0.45\textwidth}
    \centering
    \includegraphics[width=\linewidth]{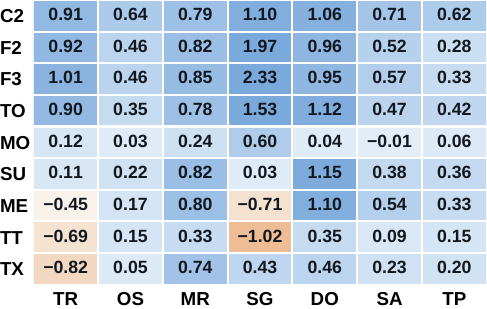}
    \caption{
    Forecast response score~\(\bar U_{m,j}\).
    }
    \label{fig:panel_U_laws}
\end{wrapfigure}
\textbf{Forecast Responses Are Often Weak or Reversed.}
In~\autoref{fig:panel_U_laws}, the median normalized aligned gain across 63 model--parameter cells is~\(0.46\), and only 15 cells lie within~\(20\%\) of the conditional reference.
Six responses remain negative across all intervention magnitudes.
No model reaches~\(0.8\) on oscillation~(\texttt{OS}), seasonal autoregression~(\texttt{SA}), or transient pulse~(\texttt{TP}), whose maxima are~\(0.64\), \(0.71\), and~\(0.62\).

\textbf{Accessibility and Forecast Response Diverge.}
Among 42 cells with accessibility above~\(0.95\), 20 have forecast response below~\(0.5\), and five respond oppositely.
Linear trend~(\texttt{TR}) provides an example: eight models have accessibility between~\(0.998\) and~\(0.999\), including three with negative responses.
Near-reference accessibility does not determine the magnitude or direction of the forecast response.

\textbf{Accessibility Explains Relative but Not Absolute Variation.}
Across models, accessibility and response have Spearman correlations of~\(0.92\) for oscillation~(\texttt{OS}) and~\(0.82\) for seasonal autoregression~(\texttt{SA}), the two laws with the largest accessibility variation.
Accessibility therefore helps distinguish models when parameter accessibility is limited, but even the strongest responses remain below the conditional reference.
Across all cells, the median centered cosine is~\(0.54\) and the median centered relative error is~\(1.06\), revealing additional disagreement in response magnitude and temporal shape.
Dose-specific and full-vector analyses appear in~\S\ref{app:response_results}.

\vspace{\mylen}
\subsubsection{Causal Geometry}
\label{sec:eval_geometry}
\label{sec:eval_geom_panel}
\vspace{\mylen}

Causal geometry explains the accessibility--response gap by comparing the hidden-state edit required for the expected response with the change induced by the parameter intervention.
We test whether the state supports the response, then compare required and input-induced changes.

\textbf{Direct Hidden-State Edits Recover Reference Responses.}
All 63 cells have a validation-selected layer where the Jacobian-derived edit passes the nonlinear response criterion.
At the calibrated step, median aligned gain is~\(1.00\) and centered error is~\(0.06\).
Projection onto the shared subspace retains median gain~\(0.93\) with centered error~\(0.18\).
Weak matched-input responses therefore coexist with effective edits that produce the reference response across models and dynamical laws alike.

\begin{wrapfigure}{r}{0.45\textwidth}
    \centering
    \includegraphics[width=\linewidth]{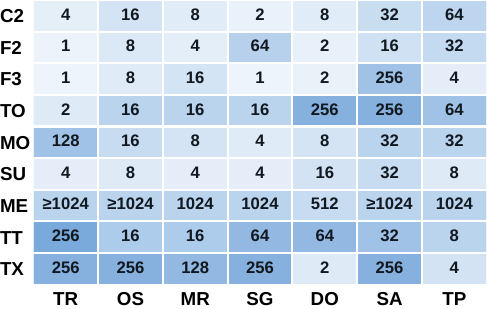}
    \caption{Shared causal-subspace rank~\(k_{90}\).}
    \label{fig:panel_k90_laws}
\end{wrapfigure}
\textbf{Effective Edits Share Compact Structure.}
\autoref{fig:panel_k90_laws} shows a median rank of~\(k_{90}=16\) for retaining~\(90\%\) of local aligned gain.
Targets shuffled across inputs within the same dynamical law have the same median rank and tie the correct targets in 51 of 63 cells.
The correct targets improve both rank and rank-curve area in only six cells.
Compactness therefore reflects shared causal-subspace structure but is not consistently tied to the correct input--response pairing.

\begin{wrapfigure}{r}{0.45\textwidth}
    \centering
    \includegraphics[width=\linewidth]{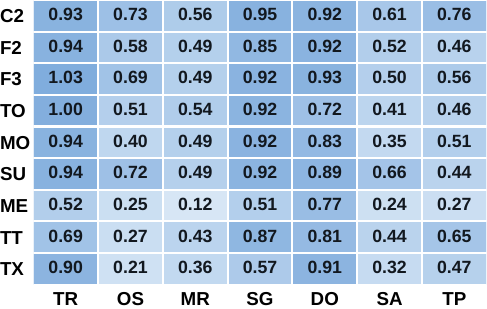}
    \caption{Deployed pooled aligned gain.}
    \label{fig:panel_deployed_laws}
\end{wrapfigure}
\textbf{Predicted Edits Partially Recover the Response.}
A readout predicts each input's edit from the original state without the corresponding test target or Jacobian.
\autoref{fig:panel_deployed_laws} shows that it outperforms the fixed-direction control in 56 of 63 cells, with median aligned gains of~\(0.58\) and~\(0.23\).
Its centered error remains~\(0.68\), versus~\(0.18\) for the projected edit.
The original hidden state therefore predicts part of the required edit.

\textbf{Input Changes Miss the Required Directions.}
Input-induced hidden-state changes have median cosine~\(0.028\) with the required local edits.
Only~\(1.1\%\) of their energy lies in the shared subspace, while~\(99.6\%\) lies in the downstream Jacobian's null space.
Their median linearized aligned gain is~\(0.28\), indicating a weak reference-aligned effect on the forecast.
Together, these results show that input changes follow a different hidden-state path from the effective edits producing the reference response across the panel.

The reference response can be produced from the selected state, but the parameter intervention often fails to induce the required hidden-state direction.
Complete causal-geometry results appear in~\S\ref{app:geometry_results}.

\vspace{\mylen}
\subsection{Multi-Parameter Laws}
\label{sec:eval_multi}
\vspace{\mylen}

\begin{figure}[t]
\centering
\subfigure[Representation accessibility.]{
\includegraphics[width=.315\textwidth]{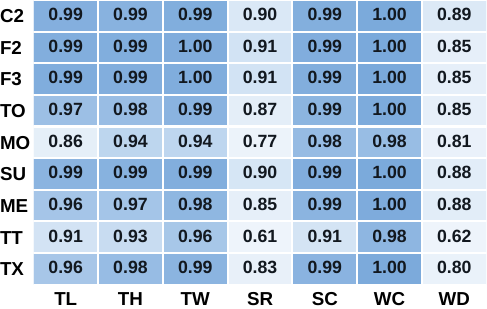}
\label{fig:multi_parameter_accessibility}}
\subfigure[Matched-input response.]{
\includegraphics[width=.315\textwidth]{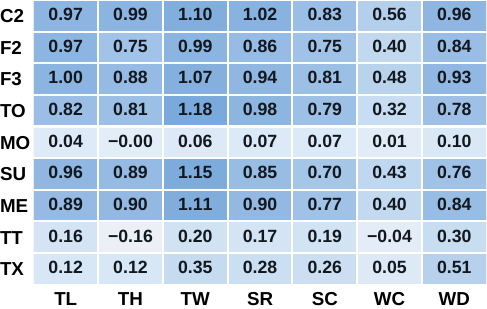}
\label{fig:multi_parameter_response}}
\subfigure[Deployed gain.]{
\includegraphics[width=.315\textwidth]{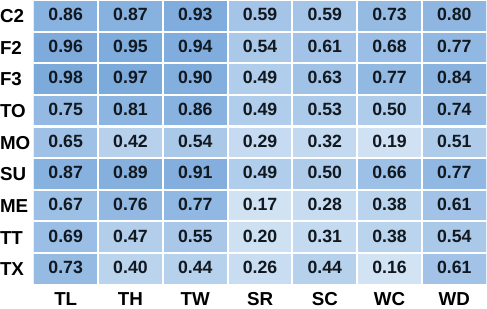}
\label{fig:multi_parameter_deployed}}
\vspace{-0.05in}
\caption{Multi-parameter accessibility, matched-input response, and deployed gain across joint laws.}
\label{fig:multi_parameter_summary}
\end{figure}

For seven parameters across three multi-parameter laws, each intervention changes one parameter while holding the others and the stochastic realization fixed.
We compare representation accessibility, matched-input forecast response, and deployed aligned gain.
\autoref{fig:multi_parameter_summary} summarizes the three measurements.

\textbf{Joint Parameters Remain Accessible.}
Median accessibility is~\(0.977\).
For linear state space, modal frequency~(\texttt{SC}) exceeds eigenvalue modulus~(\texttt{SR}) by~\(0.08\)--\(0.30\) in every model.
Adding a matched high tone~(\texttt{TH}) reduces low-tone~(\texttt{TL}) accessibility by at most~\(0.033\) in six models.
Joint components therefore produce limited interference, but accessibility still varies across parameters of the same law.

\textbf{Accessibility and Forecast Response Diverge.}
Drive frequency~(\texttt{WC}) is accessible at~\(0.98\)--\(1.00\), yet no matched-input response exceeds~\(0.56\).
Diffusion time~(\texttt{WD}) is less accessible, but six models attain responses between~\(0.76\) and~\(0.96\).
This reversal within forced diffusion further separates accessibility from forecast response under jointly varying parameter settings.

\textbf{Predicted Edits Partially Recover the Response.}
Deployed directions outperform the fixed control in 56 of 63 cells across all three multi-parameter laws.
Median centered errors are~\(0.93\), \(0.14\), and~\(0.61\) for matched-input, projected, and deployed responses.
TTM-R2~(\texttt{TT}), Timer-XL~(\texttt{TX}), and Moirai~(\texttt{MO}) attain deployed gains between~\(0.16\) and~\(0.73\) despite weak or opposing matched-input responses.
Thus, accessible parameters can support the reference response without being fully expressed by input changes or predicted edits.
Further analyses appear in~\S\ref{app:multi_results}.

\vspace{\mylen}
\subsection{Controlled Changes on Real-Series Backgrounds}
\label{sec:eval_real}
\vspace{\mylen}

The controlled-law experiments isolate parameter effects but use synthetic backgrounds.
To test whether the accessibility--response separation persists on real data, centered trend or sinusoid components are injected into ETTh1~\cite{zhou2021informer} windows.
Matched pairs share the background and differ only in the injected component.
Injection strength is the component root-mean-square (RMS) amplitude relative to context standard deviation, at~\(0.25\), \(0.5\), and~\(1.0\).
The analytic future difference defines the reference response.
All nine TSFMs are evaluated on~\(2{,}048\) test windows per setting.
No-injection and shuffled-future controls yield median pooled gains of~\(0.000\) and~\(-0.000\).

\begin{wrapfigure}{r}{0.46\linewidth}
    \centering
    \includegraphics[width=\linewidth]{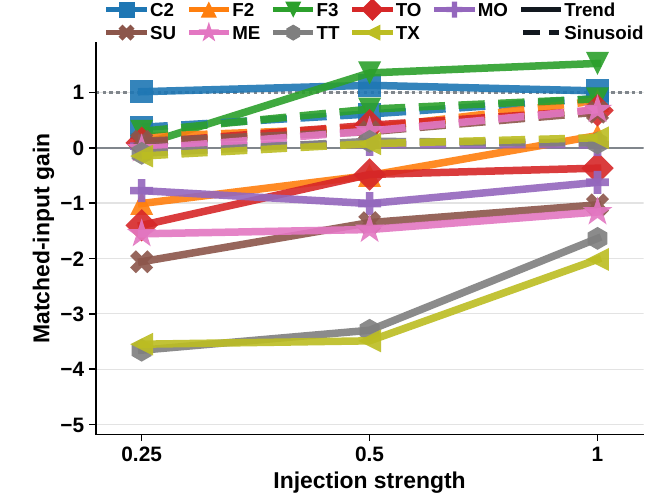}
    \caption{
    Matched-input responses on ETTh1.
    }
    \label{fig:eval_real}
\end{wrapfigure}

\textbf{Results.}
Across the evaluation, median probe correlation remains high at~\(0.94\), while pooled matched-input gain is~\(-0.50\) for linear trend and~\(0.36\) for sinusoid.
As shown in~\autoref{fig:eval_real}, ETTh1 exhibits a separation between representation accessibility and forecast response across three injection strengths, with linear trend gain remaining negative from~\(-1.40\) at strength~\(0.25\) to~\(-0.62\) at strength~\(1.0\), while sinusoid gain increases from~\(0.09\) to~\(0.67\).
The response depends on the background: at strength~\(1.0\), linear trend gain reaches~\(0.89\) on Electricity and~\(0.71\) on Weather, while Weather sinusoid gain reaches~\(0.77\).
Thus, the accessibility--response separation persists on real-series backgrounds, although its magnitude and sign vary with the background and injection strength.
These gains use the injected continuation rather than a conditional reference for the background.
The full comparison across ETTh1, Electricity, and Weather is reported in~\S\ref{app:eval_real}.

\textbf{Additional Analyses.}
The appendix reports mechanistic diagnostics in~\S\ref{app:cross_measure}, reference sensitivity in~\S\ref{app:revision_bayes}, intervention side effects in~\S\ref{app:revision_sideeffects}, computational cost in~\S\ref{app:revision_cost}, and limitations in~\S\ref{app:limitations}.
Together, these analyses characterize the robustness and scope of the main findings.

\vspace{\mylen}
\section{Conclusion}
\label{sec:conclusion}
\vspace{\mylen}

Across nine~TSFMs and thirteen controlled laws, \emph{Dynamical Parameters} are typically linearly accessible even when matched-input forecast responses are weak or reversed.
Projected hidden-state edits recover the reference responses with low error, showing that the selected hidden states can support forecast changes not elicited by the corresponding input interventions.
Input-induced changes align weakly with the required directions, while deployed edits constructed from predicted coordinates recover these responses only partially.
The analysis therefore distinguishes what a~TSFM represents, what its forecast expresses, and what its hidden states can produce under intervention.


\section*{Acknowledgments}
The research reported in this paper was sponsored in part by the DEVCOM Army Research Laboratory~(award \# W911NF1720196), the National Science Foundation (award \# 2325956), and the National Institutes of Health (award \# 1P41EB028242). The views and conclusions contained in this document are those of the authors and should not be interpreted as representing the official policies, either expressed or implied, of the funding agencies.

\bibliographystyle{iclr2027_conference}

\newpage

\setcounter{tocdepth}{2}
\renewcommand{\contentsname}{Appendix Contents}

{\tableofcontents}

\clearpage

\addtocontents{toc}{\protect\setcounter{tocdepth}{2}}

\appendix

\section{Implementation Details}
\label{app:implementation}

This appendix specifies the model checkpoints, controlled generators, reference computations, data splits, selection procedures, and intervention settings.
The organization follows representation accessibility, forecast response, and causal geometry.

\subsection{Models and Forecast Outputs}
\label{app:models}

\textbf{Checkpoints and Architectures.}
\autoref{tab:models} reports the checkpoint, parameter count, depth~\(L\), hidden width~\(d\), context length~\(T\), and architecture of each TSFM.
Parameter counts follow the published Hugging Face safetensors metadata, while~\(L\), \(d\), and~\(T\) follow our evaluation configurations.
The models span patched and masked encoders, decoder-only Transformers, a mixture-of-experts model, and an MLP-mixer, ranging from approximately~\(0.8\)M to~\(453\)M parameters and three to twenty layers.
Time-MoE's count includes all experts, with two active per token.
Context length is~\(T=256\) for seven models, \(T=288\) for Timer-XL, and~\(T=512\) for TTM-R2.

\begin{table}[t]
\centering
\caption{
Model checkpoints and configurations used throughout the evaluation.
}
\label{tab:models}
\setlength{\tabcolsep}{2.2pt}
\resizebox{\linewidth}{!}{%
\begin{tabular}{llrrrrl}
\toprule
Model & Checkpoint & Params & \(L\) & \(d\) & \(T\) & Architecture \\
\midrule
Chronos-2~\citep{ansari2025chronos2}
& \texttt{amazon/chronos-2}
& 119.5M & 12 & 768 & 256 & Patched encoder \\
TimesFM~2.5~\citep{das2024timesfm}
& \texttt{google/timesfm-2.5-200m-pytorch}
& 231.3M & 20 & 1280 & 256 & Decoder-only \\
TimesFM~3~\citep{jain2026timesfm3}
& \texttt{google/timesfm-3.0-pytorch}
& 330.7M & 20 & 1280 & 256 & Decoder-only \\
Toto-Open-Base~1.0~\citep{cohen2025toto}
& \texttt{Datadog/Toto-Open-Base-1.0}
& 151.3M & 12 & 768 & 256 & Decoder-only \\
Moirai~\citep{woo2024moirai}
& \texttt{Salesforce/moirai-1.1-R-small}
& 13.8M & 6 & 384 & 256 & Masked encoder \\
Sundial~\citep{liu2025sundial}
& \texttt{thuml/sundial-base-128m}
& 128.3M & 12 & 768 & 256 & Decoder-only \\
Time-MoE~\citep{shi2025timemoe}
& \texttt{Maple728/TimeMoE-200M}
& 453.2M & 12 & 768 & 256 & Decoder-only, MoE \\
TTM-R2~\citep{ekambaram2024ttm}
& \texttt{ibm-granite/granite-timeseries-ttm-r2}
& 805K & 3 & 192 & 512 & MLP-mixer \\
Timer-XL~\citep{liu2025timerxl}
& \texttt{thuml/timer-base-84m}
& 84.1M & 8 & 1024 & 288 & Decoder-only \\
\bottomrule
\end{tabular}
}
\end{table}

\textbf{Forecast Heads and Monte Carlo Sampling.}
\autoref{tab:sampling} summarizes the forecast statistic used for each model.
For stochastic backends, we average~64 sampled paths and use the same seed across matched arms to reduce Monte Carlo variance.

\begin{table}[t]
\centering
\caption{Forecast heads and sampling settings.}
\label{tab:sampling}
\footnotesize
\setlength{\tabcolsep}{4pt}
\renewcommand{\arraystretch}{1.12}
\begin{tabularx}{\linewidth}{
l
>{\hsize=1.3\hsize\raggedright\arraybackslash}X
c
>{\hsize=0.9\hsize\raggedright\arraybackslash}X
>{\hsize=0.8\hsize\raggedright\arraybackslash}X
}
\toprule
Model & Forecast head & Paths & Statistic & Seeding \\
\midrule
Chronos-2 & 21-quantile head & -- & Median quantile & -- \\
TimesFM~2.5 & Flip-averaged pinball head & -- & Median quantile & -- \\
TimesFM~3 & Quantile head & -- & Median quantile & -- \\
Toto-Open-Base~1.0 & Autoregressive mean & -- & Distribution mean & -- \\
Moirai & Mixture head & 64 & Path mean & Shared across arms \\
Sundial & Flow matching, 50 steps & 64 & Path mean & Shared across arms \\
Time-MoE & Greedy regression & -- & Point output & -- \\
TTM-R2 & MLP-mixer & -- & Point output & -- \\
Timer-XL & Greedy regression & -- & Point output & -- \\
\bottomrule
\end{tabularx}
\end{table}

\subsection{Controlled Dynamical Laws}
\label{app:generative_laws}

The thirteen controlled laws instantiate the state-space framework of~\S\ref{sec:framework}.
Each sequence contains a history of length~\(T\) and a future of length~\(H\).
The first ten laws each have one designated~\emph{Dynamical Parameter}, while the remaining three contain multiple parameters that are varied individually.
All other quantities are treated as nuisance variables and held fixed within each matched pair.

\subsubsection{General State-Space Form}
\label{app:general_form}

Each law is represented by a latent state~\(s_t\), input~\(u_t\), and scalar clean trajectory~\(f_t\):
\[
s_t = A s_{t-1} + B u_t + G\eta_t,
\qquad
f_t = h(s_t),
\]
where~\(\eta_t\) denotes process noise.
Eleven laws use the linear readout~\(h(s_t)=Cs_t\).
Saturating growth and transient pulse instead apply fixed nonlinear readouts to the clock state~\(s_t=s_{t-1}+1\).
The designated parameters enter the transition, input, or readout, as summarized in~\autoref{tab:general_form}.

State-space coordinates are not unique because the transformation~\(s_t\mapsto Ms_t\) preserves the observations when the transition and readout are transformed accordingly.
We therefore define the designated parameters through the generator dynamics or observed trajectory rather than arbitrary state coordinates.
Examples include eigenvalue moduli and angles, forcing frequencies and gains, and readout-shape parameters.

\begin{table}[t]
\centering
\caption{Controlled laws in the common state-space form.}
\label{tab:general_form}
\setlength{\tabcolsep}{3.5pt}
\resizebox{\linewidth}{!}{%
\begin{tabular}{llcllll}
\toprule
Law & Code & \(\dim s_t\) & Process noise & Input~\(u_t\) & Readout~\(h\) & Parameter location \\
\midrule
Linear trend & \texttt{TR} & 1 & -- & Constant & Linear & Input drift~\(S\) \\
Oscillation & \texttt{OS} & 2 & -- & -- & Linear & Transition angle~\(\omega\) \\
Mean reversion & \texttt{MR} & 1 & \checkmark & Constant & Linear & Transition eigenvalue~\(\rho\) \\
Random walk with drift & \texttt{RW} & 1 & \checkmark & Constant & Linear & Input drift~\(d\) \\
Exponential relaxation & \texttt{RX} & 1 & -- & Constant & Linear & Transition eigenvalue~\(e^{-\gamma}\) \\
Saturating growth & \texttt{SG} & 1 & -- & Constant & Logistic & Readout growth rate~\(r\) \\
Level shift & \texttt{LS} & 1 & -- & Impulse at~\(\tau\) & Linear & Input gain~\(\Delta\) \\
Damped oscillation & \texttt{DO} & 2 & -- & -- & Linear & Transition modulus~\(e^{-\gamma}\) \\
Seasonal autoregression & \texttt{SA} & \(P\) & \checkmark & Constant & Linear & Lag-\(P\) coefficient~\(\rho_s\) \\
Transient pulse & \texttt{TP} & 1 & -- & Constant & Gaussian & Readout width~\(w\) \\
\midrule
Dual tone & \texttt{DT} & 4 & -- & -- & Linear & Angles~\(\omega_L,\omega_H\), ratio~\(\kappa\) \\
Linear state space & \texttt{SS} & 2 & \checkmark & Constant & Linear & Modulus~\(r\), angle~\(\theta\) \\
Forced diffusion & \texttt{FD} & Distributed & -- & Periodic & Linear & Frequency~\(\omega\), diffusion operator~\(\Lambda\) \\
\bottomrule
\end{tabular}
}
\end{table}

\subsubsection{Individual Dynamical Laws}
\label{app:individual_laws}

\autoref{tab:law_coords} summarizes the parameter coordinates and law-identification ranges, followed by the generator definitions.
Parameters of each multi-parameter law are sampled independently.
For accessibility, we use~\(\rho,\rho_s\in\left[0.02,0.98\right]\), \(R\in\left[3,13\right]\), \(g\in\left[0.1,3\right]\) for damped oscillation, and~\(w/T\in\left[0.025,0.2\right]\) for transient pulse.
Exponential relaxation uses the same range for law identification and accessibility.

\begin{table}[t]
\centering
\caption{
Dimensionless~\emph{Dynamical Parameters} and law-identification ranges. 
}
\label{tab:law_coords}
\scriptsize
\setlength{\tabcolsep}{3.5pt}
\resizebox{\linewidth}{!}{%
\begin{tabular}{clllll}
\toprule
\# & Law & Code & Physical & Coordinate~\(z_{m,j}\) & Range \\
\midrule
1 & Linear Trend
& \texttt{TR}
& \(S\)
& \(\widetilde S=\operatorname{sign}\left(S\right)\sigma_{\mathrm{clean}}/\sigma_\varepsilon\)
& \(\pm\left[1,6\right]\) \\
2 & Oscillation
& \texttt{OS}
& \(\omega\)
& \(C=\omega T/2\pi\)
& \(\left[2,8\right]\) \\
3 & Mean Reversion
& \texttt{MR}
& \(\rho\)
& \(\rho\)
& \(\left[0.3,0.9\right]\) \\
4 & Random Walk with Drift
& \texttt{RW}
& \(d\)
& \(D=d\sqrt{T}/\sigma_\eta\)
& \(\pm\left[0.5,3\right]\) \\
5 & Exponential Relaxation
& \texttt{RX}
& \(\gamma\)
& \(g=\gamma T\)
& \(\left[0.8,2.5\right]\) \\
6 & Saturating Growth
& \texttt{SG}
& \(r\)
& \(R=rT\)
& \(\left[5,12\right]\) \\
7 & Level Shift
& \texttt{LS}
& \(\Delta\)
& \(\widetilde\Delta=\operatorname{sign}\left(\Delta\right)\sigma_{\mathrm{clean}}/\sigma_\varepsilon\)
& \(\pm\left[1,6\right]\) \\
8 & Damped Oscillation
& \texttt{DO}
& \(\gamma\)
& \(g=\gamma T\)
& \(\left[0.6,2.8\right]\) \\
9 & Seasonal Autoregression
& \texttt{SA}
& \(\rho_s\)
& \(\rho_s\)
& \(\left[0.3,0.9\right]\) \\
10 & Transient Pulse
& \texttt{TP}
& \(w\)
& \(w/T\)
& \(\left[0.04,0.2\right]\) \\
\midrule
11 & Dual Tone
& \texttt{DT}
& \(\omega_L,\omega_H,A_H/A_L\)
& \(C_L,C_H,\kappa\)
& \(\left[2,8\right],\left[10,26\right],\left[0.4,2.5\right]\) \\
& & \(\left(\texttt{TL},\texttt{TH},\texttt{TW}\right)\) & & & \\
12 & Linear State Space
& \texttt{SS}
& \(\left|\lambda\left(A\right)\right|,\arg\lambda\left(A\right)\)
& \(r,C_m\)
& \(\left[0.85,0.98\right],\left[12,32\right]\) \\
& & \(\left(\texttt{SR},\texttt{SC}\right)\) & & & \\
13 & Forced Diffusion
& \texttt{FD}
& \(\omega,z^2/\kappa_d\)
& \(C_d,\Lambda\)
& \(\left[2,8\right],\left[0.002,0.1\right]\) \\
& & \(\left(\texttt{WC},\texttt{WD}\right)\) & & & \\
\bottomrule
\end{tabular}
}
\end{table}

\textbf{1. Linear Trend~(\texttt{TR}).}
\[
x_t=b+St+\varepsilon_t,
\qquad
z_{m,1}=\widetilde S
=
\operatorname{sign}(S)
\frac{\sigma_{\mathrm{clean}}}{\sigma_\varepsilon}.
\]
Here, \(b\) is the baseline, \(S\) is the slope, and~\(\varepsilon_t\) is observation noise with standard deviation~\(\sigma_\varepsilon\).
The quantity~\(\sigma_{\mathrm{clean}}\) is the standard deviation of the noiseless history, so~\(\widetilde S\) measures signed trend strength relative to observation noise.

\textbf{2. Oscillation~(\texttt{OS}).}
\[
x_t=b+A\sin(\omega t+\phi)+\varepsilon_t,
\qquad
z_{m,1}=C=\frac{\omega T}{2\pi}.
\]
Here, \(b\) is the baseline, \(A\) the amplitude, \(\omega\) the angular frequency, \(\phi\) the phase, and~\(\varepsilon_t\) observation noise.
The designated parameter~\(C\) is the number of cycles within the observed context.
Amplitude, phase, and baseline are nuisance variables.

\textbf{3. Mean Reversion~(\texttt{MR}).}
\[
s_t=\mu+\rho(s_{t-1}-\mu)+\eta_t,
\qquad
x_t=s_t+\varepsilon_t,
\qquad
z_{m,1}=\rho.
\]
Here, \(\mu\) is the long-run mean, \(\rho\) controls persistence, \(\eta_t\) is process noise, and~\(\varepsilon_t\) is observation noise.
The latent state is initialized from its stationary distribution.
Larger~\(\rho\) produces slower reversion toward~\(\mu\) and longer-lasting effects on future values.

\textbf{4. Random Walk with Drift~(\texttt{RW}).}
\[
s_t=s_{t-1}+d+\eta_t,
\qquad
x_t=s_t+\varepsilon_t,
\qquad
z_{m,1}=D=\frac{d\sqrt{T}}{\sigma_\eta}.
\]
Here, \(d\) is the drift, \(\eta_t\) is process noise with standard deviation~\(\sigma_\eta\), and~\(\varepsilon_t\) is observation noise.
The designated parameter~\(D\) measures cumulative drift relative to diffusion over the observed context.
Unlike linear trend, process innovations accumulate through time.
This law is included in dynamical-law identification but excluded from parameter-level analyses because its conditional reference is unstable under the evaluated finite-history sampling regime.

\textbf{5. Exponential Relaxation~(\texttt{RX}).}
\[
x_t=b+A\exp(-\gamma t)+\varepsilon_t,
\qquad
z_{m,1}=g=\gamma T.
\]
Here, \(b\) is the equilibrium level, \(A\) the initial displacement, \(\gamma\) the decay rate, and~\(\varepsilon_t\) observation noise.
The designated parameter~\(g\) expresses the decay rate relative to the context length, with larger values producing faster relaxation.
The sampled range excludes trajectories that appear nearly linear or reach equilibrium before the forecast origin.

\textbf{6. Saturating Growth~(\texttt{SG}).}
\[
x_t
=
b+\frac{A}{1+\exp\!\left[-r(t-\tau)\right]}
+\varepsilon_t,
\qquad
z_{m,1}=R=rT.
\]
Here, \(b\) is the initial level, \(A\) the total increase, \(r\) the growth rate, \(\tau\) the transition midpoint, and~\(\varepsilon_t\) observation noise.
The designated parameter~\(R\) expresses growth rate relative to context length, with larger values producing a sharper transition.
The independently sampled midpoint~\(\tau\) remains fixed within each matched intervention pair.

\textbf{7. Level Shift~(\texttt{LS}).}
\[
x_t=b+\Delta\mathbf{1}[t\geq\tau]+\varepsilon_t,
\qquad
z_{m,1}=\widetilde\Delta
=
\operatorname{sign}(\Delta)
\frac{\sigma_{\mathrm{clean}}}{\sigma_\varepsilon}.
\]
Here, \(b\) is the initial level, \(\Delta\) the signed shift magnitude, \(\tau\) the change point, and~\(\varepsilon_t\) observation noise.
The change point lies within the observed context so that both levels are visible and remain distinguishable despite observation noise.
After history standardization,~\(\widetilde\Delta\) represents signed shift strength relative to observation noise rather than raw jump magnitude.
This law contributes only to dynamical-law identification, with no parameter-level scores reported.

\textbf{8. Damped Oscillation~(\texttt{DO}).}
\[
x_t
=
b+A\exp(-\gamma t)\sin(\omega t+\phi)+\varepsilon_t,
\qquad
z_{m,1}=g=\gamma T.
\]
Here, \(b\) is the baseline, \(A\) the amplitude, \(\gamma\) the decay rate, \(\omega\) the angular frequency, \(\phi\) the phase, and~\(\varepsilon_t\) observation noise.
The designated parameter~\(g\) expresses envelope decay relative to context length, with larger values producing faster damping.
Frequency, phase, amplitude, and baseline remain fixed within each matched intervention pair.

\textbf{9. Seasonal Autoregression~(\texttt{SA}).}
\[
s_t=\mu+\rho_s(s_{t-P}-\mu)+\eta_t,
\qquad
x_t=s_t+\varepsilon_t,
\qquad
z_{m,1}=\rho_s.
\]
Here, \(\mu\) is the long-run mean, \(P\) the seasonal period, \(\eta_t\) process noise, and~\(\varepsilon_t\) observation noise.
The designated parameter~\(\rho_s\) controls dependence on the state one period earlier, with larger values producing stronger seasonal persistence.
The period~\(P\) is selected so that several repetitions occur within the context and remains fixed within each matched pair.

\textbf{10. Transient Pulse~(\texttt{TP}).}
\[
x_t
=
b+A\exp\!\left[-\frac{(t-\tau)^2}{2w^2}\right]
+\varepsilon_t,
\qquad
z_{m,1}=\frac{w}{T}.
\]
Here, \(b\) is the baseline, \(A\) the pulse amplitude, \(\tau\) the pulse center, \(w\) the width, and~\(\varepsilon_t\) observation noise.
The designated parameter~\(w/T\) measures pulse duration relative to context length, with larger values producing broader transients.
The center~\(\tau\) lies near the history-to-forecast boundary and remains fixed within each matched pair.

\textbf{Multi-Parameter Laws.}
The final three laws contain multiple parameters that vary jointly across histories.
Each matched intervention changes one designated parameter while holding the remaining parameters, phases, and stochastic realization fixed.

\textbf{11. Dual Tone~(\texttt{DT}) --- Designated Parameter Codes: \texttt{TL}, \texttt{TH}, and \texttt{TW}.}
\[
x_t
=
b+A\!\left[
\cos(\omega_L t+\phi_L)
+\kappa\cos(\omega_H t+\phi_H)
\right]
+\varepsilon_t,
\qquad
z_m=(C_L,C_H,\kappa),
\]
where~\(C_L=\omega_LT/2\pi\) and~\(C_H=\omega_HT/2\pi\) are the numbers of low- and high-frequency cycles per context.
The parameter codes are low-tone frequency~\(C_L\)~(\texttt{TL}), high-tone frequency~\(C_H\)~(\texttt{TH}), and amplitude ratio~\(\kappa\)~(\texttt{TW}).
Here, \(b\) is the baseline, \(A\) the low-tone amplitude, \(\phi_L\) and~\(\phi_H\) the phases, and~\(\varepsilon_t\) observation noise.
Following dual-tone multi-frequency signaling~\citep{itu1988q23}, the frequencies are drawn from disjoint, resolvable bands.
The non-designated frequencies, phases, amplitude, and baseline remain fixed within each matched pair.

\textbf{12. Linear State Space~(\texttt{SS}) --- Designated Parameter Codes: \texttt{SR} and \texttt{SC}.}
\[
s_t=As_{t-1}+B\eta_t,
\qquad
x_t=Cs_t+\varepsilon_t,
\qquad
z_m=(r,C_m),
\]
where~\(A\) has eigenvalues~\(re^{\pm i\theta}\) and~\(C_m=\theta T/2\pi\).
The parameter codes are modal persistence~\(r\)~(\texttt{SR}) and modal frequency~\(C_m\)~(\texttt{SC}).
Here, \(A\) governs state evolution, \(B\) loads the process noise~\(\eta_t\), \(C\) maps the state to the observation, and~\(\varepsilon_t\) is observation noise.
The observed component satisfies
\[
v_t
=
2r\cos(\theta)v_{t-1}
-r^2v_{t-2}
+\eta_t
\]
and is initialized from its exact stationary distribution.
The designated parameters are invariant to changes in state coordinates.
The sampled range avoids regimes where persistence and frequency are difficult to distinguish.
The non-designated parameter remains fixed within each matched pair.

\textbf{13. Forced Diffusion~(\texttt{FD}) --- Designated Parameter Codes: \texttt{WC} and \texttt{WD}.}

The forced-diffusion law is the periodic solution of the one-dimensional heat equation in a semi-infinite medium~\citep{carslaw1959conduction}:
\[
\begin{aligned}
x_t
&=
b+A\sum_{k\in\{1,3,5\}}
\frac{e^{-\zeta\sqrt{k}}}{k}
\\[-2pt]
&\qquad\times
\sin\!\left(k(\omega t+\phi)-\zeta\sqrt{k}\right)
+\varepsilon_t,
\\
z_m&=(C_d,\Lambda).
\end{aligned}
\]
The parameter codes are drive frequency~\(C_d=\omega T/2\pi\)~(\texttt{WC}) and normalized diffusion time~\(\Lambda=z^2/(\kappa_dT)\)~(\texttt{WD}), with~\(\zeta=\sqrt{\pi C_d\Lambda}\).
Here, \(b\) is the baseline, \(A\) the drive amplitude, \(\phi\) the phase, and~\(\varepsilon_t\) observation noise.
The variable~\(z\) denotes physical depth and~\(\kappa_d\) thermal diffusivity, distinct from the parameter vector~\(z_m\).
The first three odd harmonics approximate the square-wave surface drive across the observed and forecast horizons.
Diffusion attenuates and delays harmonic~\(k\) according to~\(\zeta\sqrt{k}\), allowing~\(\Lambda\) to be identified from waveform dispersion.
This construction follows \AA ngstr\"om's periodic method for measuring thermal diffusivity~\citep{angstrom1863new}.
Drive phase, amplitude, and baseline remain fixed within each matched pair throughout drive-frequency and diffusion-time interventions.

\subsubsection{Sequence Standardization and Context-Length Matching}
\label{app:generator_standardization}

\textbf{Dataset Scope.}
The following construction applies to law identification and the parameter-accessibility datasets.
Trend and oscillation use separate raw-scale priors and fixed noise settings, while the response experiments use separate parameter ranges.
Matched shape-law interventions share the baseline arm's normalization statistics across both arms.

\textbf{History Standardization.}
Let~\(\mu_{1:T}\) and~\(\sigma_{1:T}\) denote the mean and standard deviation of the clean trajectory~\(f_m\) over the observed history.
For parameter~\(z\) and nuisance variables~\(\nu\), the standardized trajectory used to construct the observed sequence is
\[
\widetilde f_m(t,z,\nu)
=
\frac{
f_m(t,z,\nu)-\mu_{1:T}
}{
\sigma_{1:T}
}.
\]
Each clean history has zero mean and unit variance before scale, offset, and observation noise are applied.
Matched response pairs use the same baseline normalization statistics.

\textbf{Common Scale and Noise.}
For law identification, the scale variable~\(V\) and nominal signal-to-noise ratio~\(\mathrm{SNR}\) follow common distributions across laws.
The observed sequence is
\[
x_t
=
\mathrm{scale}\,\widetilde f_m(t)
+c
+\sigma_\varepsilon\varepsilon_t,
\qquad
\sigma_\varepsilon
=
\frac{V}{\sqrt{1+\mathrm{SNR}^2}},
\qquad
\mathrm{scale}
=
\mathrm{SNR}\,\sigma_\varepsilon.
\]
The offset~\(c\) is proportional to~\(V\).
This construction matches the offset, scale, and nominal~\(\mathrm{SNR}\) distributions across laws.

\textbf{Context-Length Matching.}
Dimensionless parameters are defined relative to~\(T\) when needed, giving them comparable meanings across context lengths.
Because longer noisy histories provide more evidence, the sampled signal-to-noise ratio is multiplied by
\[
\sqrt{\frac{T_{\mathrm{ref}}}{T}},
\qquad
T_{\mathrm{ref}}=256,
\]
which approximately preserves~\(\mathrm{SNR}^2T\) across models.

\textbf{Trend and Level-Shift Parameters.}
History standardization removes the raw scale of linear trend and level shift, so law identification expresses their strengths relative to observation noise.
Trend parameter analyses use separate slope and noise settings.
Level shift is included only in law identification.

\subsubsection{Scoring Coverage and Generator Constraints}
\label{app:law_constraints}

Parameter ranges keep the laws distinguishable, avoid degenerate regimes, and ensure that each designated parameter affects the forecast horizon.
All thirteen laws enter dynamical-law identification.
Random walk with drift lacks a stable parameter reference, while level shift is included only in law identification.
Exponential relaxation has an accessibility score but no response or causal-geometry score because its response reference is unstable.
\autoref{tab:law_checks} summarizes the construction criteria and parameter-level coverage for all thirteen controlled laws.

\begin{table}[t]
\centering
\caption{
Generator constraints and scoring coverage.
\texttt{LID} denotes law identification, \texttt{ACC} representation accessibility, \texttt{RSP} forecast response, and \texttt{GEO} causal geometry.
}
\label{tab:law_checks}
\scriptsize
\setlength{\tabcolsep}{2.8pt}
\renewcommand{\arraystretch}{1.12}
\begin{tabularx}{\linewidth}{@{}
>{\raggedright\arraybackslash}p{0.21\linewidth}
>{\raggedright\arraybackslash}X
cccc
@{}}
\toprule
Law & Construction criterion & \texttt{LID} & \texttt{ACC} & \texttt{RSP} & \texttt{GEO} \\
\midrule
Linear trend~(\texttt{TR}) &
Both slope signs remain visible above observation noise. &
\cmark & \cmark & \cmark & \cmark \\

Oscillation~(\texttt{OS}) &
At least two unaliased cycles occur. &
\cmark & \cmark & \cmark & \cmark \\

Mean reversion~(\texttt{MR}) &
Persistence remains separated from white noise and a unit root. &
\cmark & \cmark & \cmark & \cmark \\

Random walk with drift~(\texttt{RW}) &
Drift remains visible without dominating process innovations. &
\cmark & -- & -- & -- \\

Exponential relaxation~(\texttt{RX}) &
Relaxation remains incomplete near the forecast origin. &
\cmark & \cmark & -- & -- \\

Saturating growth~(\texttt{SG}) &
The inflection lies within the context before saturation. &
\cmark & \cmark & \cmark & \cmark \\

Level shift~(\texttt{LS}) &
Observations occur on both sides of the change point. &
\cmark & -- & -- & -- \\

Damped oscillation~(\texttt{DO}) &
Several cycles remain visible under the decaying envelope. &
\cmark & \cmark & \cmark & \cmark \\

Seasonal autoregression~(\texttt{SA}) &
Several repetitions occur for every sampled period. &
\cmark & \cmark & \cmark & \cmark \\

Transient pulse~(\texttt{TP}) &
The pulse reaches the forecast window without becoming impulsive. &
\cmark & \cmark & \cmark & \cmark \\

\midrule
Dual tone~(\texttt{DT}) &
Both tones remain visible, resolvable, and unaliased. &
\cmark & \cmark & \cmark & \cmark \\

Linear state space~(\texttt{SS}) &
The process remains stationary with an identifiable spectral mode. &
\cmark & \cmark & \cmark & \cmark \\

Forced diffusion~(\texttt{FD}) &
Surviving harmonics reveal frequency-dependent attenuation and delay. &
\cmark & \cmark & \cmark & \cmark \\
\bottomrule
\end{tabularx}
\end{table}

\subsection{Representation Accessibility Implementation}
\label{app:representation_impl}

\subsubsection{Probe Training and Model Selection}
\label{app:probe_impl}

All TSFM weights remain frozen.
\autoref{tab:decodability_settings} summarizes the probe targets, penalty grids, and conditional reference correlations.
Law-identification probes use the final exposed sequence position.
Parameter probes concatenate this position with available positions~\(1\), \(2\), \(4\), \(8\), and~\(16\) steps earlier.
This yields three positions for Toto-Open-Base~1.0 and Timer-XL, four for TimesFM~2.5 and TTM-R2, five for TimesFM~3 and Sundial, and six for Chronos-2, Moirai, and Time-MoE.
Features are standardized using training statistics before fitting any layer-specific probe models.

At each layer, the penalty maximizing validation accuracy or correlation is selected.
The best validation layer is then evaluated once on the independent test dataset.
Parameter datasets contain 131{,}070 training, 16{,}380 validation, and 32{,}770 test sequences.
Law identification uses 131{,}040 training, 16{,}380 validation, and 32{,}760 test sequences, with 2{,}520 test sequences per law.
Independent random streams generate all three splits.

Seven models use context length~\(T=256\), Timer-XL uses~\(T=288\), and TTM-R2 uses~\(T=512\).
Linear trend draws baseline~\(b\sim\mathcal{N}(0,10^2)\) and uses observation-noise standard deviations~\(10\), \(11.2\), and~\(20\), respectively.
Oscillation uses amplitude~\(A=5\), uniform phase, and noise standard deviations~\(9.43\), \(10.14\), and~\(14.71\).
At~\(T=256\), the remaining laws draw
\(\log V\sim\mathcal{U}[\log0.5,\log5]\),
\(\log\mathrm{SNR}\sim\mathcal{U}[\log1,\log6]\),
and~\(c\sim\mathcal{N}(0,V^2)\).
For other context lengths, the signal-to-noise ratio is scaled by~\(\sqrt{256/T}\) to approximately equalize finite-history statistical evidence.

The thirteen law classes are balanced, giving chance accuracy~\(1/13\).
For TTM-R2 law identification, extending the penalty grid to~\(10^{-8}\) leaves the selected accuracy unchanged.

\begin{table}[t]
\centering
\caption{Probe targets, penalty grids, and conditional reference correlations.}
\label{tab:decodability_settings}
\small
\setlength{\tabcolsep}{2.5pt}
\renewcommand{\arraystretch}{1.12}
\begin{tabularx}{\linewidth}{@{}
l
>{\raggedright\arraybackslash}X
>{\centering\arraybackslash}p{0.23\linewidth}
rrr
@{}}
\toprule
& & & \multicolumn{3}{c}{Reference~\(r^\star\)} \\
\cmidrule(l){4-6}
Task & Probe target & Penalty grid & \(T=256\) & \(T=288\) & \(T=512\) \\
\midrule
Law ID &
13 balanced classes &
\mbox{\(\lambda\in\{10^{-4},\ldots,10^3\}\)} &
-- & -- & -- \\

\texttt{TR} &
\(S\sim\mathcal{N}(0,0.02^2)\) &
\mbox{\(\alpha\in\{10^0,\ldots,10^6\}\)} &
0.922 & 0.930 & 0.959 \\

\texttt{OS} &
\(\omega=2\pi/P,\ \log P\sim\mathcal{U}[\log2.5,\log4096]\) &
\mbox{\(\alpha\in\{10^0,\ldots,10^6\}\)} &
0.988 & 0.986 & 0.971 \\

\texttt{MR} &
\(\rho\sim\mathcal{U}[0.02,0.98]\) &
\mbox{\(\alpha\in\{10^0,\ldots,10^{10}\}\)} &
0.976 & 0.978 & 0.984 \\

\texttt{RX} &
\(g\sim\mathcal{U}[0.8,2.5]\) &
\mbox{\(\alpha\in\{10^0,\ldots,10^{10}\}\)} &
0.899 & 0.901 & 0.901 \\

\texttt{SG} &
\(R\sim\mathcal{U}[3,13]\) &
\mbox{\(\alpha\in\{10^0,\ldots,10^{10}\}\)} &
0.983 & 0.983 & 0.983 \\

\texttt{DO} &
\(g\sim\mathcal{U}[0.1,3]\) &
\mbox{\(\alpha\in\{10^0,\ldots,10^{10}\}\)} &
0.986 & 0.986 & 0.986 \\

\texttt{SA} &
\(\rho_s\sim\mathcal{U}[0.02,0.98]\) &
\mbox{\(\alpha\in\{10^0,\ldots,10^{10}\}\)} &
0.976 & 0.978 & 0.984 \\

\texttt{TP} &
\(\log(w/T)\sim\mathcal{U}[\log0.025,\log0.2]\) &
\mbox{\(\alpha\in\{10^0,\ldots,10^{10}\}\)} &
0.997 & 0.997 & 0.997 \\

\midrule
\texttt{TL} &
\(C_L\sim\mathcal{U}[2,8]\) &
\mbox{\(\alpha\in\{10^0,\ldots,10^{10}\}\)} &
1.000 & 1.000 & 1.000 \\

\texttt{TH} &
\(C_H\sim\mathcal{U}[10,26]\) &
\mbox{\(\alpha\in\{10^0,\ldots,10^{10}\}\)} &
1.000 & 1.000 & 1.000 \\

\texttt{TW} &
\(\log\kappa\sim\mathcal{U}[\log0.4,\log2.5]\) &
\mbox{\(\alpha\in\{10^0,\ldots,10^{10}\}\)} &
0.991 & 0.991 & 0.991 \\

\texttt{SR} &
\(r\sim\mathcal{U}[0.85,0.98]\) &
\mbox{\(\alpha\in\{10^0,\ldots,10^{10}\}\)} &
0.894 & 0.904 & 0.944 \\

\texttt{SC} &
\(C_m\sim\mathcal{U}[12,32]\) &
\mbox{\(\alpha\in\{10^0,\ldots,10^{10}\}\)} &
0.988 & 0.987 & 0.975 \\

\texttt{WC} &
\(C_d\sim\mathcal{U}[2,8]\) &
\mbox{\(\alpha\in\{10^0,\ldots,10^{10}\}\)} &
1.000 & 1.000 & 1.000 \\

\texttt{WD} &
\(\log\Lambda\sim\mathcal{U}[\log0.002,\log0.1]\) &
\mbox{\(\alpha\in\{10^0,\ldots,10^{10}\}\)} &
0.991 & 0.991 & 0.991 \\
\bottomrule
\end{tabularx}
\end{table}

\subsubsection{Law- and Nuisance-Conditioned Parameter Reference}
\label{app:state_reference}

For law~\(m\), parameter~\(z_{m,j}\), and nuisance variables~\(\nu\), the conditional posterior estimate and reference correlation used throughout the representation-accessibility analysis below are
\[
\widehat z_{m,j}^{\star}
=
\mathbb{E}\!\left[
z_{m,j}\mid x_{1:T},m,\nu
\right],
\qquad
r_{m,j}^{\star}
=
\operatorname{corr}
\left(
\widehat z_{m,j}^{\star},
z_{m,j}
\right).
\]
The posterior mean~\(\widehat z_{m,j}^{\star}\) estimates the parameter from the observed history given the law and documented nuisance variables.
The reference~\(r_{m,j}^{\star}\) is its correlation with the true parameter across sequences and measures finite-history recoverability.
Higher values indicate that the observed history identifies the parameter more reliably.
Each accessibility score~\(D_{m,j}(\ell)\) normalizes the probe correlation by the reference computed at the corresponding model context length.

The conditional parameter reference conditions on the sequence scale, offset, noise level, and documented nuisance variables.
For multi-parameter laws, it also receives the non-designated parameters and phases.
The TSFM receives only the observed history, so~\(r_{m,j}^{\star}\) provides an informed reference for conditional parameter recoverability under the corresponding controlled generator distribution.

The trend posterior mean is computed in closed form.
Other single-parameter references use generator likelihoods evaluated over~257 parameter values.
The grids increase to~513 values for low-tone frequency~(\texttt{TL}), high-tone frequency~(\texttt{TH}), and drive frequency~(\texttt{WC}).
Oscillation uses a grid of~2{,}048 frequencies and~256 phases.
For linear state space, the stationary Gaussian second-order autoregression supplies the likelihood.
The non-designated parameter uses a~33-point lattice, while the designated parameter is integrated over its evaluation grid.

\subsection{Forecast Response Implementation}
\label{app:response_impl}

\subsubsection{Matched Parameter Interventions}
\label{app:response_pairs}

For each law~\(m\) and designated parameter~\(z_{m,j}\), a baseline vector~\(z_m^A\) is paired with an intervention vector~\(z_m^B\).
The two vectors differ only in the designated coordinate, while all nuisance quantities, initial conditions, and generator randomness remain fixed.
The resulting histories~\(x^A\) and~\(x^B\) therefore differ only through the designated parameter.

The frozen TSFM produces the forecast change
\[
\Delta\widehat y_{m,j}
=
F_\theta(x^B)-F_\theta(x^A),
\]
while the corresponding future trajectories define the reference response
\[
\Delta y_{m,j}^{\mathrm{ref}}
=
y^B-y^A.
\]
Comparing these quantities measures whether the forecast expresses the expected consequence of the parameter change in direction, magnitude, and temporal structure.

Mean reversion, seasonal autoregression, and linear state space contain future process innovations that no forecaster can anticipate.
For these laws,~\(\Delta y_{m,j}^{\mathrm{ref}}\) is computed from conditional-mean futures by setting post-origin innovations to zero in both arms.
The two histories are processed independently, while stochastic forecasting backends use the same sampling seed across arms to reduce Monte Carlo variation without changing the underlying model predictions.

\subsubsection{Intervention Magnitudes}
\label{app:response_deltas}

\autoref{tab:response_settings} summarizes the parameter ranges, signed intervention magnitudes, and conditional reference responses.
Baseline values are sampled so that both arms remain within the specified range.
Linear trend uses changes of~\(\pm0.02\) and~\(\pm0.05\), while oscillation uses~\(\pm1\), \(\pm2\), and~\(\pm4\) cycles per context.
Other parameters use~\(\pm2\%\) and~\(\pm5\%\) of their sampled ranges.
Transient pulse~(\texttt{TP}), amplitude ratio~(\texttt{TW}), and diffusion time~(\texttt{WD}) use the corresponding logarithmic ranges.
For oscillation, every intervention magnitude uses the same baseline-frequency distribution.

The reported response score averages the model and conditional-reference coefficients separately across the magnitude set~\(\mathcal{M}_{m,j}\) before computing their final normalized ratio:
\[
\bar U_{m,j}
=
\frac{
\sum_{\delta\in\mathcal{M}_{m,j}}
U_{m,j}(\delta)
}{
\sum_{\delta\in\mathcal{M}_{m,j}}
U_{m,j}^{\star}(\delta)
}.
\]
Each magnitude generally uses~32{,}768 matched pairs.
Moirai uses~131{,}072 pairs for each single-parameter response law, while Timer-XL uses~131{,}072 pairs for oscillation.
These larger datasets reduce sensitivity to heavy-tailed per-pair coefficients.

\begin{table}[t]
\centering
\caption{Intervention settings and mean conditional response references.}
\label{tab:response_settings}
\scriptsize
\setlength{\tabcolsep}{3pt}
\renewcommand{\arraystretch}{1.12}
\begin{tabularx}{\linewidth}{@{}
>{\raggedright\arraybackslash}p{0.21\linewidth}
>{\raggedright\arraybackslash}p{0.16\linewidth}
>{\raggedright\arraybackslash}X
>{\centering\arraybackslash}p{0.14\linewidth}
rrr
@{}}
\toprule
& & & & \multicolumn{3}{c}{Mean reference~\(\bar U_{m,j}^{\star}\)} \\
\cmidrule(l){5-7}
Law & Parameter & Range & Signed change & \(T=256\) & \(T=288\) & \(T=512\) \\
\midrule
Linear trend~(\texttt{TR}) &
Slope &
\(S\in[-0.45,0.45]\) &
\(\pm0.02,\pm0.05\) &
0.995 & 0.995 & 0.997 \\

Oscillation~(\texttt{OS}) &
Frequency &
\(C\in[0.0625,102.4]\) &
\(\pm1,\pm2,\pm4\) &
0.870 & 0.869 & 0.833 \\

Mean reversion~(\texttt{MR}) &
Persistence &
\(\rho\in[0.02,0.98]\) &
\(\pm2\%,\pm5\%\) &
0.760 & 0.766 & 0.755 \\

Saturating growth~(\texttt{SG}) &
Growth rate &
\(R\in[3,13]\) &
\(\pm2\%,\pm5\%\) &
0.945 & 0.950 & 0.952 \\

Damped oscillation~(\texttt{DO}) &
Decay rate &
\(g\in[0.1,3]\) &
\(\pm2\%,\pm5\%\) &
0.954 & 0.954 & 0.954 \\

Seasonal autoregression~(\texttt{SA}) &
Persistence &
\(\rho_s\in[0.02,0.98]\) &
\(\pm2\%,\pm5\%\) &
0.845 & 0.841 & 0.807 \\

Transient pulse~(\texttt{TP}) &
Pulse width &
\(w/T\in[0.025,0.2]\) &
\(\pm2\%,\pm5\%\) &
0.982 & 0.981 & 0.979 \\

\midrule
Dual tone~(\texttt{DT}) &
Low-tone frequency~(\texttt{TL}) &
\(C_L\in[2,8]\) &
\(\pm2\%,\pm5\%\) &
1.000 & 0.999 & 0.999 \\

&
High-tone frequency~(\texttt{TH}) &
\(C_H\in[10,26]\) &
\(\pm2\%,\pm5\%\) &
1.000 & 1.000 & 1.000 \\

&
Amplitude ratio~(\texttt{TW}) &
\(\kappa\in[0.4,2.5]\) &
\(\pm2\%,\pm5\%\) &
0.972 & 0.972 & 0.972 \\

\addlinespace
Linear state space~(\texttt{SS}) &
Modulus~(\texttt{SR}) &
\(r\in[0.85,0.98]\) &
\(\pm2\%,\pm5\%\) &
0.819 & 0.820 & 0.824 \\

&
Modal frequency~(\texttt{SC}) &
\(C_m\in[12,32]\) &
\(\pm2\%,\pm5\%\) &
0.866 & 0.858 & 0.831 \\

\addlinespace
Forced diffusion~(\texttt{FD}) &
Drive frequency~(\texttt{WC}) &
\(C_d\in[2,8]\) &
\(\pm2\%,\pm5\%\) &
1.000 & 1.000 & 1.000 \\

&
Diffusion time~(\texttt{WD}) &
\(\Lambda\in[0.002,0.1]\) &
\(\pm2\%,\pm5\%\) &
0.957 & 0.957 & 0.956 \\
\bottomrule
\end{tabularx}
\end{table}

\subsubsection{Conditional Posterior-Predictive Response Reference}
\label{app:u_ceiling}

A finite noisy history may not fully determine the future response to a parameter change.
We therefore define the conditional posterior-predictive forecast
\[
F_m^\star(x,\nu)
=
\mathbb{E}\!\left[y\mid x,m,\nu\right],
\]
where~\(m\) is the known law and~\(\nu\) contains the documented nuisance variables.
For the two matched histories, the conditional forecast change is
\[
\Delta y_{m,j}^{\star}
=
F_m^\star(x^B,\nu)
-
F_m^\star(x^A,\nu).
\]
This change is scored against the same generator response and over the same matched pairs as the TSFM forecast change, yielding~\(U_{m,j}^{\star}(\delta)\).
The normalized response across intervention magnitudes is
\[
\bar U_{m,j}
=
\frac{
\sum_{\delta\in\mathcal{M}_{m,j}}
U_{m,j}(\delta)
}{
\sum_{\delta\in\mathcal{M}_{m,j}}
U_{m,j}^{\star}(\delta)
}.
\]
The conditional response reference therefore corresponds to~\(\bar U_{m,j}=1\) by construction.

The conditional response reference conditions on each sequence's scale, offset, noise level, and documented nuisance variables.
For multi-parameter laws, it also receives the non-designated parameters and phases.
For mean reversion~(\texttt{MR}), seasonal autoregression~(\texttt{SA}), and the linear state-space parameters eigenvalue modulus~(\texttt{SR}) and modal frequency~(\texttt{SC}), it receives the standard deviation used to normalize the latent shape.
For saturating growth~(\texttt{SG}), transient pulse~(\texttt{TP}), and damped oscillation~(\texttt{DO}), it receives the inflection time, pulse center, and oscillation frequency and phase, respectively.

Each candidate shape is normalized using the standard deviation implied by its candidate parameter~\(q\) before comparison with either observed history.
Because the intervened history retains the baseline history's normalization scale, a candidate for the intervention arm uses the corresponding baseline parameter~\(I^{-1}(q,\delta)\), where~\(I\) is the intervention map.
This inverse is~\(q-\delta\) for additive changes and~\(q\exp(-\delta_{\log})\) for logarithmic changes.
The resulting coefficient is a nuisance-informed normalization rather than a strict upper bound on signed aligned gain.
A model can exceed this coefficient while producing greater full-vector error.
For damped oscillation, additionally inferring frequency and phase changes~\(U_{m,j}^{\star}\) by less than~\(0.005\) at every magnitude.

\textbf{Trend-Slope Reference.}
For the centered trend system
\[
x_t=M+(t-c)S+\varepsilon_t,
\qquad
c=\frac{T+1}{2},
\]
\(M\) is the level at the context midpoint, \(S\) is the slope, and~\(c\) centers the time indices.
Define the observation vector~\(\mathbf{x}=(x_1,\ldots,x_T)^\top\) and centered time vector~\(\mathbf{g}\), with~\(g_t=t-c\).
Because~\(\sum_t g_t=0\), the slope can be estimated independently of the unknown level~\(M\).
Its least-squares sufficient statistic under this centered finite-history regression model is
\[
\widehat S
=
\frac{\mathbf{g}^\top\mathbf{x}}
{\mathbf{g}^\top\mathbf{g}},
\qquad
\mathbf{g}^\top\mathbf{g}
=
\sum_{t=1}^{T}(t-c)^2.
\]

For independent observation noise~\(\varepsilon_t\sim\mathcal{N}(0,\sigma_\varepsilon^2)\), the estimator satisfies
\[
\widehat S\mid S
\sim
\mathcal{N}
\left(
S,\sigma_{\widehat S}^2
\right),
\qquad
\sigma_{\widehat S}
=
\frac{\sigma_\varepsilon}
{\sqrt{\mathbf{g}^\top\mathbf{g}}}.
\]
Thus,~\(\sigma_{\widehat S}\) quantifies the uncertainty in the slope estimate induced by finite context length and observation noise for the assumed independent Gaussian observations.

Under the uniform prior~\(S\in[-s_{\max},s_{\max}]\), the Gaussian likelihood is truncated at the prior boundaries.
Define the standardized lower and upper boundaries
\[
a_-
=
\frac{-s_{\max}-\widehat S}
{\sigma_{\widehat S}},
\qquad
a_+
=
\frac{s_{\max}-\widehat S}
{\sigma_{\widehat S}}.
\]
The posterior mean is
\[
\mathbb{E}[S\mid\mathbf{x}]
=
\widehat S
+
\sigma_{\widehat S}
\frac{
\phi(a_-)-\phi(a_+)
}{
\Phi(a_+)-\Phi(a_-)
},
\]
where~\(\phi\) and~\(\Phi\) denote the standard normal density and cumulative distribution functions.
The correction term pulls estimates near either prior boundary toward the interior of the allowed slope range while leaving interior estimates nearly unchanged.

A matched slope intervention changes~\(S\) by~\(\delta\) while preserving the history mean.
Because the centered statistic changes by the same amount,~\(\widehat S^B=\widehat S^A+\delta\).
The conditional-reference coefficient for the corresponding matched slope intervention is therefore
\[
U_{\mathrm{TR}}^\star(\delta)
=
\frac{
\mathbb{E}[S\mid\mathbf{x}^B]
-
\mathbb{E}[S\mid\mathbf{x}^A]
}{
\delta
}.
\]
A coefficient near one indicates that the posterior mean preserves the imposed slope change.
At~\(T=256\) and~\(s_{\max}=0.45\), the average coefficient is~\(1.000\) when~\(\lvert S^A\rvert<0.20\) and ranges from~\(0.986\) to~\(0.994\) at larger baseline magnitudes because of truncation near the prior boundaries.

\textbf{References Across Context Lengths.}
At a fixed context length~\(T\), the matched-pair distribution and conditional response reference do not depend on the TSFM.
The seven models using~\(T=256\) therefore share the same conditional reference at each intervention magnitude.
Timer-XL and TTM-R2 use separately computed references at~\(T=288\) and~\(T=512\), respectively.
\autoref{tab:response_settings} reports the mean reference for each context length under the corresponding matched intervention distributions.

\subsection{Causal Geometry Implementation}
\label{app:geometry_impl}

\subsubsection{Fitting, Validation, and Test Datasets}
\label{app:geometry_splits}

The causal-geometry analysis uses separate datasets for basis estimation, model selection, and final evaluation.
The fitting dataset~\(\mathcal{D}_{\mathrm{disc}}\) estimates the shared causal basis and fits the coordinate readout.
The validation dataset selects admissible layers, subspace ranks, and readout penalties.
The selected configuration is evaluated once on an independent test dataset.
No test response contributes to basis estimation, readout fitting, or configuration selection.

Each of the~126 model--parameter cells uses~16{,}384 fitting, 4{,}096 validation, and~4{,}096 test matched pairs.
The datasets are generated at the model's context length with horizon~\(H=16\) and the intervention magnitude reported in~\autoref{tab:geometry_settings}.
Independent random streams are used across the three splits.
Generator streams are shared across models for a given parameter, except for Moirai on seasonal autoregression, whose streams are offset by one while preserving the generator and sample sizes.
Among admissible layers, selection maximizes the validation projected response over the evaluated ranks.
All reported responses are then computed on the test dataset.

\begin{table}[t]
\centering
\caption{
Hidden-intervention calibration across parameter targets for every evaluated model and layer.
The final two columns report layer counts out of~90.
}
\label{tab:geometry_settings}
\footnotesize
\setlength{\tabcolsep}{3.5pt}
\renewcommand{\arraystretch}{1.12}
\begin{tabularx}{\linewidth}{@{}
>{\raggedright\arraybackslash}p{0.26\linewidth}
>{\raggedright\arraybackslash}X
c
cc
cc
@{}}
\toprule
&
&
&
\multicolumn{2}{c}{Calibrated step~\(\alpha_\ell\)}
&
\multicolumn{2}{c}{Layer count}
\\
\cmidrule(lr){4-5}
\cmidrule(l){6-7}
Law
& Parameter
& \(\delta\)
& Median
& Range
& \(\alpha_\ell<1\)
& Converged \\
\midrule
Linear trend~(\texttt{TR})
& Slope
& 0.02 & 0.58 & 0.021--1.00 & 86/90 & 85/90 \\

Oscillation~(\texttt{OS})
& Frequency
& 2 & 0.19 & 0.037--0.76 & 90/90 & 90/90 \\

Mean reversion~(\texttt{MR})
& Persistence
& 5\% & 1.00 & 0.19--1.00 & 26/90 & 90/90 \\

Saturating growth~(\texttt{SG})
& Growth rate
& 5\% & 0.58 & 0.084--1.00 & 57/90 & 90/90 \\

Damped oscillation~(\texttt{DO})
& Decay rate
& 5\% & 1.00 & 0.33--1.00 & 42/90 & 90/90 \\

Seasonal autoregression~(\texttt{SA})
& Persistence
& 5\% & 0.11 & 0.012--1.00 & 86/90 & 83/90 \\

Transient pulse~(\texttt{TP})
& Pulse width
& 5\% & 0.25 & 0.11--0.76 & 90/90 & 90/90 \\

\midrule
Dual tone~(\texttt{DT})
& Low-tone frequency~(\texttt{TL})
& 5\% & 1.00 & 0.76--1.00 & 19/90 & 90/90 \\

&
High-tone frequency~(\texttt{TH})
& 5\% & 1.00 & 0.22--1.00 & 34/90 & 90/90 \\

&
Amplitude ratio~(\texttt{TW})
& 5\% & 1.00 & 0.33--1.00 & 21/90 & 90/90 \\

\addlinespace
Linear state space~(\texttt{SS})
& Modulus~(\texttt{SR})
& 5\% & 1.00 & 0.19--1.00 & 32/90 & 90/90 \\

&
Modal frequency~(\texttt{SC})
& 5\% & 0.19 & 0.064--0.44 & 90/90 & 90/90 \\

\addlinespace
Forced diffusion~(\texttt{FD})
& Drive frequency~(\texttt{WC})
& 5\% & 1.00 & 0.15--1.00 & 28/90 & 90/90 \\

&
Diffusion time~(\texttt{WD})
& 5\% & 0.76 & 0.25--1.00 & 49/90 & 90/90 \\
\bottomrule
\end{tabularx}
\end{table}

\subsubsection{Local Causal Directions and Step Calibration}
\label{app:geometry_local_impl}

\autoref{tab:geometry_settings} summarizes the parameter changes, calibrated hidden-state steps, and layer-level calibration outcomes for every evaluated model and target.

\textbf{Regularized Local Directions.}
The regularized Jacobian inversion in~\autoref{eq:local_direction} uses
\[
\lambda
=
10^{-4}
\frac{
\operatorname{tr}\!\left(J_i^\ell J_i^{\ell\top}\right)
}{
H
}.
\]
This scaling adapts the regularization strength to the average Jacobian row energy at each input and layer.
The direction is computed through the output-space system
\[
v_i^\ell
=
J_i^{\ell\top}
\left(
J_i^\ell J_i^{\ell\top}
+
\lambda I_H
\right)^{-1}
\Delta y_i^{\mathrm{ref}}.
\]
The inversion targets the complete uncentered reference response.
Centering is applied only when the realized nonlinear response is scored using~\autoref{eq:pooled_response}.

\textbf{Intervention-Step Calibration.}
At each layer, the intervention step~\(\alpha_\ell\) is selected on validation data through geometric search followed by log-midpoint bisection.
The search begins from the step accepted at the previous layer, or from~\(\alpha_\ell=1\) at the first measured layer, and evaluates at most nine steps.
Seasonal-autoregression runs for Chronos-2, Moirai, Timer-XL, Toto-Open-Base~1.0, and TTM-R2 instead restart from~\(\alpha_\ell=1\) at every layer.
The selected step minimizes the larger deviation of the pooled and median per-pair local responses from one, with preference given to steps placing both responses within~\(0.03\) of one.

\textbf{Direction Capping and Layer Selection.}
Directions whose root-mean-square norm exceeds three times the fitting-dataset median at that layer are rescaled to this threshold.
The same factor is applied to the corresponding reference response, preserving the direction--target relationship.
A layer is admissible when its pooled and median per-pair local responses lie within~\([0.85,1.05]\) and its projected response does not exceed
\[
\max\!\left\{1.05,U_{\mathrm{local}}\right\}
\]
on validation data.
The median is computed over pairs whose capped reference-response norm is at least~\(5\%\) of the median norm for that model, target, and layer.

\textbf{Evaluated Layers.}
Blocks~1 through~\(L\) are evaluated for each model, except TimesFM~2.5, for which layers~1, 5, 9, 12, 16, and~20 are evaluated.
Time-MoE's final block is excluded because its exposed hidden state follows the final normalization and cannot be modified through the intervention interface.
These choices govern all causal-geometry measurements.

\textbf{Connecting Hidden Edits to Input-Induced Changes.}
For
\[
\Delta h_i^\ell
=
h^\ell(x_i^B)-h^\ell(x_i^A),
\]
the local approximation
\[
\Delta F_i
\approx
J_i^\ell\Delta h_i^\ell
\]
holds with downstream bypass quantities fixed.
The actual input intervention can also change the bypass quantities~\(b_i\) and move the model beyond the local linear regime.
The local causal direction~\(v_i^\ell\) is constructed to produce the reference response, whereas~\(\Delta h_i^\ell\) is the state change induced by the parameter intervention.
Their alignment, row- and null-space energy, and induced gains indicate whether the input change follows a forecast-sensitive hidden-state direction.

\subsubsection{Shared Causal Subspace and Rank Selection}
\label{app:geometry_space_impl}

The shared basis and projected directions follow~\autoref{eq:causal_space}.
For each layer, the numerical rank~\(r_\ell\) counts squared singular values greater than~\(10^{-10}\) times the largest.
The evaluated ranks are
\[
\mathcal{K}_\ell
=
\left\{2^0,2^1,\ldots,2^{10}\right\}
\cap
\left[1,r_\ell\right].
\]
Thus, the largest evaluated rank is~\(1{,}024\), or the largest admissible power of two when~\(r_\ell<1{,}024\).
The compactness statistic~\(k_{90}\left(\ell\right)\) is the smallest evaluated rank that preserves at least~\(90\%\) of~\(U_{\mathrm{local}}\left(\ell\right)\) on validation data.
If no rank reaches the threshold, the largest evaluated rank is reported as a lower bound for that layer and parameter target.

\textbf{State-Dimension Normalization.}
Panel-wide comparisons normalize~\(k_{90}\) by the intervened-state dimension~\(d_h\), defined as the hidden width multiplied by the number of modified token positions.
The intervention modifies~3 positions for Timer-XL~(\texttt{TX}), 4 for Toto-Open-Base~1.0~(\texttt{TO}), 8 for TimesFM~2.5~(\texttt{F2}) and TTM-R2~(\texttt{TT}), 10 for TimesFM~3~(\texttt{F3}), 16 for Sundial~(\texttt{SU}), 17 for Moirai~(\texttt{MO}), 18 for Chronos-2~(\texttt{C2}), and 256 for Time-MoE~(\texttt{ME}).
The resulting dimension~\(d_h\) ranges from~\(1{,}536\) for TTM-R2~(\texttt{TT}) to~\(196{,}608\) for Time-MoE~(\texttt{ME}).
Consequently, a large absolute~\(k_{90}\) can still occupy a small fraction of a model's intervened hidden state.

\subsubsection{Causal-Coordinate Readout and Oracle Fixed-Direction Control}
\label{app:geometry_access_impl}

\textbf{Causal-Coordinate Readout.}
For layer~\(\ell\) and rank~\(k\), the coordinate readout in~\autoref{eq:deployed_direction} is fitted by ridge regression on standardized fitting-dataset representations:
\[
\left(
\widetilde P_{\ell,k},
\widetilde b_{\ell,k}^{c}
\right)
=
\arg\min_{P,b}
\left\{
\sum_{i\in\mathcal{D}_{\mathrm{disc}}}
\left\|
c_i^{\ell,k}
-
P D_\ell^{-1}
\left(
h_i^\ell-\mu_\ell
\right)
-
b
\right\|_2^2
+
\lambda_c\|P\|_F^2
\right\}.
\]
Here,~\(c_i^{\ell,k}\) contains the coordinates of the local causal direction in the rank-\(k\) shared basis.
The vector~\(\mu_\ell\) and diagonal matrix~\(D_\ell\) contain the coordinate-wise mean and standard deviation of the fitting representations.
The intercept is not penalized.

Transforming the fitted coefficients back to the original representation scale gives
\[
P_{\ell,k}
=
\widetilde P_{\ell,k}D_\ell^{-1},
\qquad
b_{\ell,k}^{c}
=
\widetilde b_{\ell,k}^{c}
-
P_{\ell,k}\mu_\ell.
\]
Regularization therefore applies to the standardized-feature coefficients~\(\widetilde P_{\ell,k}\).

\textbf{Penalty and Rank Selection.}
The penalty~\(\lambda_c\) is selected independently at each rank from~33 values logarithmically spaced between~\(10^{-4}\) and~\(10^{12}\).
Selection maximizes validation~\(R^2\) pooled across the predicted coordinates.
For mean reversion~(\texttt{MR}), Timer-XL~(\texttt{TX}) and TTM-R2~(\texttt{TT}) use~25 values between~\(10^{-4}\) and~\(10^8\).
For seasonal autoregression~(\texttt{SA}), Chronos-2~(\texttt{C2}), Timer-XL~(\texttt{TX}), Toto-Open-Base~1.0~(\texttt{TO}), and TTM-R2~(\texttt{TT}) use the same reduced grid.
Moirai~(\texttt{MO}) on seasonal autoregression~(\texttt{SA}) uses~11 values between~\(10^1\) and~\(10^6\).

The readout rank~\(\widehat k(\ell)\) is the smallest evaluated rank whose validation deployed response retains at least~\(90\%\) of the local response.
If no evaluated rank reaches this threshold, the largest rank is selected.
The test deployed response and projected-response validity condition are then evaluated at the selected rank for each layer and parameter target.

At validation and test time, the deployed direction requires only the input's hidden representation and quantities estimated from the fitting dataset.
It receives neither the input-specific Jacobian nor the reference response.
The shared basis, standardization statistics, and readout coefficients use only fitting data, while the penalty and rank use only validation data.

\textbf{Oracle Fixed-Direction Control.}
The fixed-direction control is deliberately oracle-assisted.
For an evaluation dataset~\(\mathcal{D}_{\mathrm{eval}}\), its common direction is the mean local causal direction
\[
\overline v^\ell
=
\frac{1}{
\left|
\mathcal{D}_{\mathrm{eval}}
\right|
}
\sum_{i\in\mathcal{D}_{\mathrm{eval}}}
v_i^\ell.
\]
For each example~\(i\), this direction is rescaled to the norm of its oracle local direction:
\[
v_i^{\ell,\mathrm{fixed}}
=
\left\|
v_i^\ell
\right\|_2
\frac{
\overline v^\ell
}{
\left\|
\overline v^\ell
\right\|_2
}.
\]
The control therefore receives both the evaluation-dataset mean oracle direction and each example's oracle perturbation norm.
The comparison asks whether an input-dependent readout improves upon a strong input-independent direction despite this oracle assistance.
Because the deployed readout predicts direction and magnitude jointly, the comparison does not isolate directional inference from norm prediction.
Both components determine its reported performance.

\section{Detailed Dynamical-Parameter Results}
\label{app:results}

This appendix presents layer-wise and model-level results for law identification, representation accessibility, forecast response, and causal geometry.
Single-parameter results are followed by multi-parameter and cross-measure analyses.

\textbf{Common Protocol.}
Matched-input analyses use paired datasets spanning signed intervention magnitudes.
Causal geometry uses independent fitting, validation, and test datasets.
Each forecast horizon is evaluated separately because some forecasting backends are not prefix-consistent.

\subsection{Dynamical-Law Identification}
\label{app:law_results}
\label{app:revision_lawid}

Complementing the validation-selected test results in~\S\ref{sec:eval_law}, we examine layer-wise validation profiles and test-set confusion patterns.
The multinomial probe uses the final exposed sequence position, and the balanced test dataset contains~2{,}520 sequences per law.
Each multi-parameter law is treated as one class, and law decodability scores report normalized above-chance accuracy.

\subsubsection{Layer-Wise Dynamical-Law Decodability}
\label{app:law_layers}

\begin{figure}[t]
\centering
\includegraphics[width=0.325\linewidth]{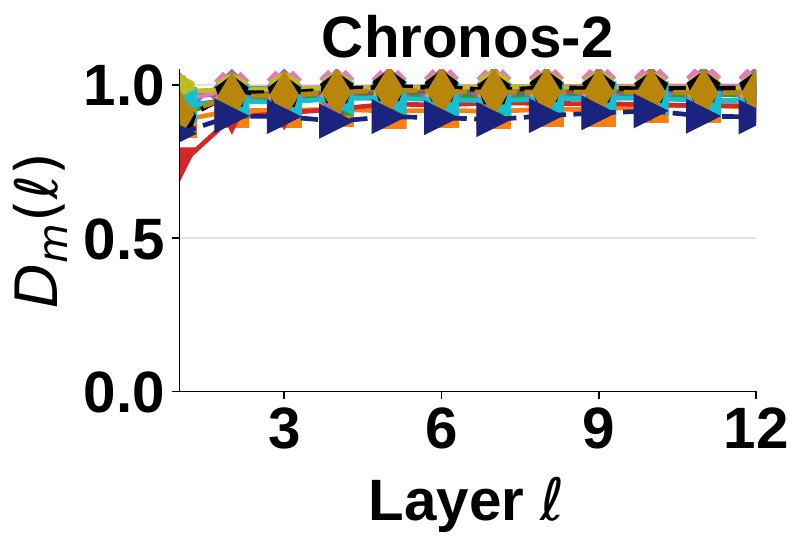}
\hfill
\includegraphics[width=0.325\linewidth]{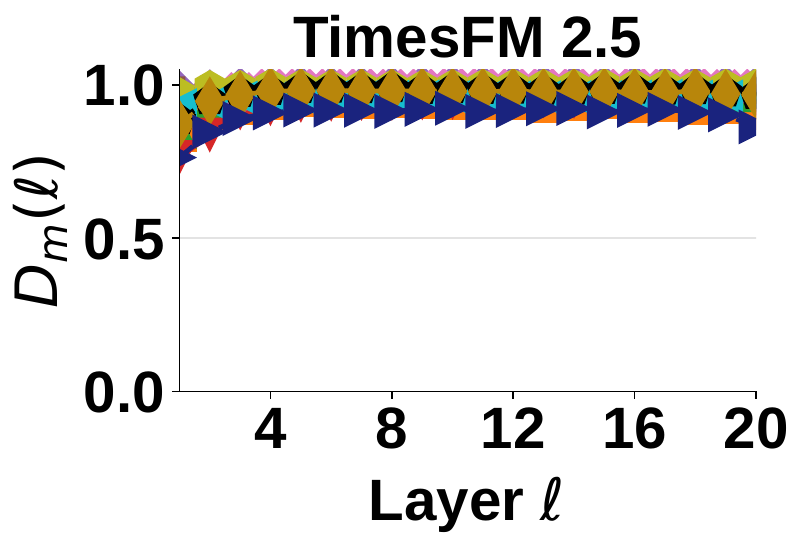}
\hfill
\includegraphics[width=0.325\linewidth]{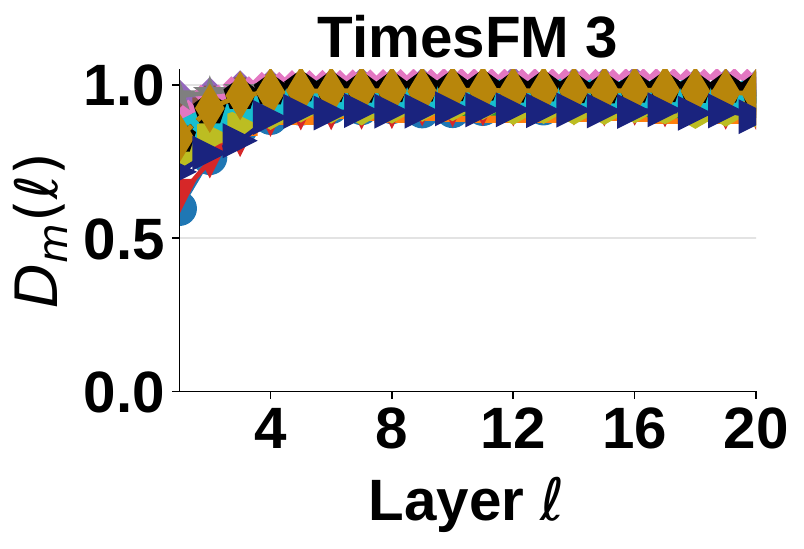}
\\[2pt]
\includegraphics[width=0.325\linewidth]{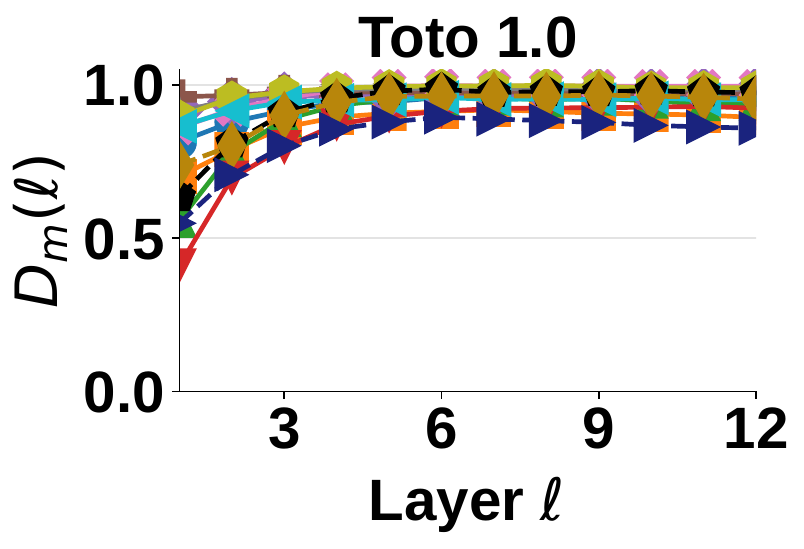}
\hfill
\includegraphics[width=0.325\linewidth]{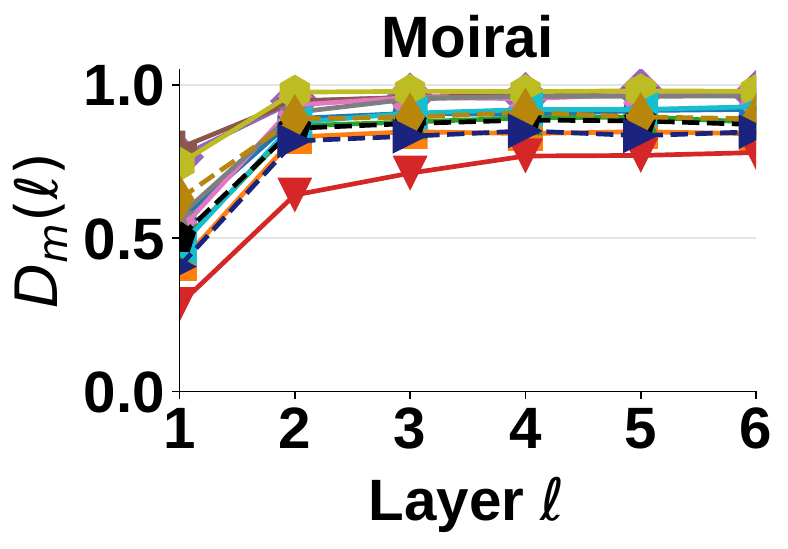}
\hfill
\includegraphics[width=0.325\linewidth]{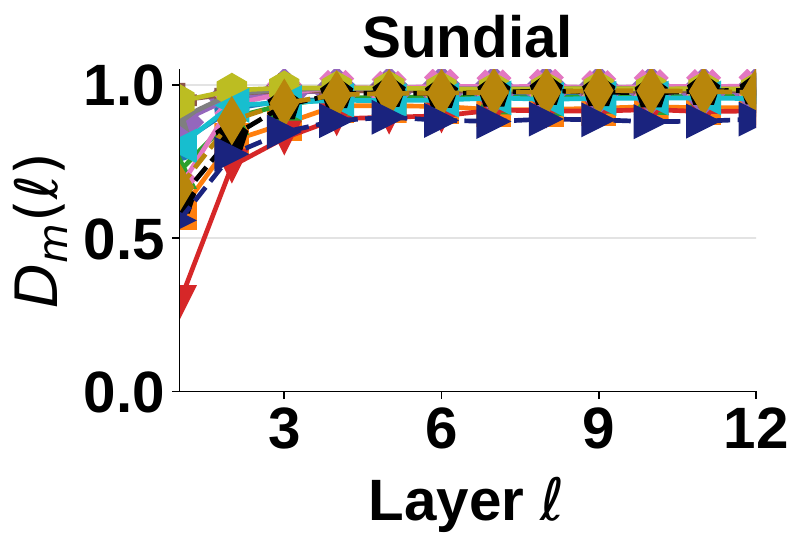}
\\[2pt]
\includegraphics[width=0.325\linewidth]{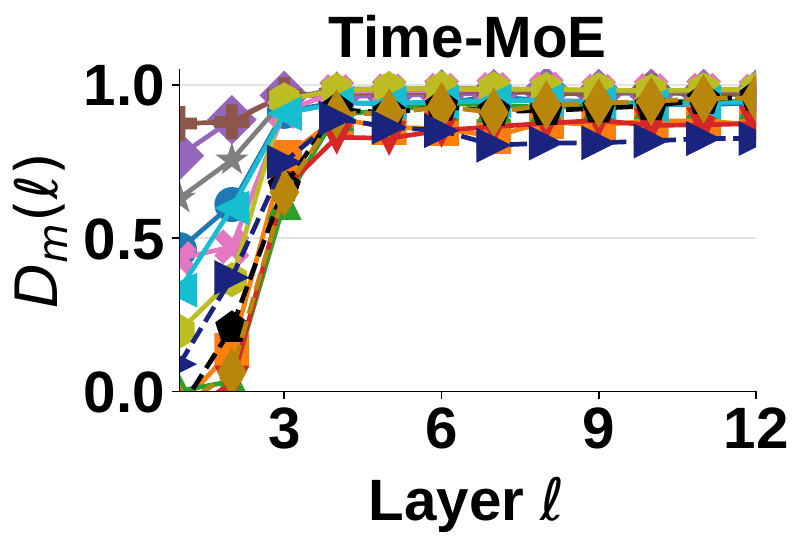}
\hfill
\includegraphics[width=0.325\linewidth]{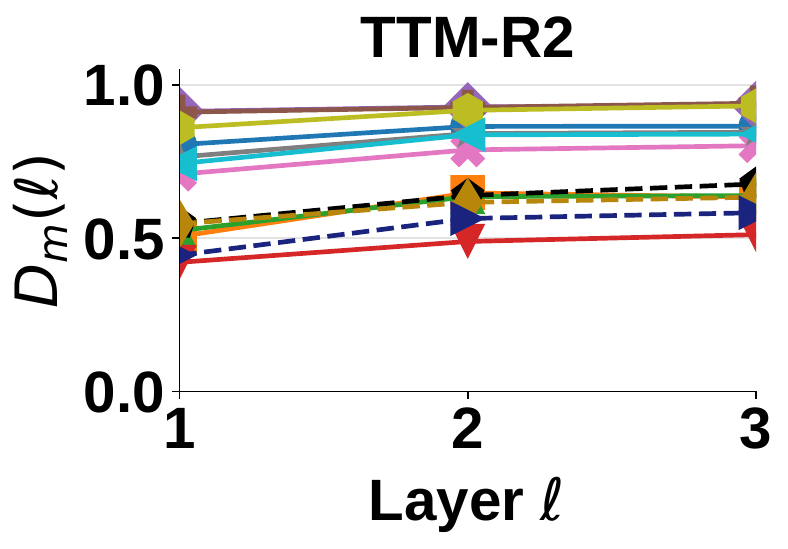}
\hfill
\includegraphics[width=0.325\linewidth]{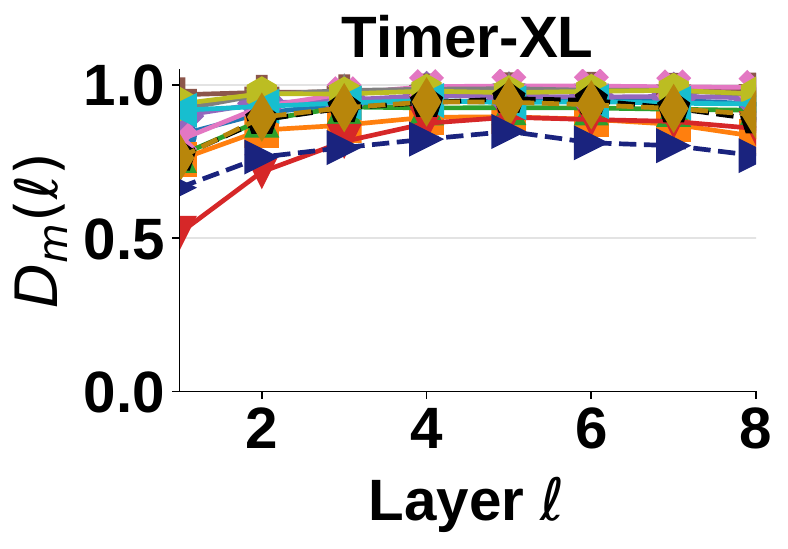}
\\[2pt]
\includegraphics[width=\linewidth]{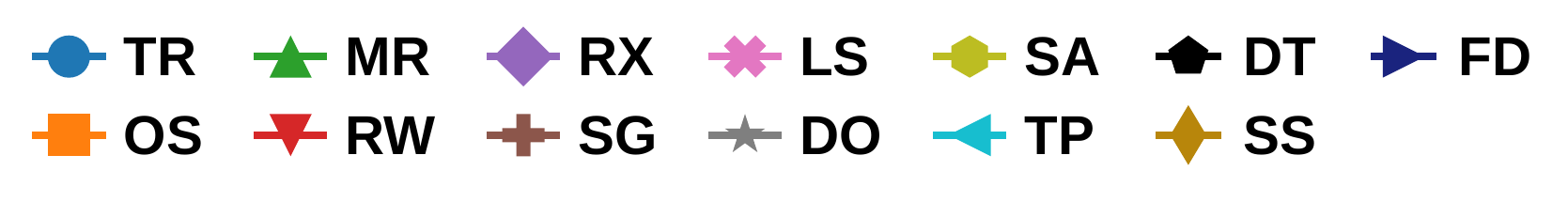}
\caption{
Validation law decodability across model depth.
Panels show all nine frozen TSFMs.
Dashed curves denote multi-parameter laws.
}
\label{fig:lawid_layers}
\end{figure}

For each~TSFM, we evaluate~\(D_m(\ell)\) across all layers and thirteen dynamical laws.
The layer-wise curves use validation data because layer selection is also performed on the validation dataset.
All panels share the same vertical range, allowing direct comparison across models.
\autoref{tab:layer_depth} summarizes emergence depth, while~\autoref{fig:lawid_layers} presents the complete layer-wise profiles.

For each model and law, emergence depth is the first layer at which~\(D_m(\ell)\) reaches~\(95\%\) of its final-layer value, expressed as a fraction of model depth.
\autoref{tab:layer_depth} also reports mean first-layer decodability to characterize how early law information appears.

\begin{table}[t]
\centering
\caption{
Normalized law-emergence depth and first-layer decodability across models.
}
\label{tab:layer_depth}
\scriptsize
\setlength{\tabcolsep}{2.6pt}
\resizebox{\linewidth}{!}{%
\begin{tabular}{@{}l*{15}{r}@{}}
\toprule
Model & TR & OS & MR & RW & RX & SG & LS & DO & SA & TP & DT & SS & FD & Mean & \(D_m(1)\) \\
\midrule
Chronos-2~(\texttt{C2})
& 0.08 & 0.08 & 0.17 & 0.17 & 0.08 & 0.08 & 0.08 & 0.08 & 0.08 & 0.08 & 0.17 & 0.17 & 0.17 & 0.12 & 0.91 \\

TimesFM~2.5~(\texttt{F2})
& 0.10 & 0.10 & 0.10 & 0.15 & 0.05 & 0.05 & 0.10 & 0.05 & 0.05 & 0.05 & 0.10 & 0.10 & 0.10 & 0.08 & 0.90 \\

TimesFM~3~(\texttt{F3})
& 0.20 & 0.15 & 0.15 & 0.20 & 0.05 & 0.10 & 0.15 & 0.05 & 0.15 & 0.15 & 0.15 & 0.15 & 0.20 & 0.14 & 0.81 \\

Toto-Open-Base~1.0~(\texttt{TO})
& 0.25 & 0.25 & 0.33 & 0.42 & 0.17 & 0.08 & 0.25 & 0.17 & 0.17 & 0.17 & 0.33 & 0.33 & 0.33 & 0.25 & 0.75 \\

Moirai~(\texttt{MO})
& 0.33 & 0.33 & 0.33 & 0.67 & 0.33 & 0.33 & 0.33 & 0.50 & 0.33 & 0.50 & 0.33 & 0.33 & 0.33 & 0.38 & 0.55 \\

Sundial~(\texttt{SU})
& 0.25 & 0.33 & 0.25 & 0.33 & 0.17 & 0.08 & 0.25 & 0.17 & 0.08 & 0.17 & 0.25 & 0.25 & 0.25 & 0.22 & 0.72 \\

Time-MoE~(\texttt{ME})
& 0.25 & 0.33 & 0.33 & 0.50 & 0.25 & 0.25 & 0.33 & 0.25 & 0.25 & 0.25 & 0.33 & 0.50 & 0.33 & 0.32 & 0.27 \\

TTM-R2~(\texttt{TT})
& 0.67 & 0.67 & 0.67 & 0.67 & 0.33 & 0.33 & 0.67 & 0.67 & 0.67 & 0.67 & 1.00 & 0.67 & 0.67 & 0.64 & 0.67 \\

Timer-XL~(\texttt{TX})
& 0.25 & 0.25 & 0.25 & 0.38 & 0.25 & 0.12 & 0.38 & 0.25 & 0.12 & 0.12 & 0.25 & 0.25 & 0.25 & 0.24 & 0.81 \\
\midrule
Mean
& 0.26 & 0.28 & 0.29 & 0.39 & 0.19 & 0.16 & 0.28 & 0.24 & 0.21 & 0.24 & 0.32 & 0.31 & 0.29 & 0.27 & \\
\bottomrule
\end{tabular}
}
\end{table}

The layer-wise profiles reveal three properties not captured by the validation-selected summary in~\S\ref{sec:eval_law} because it reports only one layer.

\textbf{Laws Become Decodable at Different Depths.}
Random walk with drift~(\texttt{RW}) emerges latest, at~\(0.39\) of model depth on average, compared with~\(0.16\)--\(0.32\) for the remaining laws.
It is the last or joint-last law to emerge in eight of nine models.
For TTM-R2~(\texttt{TT}), whose three layers make the depth measure coarse, dual tone~(\texttt{DT}) emerges last.
The three multi-parameter laws emerge between~\(0.29\) and~\(0.32\) of model depth on average.
These jointly varying laws therefore become linearly distinguishable relatively late, although law classification does not require recovery of every parameter.
Saturating growth~(\texttt{SG}), exponential relaxation~(\texttt{RX}), and seasonal autoregression~(\texttt{SA}) emerge earliest on average.
The models broadly agree on which laws require greater depth before becoming linearly accessible across their otherwise distinct hidden representations.

\textbf{Depth Requirements Vary More Than Final Decodability.}
Mean first-layer decodability ranges from~\(0.27\) for Time-MoE~(\texttt{ME}) to~\(0.91\) for Chronos-2~(\texttt{C2}).
Chronos-2~(\texttt{C2}) and TimesFM~2.5~(\texttt{F2}) gain only~\(0.05\) and~\(0.06\) across their layers, whereas Time-MoE~(\texttt{ME}) gains~\(0.67\).
Models with similar final decodability can therefore differ substantially in where law information becomes accessible.

\textbf{Selected Layers Typically Lie Beyond Saturation.}
For every model, the validation-selected layer lies at or beyond the point where mean law decodability first reaches~\(95\%\) of its final-layer value.
Mean decodability at the selected layer is within~\(0.026\) of its final-layer value.
The selected layer therefore represents a state supporting strong law identification rather than the precise point at which law identity first becomes accessible.

\subsubsection{Random-Walk Confusions Across Drift Magnitudes}
\label{app:law_rw_confusions}

Random walk with drift~(\texttt{RW}) and forced diffusion~(\texttt{FD}) have the lowest average identification scores in~\autoref{fig:lawid}.
We examine which laws receive the random-walk errors and whether those errors vary with the amount of drift.
This isolates dominant finite-history ambiguity patterns.

\textbf{Random Walk Is Primarily Confused with Other Drifting Laws.}
\autoref{tab:lawid_confusion_rw} reports the classifications assigned to test random-walk sequences.
Errors concentrate on laws whose finite histories can depart persistently from their initial levels.
Linear trend~(\texttt{TR}) receives between~\(2.1\%\) and~\(12.6\%\) of these sequences, while exponential relaxation~(\texttt{RX}) receives between~\(1.0\%\) and~\(6.4\%\).
TTM-R2~(\texttt{TT}) additionally assigns~\(14.0\%\) to level shift~(\texttt{LS}).
Linear trend is the largest incorrect classification for seven models.
TimesFM~3~(\texttt{F3}) instead assigns marginally more sequences to mean reversion~(\texttt{MR}), while TTM-R2~(\texttt{TT}) assigns more to level shift~(\texttt{LS}).
The dominant confusions therefore remain concentrated among laws with similar finite-history trajectories.

\begin{table}[t]
\centering
\caption{
Test-set classifications of random-walk sequences at each model's validation-selected layer.
Correct classifications are bold.
}
\label{tab:lawid_confusion_rw}
\scriptsize
\setlength{\tabcolsep}{3.2pt}
\resizebox{\linewidth}{!}{%
\begin{tabular}{lrrrrrrrrrrrrrc}
\toprule
& \multicolumn{13}{c}{True random walk classified as~(\%)} & \\
\cmidrule(lr){2-14}
Model & TR & OS & MR & RW & RX & SG & LS & DO & SA & TP & DT & SS & FD & Layer \\
\midrule
Chronos-2~(\texttt{C2})
& 2.3 & 0.0 & 1.2 & \textbf{94.0} & 1.0 & 0.3 & 0.3 & 0.1 & 0.0 & 0.6 & 0.0 & 0.0 & 0.0 & 10/12 \\

TimesFM~2.5~(\texttt{F2})
& 2.5 & 0.0 & 1.0 & \textbf{94.4} & 1.1 & 0.2 & 0.2 & 0.0 & 0.0 & 0.5 & 0.0 & 0.0 & 0.0 & 10/20 \\

TimesFM~3~(\texttt{F3})
& 2.1 & 0.0 & 2.5 & \textbf{92.7} & 1.3 & 0.2 & 0.2 & 0.2 & 0.1 & 0.5 & 0.0 & 0.1 & 0.1 & 16/20 \\

Toto-Open-Base~1.0~(\texttt{TO})
& 2.9 & 0.0 & 1.2 & \textbf{92.7} & 1.5 & 0.3 & 0.6 & 0.2 & 0.0 & 0.5 & 0.0 & 0.0 & 0.1 & 6/12 \\

Moirai~(\texttt{MO})
& 10.8 & 0.6 & 1.1 & \textbf{76.2} & 4.2 & 1.5 & 2.5 & 1.1 & 0.0 & 1.4 & 0.2 & 0.0 & 0.4 & 5/6 \\

Sundial~(\texttt{SU})
& 3.4 & 0.0 & 1.0 & \textbf{92.6} & 1.2 & 0.4 & 0.5 & 0.1 & 0.0 & 0.6 & 0.0 & 0.0 & 0.1 & 12/12 \\

Time-MoE~(\texttt{ME})
& 4.9 & 0.0 & 1.5 & \textbf{88.3} & 2.2 & 0.6 & 1.5 & 0.2 & 0.0 & 0.6 & 0.0 & 0.0 & 0.1 & 12/12 \\

TTM-R2~(\texttt{TT})
& 12.6 & 1.0 & 0.8 & \textbf{55.0} & 6.4 & 2.4 & 14.0 & 1.5 & 0.1 & 3.7 & 1.3 & 0.1 & 1.2 & 3/3 \\

Timer-XL~(\texttt{TX})
& 4.0 & 0.0 & 0.9 & \textbf{90.7} & 2.0 & 0.6 & 1.0 & 0.2 & 0.0 & 0.6 & 0.1 & 0.0 & 0.1 & 5/8 \\
\bottomrule
\end{tabular}
}
\end{table}

\textbf{Confusions Increase with Drift Magnitude.}
\autoref{tab:lawid_rw_bins} divides random-walk sequences into three drift-magnitude bins for two models with strong identification and two with weaker identification.
Recall decreases as drift increases for every model.
It falls from~\(95.4\%\) to~\(89.6\%\) for Chronos-2~(\texttt{C2}), from~\(94.7\%\) to~\(86.1\%\) for Toto-Open-Base~1.0~(\texttt{TO}), from~\(83.1\%\) to~\(59.4\%\) for Moirai~(\texttt{MO}), and from~\(60.3\%\) to~\(44.8\%\) for TTM-R2~(\texttt{TT}).
Most of the lost recall shifts to linear trend~(\texttt{TR}) and then exponential relaxation~(\texttt{RX}), consistent with a strongly drifting random walk resembling a noisy line over a finite context.
TTM-R2's confusion with level shift~(\texttt{LS}) remains between~\(12\%\) and~\(15\%\) across drift bins.
This persistent confusion is consistent with its level-shift law decodability score of~\(0.83\) in~\autoref{fig:lawid}, compared with~\(0.97\) or higher for every other model.

\begin{table}[t]
\centering
\caption{
Test-set classifications of random-walk sequences by drift magnitude for two stronger and two weaker models.
All reported class entries are percentages.
}
\label{tab:lawid_rw_bins}
\small
\setlength{\tabcolsep}{4.5pt}
\begin{tabular}{llrrrrrr}
\toprule
Model & \(\lvert D\rvert\) & \(n\) & Random walk & Trend & Relaxation & Shift & Other \\
\midrule
Chronos-2~(\texttt{C2})
& \([0.5,1)\) & 950 & 95.4 & 1.2 & 0.6 & 0.4 & 2.4 \\
& \([1,2)\) & 981 & 95.4 & 1.3 & 0.7 & 0.3 & 2.2 \\
& \([2,3]\) & 589 & 89.6 & 5.9 & 2.2 & 0.2 & 2.0 \\

\addlinespace
Toto-Open-Base~1.0~(\texttt{TO})
& \([0.5,1)\) & 950 & 94.7 & 0.9 & 0.6 & 0.8 & 2.8 \\
& \([1,2)\) & 981 & 94.6 & 1.7 & 1.1 & 0.5 & 2.0 \\
& \([2,3]\) & 589 & 86.1 & 8.1 & 3.4 & 0.3 & 2.0 \\

\addlinespace
Moirai~(\texttt{MO})
& \([0.5,1)\) & 950 & 83.1 & 3.8 & 2.1 & 3.2 & 7.9 \\
& \([1,2)\) & 981 & 79.6 & 9.0 & 3.9 & 2.3 & 5.2 \\
& \([2,3]\) & 589 & 59.4 & 25.3 & 8.0 & 1.7 & 5.6 \\

\addlinespace
TTM-R2~(\texttt{TT})
& \([0.5,1)\) & 925 & 60.3 & 5.6 & 3.8 & 14.6 & 15.7 \\
& \([1,2)\) & 1{,}010 & 56.0 & 11.7 & 6.5 & 14.7 & 11.1 \\
& \([2,3]\) & 585 & 44.8 & 25.3 & 10.3 & 12.0 & 7.7 \\
\bottomrule
\end{tabular}
\end{table}

\subsection{Single-Parameter Representation Accessibility}
\label{app:representation_results}

This subsection examines how parameter accessibility develops across depth and compares representations with raw and matched-input controls.
It compares ridge probes with shallow nonlinear probes for oscillation frequency, seasonal dependence, and the TimesFM~3 trend-slope result.

\subsubsection{Layer-Wise Parameter Accessibility}
\label{app:representation_layers}

Parameter accessibility is evaluated for eight of the ten single-parameter laws.
Random walk with drift~(\texttt{RW}) is excluded because its finite-history conditional parameter reference is unstable.
Level shift~(\texttt{LS}) is retained only for law identification because its standardized construction does not provide a comparable continuous target for parameter-level scoring.
The remaining eight laws have stable conditional references and continuous parameter targets.

\begin{figure}[t]
\centering

\includegraphics[width=0.325\linewidth]{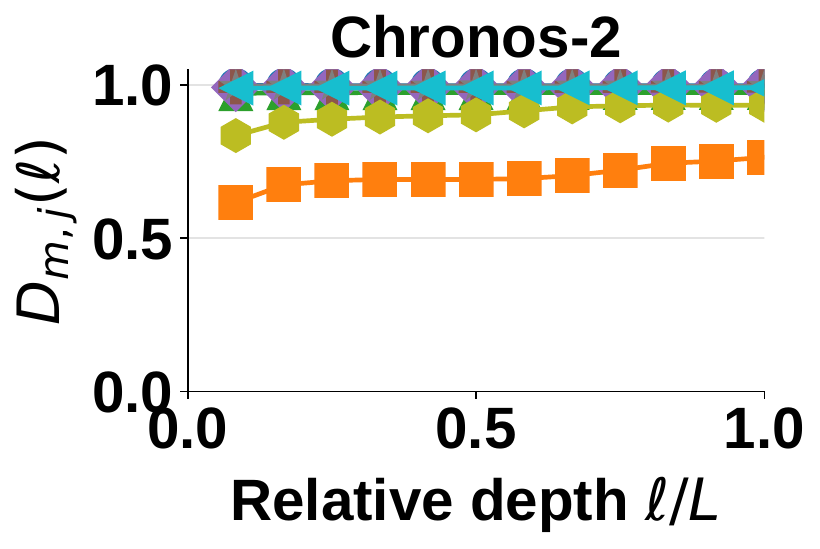}
\hfill
\includegraphics[width=0.325\linewidth]{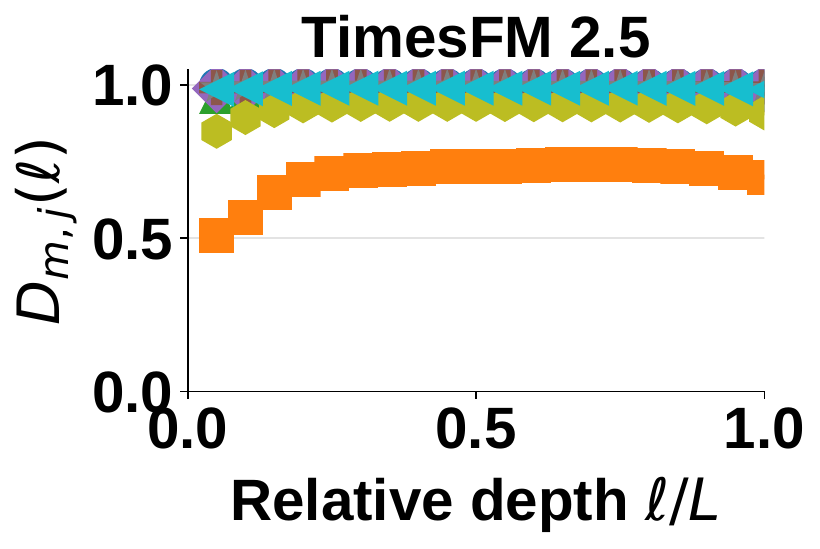}
\hfill
\includegraphics[width=0.325\linewidth]{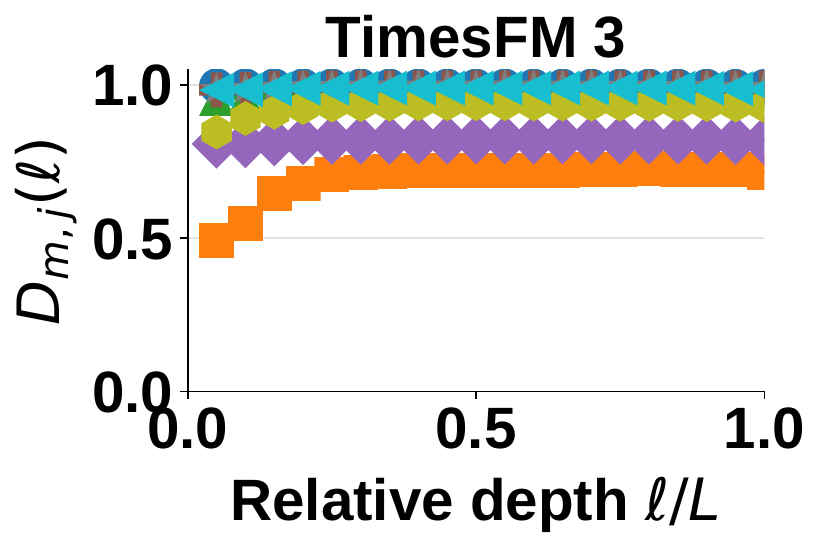}

\vspace{2pt}

\includegraphics[width=0.325\linewidth]{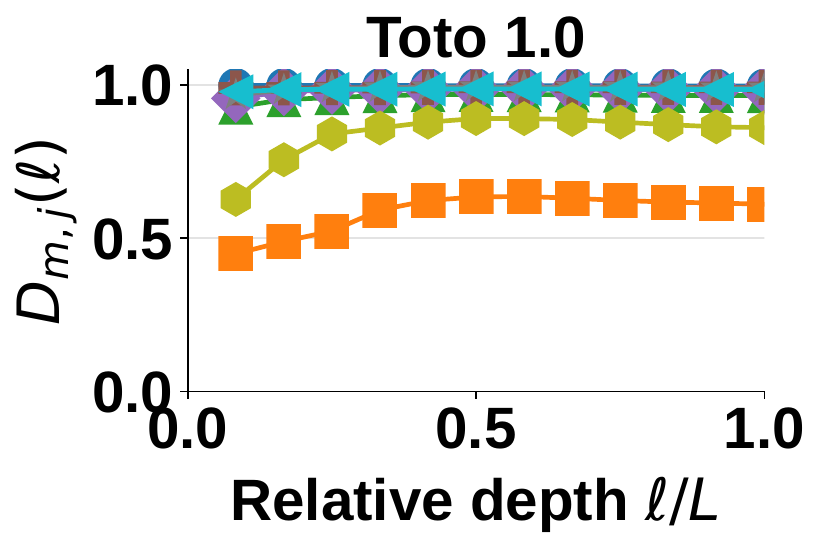}
\hfill
\includegraphics[width=0.325\linewidth]{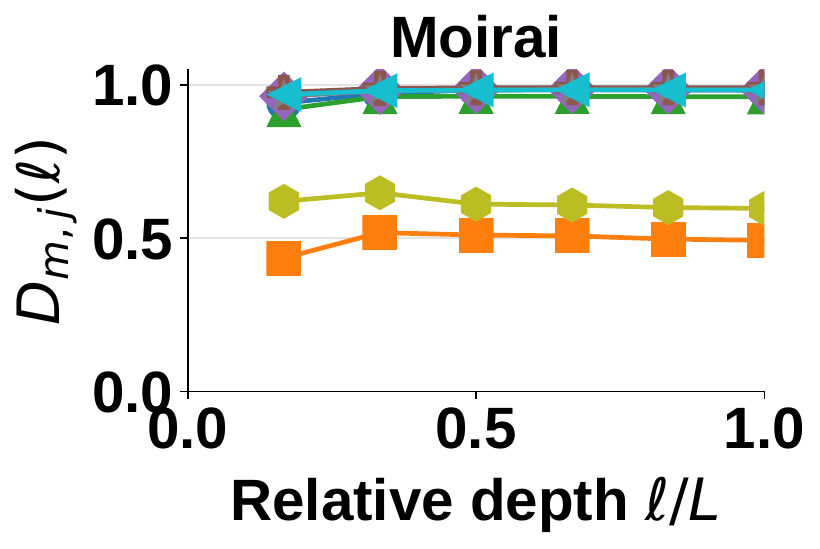}
\hfill
\includegraphics[width=0.325\linewidth]{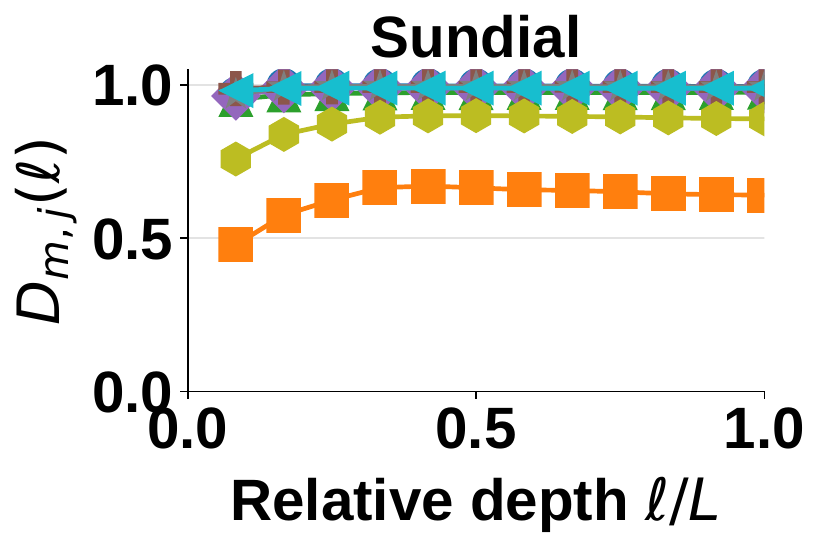}

\vspace{2pt}

\includegraphics[width=0.325\linewidth]{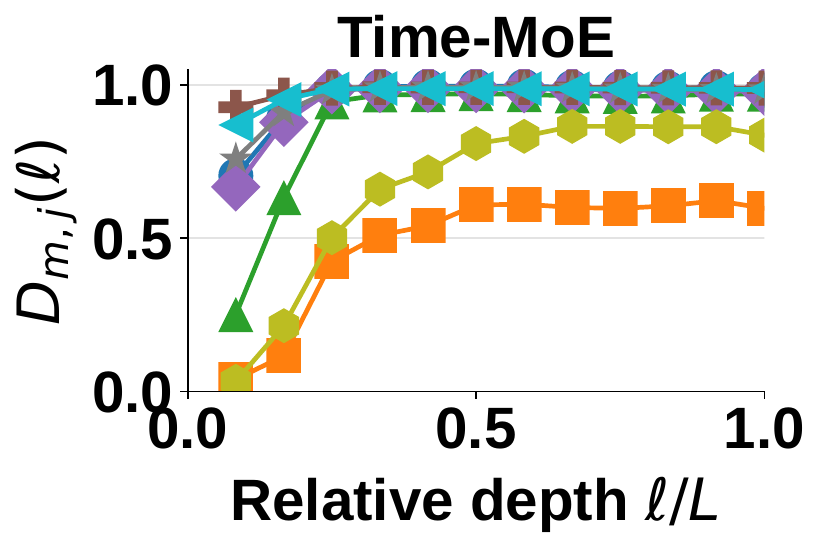}
\hfill
\includegraphics[width=0.325\linewidth]{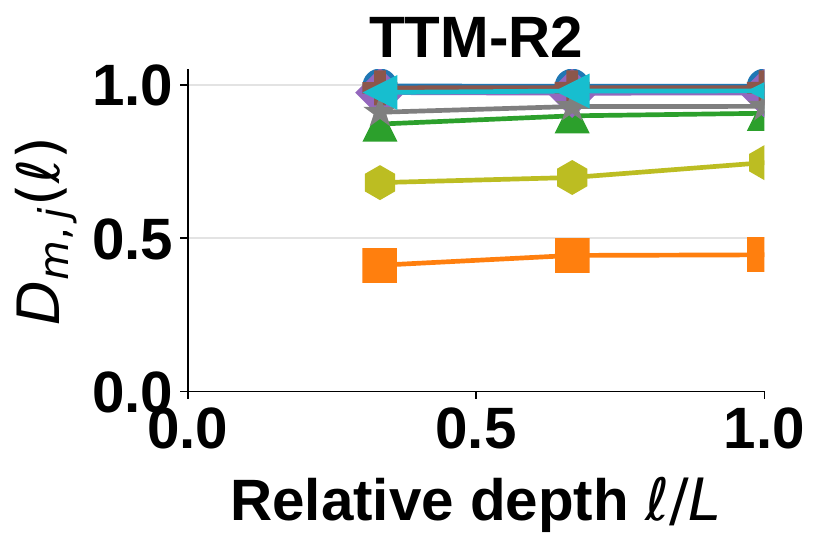}
\hfill
\includegraphics[width=0.325\linewidth]{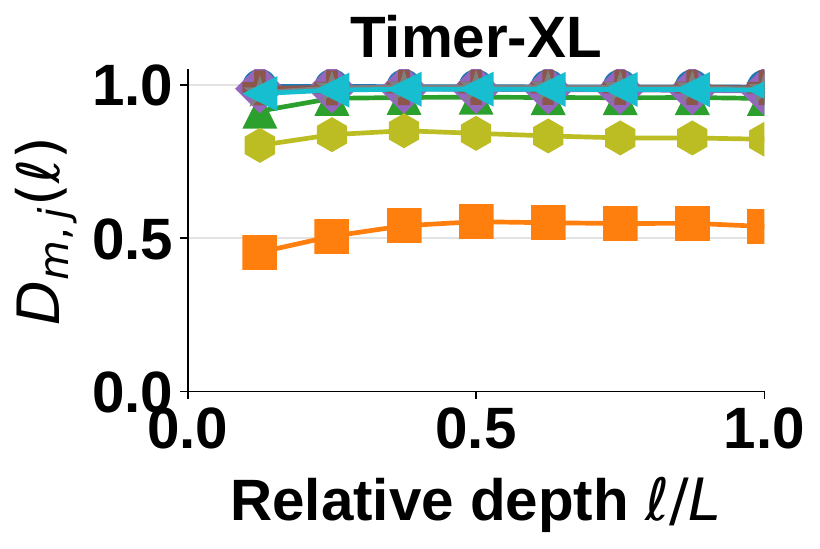}

\vspace{2pt}

\includegraphics[width=\linewidth]{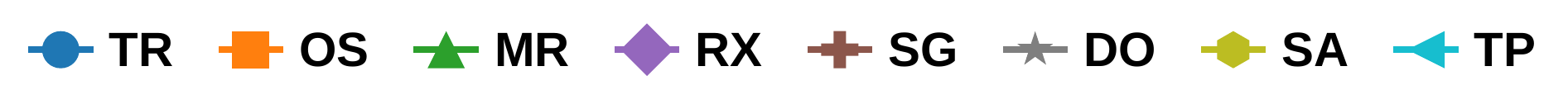}

\caption{
Test-set parameter representation accessibility across relative depth for eight single-parameter laws and nine evaluated TSFM architectures.
}
\label{fig:law_D_layers}
\vspace{-0.1in}
\end{figure}

\autoref{tab:representation_diag} summarizes conditional recoverability, emergence depth, and input controls, while~\autoref{fig:law_D_layers} shows parameter accessibility across model depth.

The main comparison in~\S\ref{sec:eval_representation} uses histories from the forecast-response regime.
Here, we retain the original accessibility-prior depth analysis for the eight laws with parameter-accessibility scores, including exponential relaxation~(\texttt{RX}).
Random walk with drift~(\texttt{RW}) lacks a stable conditional parameter reference, while level shift~(\texttt{LS}) is retained only for law identification.
Relative to the main comparison, this analysis uses a narrower trend prior and the original law-specific sampling settings.

For each law, the table reports the range of~\(D_{m,j}\) across nine models, first-layer accessibility, emergence depth, largest subsequent decline, and input controls.
The controls are properties of the generator rather than individual models.
Their ranges reflect the three context lengths used in the panel.
Seven models use~\(T=256\), Timer-XL~(\texttt{TX}) uses~\(T=288\), and TTM-R2~(\texttt{TT}) uses~\(T=512\).

\begin{table}[t]
\centering
\caption{
Single-parameter references, accessibility profiles, and input controls across nine~TSFMs.
Brackets show minima and maxima across models or context lengths.
}
\label{tab:representation_diag}
\scriptsize
\setlength{\tabcolsep}{3pt}
\renewcommand{\arraystretch}{1.12}
\resizebox{\linewidth}{!}{%
\begin{tabular}{@{}lccccccc@{}}
\toprule
&
\multicolumn{3}{c}{Representation accessibility}
&
\multicolumn{2}{c}{Depth profile}
&
\multicolumn{2}{c}{Input control}
\\
\cmidrule(lr){2-4}
\cmidrule(lr){5-6}
\cmidrule(l){7-8}
Law
& Reference \(r^\star\)
& Selected \(D\)
& First layer \(D(\ell_1)\)
& Emergence
& Decline
& Raw
& Matched \\
\midrule
Linear trend~(\texttt{TR})
& \([0.92,0.96]\)
& \([0.985,0.997]\)
& \([0.70,1.00]\)
& 0.08
& 0.013
& 1.00
& N/A \\

Oscillation~(\texttt{OS})
& \([0.97,0.99]\)
& \([0.444,0.763]\)
& \([0.04,0.62]\)
& 0.38
& 0.041
& \([0.00,0.01]\)
& \([0.65,0.72]\) \\

Mean reversion~(\texttt{MR})
& 0.98
& \([0.907,0.976]\)
& \([0.25,0.97]\)
& 0.08
& 0.009
& 0.00
& 0.00 \\

Exponential relaxation~(\texttt{RX})
& 0.90
& \([0.821,0.993]\)
& \([0.67,0.99]\)
& 0.08
& 0.009
& \([0.83,0.84]\)
& 0.99 \\

Saturating growth~(\texttt{SG})
& 0.98
& \([0.990,0.993]\)
& \([0.93,0.99]\)
& 0.08
& 0.003
& 0.83
& \([0.98,0.99]\) \\

Damped oscillation~(\texttt{DO})
& 0.99
& \([0.929,0.996]\)
& \([0.76,1.00]\)
& 0.08
& 0.008
& \([-0.01,0.00]\)
& \([0.00,0.01]\) \\

Seasonal autoregression~(\texttt{SA})
& 0.98
& \([0.648,0.938]\)
& \([0.03,0.85]\)
& 0.25
& 0.051
& \(-0.01\)
& \([0.00,0.01]\) \\

Transient pulse~(\texttt{TP})
& 1.00
& \([0.980,0.990]\)
& \([0.87,0.99]\)
& 0.08
& 0.004
& \([-0.01,0.01]\)
& \([-0.01,0.00]\) \\
\bottomrule
\end{tabular}
}
\end{table}

Three properties of the layer-wise profiles are not visible in the validation-selected summary of~\autoref{fig:panel_D_laws}, which contains one layer per TSFM.

\textbf{Six Parameters Are Accessible Early, While Two Require Greater Depth.}
Linear trend~(\texttt{TR}), mean reversion~(\texttt{MR}), exponential relaxation~(\texttt{RX}), saturating growth~(\texttt{SG}), damped oscillation~(\texttt{DO}), and transient pulse~(\texttt{TP}) reach~\(95\%\) of their maximum by the second layer in eight of nine models.
For most corresponding model--law cells, first-layer accessibility is already between~\(0.91\) and~\(1.00\).
Oscillation frequency~(\texttt{OS}) and seasonal autoregression~(\texttt{SA}) are the exceptions, reaching~\(95\%\) of their maxima at median relative depths of~\(0.38\) and~\(0.25\).
Oscillation reaches this threshold only at~\(0.83\) of Chronos-2's~(\texttt{C2}) depth, while seasonal autoregression reaches it at the final layer of TTM-R2~(\texttt{TT}).
These two parameters therefore exhibit larger depth-dependent gains in linear accessibility.

\textbf{Time-MoE Begins with Lower Linear Accessibility.}
Time-MoE's~(\texttt{ME}) first-layer accessibility averages~\(0.53\) across the eight laws, compared with~\(0.85\)--\(0.92\) for every other model.
Its first-layer scores are near chance for oscillation frequency at~\(0.04\) and seasonal autoregression at~\(0.03\).
Accessibility then increases by~\(0.40\) on average, including gains of~\(0.83\) for seasonal autoregression and~\(0.72\) for mean reversion.
Six laws reach the range of the remaining models by one quarter of the network depth.
Seasonal autoregression and oscillation frequency reach~\(95\%\) of their maxima only at relative depths of~\(0.58\) and~\(0.50\).
The law-identification profiles above show a similar delayed emergence, indicating a broader property of Time-MoE's early layers.

\textbf{Accessibility Profiles Generally Saturate Rather Than Peak.}
No curve declines by more than~\(0.051\) after reaching its maximum.
The largest decline occurs for Moirai~(\texttt{MO}) on seasonal autoregression, followed by~\(0.041\) for TimesFM~2.5~(\texttt{F2}) on oscillation frequency.
Linear accessibility is therefore largely retained across later measured layers.
TimesFM~3~(\texttt{F3}) provides the principal flat outlier: exponential-relaxation accessibility remains between~\(0.81\) and~\(0.82\) across all~20 layers.
The low end of the exponential-relaxation range in~\autoref{tab:representation_diag} therefore does not result from layer selection.

Together, the profiles show early and persistent linear accessibility for most parameters, with later emergence for oscillation frequency and seasonal dependence.

\subsubsection{Linear and Nonlinear Probe Comparison}
\label{app:revision_probes}

\textbf{Probe Setting.}
We compare the ridge probe with a one-hidden-layer GELU MLP using the same features, targets, datasets, and selected layers.
The MLP uses hidden width~64 or~256, with dropout and weight decay selected on validation data.
The comparison contains~19 distinct model--law cells: nine for oscillation frequency~(\texttt{OS}), nine for seasonal dependence~(\texttt{SA}), and one for TimesFM~3~(\texttt{F3}) trend slope~(\texttt{TR}), using one independent result per cell.

For cell~\(i\), the nonlinear accessibility difference is
\[
\Delta D_i
=
D_{i,\mathrm{MLP}}
-
D_{i,\mathrm{linear}}.
\]
A positive value indicates higher accessibility under the MLP, while a negative value favors the linear probe.
The differences are computed for individual cells before aggregation.
\autoref{tab:revision_probe_headroom} reports the median linear and nonlinear accessibility scores together with the median and maximum cellwise differences across the evaluated models and parameters.

\begin{table}[t]
\centering
\caption{
Linear and nonlinear probe accessibility across~19 model--law cells.
Accessibility differences are computed separately for each cell.
}
\label{tab:revision_probe_headroom}
\footnotesize
\setlength{\tabcolsep}{3pt}
\renewcommand{\arraystretch}{1.1}
\begin{tabularx}{\linewidth}{@{}
>{\raggedright\arraybackslash}p{0.25\linewidth}
>{\raggedright\arraybackslash}p{0.15\linewidth}
c
>{\hsize=0.90\hsize\centering\arraybackslash}X
>{\hsize=0.90\hsize\centering\arraybackslash}X
>{\hsize=0.85\hsize\centering\arraybackslash}X
>{\hsize=1.35\hsize\centering\arraybackslash}X
@{}}
\toprule
&
&
&
\multicolumn{2}{c}{Median Accessibility}
&
\multicolumn{2}{c}{Accessibility Difference}
\\
\cmidrule(lr){4-5}
\cmidrule(l){6-7}
Parameter
& Models
& Cells
& Linear \(D\)
& MLP \(D\)
& \mbox{Median \(\Delta D\)}
& \mbox{Maximum \(\Delta D\)} \\
\midrule
Oscillation frequency~(\texttt{OS})
& All nine
& 9
& 0.636
& 0.708
& \(+0.046\)
& \(+0.078\) \\

Seasonal dependence~(\texttt{SA})
& All nine
& 9
& 0.890
& 0.919
& \(+0.029\)
& \(+0.115\) \\

Trend slope~(\texttt{TR})
& TimesFM~3~(\texttt{F3})
& 1
& 0.372
& 0.363
& \(-0.010\)
& \(-0.010\) \\
\bottomrule
\end{tabularx}
\end{table}

\textbf{Control Probes.}
Random-label and permutation controls select their probe configurations using their corresponding validation labels.
The largest absolute correlation between control predictions and control labels is~\(0.1422\).
The maximum correlation between control predictions and the true labels is~\(0.2905\) and serves as a separate leakage diagnostic.

\textbf{Nonlinear Accessibility Differences.}
The MLP increases median accessibility by~\(0.046\) for oscillation frequency~(\texttt{OS}) and by~\(0.029\) for seasonal dependence~(\texttt{SA}).
The maximum cellwise improvements are~\(0.078\) and~\(0.115\), respectively.
For TimesFM~3~(\texttt{F3}) trend slope~(\texttt{TR}), accessibility instead changes from~\(0.372\) to~\(0.363\).
This MLP therefore recovers modest additional accessibility for oscillation and seasonal dependence but does not improve the wide-slope trend result.
The main accessibility--response comparison retains the matched-regime ridge scores.

\textbf{Sampling Uncertainty.}
Across~71 original-regime model--law cells, accessibility standard errors have median~\(0.0005\) and maximum~\(0.0034\).
These values summarize within-cell sampling variation for the original accessibility datasets.
The main matched-regime analysis contains a separate set of~63 model--law cells with its own response-regime sampling design.

\subsection{Single-Parameter Forecast Response}
\label{app:response_results}

Forecast response is evaluated for seven of the ten single-parameter laws.
Random walk with drift\~(\texttt{RW}) and level shift\~(\texttt{LS}) have no parameter-level score, while exponential relaxation\~(\texttt{RX}) is excluded because its conditional response reference is unstable.
This subsection relates representation accessibility to matched-input forecast response under a common generative regime, examines stability across intervention magnitudes, evaluates alternative response estimators, and audits the exclusion of exponential relaxation through complementary robustness and sensitivity analyses.

\begin{table}[t]
\centering
\caption{
Parameter accessibility on the original and response-regime datasets.
Each model entry reports original\(\rightarrow\)response-regime accessibility, and the final row reports the corresponding conditional reference~\(r^\star\) computed from the same history distribution.
}
\label{tab:matched}
\footnotesize
\setlength{\tabcolsep}{3pt}
\renewcommand{\arraystretch}{1.12}
\resizebox{\linewidth}{!}{%
\begin{tabular}{@{}lccccccc@{}}
\toprule
Model
& \texttt{TR}
& \texttt{OS}
& \texttt{MR}
& \texttt{SG}
& \texttt{DO}
& \texttt{SA}
& \texttt{TP} \\
\midrule
Chronos-2~(\texttt{C2})
& \(0.997\rightarrow0.999\)
& \(0.763\rightarrow0.796\)
& \(0.975\rightarrow0.972\)
& \(0.993\rightarrow0.992\)
& \(0.996\rightarrow0.996\)
& \(0.934\rightarrow0.927\)
& \(0.990\rightarrow0.989\) \\

TimesFM~2.5~(\texttt{F2})
& \(0.997\rightarrow0.999\)
& \(0.738\rightarrow0.770\)
& \(0.976\rightarrow0.973\)
& \(0.993\rightarrow0.992\)
& \(0.996\rightarrow0.995\)
& \(0.936\rightarrow0.931\)
& \(0.989\rightarrow0.987\) \\

TimesFM~3~(\texttt{F3})
& \(0.997\rightarrow0.372\)
& \(0.726\rightarrow0.764\)
& \(0.976\rightarrow0.973\)
& \(0.990\rightarrow0.989\)
& \(0.996\rightarrow0.995\)
& \(0.938\rightarrow0.934\)
& \(0.989\rightarrow0.988\) \\

Toto-Open-Base~1.0~(\texttt{TO})
& \(0.997\rightarrow0.999\)
& \(0.636\rightarrow0.686\)
& \(0.968\rightarrow0.965\)
& \(0.992\rightarrow0.991\)
& \(0.992\rightarrow0.991\)
& \(0.890\rightarrow0.884\)
& \(0.986\rightarrow0.985\) \\

Moirai~(\texttt{MO})
& \(0.985\rightarrow0.998\)
& \(0.518\rightarrow0.592\)
& \(0.962\rightarrow0.959\)
& \(0.990\rightarrow0.989\)
& \(0.984\rightarrow0.981\)
& \(0.648\rightarrow0.629\)
& \(0.983\rightarrow0.982\) \\

Sundial~(\texttt{SU})
& \(0.997\rightarrow0.999\)
& \(0.670\rightarrow0.719\)
& \(0.976\rightarrow0.973\)
& \(0.993\rightarrow0.992\)
& \(0.995\rightarrow0.994\)
& \(0.899\rightarrow0.894\)
& \(0.989\rightarrow0.988\) \\

Time-MoE~(\texttt{ME})
& \(0.995\rightarrow0.999\)
& \(0.622\rightarrow0.680\)
& \(0.971\rightarrow0.968\)
& \(0.992\rightarrow0.991\)
& \(0.992\rightarrow0.991\)
& \(0.864\rightarrow0.856\)
& \(0.988\rightarrow0.986\) \\

TTM-R2~(\texttt{TT})
& \(0.996\rightarrow0.998\)
& \(0.444\rightarrow0.489\)
& \(0.907\rightarrow0.902\)
& \(0.990\rightarrow0.990\)
& \(0.929\rightarrow0.927\)
& \(0.746\rightarrow0.733\)
& \(0.980\rightarrow0.980\) \\

Timer-XL~(\texttt{TX})
& \(0.995\rightarrow0.999\)
& \(0.550\rightarrow0.621\)
& \(0.959\rightarrow0.956\)
& \(0.991\rightarrow0.990\)
& \(0.993\rightarrow0.992\)
& \(0.850\rightarrow0.841\)
& \(0.985\rightarrow0.985\) \\
\midrule
Conditional reference~\(r^\star\)
& \(0.922\rightarrow0.999\)
& \(0.988\rightarrow0.990\)
& \(0.976\rightarrow0.975\)
& \(0.983\rightarrow0.982\)
& \(0.986\rightarrow0.986\)
& \(0.976\rightarrow0.975\)
& \(0.997\rightarrow0.997\) \\
\bottomrule
\end{tabular}
}
\end{table}

\subsubsection{Accessibility and Response under a Common Regime}
\label{app:response_accessibility}
\label{app:matched}

Accessibility predicts relative response primarily where it varies across models.
For oscillation frequency~(\texttt{OS}) and seasonal autoregression~(\texttt{SA}), the Spearman correlations are~\(0.92\) and~\(0.82\), although their largest normalized responses reach only~\(0.64\) and~\(0.71\).
Among the~42 cells with accessibility above~\(0.95\), 20 have response below~\(0.5\), and five respond oppositely.
Linear trend~(\texttt{TR}) provides the clearest example: eight models have accessibility between~\(0.998\) and~\(0.999\), including all three with negative trend responses.

For a distribution-matched comparison, we repeat the accessibility probes on histories from the response experiments using the original splits and validation-based model selection.
For the six paired laws, both arms enter the accessibility dataset with their respective parameter values.
The accessibility and response experiments otherwise share the generator, parameter and nuisance distributions, noise, context length, and horizon.
Linear trend instead uses independent histories from the wider response-regime slope distribution, whose standard deviation is~\(0.260\) rather than~\(0.020\).
Each score is normalized by the conditional reference~\(r^\star\) computed on the corresponding history distribution.
\autoref{tab:matched} reports the resulting scores for all models and laws.

\textbf{Five Paired Laws Retain Their Accessibility.}
Across mean reversion~(\texttt{MR}), saturating growth~(\texttt{SG}), damped oscillation~(\texttt{DO}), seasonal autoregression~(\texttt{SA}), and transient pulse~(\texttt{TP}), the largest changes are~\(0.019\) for seasonal autoregression and~\(0.005\) for mean reversion.
All remaining changes are at most~\(0.003\), and model ordering is effectively unchanged.
The accessibility--response correlations remain~\(0.92\) for mean reversion and~\(0.82\) for seasonal autoregression.

\textbf{Oscillation Becomes Easier without Changing Model Order.}
Every model gains between~\(0.032\) and~\(0.074\) on oscillation frequency~(\texttt{OS}), while~\(r^\star\) changes by at most~\(0.002\).
The response histories contain fewer examples with less than one observed cycle, increasing accessibility without changing model order.
The accessibility--response correlation therefore remains~\(0.92\).

\textbf{Trend Remains Nearly Complete Except for TimesFM~3.}
Under the response-regime slope distribution, eight models reach accessibility between~\(0.998\) and~\(0.999\).
These include Time-MoE~(\texttt{ME}), TTM-R2~(\texttt{TT}), and Timer-XL~(\texttt{TX}), whose trend responses are negative.
TimesFM~3~(\texttt{F3}) instead declines from~\(0.997\) on the original dataset to~\(0.372\) under the response regime, remaining low across all layers despite having the strongest trend response at~\(1.01\).
Its separate detrending pathway distinguishes selected-state accessibility from information used elsewhere in the forecasting pipeline.

Therefore, the common-regime results confirm that high parameter accessibility can coexist with weak or reversed forecast response across models and controlled dynamical settings.

\begin{figure}[t]
\centering
\includegraphics[width=0.495\linewidth]{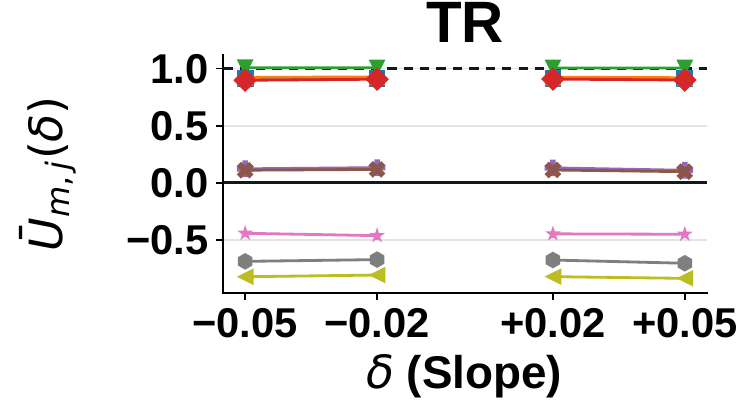}
\hfill
\includegraphics[width=0.495\linewidth]{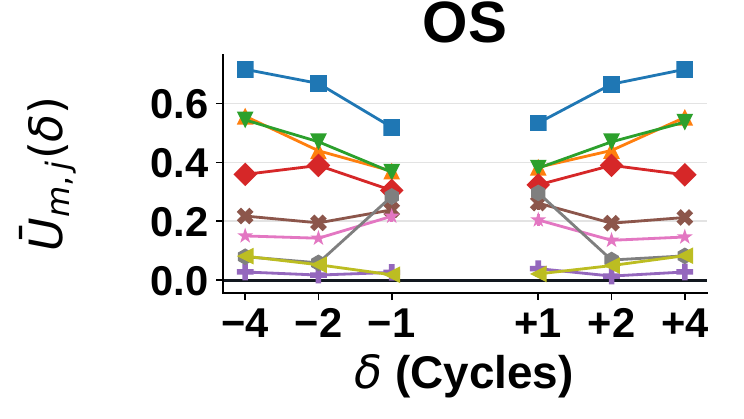}
\\[3pt]
\includegraphics[width=0.495\linewidth]{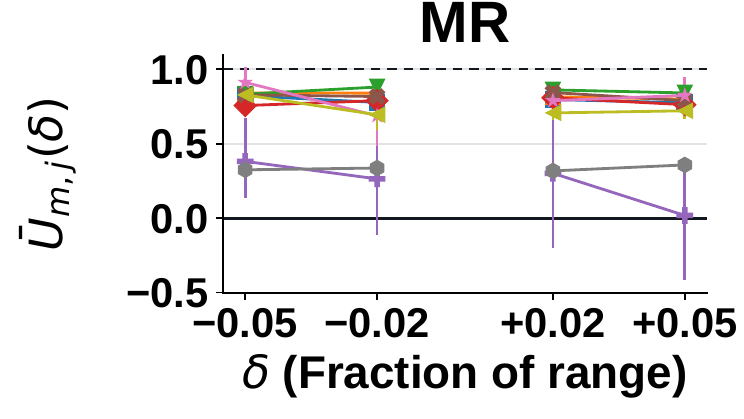}
\hfill
\includegraphics[width=0.495\linewidth]{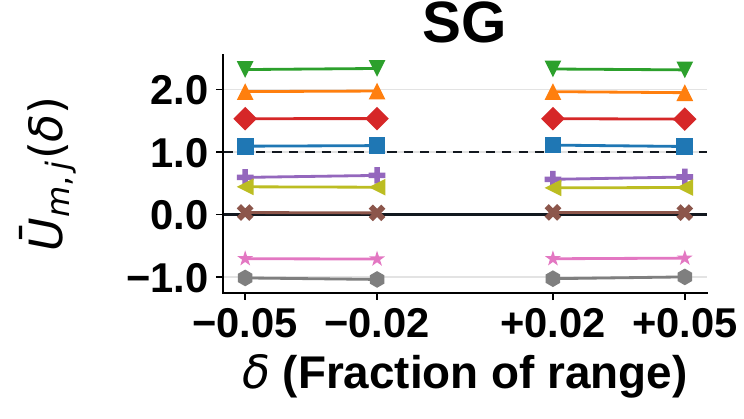}
\\[3pt]
\includegraphics[width=0.495\linewidth]{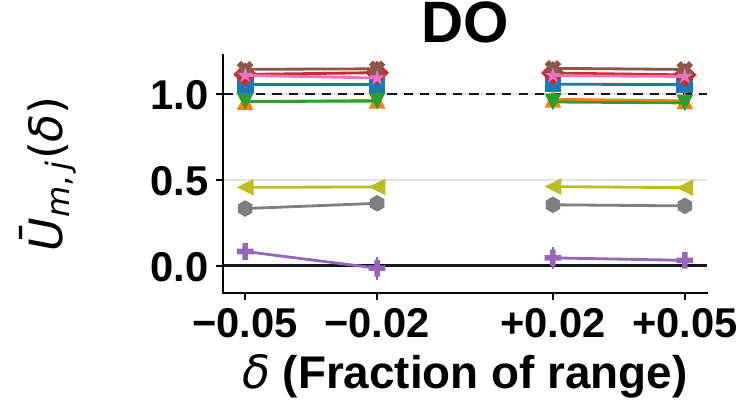}
\hfill
\includegraphics[width=0.495\linewidth]{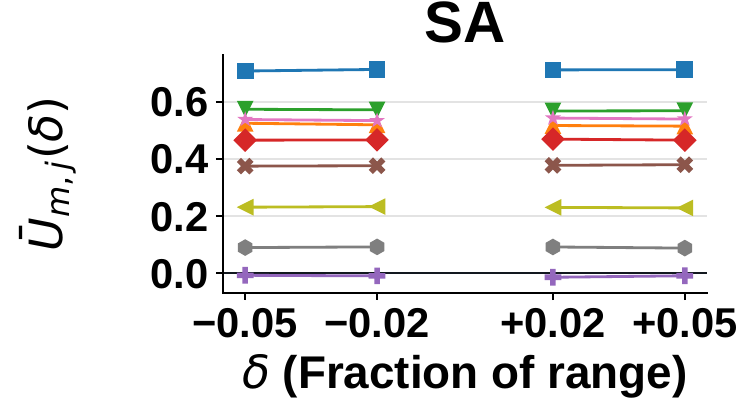}
\\[3pt]
\includegraphics[width=0.495\linewidth]{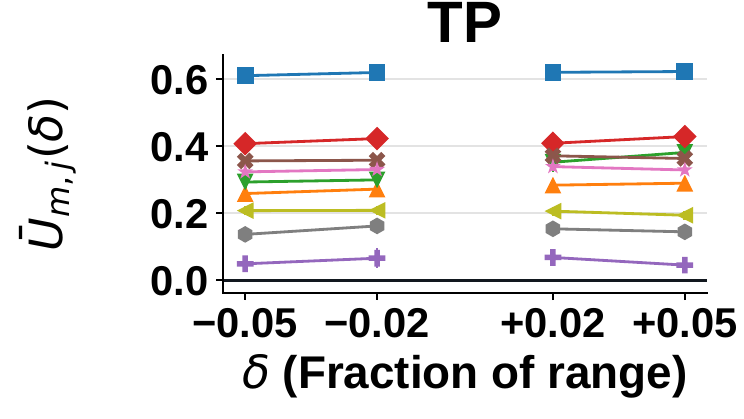}
\hfill
\includegraphics[width=0.495\linewidth]{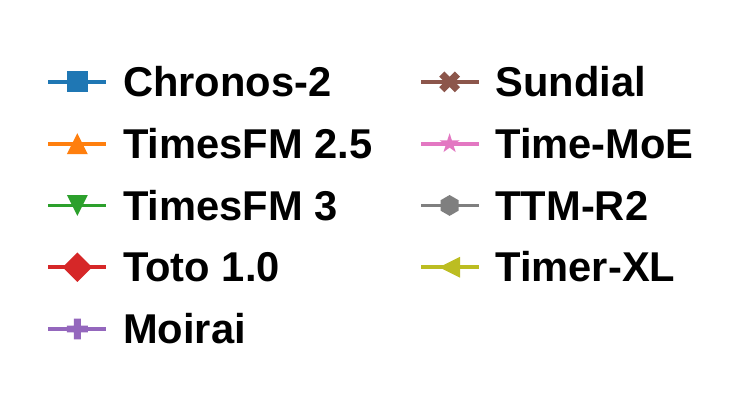}
\caption{
Conditional-reference-normalized response across signed intervention magnitudes.
The dashed line and bands mark the conditional reference and~\(95\%\) bootstrap intervals.
}
\label{fig:law_U_dose}
\end{figure}

\subsubsection{Response Stability Across Intervention Magnitudes}
\label{app:response_dose}

\autoref{fig:law_U_dose} reports the conditional-reference-normalized response at every signed intervention magnitude for the seven scored laws.
The main score~\(\bar U_{m,j}\) averages the model and conditional-reference coefficients separately across these magnitudes.

\textbf{Six Laws Are Stable Across Magnitudes.}
For linear trend~(\texttt{TR}), mean reversion~(\texttt{MR}), saturating growth~(\texttt{SG}), damped oscillation~(\texttt{DO}), seasonal autoregression~(\texttt{SA}), and transient pulse~(\texttt{TP}), the median response spread ranges from~\(0.006\) to~\(0.037\).
After conditional normalization, the spread ranges from~\(0.006\) to~\(0.052\), while the median fitted change from the smallest to largest magnitude remains within~\(0.01\) of zero.
The negative trend responses of Time-MoE~(\texttt{ME}), TTM-R2~(\texttt{TT}), and Timer-XL~(\texttt{TX}) persist at every magnitude, as do the negative saturating-growth responses of Time-MoE~(\texttt{ME}) and TTM-R2~(\texttt{TT}), confirming these effects are not dose-specific.

\textbf{Oscillation Frequency Is the Main Exception.}
Oscillation frequency~(\texttt{OS}) has a median spread of~\(0.073\) before and~\(0.085\) after conditional normalization.
Chronos-2~(\texttt{C2}), TimesFM~2.5~(\texttt{F2}), and TimesFM~3~(\texttt{F3}) respond more strongly to larger frequency shifts, whereas TTM-R2~(\texttt{TT}) falls from~\(0.29\) to~\(0.06\) between shifts of one and two cycles.
These model-specific trends largely cancel, leaving the mean response near~\(0.27\) at one cycle and~\(0.30\) at four cycles despite changes in model ordering across the evaluated frequency intervention magnitudes.

\textbf{Signs Are Stable under Alternative Aggregations.}
Among the~43 cells whose intervals exclude zero at every magnitude, 34 differ by less than~\(0.02\) between positive and negative interventions.
Most larger asymmetries shrink under a~\(10\%\)-trimmed average, indicating sensitivity to heavy-tailed per-pair coefficients.
Across the five laws with stored per-pair coefficients, trimmed means preserve the sign and remain within~\(0.1\) of~\(\bar U_{m,j}\) in 36 of~45 cells, compared with 28 cells under median aggregation.
Energy-weighted pooled ratios differ more often, especially for transient pulse, because they emphasize pairs with the largest reference responses.

Overall, weak and reversed responses persist across intervention magnitudes and reasonable pair-weighted aggregations across evaluated models and dynamical laws.

\begin{table}[t]
\centering
\caption{
Scale-aware and unpaired response diagnostics.
Panel A summarizes seven laws across nine models, while Panel B reports cells with~\(\bar U_{m,j}<0\) or~\(\bar U_{m,j}>1.5\).
Bracketed entries report model minima and maxima; unbracketed~\(\gamma\), \(\bar\rho\), and~\(R^2\) entries in Panel A report medians.
}
\label{tab:response_diag}
\footnotesize
\setlength{\tabcolsep}{3pt}
\renewcommand{\arraystretch}{1.12}

\par
{\raggedright\textbf{\MakeUppercase{Panel A: Law-Level Summaries}}\par}
\smallskip

\resizebox{\linewidth}{!}{%
\begin{tabular}{@{}lrrrrrrr@{}}
\toprule
& \multicolumn{5}{c}{Paired Intervention}
& \multicolumn{2}{c}{Unpaired Check} \\
\cmidrule(lr){2-6}
\cmidrule(l){7-8}
Law
& \(\bar U_{m,j}\)
& \(U^{(3)}\)
& \(\gamma\)
& \(\bar\rho\)
& \(U^{\star(3)}/U^\star\)
& Coefficient
& \(R^2\) \\
\midrule
Linear trend~(\texttt{TR})
& \([-0.82,1.01]\)
& \([-0.87,1.22]\)
& \(-0.021\)
& --
& 0.93
& \([-1.54,0.89]\)
& -- \\

Oscillation~(\texttt{OS})
& \([0.03,0.64]\)
& \([0.02,0.49]\)
& \(-0.100\)
& 0.42
& 0.82
& \([-0.15,0.48]\)
& 0.52 \\

Mean reversion~(\texttt{MR})
& \([0.24,0.85]\)
& \([0.24,0.79]\)
& \(-0.007\)
& 0.68
& 0.81
& \([0.59,1.26]\)
& 0.78 \\

Saturating growth~(\texttt{SG})
& \([-1.02,2.33]\)
& \([-0.99,0.63]\)
& \(-0.002\)
& 0.85
& 1.01
& \([-1.03,1.14]\)
& 0.87 \\

Damped oscillation~(\texttt{DO})
& \([0.04,1.15]\)
& \([0.02,0.91]\)
& \(-0.005\)
& 0.97
& 1.01
& \([0.04,1.22]\)
& 0.95 \\

Seasonal autoregression~(\texttt{SA})
& \([-0.01,0.71]\)
& \([-0.01,0.43]\)
& \(-0.003\)
& 0.33
& 0.73
& \([-0.04,0.36]\)
& 0.73 \\

Transient pulse~(\texttt{TP})
& \([0.06,0.62]\)
& \([0.04,0.45]\)
& \(-0.006\)
& 0.91
& 1.08
& \([0.02,0.50]\)
& 0.67 \\
\bottomrule
\end{tabular}
}

\medskip
\par
{\raggedright\textbf{\MakeUppercase{Panel B: Weak, Reversed, or Amplified Responses}}\par}
\smallskip

\resizebox{\linewidth}{!}{%
\begin{tabular}{@{}llrrrrrr@{}}
\toprule
Law
& Model
& \(\bar U_{m,j}\)
& \(U^{(3)}\)
& \(\gamma\)
& \(\bar\rho\)
& Coefficient
& \(R^2\) \\
\midrule
Linear trend~(\texttt{TR})
& Time-MoE~(\texttt{ME})
& \(-0.45\)
& \(-0.19\)
& \(-0.103\)
& --
& \(-1.41\)
& -- \\

Linear trend~(\texttt{TR})
& TTM-R2~(\texttt{TT})
& \(-0.69\)
& \(-0.75\)
& \(-0.003\)
& --
& \(-1.27\)
& -- \\

Linear trend~(\texttt{TR})
& Timer-XL~(\texttt{TX})
& \(-0.82\)
& \(-0.87\)
& \(-0.021\)
& --
& \(-1.54\)
& -- \\
\addlinespace[2pt]

Saturating growth~(\texttt{SG})
& TimesFM~2.5~(\texttt{F2})
& 1.97
& 0.36
& \(-0.003\)
& 0.99
& 1.02
& 0.99 \\

Saturating growth~(\texttt{SG})
& TimesFM~3~(\texttt{F3})
& 2.33
& 0.39
& \(-0.005\)
& 0.99
& 1.14
& 0.99 \\

Saturating growth~(\texttt{SG})
& Toto-Open-Base~1.0~(\texttt{TO})
& 1.53
& 0.40
& \(-0.002\)
& 0.99
& 0.96
& 0.98 \\

Saturating growth~(\texttt{SG})
& Time-MoE~(\texttt{ME})
& \(-0.71\)
& \(-0.30\)
& \(-0.009\)
& 0.85
& \(-0.21\)
& 0.87 \\

Saturating growth~(\texttt{SG})
& TTM-R2~(\texttt{TT})
& \(-1.02\)
& \(-0.99\)
& 0.000
& 0.80
& \(-1.03\)
& 0.70 \\
\addlinespace[2pt]

Seasonal autoregression~(\texttt{SA})
& Moirai~(\texttt{MO})
& \(-0.01\)
& \(-0.01\)
& \(-0.165\)
& 0.19
& \(-0.04\)
& 0.07 \\
\bottomrule
\end{tabular}
}

\smallskip
\parbox{\linewidth}{
\scriptsize
For linear trend~(\texttt{TR}), the reference forecast and reference change are collinear across the horizon, so the refit does not uniquely separate shape from scale.
Dashes denote unavailable or non-identifiable diagnostics.
}
\end{table}

\subsubsection{Response-Metric Diagnostics}
\label{app:response_diag}

We evaluate whether the matched-input score reflects the designated response rather than rescaling of the existing forecast or an artifact of the paired construction.
For pair~\(i\), let~\(q_i\) denote the model forecast change, \(r_i=\Delta y_i^{\mathrm{ref}}\) the reference change, and \(b_i=\mathcal{C}F_m(x_i^A)\) the centered arm-A forecast.
To separate these two contributions explicitly, the scale-aware fit is
\begin{equation}
(a_i,U_i^{(3)},\gamma_i)
\in
\arg\min_{a,u,\gamma}
\left\|
q_i-a\mathbf{1}-u r_i-\gamma b_i
\right\|_2^2.
\label{eq:scale_refit}
\end{equation}
Here, \(U_i^{(3)}\) measures the reference-aligned response after controlling for a component proportional to the existing forecast.
The unpaired diagnostic separately compares each model forecast with the conditional forecast for the same history, without matched pairs or response normalization.
\autoref{tab:response_diag} summarizes both diagnostics across evaluated model and law combinations.

\textbf{Opposing Trend Responses Persist.}
Under the scale-aware refit, the negative trend responses remain~\(-0.19\), \(-0.75\), and~\(-0.87\) for Time-MoE~(\texttt{ME}), TTM-R2~(\texttt{TT}), and Timer-XL~(\texttt{TX}), respectively.
Their fitted forecast-slope ratios in the unpaired noisy-ramp check are~\(-1.41\), \(-1.27\), and~\(-1.54\).
The opposing responses therefore persist without matched pairs or conditional normalization.

\textbf{Saturating Growth Separates Alignment from Scale.}
Chronos-2~(\texttt{C2}), Toto-Open-Base~1.0~(\texttt{TO}), TimesFM~2.5~(\texttt{F2}), and TimesFM~3~(\texttt{F3}) have saturating-growth matched-input responses between~\(1.10\) and~\(2.33\).
The scale-aware refit places the same models between~\(0.36\) and~\(0.40\).
The two regressors have median absolute correlation~\(\bar\rho=0.99\), whereas the unpaired coefficients remain between~\(0.96\) and~\(1.14\), with shape~\(R^2\) between~\(0.98\) and~\(0.99\).
These models therefore track the reference response shape but overstate its matched-input magnitude, while the scale-aware refit removes part of the aligned component.

Across the remaining laws, the paired and unpaired diagnostics generally preserve the direction of the primary response.
Alternative estimators therefore do not remove the separation between parameter accessibility and forecast expression.

\subsubsection{Excluded-Law Reference Audit}
\label{app:revision_excluded}

\begin{figure}[t]
\centering
\includegraphics[width=\linewidth]{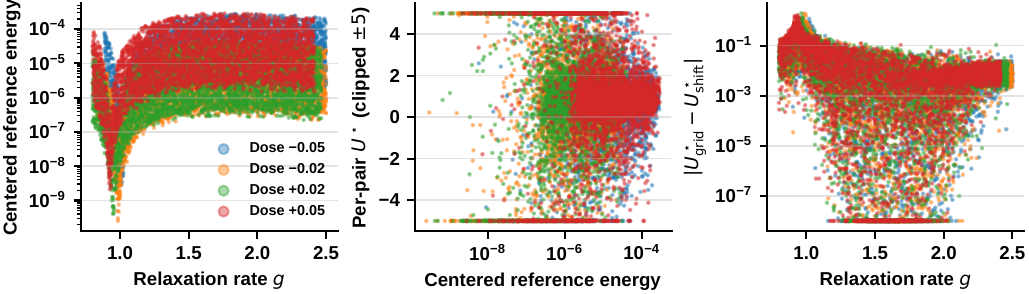}
\caption{
Exponential-relaxation reference stability across the original and restricted parameter ranges under multiple numerical robustness checks.
}
\label{fig:revision_er}
\end{figure}

Exponential relaxation~(\texttt{RX}) remains in law identification and representation accessibility but is excluded from normalized forecast-response scoring because many matched futures contain little centered reference change over the forecast horizon.
As shown in~\autoref{fig:revision_er}, \(33.5\%\) of pairs on the original parameter range fail the relative nondegeneracy criterion.
Across quadrature budgets, grid shifts, independent datasets, and aggregation rules, the reference summary changes by as much as~\(0.228\).
Restricting the range to its central~\(80\%\) and~\(60\%\) reduces this variation to~\(0.054\) and~\(0.034\), but still leaves~\(29.4\%\) and~\(22.2\%\) degenerate pairs.
No evaluated range reaches the required~\(90\%\) nondegenerate-pair threshold with a converged reference.
The exclusion therefore reflects an unstable response reference rather than unfavorable model performance across tested numerical and range restrictions in every evaluated robustness configuration considered.

\subsection{Single-Parameter Causal Geometry}
\label{app:geometry_results}

Causal geometry uses the same seven laws as the forecast-response analysis to explain why accessible parameters can remain weakly expressed in matched-input forecasts.
We first validate local edits that produce the reference response, characterize their shared subspaces and target specificity, compare the required edits with input-induced hidden-state changes, and evaluate input-dependent coordinate prediction.
The final analysis measures full-vector fidelity for matched-input, projected, and deployed responses across every evaluated model and parameter.

\begin{figure}[t]
\centering
\subfigure[Selected-layer test response.]{
\includegraphics[width=0.45\linewidth]{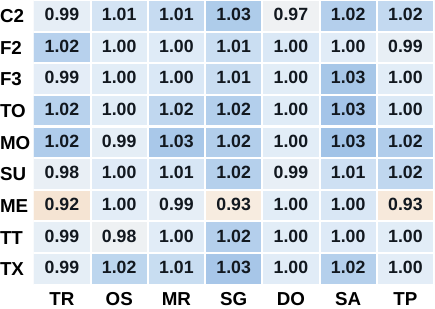}}
\hfill
\subfigure[Validation-admissible layers.]{
\includegraphics[width=0.45\linewidth]{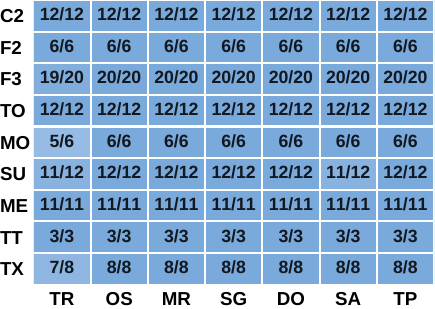}}
\caption{
Selected-layer test response and validation-admissible layer fraction for Jacobian-derived hidden-state edits across seven single-parameter laws.
}
\label{fig:local_validity}
\end{figure}

\begin{figure}[t]
\centering
\includegraphics[width=0.325\linewidth]{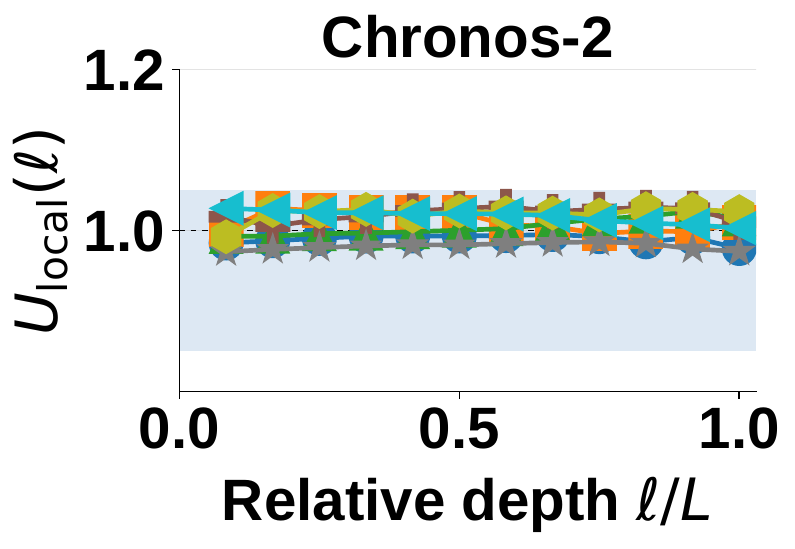}
\hfill
\includegraphics[width=0.325\linewidth]{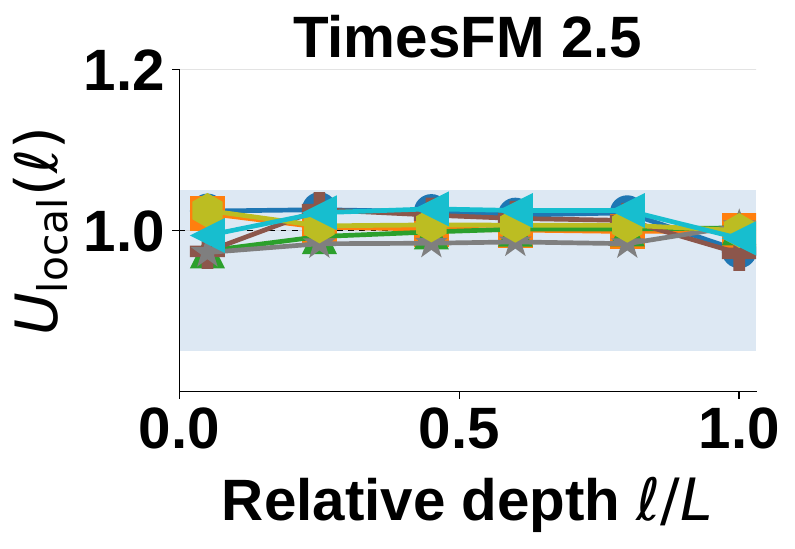}
\hfill
\includegraphics[width=0.325\linewidth]{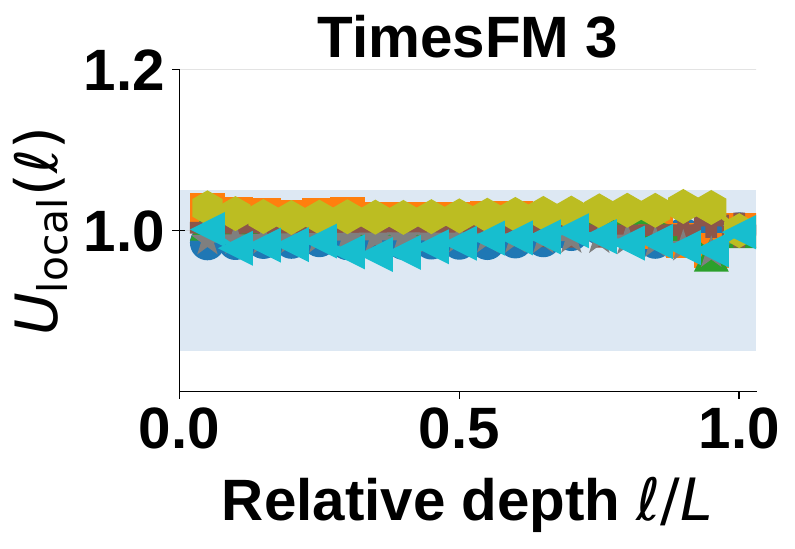}
\\[2pt]
\includegraphics[width=0.325\linewidth]{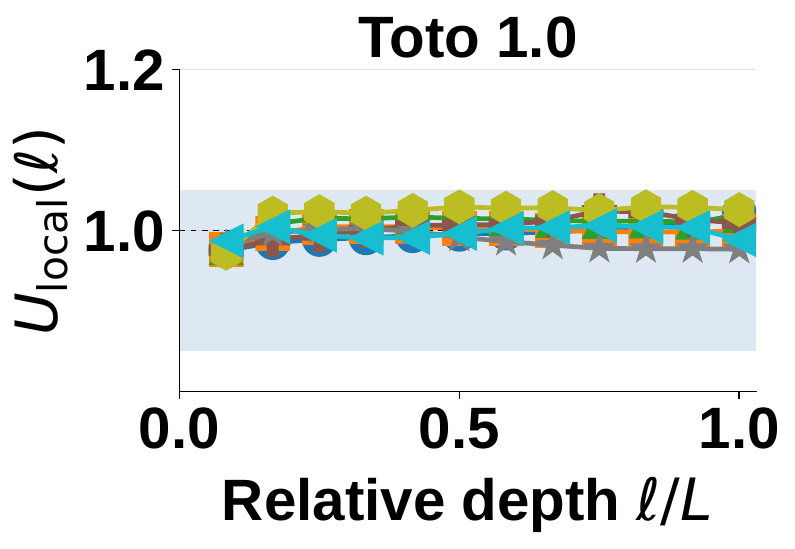}
\hfill
\includegraphics[width=0.325\linewidth]{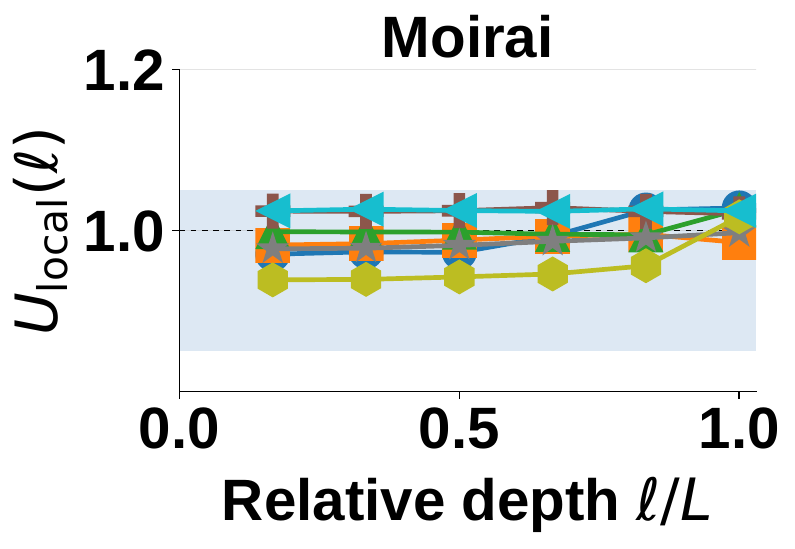}
\hfill
\includegraphics[width=0.325\linewidth]{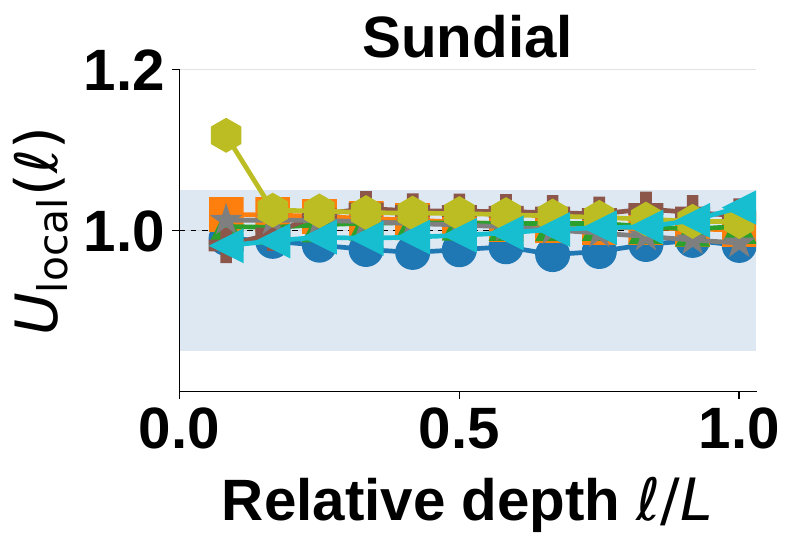}
\\[2pt]
\includegraphics[width=0.325\linewidth]{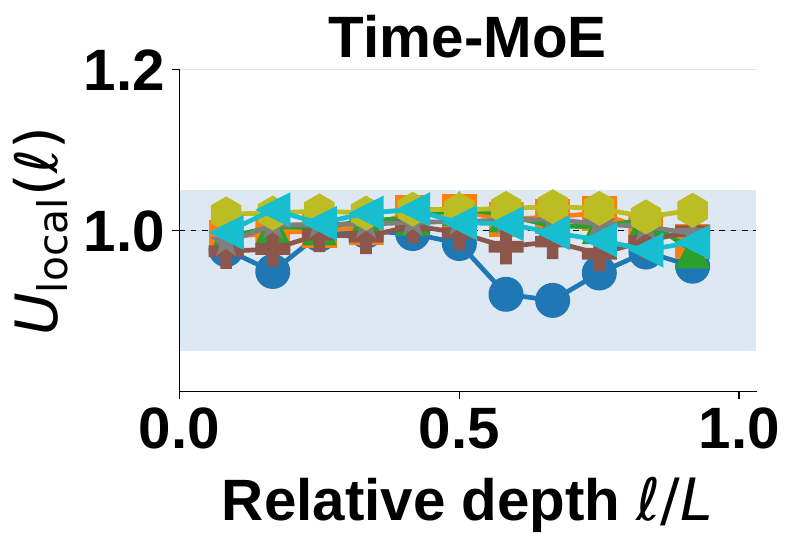}
\hfill
\includegraphics[width=0.325\linewidth]{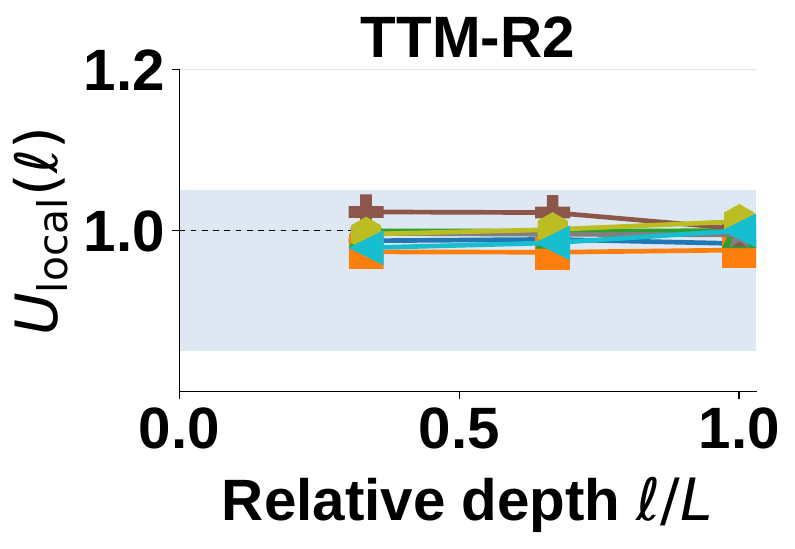}
\hfill
\includegraphics[width=0.325\linewidth]{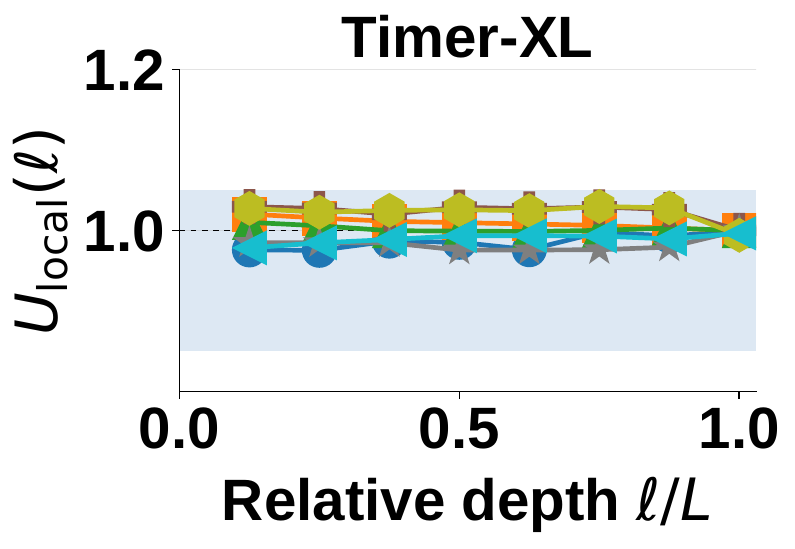}
\\[2pt]
\includegraphics[width=\linewidth]{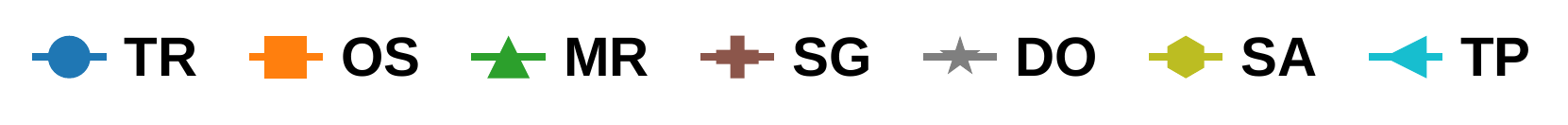}
\caption{
Validation local response across model depth for nine TSFMs and seven single-parameter laws.
The shaded band denotes the admissible interval~\([0.85,1.05]\).
}
\label{fig:local_layers}
\end{figure}

\begin{figure}[t]
\centering
\includegraphics[width=\linewidth]{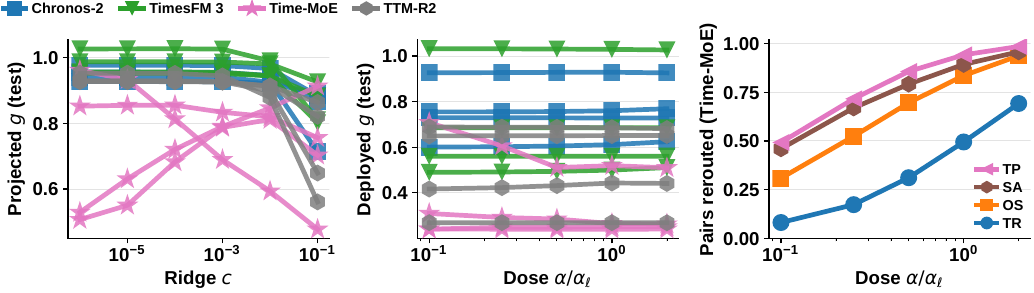}
\caption{
Numerical sensitivity of causal-geometry interventions.
The left and center panels vary the inversion ridge and relative hidden-state dose.
The right panel reports Time-MoE~(\texttt{ME}) expert rerouting for linear trend~(\texttt{TR}), oscillation~(\texttt{OS}), seasonal autoregression~(\texttt{SA}), and transient pulse~(\texttt{TP}) across the evaluated intervention strengths.
}
\label{fig:revision_numerical}
\end{figure}

\begin{table}[t]
\centering
\caption{
Time-MoE~(\texttt{ME}) linear-trend~(\texttt{TR}) local response and expert rerouting under live and frozen routing at the calibrated step~\(\alpha_\ell\) and one-tenth of that step.
}
\label{tab:timemoe_routing}
\small
\setlength{\tabcolsep}{4pt}
\renewcommand{\arraystretch}{1.08}
\begin{tabular}{@{}ccrrrrrr@{}}
\toprule
&
&
\multicolumn{2}{c}{Response at~\(\alpha_\ell\)}
&
\multicolumn{2}{c}{Response at~\(0.1\alpha_\ell\)}
&
\multicolumn{2}{c}{Rerouted Pairs} \\
\cmidrule(lr){3-4}
\cmidrule(lr){5-6}
\cmidrule(l){7-8}
Layer
& \(\alpha_\ell\)
& Live
& Frozen
& Live
& Frozen
& \(\alpha_\ell\)
& \(0.1\alpha_\ell\) \\
\midrule
1  & 0.0370 & 0.79 & 0.96 & 0.39 & 0.97 & 94\% & 37\% \\
2  & 0.0642 & 0.84 & 0.98 & 0.70 & 0.97 & 97\% & 47\% \\
3  & 0.0214 & 0.67 & 0.92 & 0.55 & 0.98 & 75\% & 19\% \\
4  & 0.0281 & 0.75 & 0.94 & 0.44 & 0.98 & 77\% & 20\% \\
5  & 0.0281 & 0.79 & 0.93 & 0.65 & 0.98 & 74\% & 18\% \\
6  & 0.0642 & 0.81 & 0.91 & 1.06 & 0.97 & 89\% & 29\% \\
7  & 0.0642 & 0.77 & 0.90 & 0.75 & 0.97 & 86\% & 27\% \\
8  & 0.0642 & 0.85 & 0.90 & 1.01 & 0.96 & 79\% & 20\% \\
9  & 0.0642 & 0.87 & 0.90 & 0.99 & 0.96 & 73\% & 18\% \\
10 & 0.0642 & 0.92 & 0.89 & 0.98 & 0.96 & 65\% & 13\% \\
11 & 0.0559 & 0.92 & 0.92 & 1.01 & 0.95 & 49\% & 8\% \\
\bottomrule
\end{tabular}
\end{table}

\subsubsection{Local Causal Directions and Numerical Robustness}
\label{app:geometry_local}
\label{app:revision_robust}

The Jacobian-derived local direction is constructed to produce the reference forecast change under a first-order approximation.
The nonlinear response~\(U_{\mathrm{local}}(\ell)\) tests whether the resulting edit remains effective in the full TSFM.
\autoref{fig:local_validity} reports selected-layer validity and the fraction of validation-admissible layers, while~\autoref{fig:local_layers} shows the complete depth profiles.

\textbf{Validation and Test Agree at Most Layers.}
Validation and test admissibility differ at~11 of~630 measured layers.
Seven discrepancies occur at Time-MoE~(\texttt{ME}) trend-slope layers~1--7.
The remaining four occur at early transient-pulse~(\texttt{TP}) layers whose test local responses lie between~\(1.05\) and~\(1.08\).
Using test rather than validation admissibility would not change any selected layer.

\textbf{Expert Routing Explains Much of the Time-MoE Trend Exception.}
\autoref{tab:timemoe_routing} reports the layer-wise comparison between live and frozen expert routing.
At the calibrated step, freezing Time-MoE~(\texttt{ME}) expert selections raises its linear-trend~(\texttt{TR}) local response from between~\(0.67\) and~\(0.92\) to between~\(0.89\) and~\(0.98\).
At one-tenth of the step, frozen-routing responses remain between~\(0.95\) and~\(0.98\).
The calibrated edit reroutes between~\(49\%\) and~\(97\%\) of pairs across the measured layers.

\textbf{Numerical Sensitivity Is Cell-Dependent.}
For~16 cells formed by Chronos-2~(\texttt{C2}), TimesFM~3~(\texttt{F3}), Time-MoE~(\texttt{ME}), and TTM-R2~(\texttt{TT}) crossed with linear trend~(\texttt{TR}), oscillation~(\texttt{OS}), seasonal autoregression~(\texttt{SA}), and transient pulse~(\texttt{TP}), we vary the inversion ridge over six values and the hidden-state step over five multiples of its calibrated value.
As shown in~\autoref{fig:revision_numerical}, projected and deployed responses remain stable in some cells but vary substantially in others.
Projected response ranges from~\(0.89\) to~\(0.93\) for Chronos-2~(\texttt{C2}) on linear trend~(\texttt{TR}) and from~\(0.48\) to~\(0.96\) for Time-MoE~(\texttt{ME}) on oscillation~(\texttt{OS}).
Larger Time-MoE~(\texttt{ME}) interventions also reroute more pairs, limiting extrapolation beyond the calibrated local step.
These sweeps are diagnostic, and the primary results retain the validation-selected settings used throughout all reported causal-geometry comparisons.

In summary, the selected layers therefore support calibrated edits that produce the reference response, while routing boundaries account for the principal local-validity exception.

\begin{figure}[t]
\centering
\includegraphics[width=0.325\linewidth]{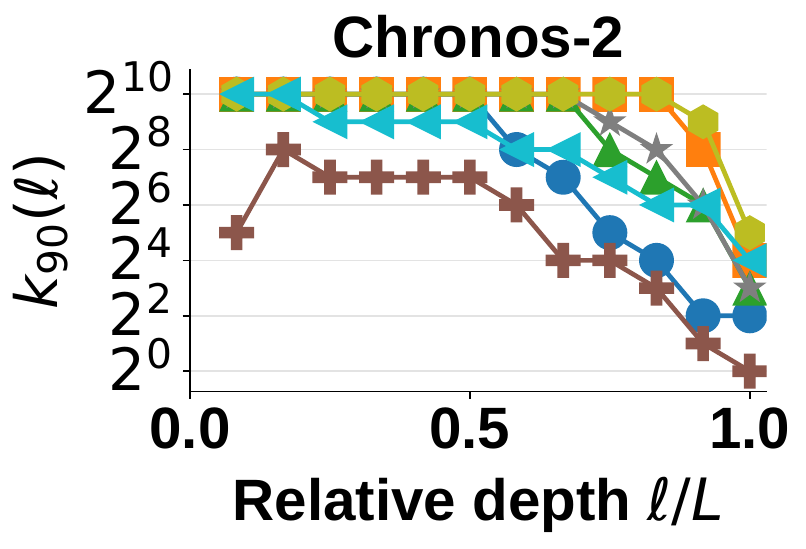}
\hfill
\includegraphics[width=0.325\linewidth]{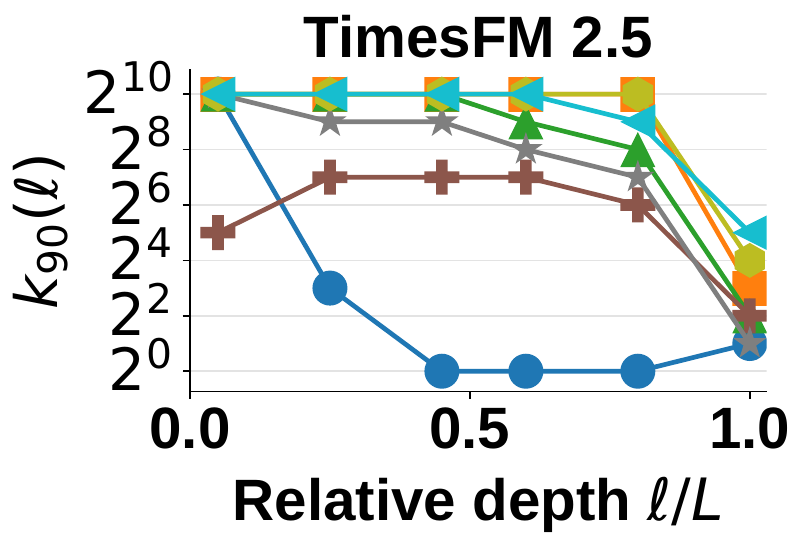}
\hfill
\includegraphics[width=0.325\linewidth]{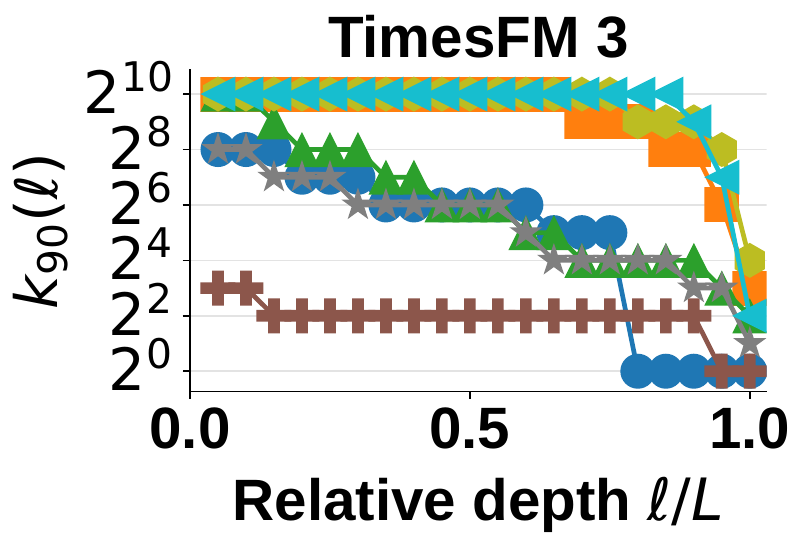}
\\[2pt]
\includegraphics[width=0.325\linewidth]{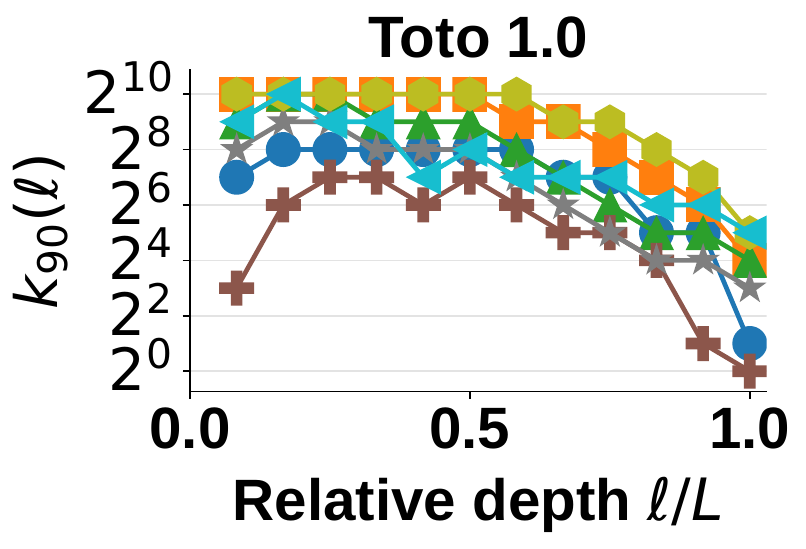}
\hfill
\includegraphics[width=0.325\linewidth]{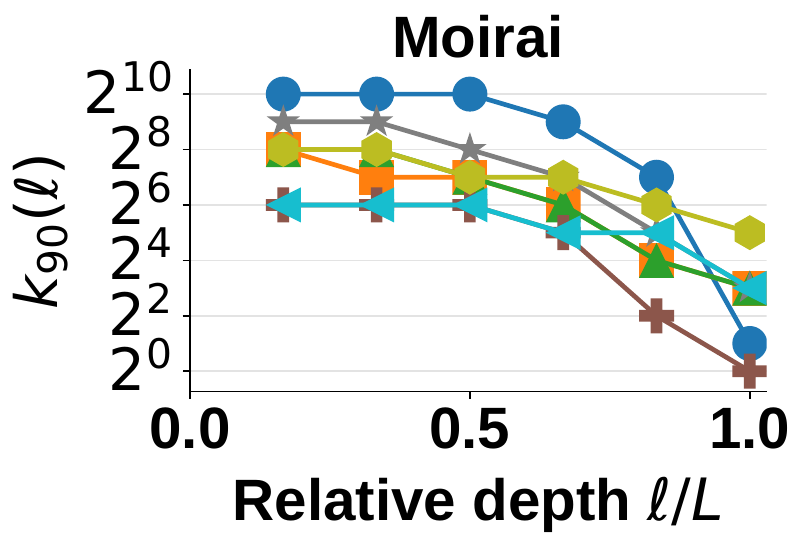}
\hfill
\includegraphics[width=0.325\linewidth]{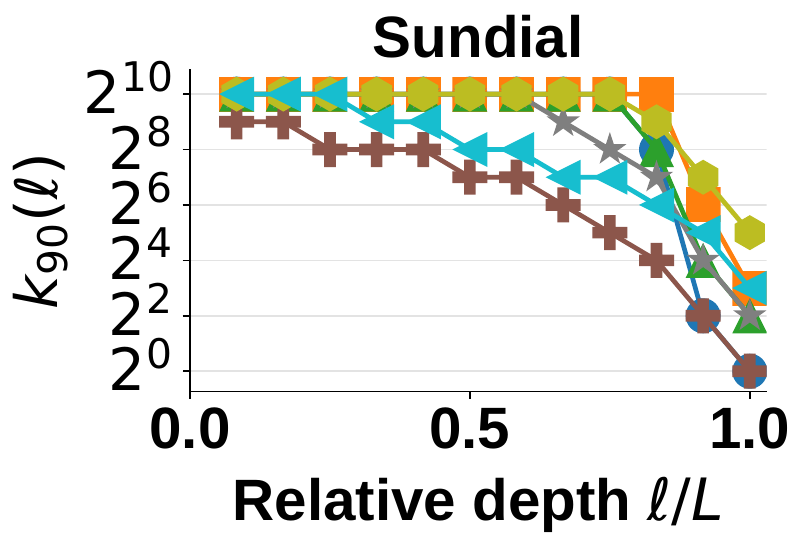}
\\[2pt]
\includegraphics[width=0.325\linewidth]{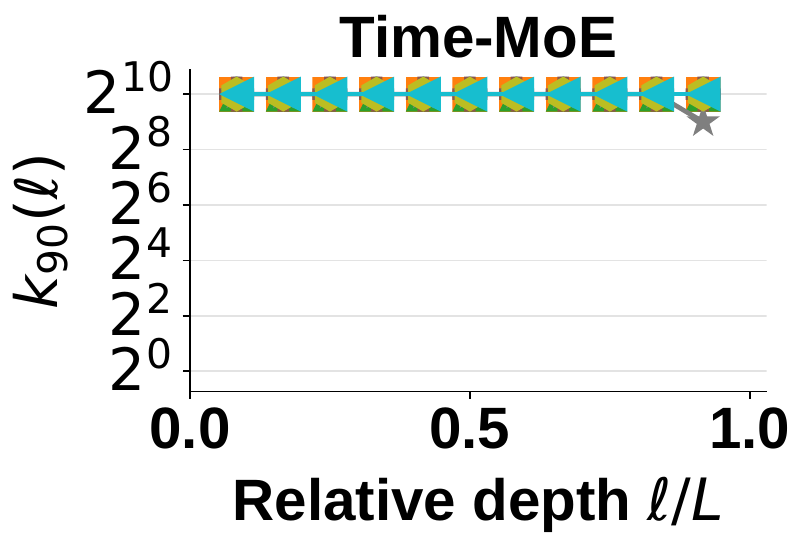}
\hfill
\includegraphics[width=0.325\linewidth]{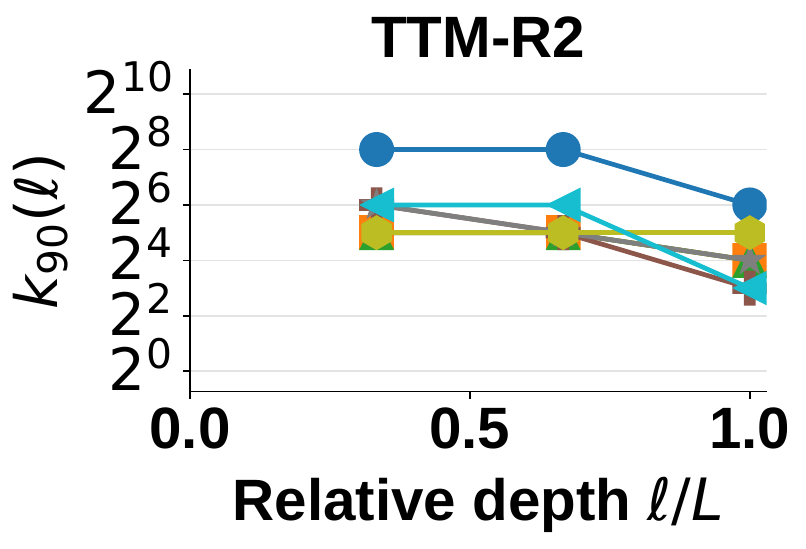}
\hfill
\includegraphics[width=0.325\linewidth]{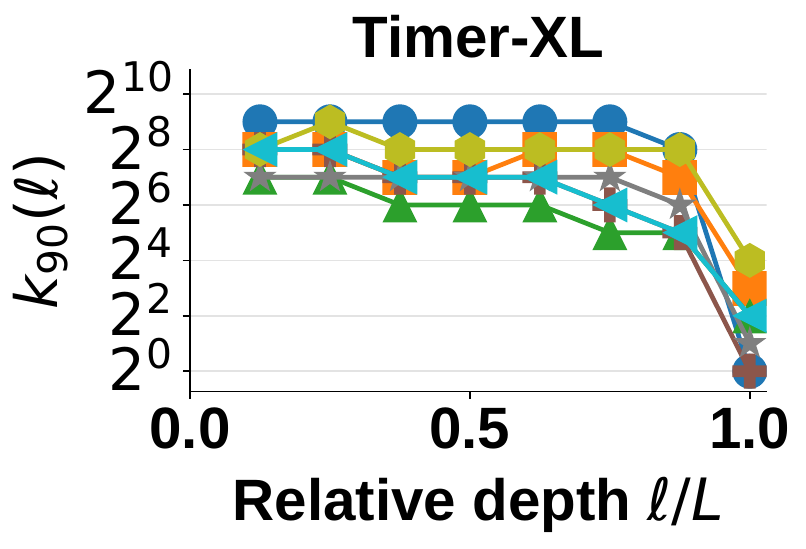}
\\[2pt]
\includegraphics[width=\linewidth]{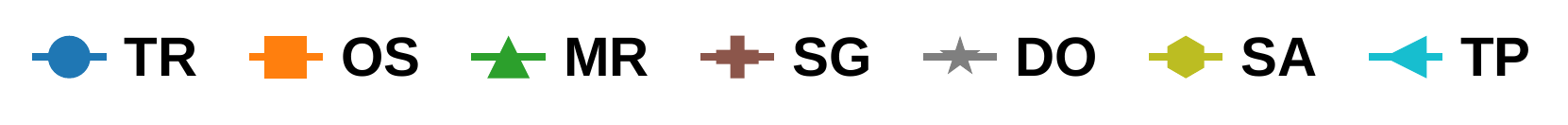}
\caption{
Shared causal subspace rank~\(k_{90}\) across relative model depth for nine TSFMs and seven single-parameter laws.
Unreached~\(90\%\) response-retention thresholds are shown as lower bounds.
}
\label{fig:jspace_compact}
\end{figure}

\begin{figure}[t]
\centering
\includegraphics[width=\linewidth]{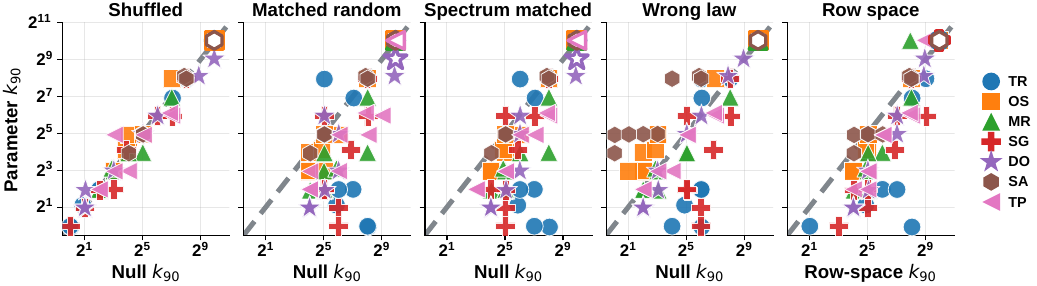}
\caption{
Required subspace dimension~\(k_{90}\) for parameter targets relative to shuffled, matched-random, spectrum-matched, wrong-law, and generic row-space controls.
Hollow markers indicate unreached~\(90\%\) response-retention thresholds and corresponding lower bounds.
}
\label{fig:revision_nulls}
\end{figure}

\begin{figure}[t]
\centering
\includegraphics[width=\linewidth]{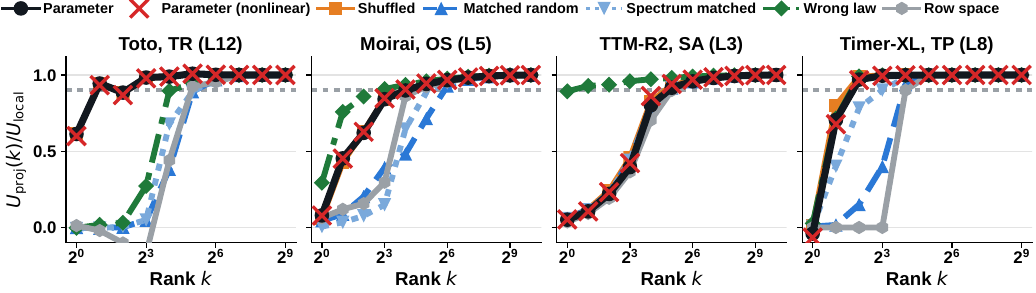}
\caption{
Linearized response retention across subspace dimension~\(k\) for parameter targets, control targets, and the generic Jacobian row-space basis in four representative model--parameter cells.
}
\label{fig:revision_rank_curves}
\end{figure}

\begin{figure}[t]
\centering
\includegraphics[width=\linewidth]{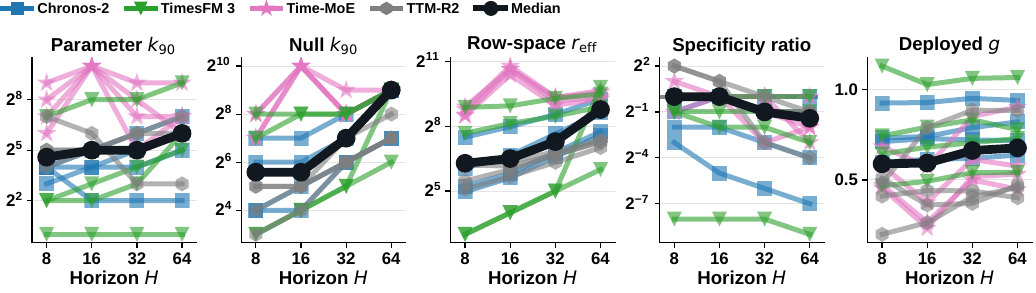}
\caption{
Parameter-target and matched-control subspace dimensions, specificity, and deployed gain across four forecast horizons.
}
\label{fig:revision_horizon}
\end{figure}

\subsubsection{Shared Causal Subspace and Target Controls}
\label{app:geometry_space}
\label{app:revision_geometry}

This subsection examines how compact shared causal subspaces develop across model depth, whether their compactness is specific to the correct parameter target, and how both properties vary with forecast horizon.
\autoref{fig:jspace_compact} presents the depth profiles, \autoref{fig:revision_nulls} compares parameter targets with controls, \autoref{fig:revision_rank_curves} shows representative response-retention curves, and \autoref{fig:revision_horizon} reports horizon sensitivity across all evaluated models and laws.

For each admissible layer, the shared basis is estimated from the fitting dataset.
The required subspace dimension~\(k_{90}(\ell)\) is the smallest evaluated rank satisfying
\[
\frac{U_{\mathrm{projected}}(\ell,k)}
     {U_{\mathrm{local}}(\ell)}
\geq
0.9.
\]
Because intervened-state dimensions differ across models, panel-wide comparisons also report~\(k_{90}(\ell)/d_h\).
\autoref{fig:jspace_compact} traces the required dimension or corresponding lower bound across model depth for every evaluated law and model.

\textbf{Required Subspace Dimensions Usually Contract with Depth.}
Across the~63 model--parameter cells, endpoint comparisons give~56 decreases and seven unchanged outcomes.
Seven models contract on all seven laws, while TTM-R2~(\texttt{TT}) contracts on six.
Time-MoE~(\texttt{ME}) is the main exception: only damped oscillation~(\texttt{DO}) contracts, and no evaluated dimension reaches the threshold for linear trend~(\texttt{TR}), oscillation~(\texttt{OS}), or seasonal autoregression~(\texttt{SA}).
At the selected layers, the median is~\(k_{90}=16\), corresponding to~\(0.26\%\) of the intervened-state dimension.

\textbf{Compactness Is Not Consistently Parameter Specific.}
The target-control experiment holds the selected layer, Jacobian, datasets, regularization, edit cap, and evaluated dimensions fixed while changing the forecast-space target.
Controls include shuffled input--target pairings, norm-matched random targets, spectrum-matched targets, wrong-law targets, and a generic basis constructed from pooled Jacobian row spaces.
\autoref{fig:revision_nulls} compares the required dimensions across all target-control cells, while~\autoref{fig:revision_rank_curves} shows the complete response-retention curves for four representative cells.
The correct parameter target improves both required dimension and rank-curve area over matched-random, spectrum-matched, and generic row-space controls in~40, 41, and~44 of~63 cells.
It improves both measures over the shuffled target in only six cells and ties its~\(k_{90}\) in~51 cells.
The compact dimensions therefore reflect substantial generic shared causal-subspace structure beyond the correct input--response pairing across most evaluated models and laws.

\textbf{Longer Horizons Increase Dimension but Can Sharpen the Control Comparison.}
\autoref{fig:revision_horizon} summarizes the parameter-target and matched-control dimensions, specificity, and deployed gain at horizons~\(8\), \(16\), \(32\), and~\(64\).
The median parameter and matched-random dimensions are~\(24/48\), \(32/48\), \(32/128\), and~\(64/512\), respectively.
The parameter target has lower required dimension than its matched-random control in~\(44\%\), \(31\%\), \(69\%\), and~\(81\%\) of runs.
Median deployed gain rises from~\(0.59\) to~\(0.68\), while centered error falls from~\(0.70\) to~\(0.55\).
Both the absolute subspace dimension and the parameter--control comparison therefore depend on forecast horizon.

\subsubsection{Input-Induced Hidden-State Changes}
\label{app:revision_natural}

\begin{figure}[t]
\centering
\includegraphics[width=\linewidth]{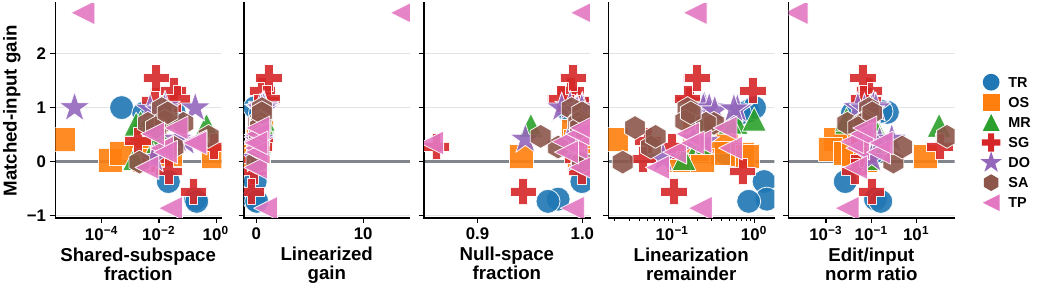}
\caption{
Matched-input aligned gain relative to geometric diagnostics of input-induced hidden-state changes across~63 model--parameter cells.
The panels report shared-subspace fraction, linearized gain, Jacobian-null-space fraction, nonlinear remainder, and intervention size.
}
\label{fig:revision_natural}
\end{figure}

A successful reference-derived edit does not establish that changing the input follows the same hidden-state route.
At each selected layer, we form
\[
\Delta h_i
=
h^\ell(x_i^B)-h^\ell(x_i^A)
\]
on~4{,}096 test matched pairs.
We measure its cosine with the required local edit, its energy in the shared \(k_{90}\)-dimensional basis, and its row- and null-space energy under the baseline Jacobian~\(J_i\).
The gain of~\(J_i\Delta h_i\) measures the forecast change predicted by the local linearization.
\autoref{fig:revision_natural} relates matched-input gain to the shared-subspace fraction, linearized gain, null-space fraction, nonlinear remainder, and intervention size for every evaluated model and parameter.

\textbf{Input Changes Mostly Follow Forecast-Insensitive Directions.}
Input-induced changes place median energy of~\(99.6\%\) in the Jacobian null space and only~\(1.1\%\) in the learned shared subspace.
Their median linearized aligned gain is~\(0.28\).
Among the~32 cells with matched-input gain below~\(0.5\), 31 have either linearized gain below~\(0.5\) or nonlinear remainder above~\(0.5\).
The accessibility--response gap is therefore accompanied by a mismatch between input-induced motion and forecast-sensitive motion across all seven scored single-parameter laws.

\begin{figure}[t]
\centering
\includegraphics[width=0.325\linewidth]{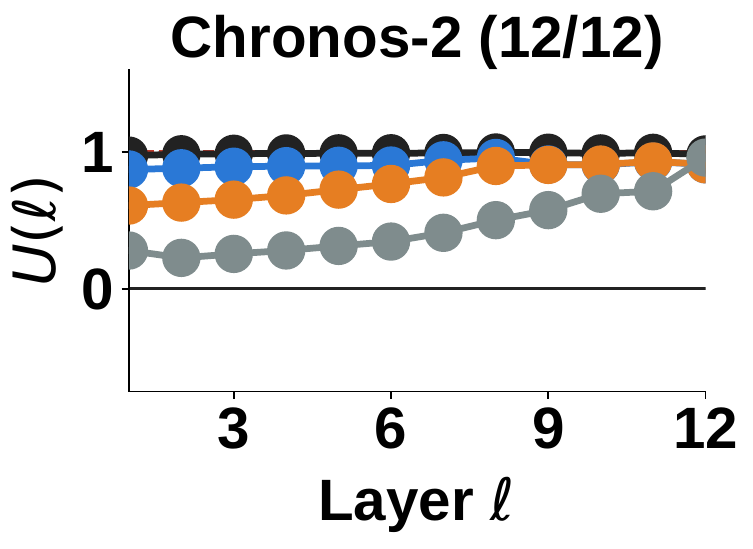}
\hfill
\includegraphics[width=0.325\linewidth]{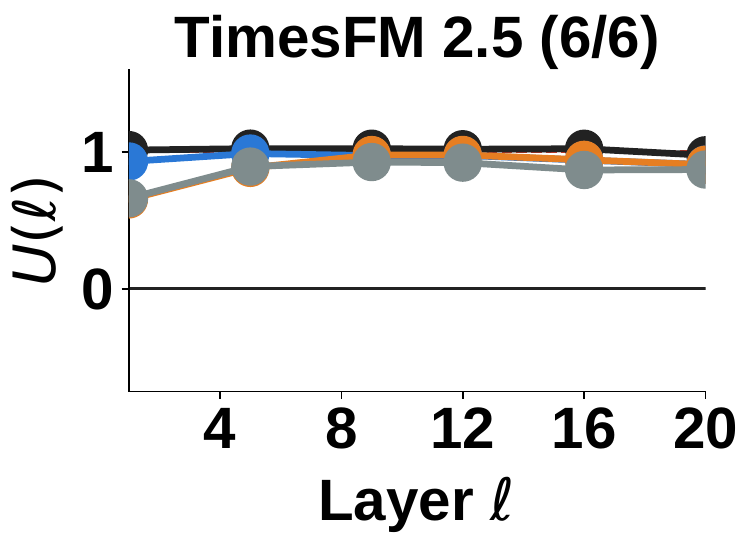}
\hfill
\includegraphics[width=0.325\linewidth]{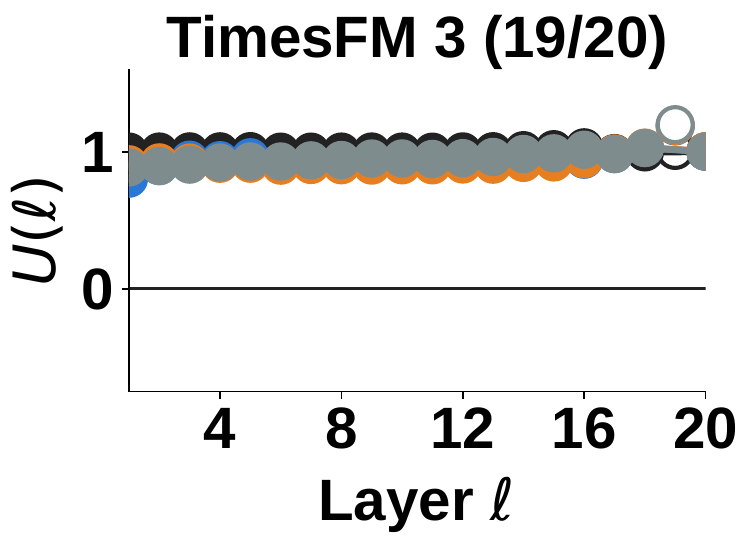}
\\[2pt]
\includegraphics[width=0.325\linewidth]{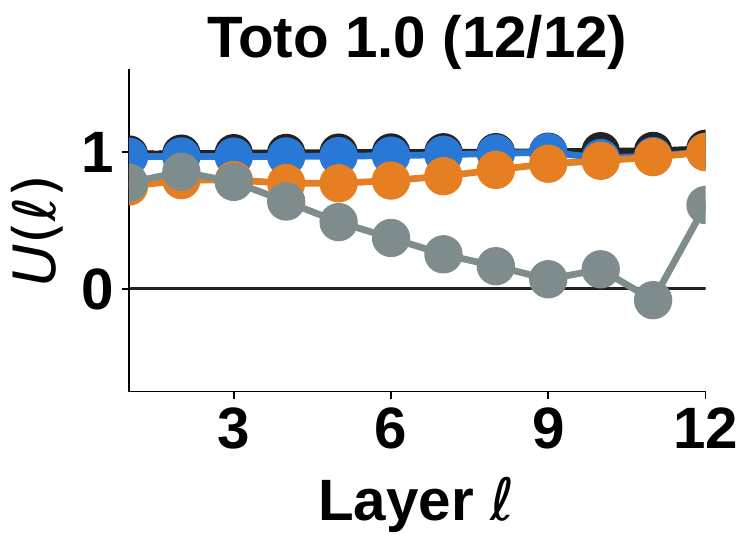}
\hfill
\includegraphics[width=0.325\linewidth]{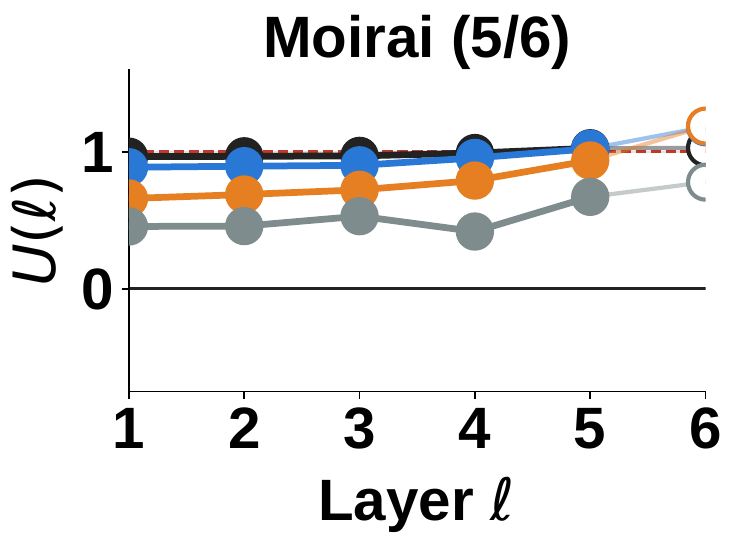}
\hfill
\includegraphics[width=0.325\linewidth]{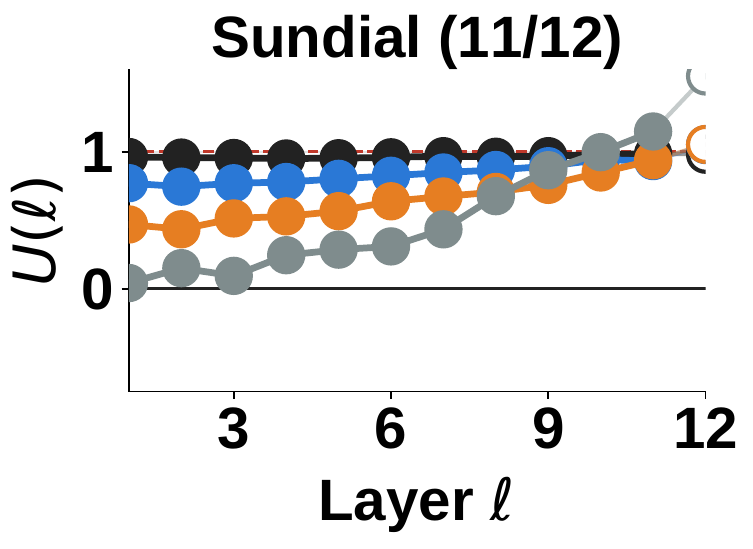}
\\[2pt]
\includegraphics[width=0.325\linewidth]{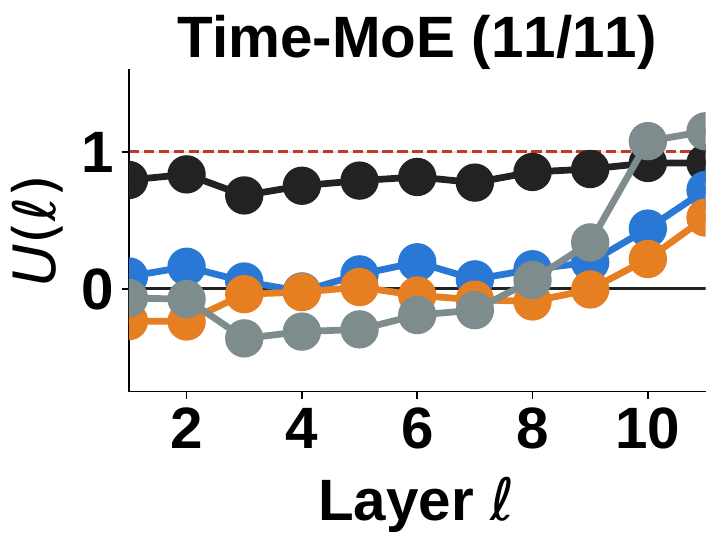}
\hfill
\includegraphics[width=0.325\linewidth]{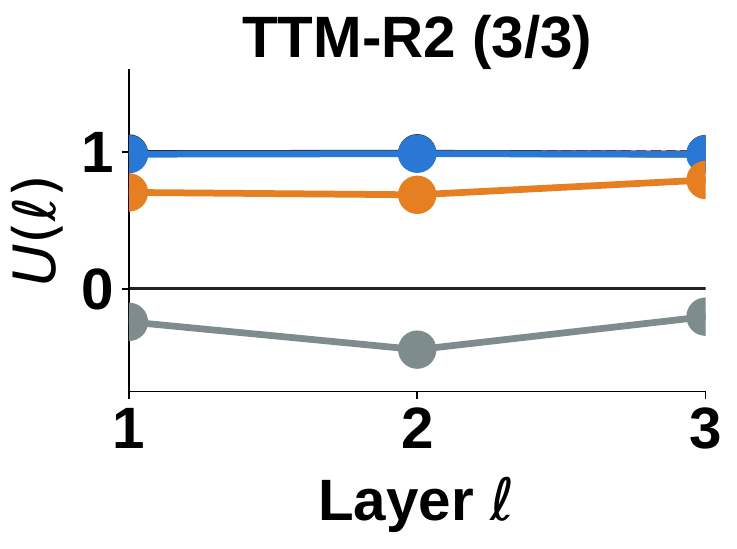}
\hfill
\includegraphics[width=0.325\linewidth]{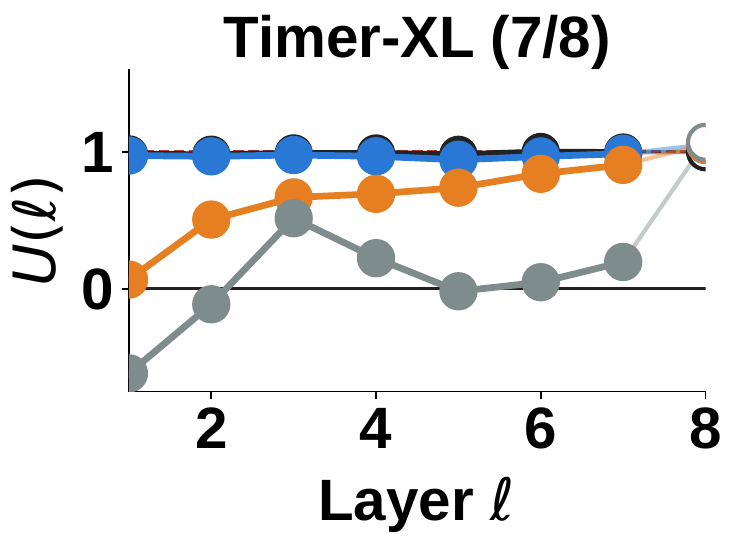}
\\[2pt]
\includegraphics[width=\linewidth]{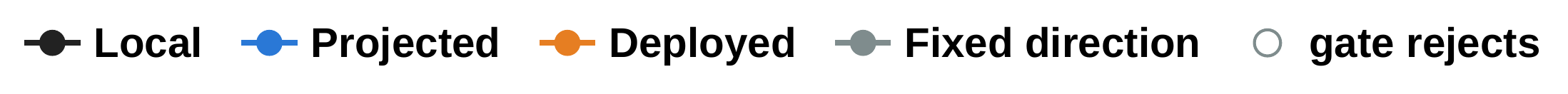}
\caption{
Local, projected, deployed, and oracle-assisted fixed-direction responses for linear trend~(\texttt{TR}) across model depth in nine TSFMs.
Hollow markers denote validation-rejected layers, and parentheses report admissible and evaluated layer counts.
}
\label{fig:jspace_depth}
\end{figure}

\begin{figure}[t]
\centering
\includegraphics[width=0.325\linewidth]{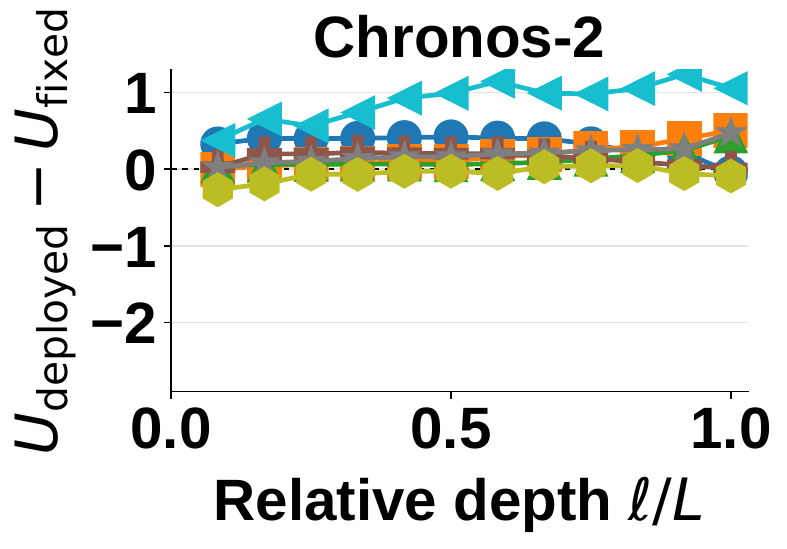}
\hfill
\includegraphics[width=0.325\linewidth]{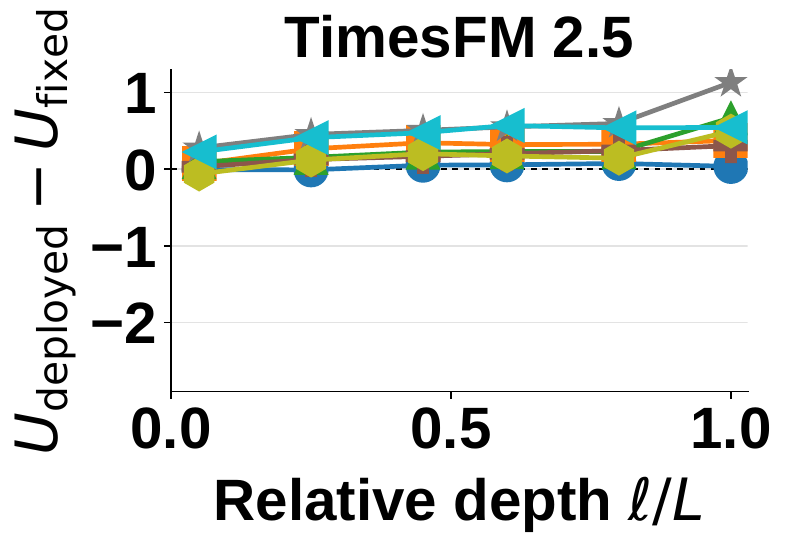}
\hfill
\includegraphics[width=0.325\linewidth]{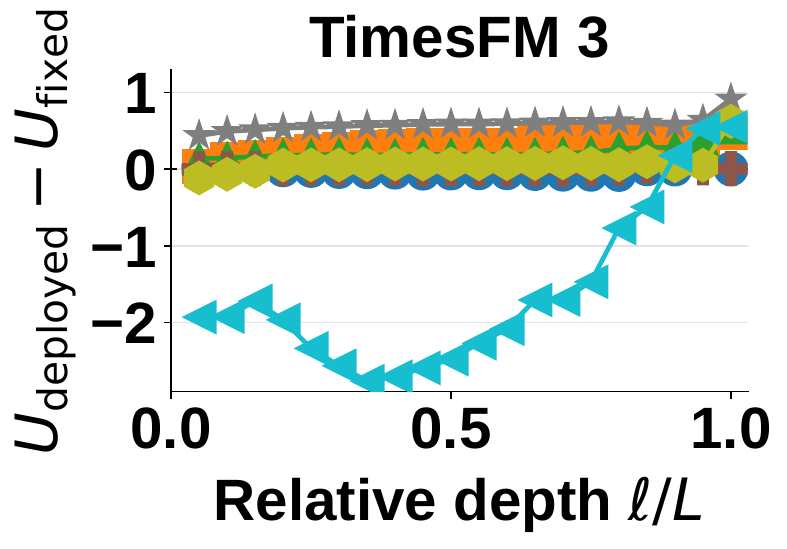}
\\[2pt]
\includegraphics[width=0.325\linewidth]{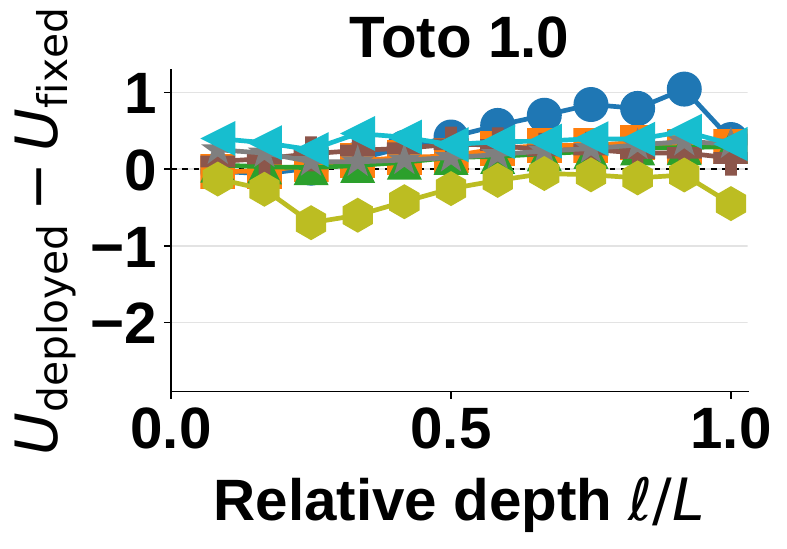}
\hfill
\includegraphics[width=0.325\linewidth]{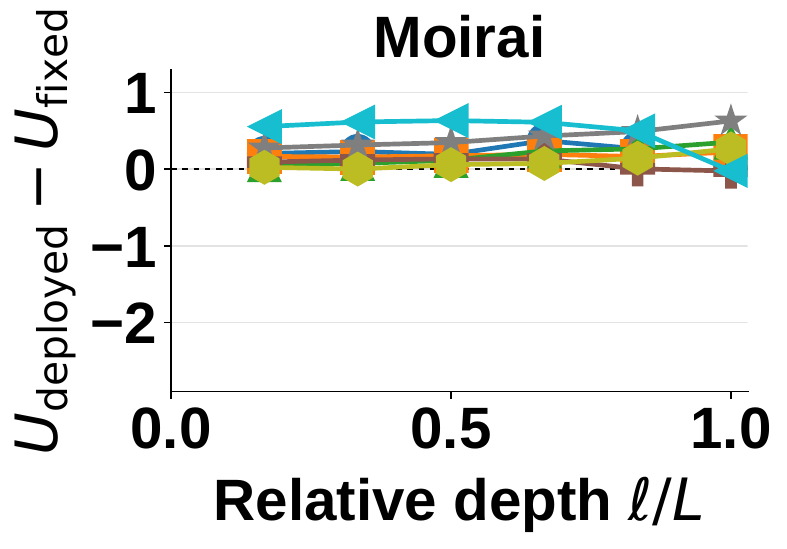}
\hfill
\includegraphics[width=0.325\linewidth]{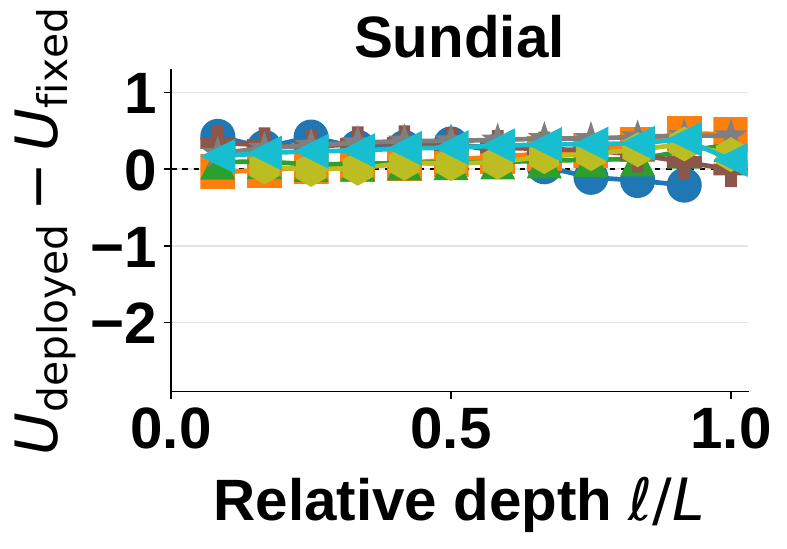}
\\[2pt]
\includegraphics[width=0.325\linewidth]{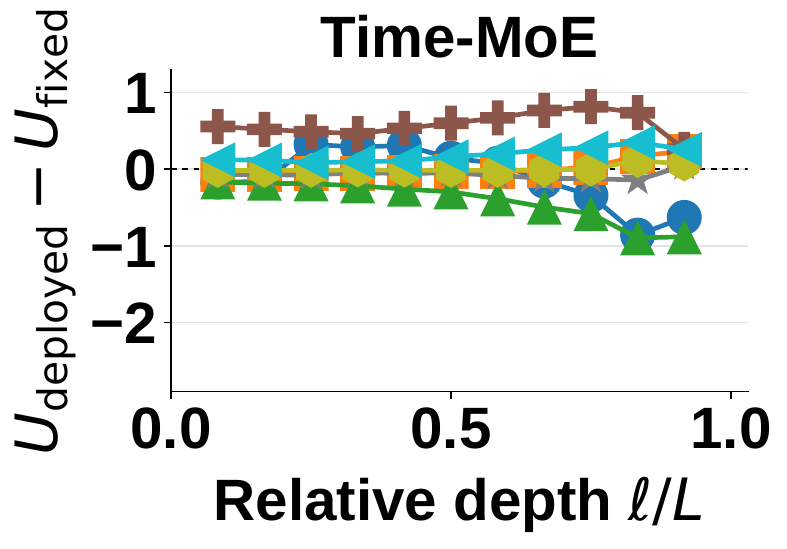}
\hfill
\includegraphics[width=0.325\linewidth]{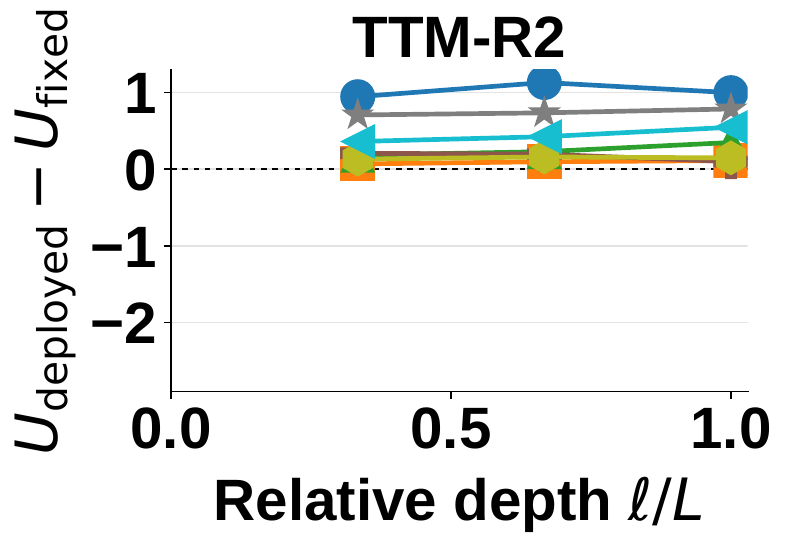}
\hfill
\includegraphics[width=0.325\linewidth]{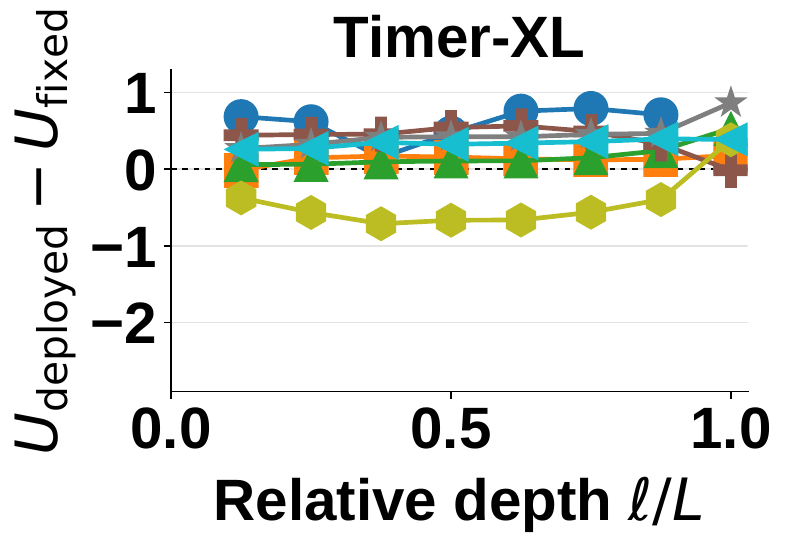}
\\[2pt]
\includegraphics[width=\linewidth]{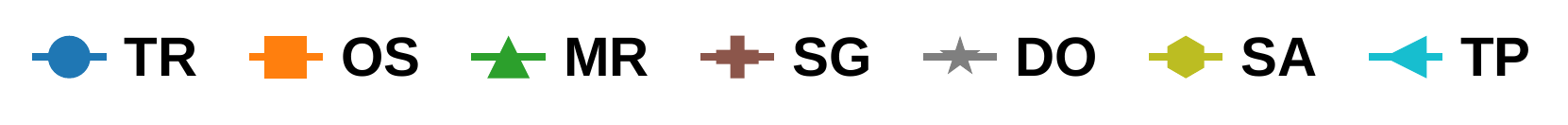}
\caption{
Difference between deployed and oracle-assisted fixed-direction responses across validation-admissible layers for nine TSFMs and seven single-parameter laws.
Positive values indicate an advantage for the input-specific coordinate readout.
}
\label{fig:jspace_gain}
\end{figure}

\subsubsection{Causal-Coordinate Readout and Transfer}
\label{app:geometry_access}
\label{app:revision_transfer}

\begin{figure}[t]
\centering

\subfigure[Aligned gain~\(g\).]{
\includegraphics[width=0.31\linewidth]
{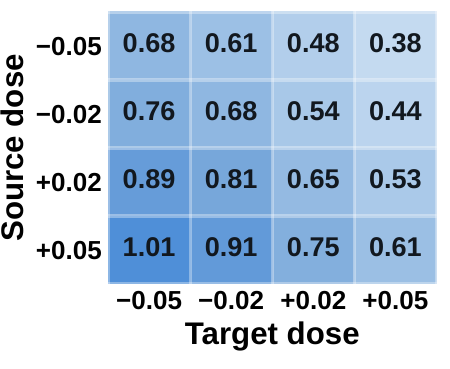}}
\hfill
\subfigure[Cosine similarity~\(s\).]{
\includegraphics[width=0.31\linewidth]
{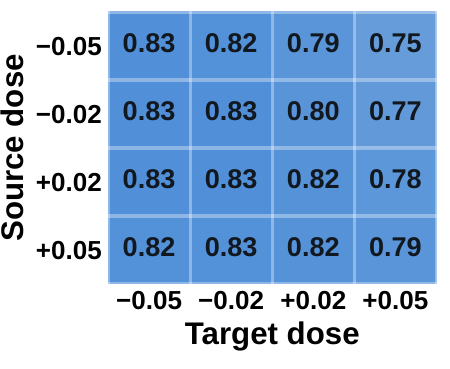}}
\hfill
\subfigure[Centered error~\(e\).]{
\includegraphics[width=0.31\linewidth]
{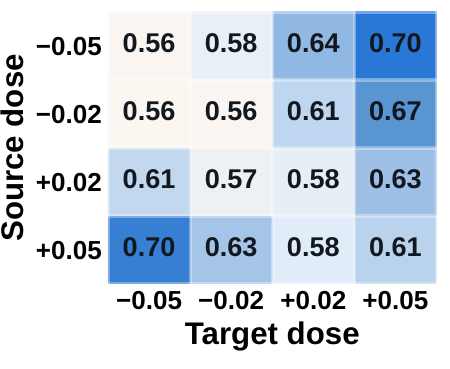}}
\par\vspace{-8pt}\noindent
\subfigure[Gain over the fixed control.]{
\includegraphics[width=0.31\linewidth]
{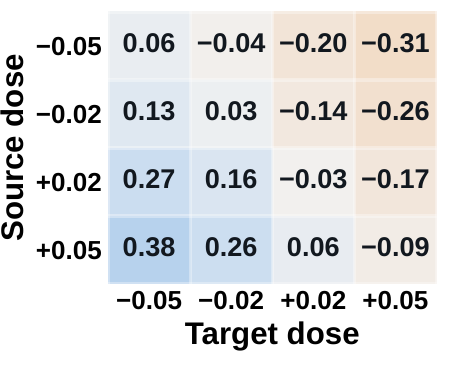}}
\hfill
\subfigure[Transfer across strata.]{
\includegraphics[width=0.31\linewidth]
{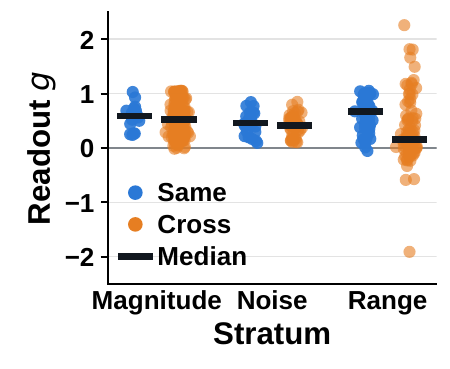}}
\hfill
\begin{minipage}{0.31\linewidth}
\mbox{}
\end{minipage}
\caption{
Causal-coordinate transfer across intervention magnitude, noise, and parameter-range shifts without target-regime refitting.
The top row reports aligned gain, cosine similarity, and centered error for Chronos-2~(\texttt{C2}) on seasonal autoregression~(\texttt{SA}).
The second row reports improvement over the oracle-assisted fixed-direction control and readout gain across all evaluated transfer cells.
}
\label{fig:revision_transfer}
\vspace{-0.1in}
\end{figure}

The deployed intervention predicts input-specific coordinates in the shared causal basis from the unperturbed hidden state.
The oracle-assisted fixed-direction control instead applies the mean local direction to every input while receiving each example's local-direction norm.
This subsection examines behavior across model depth, improvement over the fixed-direction control, and transfer across distribution shifts.
The depth, fixed-control, and transfer results appear in~\autoref{fig:jspace_depth}, \autoref{fig:jspace_gain}, and~\autoref{fig:revision_transfer}, respectively, for all evaluated model--parameter settings.

\textbf{Deployed Trend Response Generally Improves with Depth.}
\autoref{fig:jspace_depth} compares local, projected, deployed, and fixed-direction responses for linear trend~(\texttt{TR}) across model depth.
For Chronos-2~(\texttt{C2}), Toto-Open-Base~1.0~(\texttt{TO}), Moirai~(\texttt{MO}), Sundial~(\texttt{SU}), and Timer-XL~(\texttt{TX}), deployed response rises substantially across admissible layers and reaches between~\(0.90\) and~\(1.00\).
TimesFM~2.5~(\texttt{F2}) and TimesFM~3~(\texttt{F3}) remain near~\(0.9\) or above at later layers, while TTM-R2~(\texttt{TT}) remains between~\(0.69\) and~\(0.79\).
Time-MoE~(\texttt{ME}) remains near zero through layer~9 and reaches~\(0.52\) at layer~11.

\textbf{Input-Specific Coordinates Usually Improve upon a Fixed Direction.}
As shown in~\autoref{fig:jspace_gain}, the deployed response exceeds the fixed-direction response in~56 of~63 cells at the selected layer and subspace dimension, with a median difference of~\(0.26\).
Across all~625 validation-admissible layers, the median difference is~\(0.18\), and the deployed response remains higher at the final admissible layer in~53 cells.
The improvement is most consistent for mean reversion~(\texttt{MR}) and damped oscillation~(\texttt{DO}) and least consistent for seasonal autoregression~(\texttt{SA}).

\textbf{The Coordinate Readout Transfers Partially.}
\autoref{fig:revision_transfer} evaluates transfer across intervention magnitude, noise, and parameter-range shifts without target-regime refitting.
Panels~(a)--(d) show Chronos-2~(\texttt{C2}) on seasonal autoregression~(\texttt{SA}) across source and target intervention magnitudes.
Panel~(a) shows that aligned gain varies from~\(0.38\) to~\(1.01\), with stronger transfer when the target remains near the source intervention regime.
Panel~(b) shows comparatively stable cosine similarity between~\(0.75\) and~\(0.83\), indicating that response direction transfers more consistently than response magnitude.
Panel~(c) reports centered error between~\(0.56\) and~\(0.70\), demonstrating that substantial full-vector error remains even when directional alignment is preserved.
Panel~(d) compares the deployed readout with the oracle-assisted fixed-direction control.
The difference ranges from~\(-0.31\) to~\(0.38\), so the input-specific coordinates help most for nearby or same-sign target interventions but do not dominate uniformly across all evaluated intervention combinations.

Panel~(e) aggregates aligned gain across all~16 evaluated model--parameter cells.
Median deployed gain is~\(0.59\) in regime, \(0.51\) across intervention magnitude, \(0.42\) across noise strata, and~\(0.16\) across parameter-range strata.
The corresponding fixed-direction gains are~\(0.20\), \(0.09\), \(0.11\), and~\(0.07\).
A direct hidden-to-output map reaches~\(0.45\) in regime and~\(0.52\) across magnitude, while a probe-derived direction reaches~\(0.03\) in regime.
The coordinate readout therefore retains an advantage over the fixed-direction control under the tested shifts, although its magnitude fidelity and absolute response decrease outside the source parameter range for the most challenging transfers.

Overall, the unperturbed hidden state predicts useful input-specific edit coordinates that generally outperform the oracle-assisted fixed-direction control.
However, the predicted edits remain incomplete and degrade under distribution shifts, particularly outside the source parameter range.

\begin{table}[t]
\centering
\caption{
Median pooled response-fidelity metrics across~63 model--parameter cells.
Hidden-state interventions use calibrated steps and capped targets.
}
\label{tab:revision_fidelity}
\fontsize{9}{10.5}\selectfont
\setlength{\tabcolsep}{2.8pt}
\renewcommand{\arraystretch}{1.08}
\begin{tabular}{@{}lrrrrrrr@{}}
\toprule
& \multicolumn{5}{c}{Centered fidelity}
& \multicolumn{2}{c}{Raw fidelity} \\
\cmidrule(lr){2-6}
\cmidrule(l){7-8}
Condition
& \(g\)
& \(s\)
& \(e\)
& \(e_\perp\)
& \(e_{\mathrm{level}}\)
& \(g^{\mathrm{raw}}\)
& \(e^{\mathrm{raw}}\) \\
\midrule
Matched input
& 0.42 & 0.54 & 1.06 & 0.55 & 1.70 & 0.84 & 0.54 \\
Local
& 1.00 & 1.00 & 0.06 & 0.04 & 0.09 & 1.00 & 0.03 \\
Projected
& 0.93 & 0.99 & 0.18 & 0.12 & 0.22 & 0.96 & 0.10 \\
Deployed
& 0.58 & 0.75 & 0.68 & 0.29 & 0.77 & 0.77 & 0.48 \\
Fixed direction
& 0.23 & 0.34 & 1.06 & 0.54 & 1.54 & 0.52 & 0.93 \\
Conditional reference
& 0.97 & 0.99 & 0.14 & 0.10 & 0.84 & 0.97 & 0.14 \\
\bottomrule
\end{tabular}
\end{table}

\begin{figure}[t]
\centering
\includegraphics[width=\linewidth]{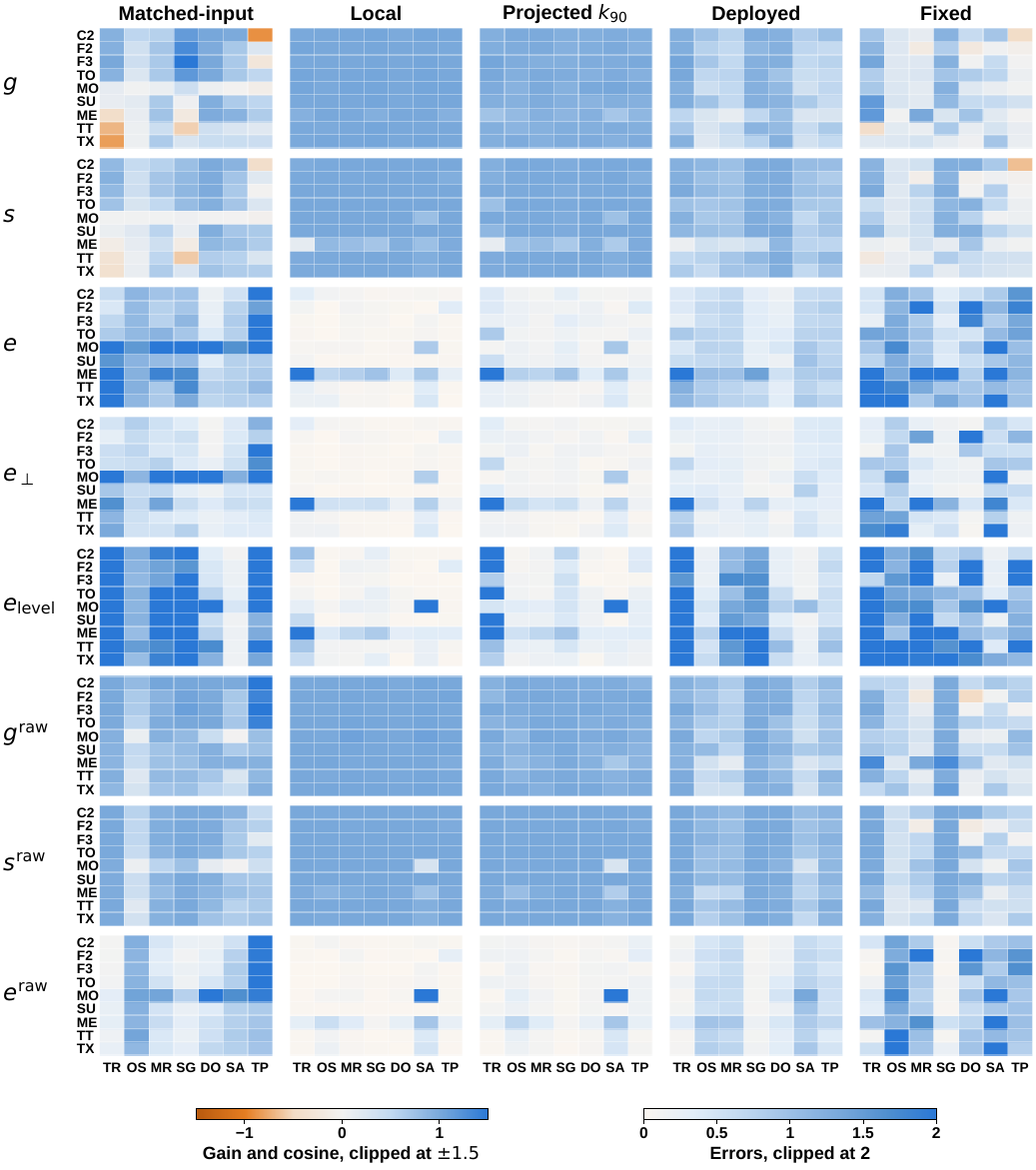}
\caption{
Centered and raw response-fidelity metrics across the single-parameter panel.
Columns compare matched-input, local, projected, deployed, and fixed-direction responses.
Centered metrics capture response alignment and trajectory error, while raw metrics use uncentered forecast changes.
}
\label{fig:revision_rf_heatmaps}
\end{figure}

\begin{figure}[t]
\centering
\includegraphics[width=\linewidth]{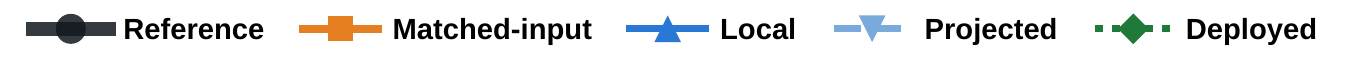}
\begin{minipage}[t]{0.32\linewidth}
\centering
\includegraphics[width=\linewidth]{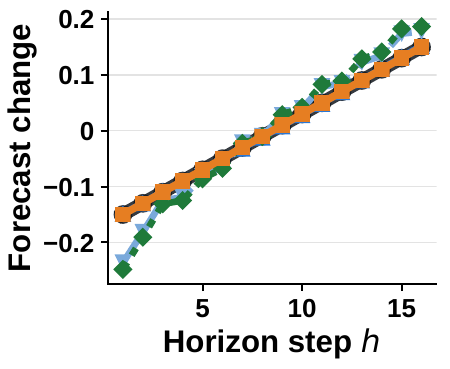}
\\[-0.2em]
{\small (a) Coherent: TimesFM 3 (\texttt{F3})}
\\[-0.1em]
{\scriptsize $g=1.00$, $s=1.00$, $e=0.00$, $e_{\perp}=0.00$, $e_{\mathrm{level}}=0.00$}
\end{minipage}
\hfill
\begin{minipage}[t]{0.32\linewidth}
\centering
\includegraphics[width=\linewidth]{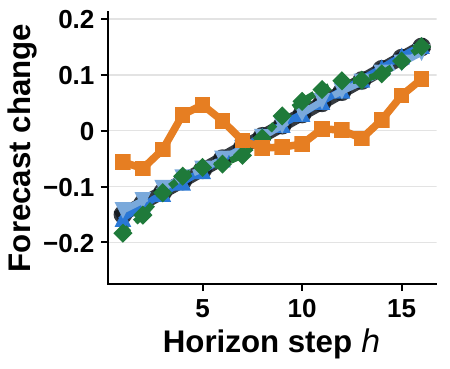}
\\[-0.2em]
{\small (b) Weak: Moirai (\texttt{MO})}
\\[-0.1em]
{\scriptsize $g=0.27$, $s=0.60$, $e=0.81$, $e_{\perp}=0.36$, $e_{\mathrm{level}}=0.48$}
\end{minipage}
\hfill
\begin{minipage}[t]{0.32\linewidth}
\centering
\includegraphics[width=\linewidth]{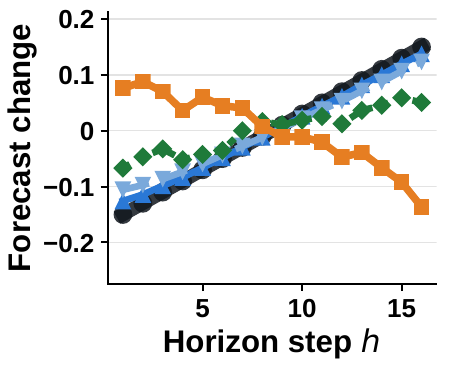}
\\[-0.2em]
{\small (c) Reversed: TTM-R2 (\texttt{TT})}
\\[-0.1em]
{\scriptsize $g=-0.66$, $s=-0.97$, $e=1.67$, $e_{\perp}=0.17$, $e_{\mathrm{level}}=0.01$}
\end{minipage}
\vspace{0.05in}
\caption{
Matched linear-trend (\texttt{TR}) response examples illustrating coherent, weak, and reversed matched-input behavior across TimesFM 3 (\texttt{F3}), Moirai (\texttt{MO}), and TTM-R2 (\texttt{TT}).
Each panel compares the reference forecast response with the matched-input, local, projected, and deployed responses across representative model-specific hidden-state response patterns.
}
\label{fig:revision_rf_example}
\vspace{-0.2in}
\end{figure}

\begin{figure}[t]
\centering

\includegraphics[width=\linewidth]{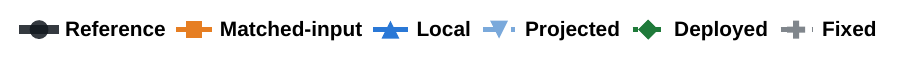}

\begin{minipage}[t]{0.45\linewidth}
\centering
\includegraphics[width=\linewidth]{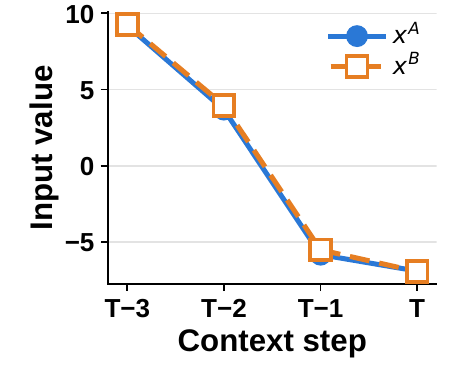}
\\[-0.2em]
{\small (a) Context tail only}
\end{minipage}
\hfill
\begin{minipage}[t]{0.45\linewidth}
\centering
\includegraphics[width=\linewidth]{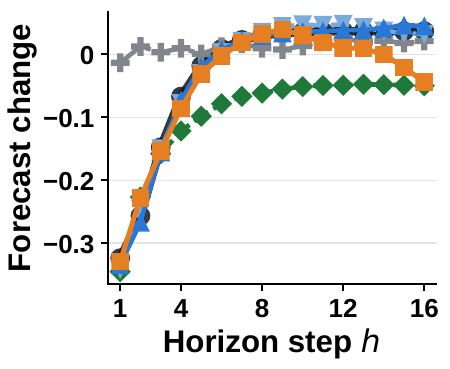}
\\[-0.2em]
{\small (b) Forecast-change coordinates}
\end{minipage}

\caption{
One Chronos-2~(\texttt{C2}) mean-reversion~(\texttt{MR}) pair at layer~12, with~\(\rho^{A}=0.4006\), \(\rho^{B}=0.4486\), and~\(\delta q=+0.0480\), traced through the matched-input and hidden-state intervention responses.
}
\label{fig:revision_integrated_example}
\end{figure}

\subsubsection{Full-Vector Response Fidelity and Centering}
\label{app:revision_fidelity}

Aligned gain in~\S\ref{sec:response} measures only the component of a forecast change along the reference response.
We therefore compare the complete response vectors for matched-input and hidden-state intervention responses to characterize their full-trajectory fidelity comprehensively.

We evaluate six conditions: matched input, reference-derived local edit, rank-restricted projected edit, deployed coordinate-readout edit, fixed-direction control, and conditional posterior-predictive reference.
Matched-input and conditional-reference responses use the response test dataset.
Hidden-edit responses use the independent geometry test dataset at the validation-selected layer, calibrated step, and capped target for consistent comparison across intervention conditions.

For response~\(q\) and reference~\(r\), let~\(\widetilde q\) and~\(\widetilde r\) subtract their respective horizon means.
We report centered aligned gain~\(g\), centered cosine similarity~\(s\), total relative error~\(e\), orthogonal error~\(e_\perp\), and level error~\(e_{\mathrm{level}}\) to capture complementary aspects of fidelity:
\[
\begin{aligned}
g
&=
\frac{
\left\langle \widetilde q,\widetilde r \right\rangle
}{
\left\|\widetilde r\right\|_2^2
},
\qquad
s
=
\frac{
\left\langle \widetilde q,\widetilde r \right\rangle
}{
\left\|\widetilde q\right\|_2
\left\|\widetilde r\right\|_2
},
\\
e
&=
\frac{
\left\|\widetilde q-\widetilde r\right\|_2
}{
\left\|\widetilde r\right\|_2
},
\qquad
e_\perp
=
\frac{
\left\|\widetilde q-g\widetilde r\right\|_2
}{
\left\|\widetilde r\right\|_2
},
\qquad
e_{\mathrm{level}}
=
\frac{
\sqrt{H}
\left|
\operatorname{mean}\left(q\right)
-
\operatorname{mean}\left(r\right)
\right|
}{
\left\|\widetilde r\right\|_2
}.
\end{aligned}
\]
The raw metrics~\(g^{\mathrm{raw}}\) and~\(e^{\mathrm{raw}}\) use the uncentered vectors.
The identity~\(e^2=\left(g-1\right)^2+e_\perp^2\) separates aligned-magnitude error from off-direction error.

\textbf{Matched-Input Gain Does Not Imply Full-Trajectory Agreement.}
As shown in~\autoref{tab:revision_fidelity} and~\autoref{fig:revision_rf_heatmaps}, matched-input responses have median centered gain~\(0.42\) and centered error~\(1.06\).
The heatmaps further show substantial variation across model--parameter cells, while local and projected responses remain consistently closer to the reference across the panel.
Deployed responses provide an intermediate level of fidelity, whereas the fixed-direction control generally retains larger errors.
For matched input, the median raw gain rises to~\(0.84\), showing that level agreement is substantially stronger than centered dynamical-shape agreement.
Only~\(8\%\) of matched-input pairs satisfy all three conditions:~\(0.8\leq g\leq1.2\), \(s\geq0.9\), and~\(e\leq0.5\).

\textbf{Local and Projected Edits Recover the Full Target Most Closely.}
As shown in~\autoref{tab:revision_fidelity} and~\autoref{fig:revision_rf_heatmaps}, local edits achieve median centered gain~\(1.00\) and error~\(0.06\), while projected edits retain gain~\(0.93\) with error~\(0.18\).
The heatmaps show that this high fidelity is broadly consistent across model--parameter cells rather than being driven by a small subset of cases.
The deployed edit partially recovers the target at gain~\(0.58\) and error~\(0.68\), outperforming the fixed direction at gain~\(0.23\) and error~\(1.06\).
The simultaneous fidelity criterion is met by~\(100\%\) of local pairs,~\(88\%\) of projected pairs,~\(21\%\) of deployed pairs, and~\(3\%\) of fixed-direction pairs.
Together, these results separate the high fidelity of reference-derived edits from the partial recovery achieved by predicted coordinates.

\textbf{Representative Trend Responses Show Distinct Matched-Input Behavior.}
\autoref{fig:revision_rf_example} provides three representative examples for the linear trend~(\texttt{TR}) law.
Panel~(a) shows a coherent matched-input response for TimesFM 3~(\texttt{F3}), panel~(b) shows a weak response for Moirai~(\texttt{MO}), and panel~(c) shows a reversed response for TTM-R2~(\texttt{TT}).
The centered fidelity metrics below each panel quantify these differences in gain, direction, trajectory error, and level error.
Despite the distinct matched-input behavior, the local and projected responses remain close to the reference in all three panels.
The deployed responses recover the reference to varying degrees, indicating that predicted hidden-state coordinates recover only part of the reference response.
These examples illustrate how similar hidden-state response capacity can coexist with substantially different responses to the corresponding input change across models with distinct forecasting behaviors.

\textbf{A Single-Pair Trace Separates the Intervention Stages.}
\autoref{fig:revision_integrated_example} follows one Chronos-2~(\texttt{C2}) mean-reversion~(\texttt{MR}) pair through the matched-input, local, projected, deployed, and fixed responses.
Panel~(a) shows the matched context tails~\(x^A\) and~\(x^B\) generated by the change from~\(\rho^A=0.4006\) to~\(\rho^B=0.4486\).
Panel~(b) compares the corresponding forecast-change trajectories against the reference response.
The matched-input response captures part of the target, while the local and projected traces closely follow the reference across the forecast horizon.
The deployed trace preserves much of the response direction but retains a noticeable level offset, whereas the fixed direction largely misses the target change.
For the matched-input, local, projected, deployed, and fixed responses, respectively, the raw aligned gains are~\(0.941\), \(1.041\), \(1.029\), \(0.883\), and~\(0.029\).
The corresponding centered gains are~\(0.897\), \(1.042\), \(1.032\), \(0.714\), and~\(0.061\).
The differences between raw and centered gains illustrate why response alignment should be interpreted together with trajectory, orthogonal, and level errors across the distinct hidden-state intervention stages.

Together, these analyses show that the selected hidden states can produce reference-aligned forecast changes even when the corresponding input-induced changes fail to produce the same response.
Local and projected interventions recover the target with high full-vector fidelity, while the predicted coordinate readout recovers only part of the required response.
This separation distinguishes response capacity in the hidden state from the response naturally induced by changing the input.

\subsection{Multi-Parameter Laws}
\label{app:multi_results}

\S\ref{sec:eval_multi} summarizes representation accessibility, matched-input response, and deployed response for seven parameters across three multi-parameter laws.
This subsection provides the detailed analyses supporting those results.
We first examine parameter accessibility across models and depth, together with a matched single-tone control.
We then evaluate matched-input response across intervention magnitudes and full-vector response fidelity.
Finally, we analyze causal geometry through local intervention validity, shared causal-subspace structure, and the advantage of input-specific deployed directions over the fixed-direction control.
The parameter ranges, conditional references, and calibrated intervention settings are summarized in~\autoref{tab:decodability_settings}, \autoref{tab:response_settings}, and~\autoref{tab:geometry_settings}.

\begin{table}[t]
\centering
\caption{
Multi-parameter summary across nine TSFMs.
Entries for representation accessibility, matched-input response, and deployed aligned gain report ranges across models.
The six-model group contains all models except Moirai~(\texttt{MO}), TTM-R2~(\texttt{TT}), and Timer-XL~(\texttt{TX}), which form the three-model group.
Compact cells report the descriptive compactness count from the run summary.
}
\label{tab:multilaw_summary}
\footnotesize
\setlength{\tabcolsep}{2.5pt}
\renewcommand{\arraystretch}{1.08}

\resizebox{\linewidth}{!}{%
\begin{tabular}{@{}lrrrrrr@{}}
\toprule
& \multicolumn{1}{c}{Accessibility}
& \multicolumn{2}{c}{Matched-Input Response~\(\bar U_{m,j}\)}
& \multicolumn{1}{c}{Compact}
& \multicolumn{1}{c}{Deployed}
& \multicolumn{1}{c}{Deployed \(>\) Fixed} \\
\cmidrule(lr){2-2}
\cmidrule(lr){3-4}
\cmidrule(lr){5-5}
\cmidrule(lr){6-6}
\cmidrule(l){7-7}
Parameter
& \(D_{m,j}\)
& Six models
& Three models
& Cells
& Gain
& Cells \\
\midrule

\multicolumn{7}{@{}l}{\textit{Dual Tone}} \\
Low-tone frequency (\texttt{TL})
& 0.86--0.99
& 0.82--1.00
& 0.04--0.16
& 7/9
& 0.65--0.98
& 9/9 \\

High-tone frequency (\texttt{TH})
& 0.93--0.99
& 0.75--0.99
& \(-0.16\)--0.12
& 6/9
& 0.40--0.97
& 9/9 \\

Amplitude ratio (\texttt{TW})
& 0.94--1.00
& 0.99--1.18
& 0.06--0.35
& 6/9
& 0.44--0.94
& 8/9 \\

\addlinespace[0.3em]
\multicolumn{7}{@{}l}{\textit{Linear State Space}} \\
Eigenvalue angle (\texttt{SC})
& 0.91--0.99
& 0.70--0.83
& 0.07--0.26
& 7/9
& 0.28--0.63
& 8/9 \\

Eigenvalue modulus (\texttt{SR})
& 0.61--0.91
& 0.85--1.02
& 0.07--0.28
& 6/9
& 0.17--0.59
& 5/9 \\

\addlinespace[0.3em]
\multicolumn{7}{@{}l}{\textit{Forced Diffusion}} \\
Drive frequency (\texttt{WC})
& 0.98--1.00
& 0.32--0.56
& \(-0.04\)--0.05
& 6/9
& 0.16--0.77
& 9/9 \\

Diffusion time (\texttt{WD})
& 0.62--0.89
& 0.76--0.96
& 0.10--0.51
& 7/9
& 0.51--0.84
& 8/9 \\

\bottomrule
\end{tabular}%
}
\end{table}

\begin{table}[t]
\centering
\caption{
Multi-parameter recoverability, accessibility, depth profiles, and input controls across nine~TSFMs.
Brackets indicate minima and maxima across models or context lengths.
}
\label{tab:representation_diag_multi}
\footnotesize
\setlength{\tabcolsep}{2.5pt}
\renewcommand{\arraystretch}{1.08}

\resizebox{\linewidth}{!}{%
\begin{tabular}{@{}lrrrrrrr@{}}
\toprule
&
\multicolumn{2}{c}{Accessibility}
&
\multicolumn{1}{c}{First Layer}
&
\multicolumn{2}{c}{Depth Profile}
&
\multicolumn{2}{c}{Input Control}
\\
\cmidrule(lr){2-3}
\cmidrule(lr){4-4}
\cmidrule(lr){5-6}
\cmidrule(l){7-8}
Parameter
& \(r^{\star}\)
& \(D_{m,j}\)
& \(D_{m,j}\left(\ell_1\right)\)
& Emergence
& Decline
& Raw
& Matched
\\
\midrule

\multicolumn{8}{@{}l}{\textit{Dual Tone}} \\
Low-tone frequency (\texttt{TL})
& 1.00
& \([0.857, 0.995]\)
& \([0.10, 0.96]\)
& 0.25
& 0.039
& 0.00
& \([0.00, 0.01]\)
\\

High-tone frequency (\texttt{TH})
& 1.00
& \([0.927, 0.994]\)
& \([0.07, 0.97]\)
& 0.17
& 0.028
& \([0.01, 0.02]\)
& \([0.00, 0.01]\)
\\

Amplitude ratio (\texttt{TW})
& 0.99
& \([0.940, 0.996]\)
& \([0.08, 0.99]\)
& 0.17
& 0.037
& 0.01
& 0.00
\\

\addlinespace[0.3em]
\multicolumn{8}{@{}l}{\textit{Linear State Space}} \\
Eigenvalue angle (\texttt{SC})
& \([0.98, 0.99]\)
& \([0.909, 0.995]\)
& \([0.09, 0.99]\)
& 0.08
& 0.012
& \([-0.02, 0.00]\)
& \([-0.01, 0.00]\)
\\

Eigenvalue modulus (\texttt{SR})
& \([0.89, 0.94]\)
& \([0.608, 0.912]\)
& \([0.17, 0.84]\)
& 0.25
& 0.064
& \([-0.01, 0.00]\)
& \([-0.01, 0.01]\)
\\

\addlinespace[0.3em]
\multicolumn{8}{@{}l}{\textit{Forced Diffusion}} \\
Drive frequency (\texttt{WC})
& 1.00
& \([0.976, 0.999]\)
& \([0.18, 1.00]\)
& 0.12
& 0.004
& 0.00
& \([0.00, 0.01]\)
\\

Diffusion time (\texttt{WD})
& 0.99
& \([0.623, 0.895]\)
& \([0.36, 0.88]\)
& 0.25
& 0.076
& \([-0.01, 0.00]\)
& 0.00
\\

\bottomrule
\end{tabular}%
}
\end{table}

\begin{figure}[t]
\centering

\begin{minipage}[t]{0.325\linewidth}
\centering
\includegraphics[width=\linewidth]{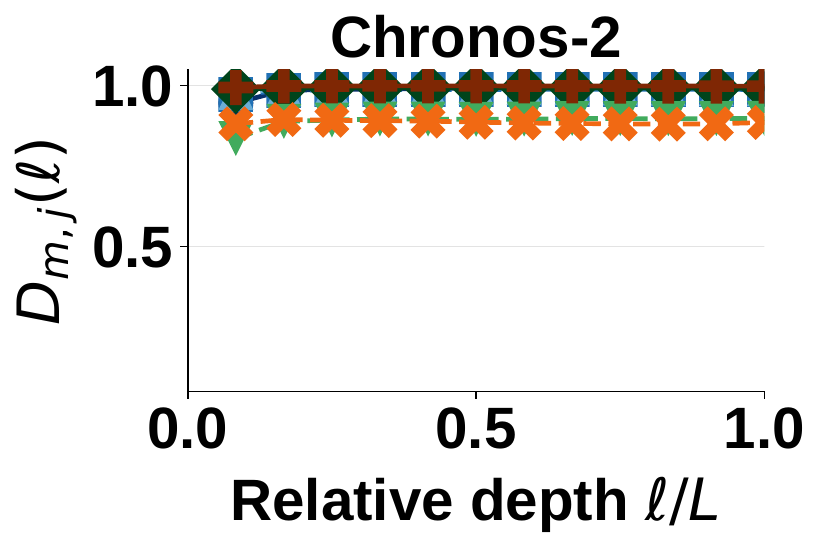}
\end{minipage}
\hfill
\begin{minipage}[t]{0.325\linewidth}
\centering
\includegraphics[width=\linewidth]{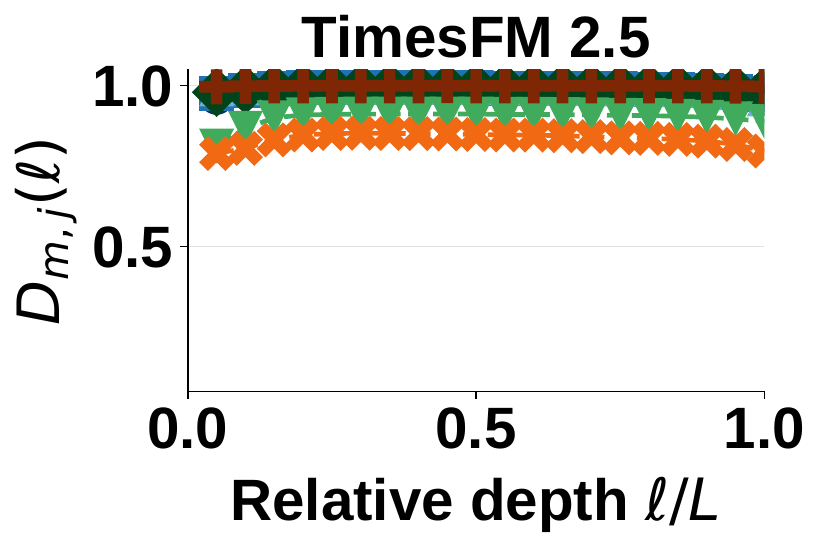}
\end{minipage}
\hfill
\begin{minipage}[t]{0.325\linewidth}
\centering
\includegraphics[width=\linewidth]{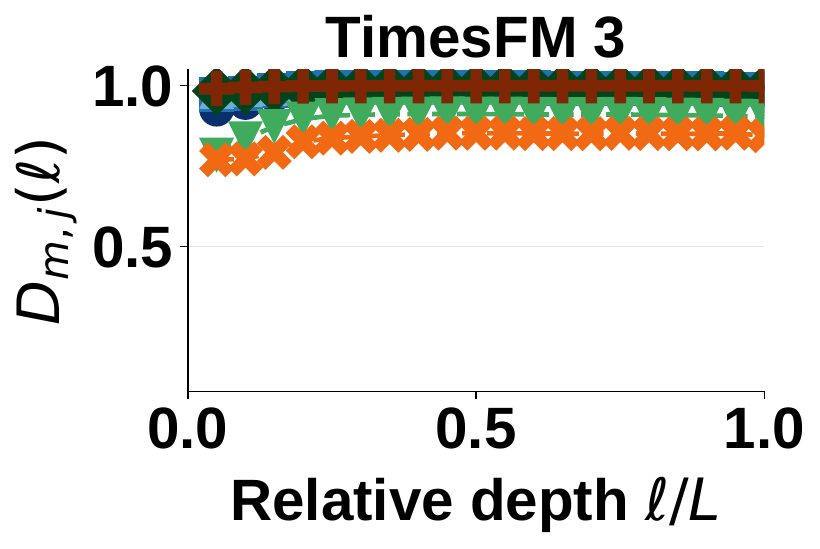}
\end{minipage}

\vspace{2pt}

\begin{minipage}[t]{0.325\linewidth}
\centering
\includegraphics[width=\linewidth]{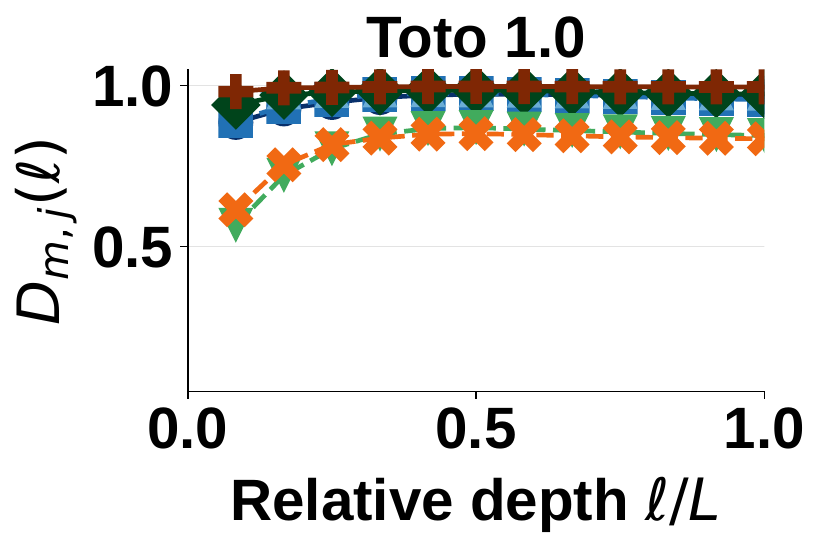}
\end{minipage}
\hfill
\begin{minipage}[t]{0.325\linewidth}
\centering
\includegraphics[width=\linewidth]{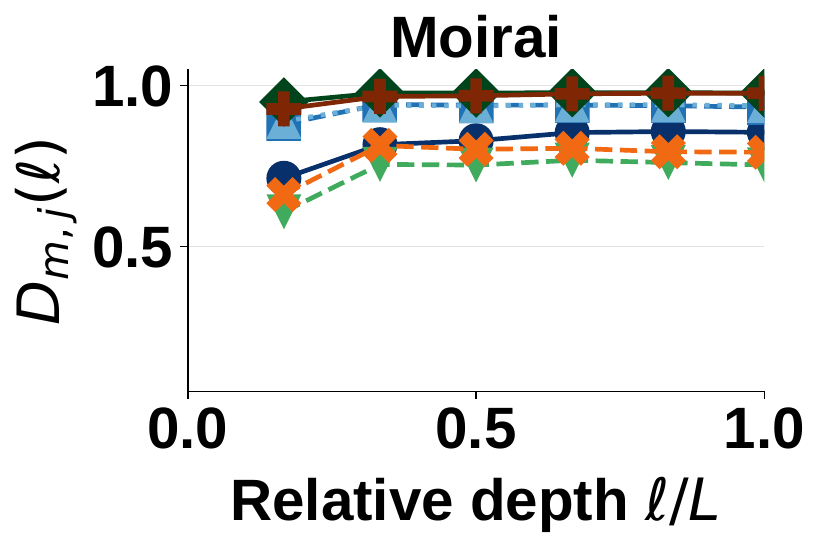}
\end{minipage}
\hfill
\begin{minipage}[t]{0.325\linewidth}
\centering
\includegraphics[width=\linewidth]{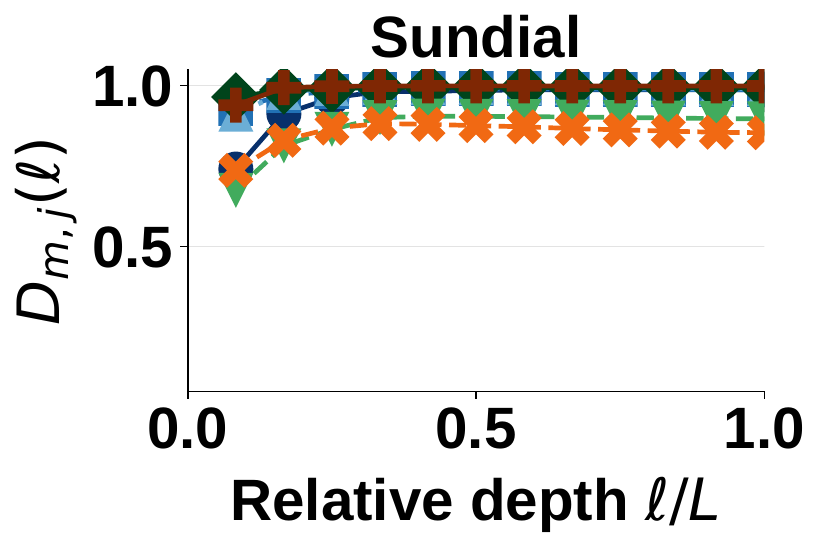}
\end{minipage}

\vspace{2pt}

\begin{minipage}[t]{0.325\linewidth}
\centering
\includegraphics[width=\linewidth]{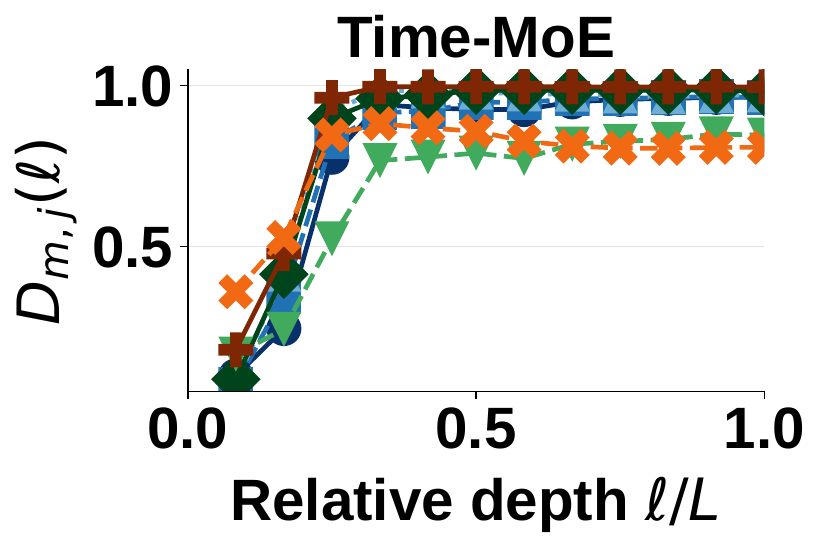}
\end{minipage}
\hfill
\begin{minipage}[t]{0.325\linewidth}
\centering
\includegraphics[width=\linewidth]{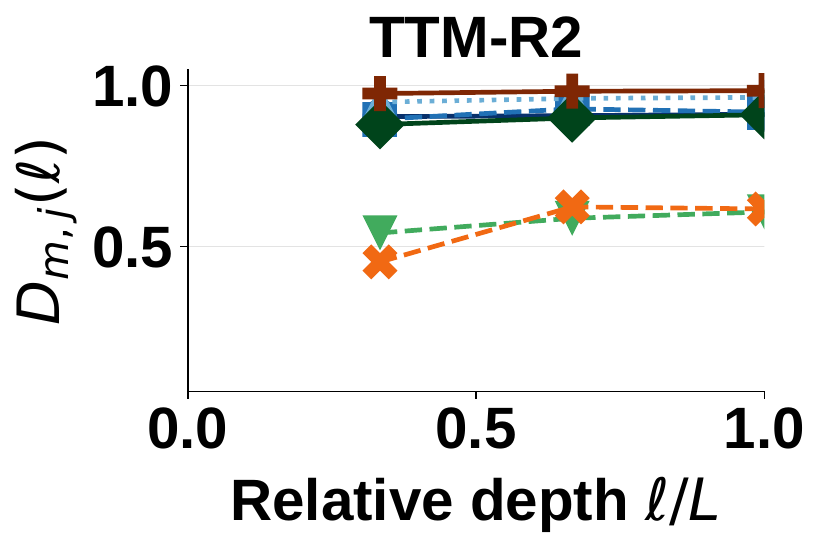}
\end{minipage}
\hfill
\begin{minipage}[t]{0.325\linewidth}
\centering
\includegraphics[width=\linewidth]{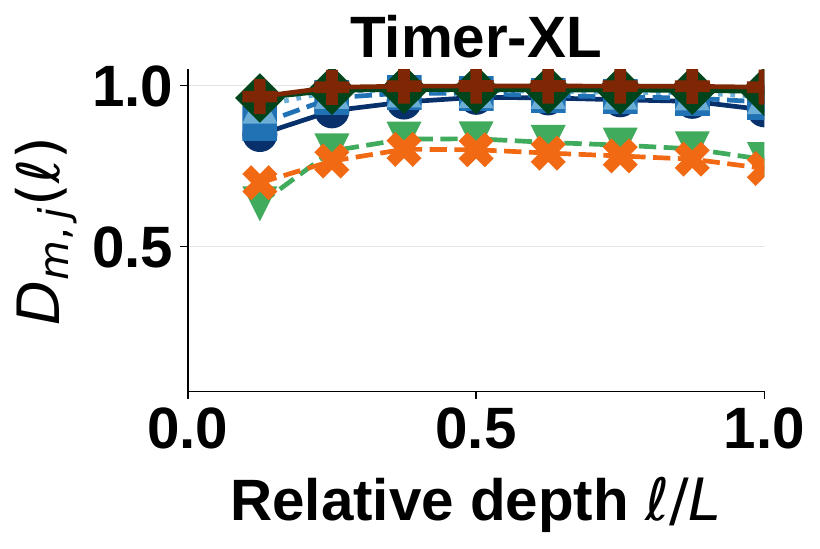}
\end{minipage}

\vspace{2pt}

\includegraphics[width=\linewidth]{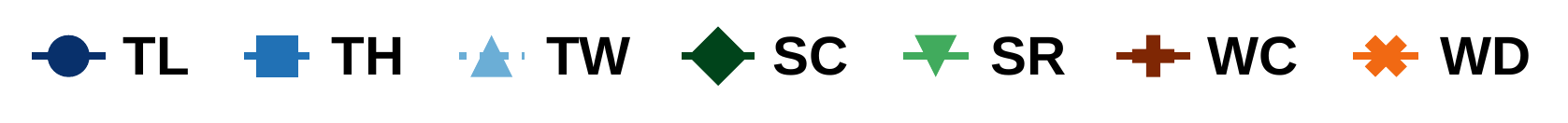}

\caption{
Test-set parameter accessibility across relative depth for seven jointly identified parameters.
Panels show all nine frozen TSFMs.
}
\label{fig:law_D_layers_multi}
\vspace{-0.1in}
\end{figure}

\begin{table}[t]
\centering
\caption{
Matched single-tone control for the dual-tone low-frequency parameter~(\texttt{TL}).
Accessibility and matched-input response compare the single-tone setting with~\texttt{TL}, with~\(\Delta=\texttt{TL}-\text{single tone}\).
}
\label{tab:multilaw_control}
\footnotesize
\setlength{\tabcolsep}{3.5pt}
\renewcommand{\arraystretch}{1.08}
\begin{tabular}{@{}lrrrrrr@{}}
\toprule
& \multicolumn{3}{c}{Accessibility~\(D_{m,j}\)}
& \multicolumn{3}{c}{Matched-Input Response~\(\bar U_{m,j}\)} \\
\cmidrule(lr){2-4}
\cmidrule(l){5-7}
Model
& Single
& \texttt{TL}
& \(\Delta\)
& Single
& \texttt{TL}
& \(\Delta\) \\
\midrule
Chronos-2 (\texttt{C2})
& 0.999 & 0.992 & \(-0.007\)
& 1.02 & 0.97 & \(-0.05\) \\

TimesFM 2.5 (\texttt{F2})
& 1.000 & 0.994 & \(-0.006\)
& 0.97 & 0.97 & 0.00 \\

TimesFM 3 (\texttt{F3})
& 1.000 & 0.995 & \(-0.005\)
& 1.01 & 1.00 & 0.00 \\

Toto-Open-Base 1.0 (\texttt{TO})
& 0.998 & 0.973 & \(-0.025\)
& 1.03 & 0.82 & \(-0.21\) \\

Moirai (\texttt{MO})
& 0.983 & 0.857 & \(-0.126\)
& 0.08 & 0.04 & \(-0.04\) \\

Sundial (\texttt{SU})
& 0.999 & 0.987 & \(-0.012\)
& 1.03 & 0.96 & \(-0.08\) \\

Time-MoE (\texttt{ME})
& 0.998 & 0.965 & \(-0.033\)
& 0.99 & 0.89 & \(-0.10\) \\

TTM-R2 (\texttt{TT})
& 0.989 & 0.911 & \(-0.078\)
& 0.22 & 0.16 & \(-0.06\) \\

Timer-XL (\texttt{TX})
& 0.999 & 0.963 & \(-0.036\)
& 0.27 & 0.12 & \(-0.15\) \\
\bottomrule
\end{tabular}%
\end{table}

\subsubsection{Representation Accessibility}
\label{app:multi_rep}

This analysis examines how representation accessibility for the seven jointly identified parameters varies across models, depth, and input controls.
\autoref{tab:multilaw_summary} summarizes parameter-level measurements, while~\autoref{tab:representation_diag_multi} reports recoverability, depth profiles, and input controls.
\autoref{tab:multilaw_control} compares the dual-tone low-frequency parameter with a matched single-tone control.
Finally, \autoref{fig:law_D_layers_multi} shows accessibility across model depth for all seven parameters.
These analyses further examine the differences summarized in~\S\ref{sec:eval_multi} across the three multi-parameter dynamical laws.

\textbf{Frequencies Are Accessible Early, While Rates Remain Lower.}
As shown in~\autoref{tab:multilaw_summary}, \autoref{tab:representation_diag_multi}, and~\autoref{fig:law_D_layers_multi}, the four frequency parameters have strong accessibility across models and are already accessible at early layers.
Their median first-layer accessibility is~\(0.886\) for~\texttt{TL}, \(0.895\) for~\texttt{TH}, \(0.961\) for~\texttt{SC}, and~\(0.975\) for~\texttt{WC}.
They reach~\(95\%\) of their maximum accessibility by median relative depths between~\(0.08\) and~\(0.25\).
The two rate parameters start lower, at~\(0.633\) for~\texttt{SR} and~\(0.698\) for~\texttt{WD}, and their median peak accessibility values of~\(0.869\) and~\(0.852\) remain below the median first-layer accessibility of every frequency parameter.
The gap between modal frequency and eigenvalue modulus reported in~\S\ref{sec:eval_multi} is therefore present throughout model depth rather than arising from layer selection.
Time-MoE~(\texttt{ME}) has the lowest first-layer accessibility, ranging from~\(0.07\) to~\(0.36\), and exhibits the largest increases across the seven parameters, ranging from~\(0.52\) to~\(0.91\).
The depth profiles are otherwise largely stable after reaching their maxima, with the largest subsequent decline equal to~\(0.076\) across all evaluated model--parameter combinations.

\textbf{The Second Tone Has Limited Effect on Accessibility but Reduces Response.}
The matched single-tone control in~\autoref{tab:multilaw_control} isolates the effect of adding the second tone, while~\autoref{tab:multilaw_summary} places the resulting low-tone accessibility and response within the full multi-parameter panel.
Relative to the single-tone setting, low-tone frequency~(\texttt{TL}) loses between~\(0.005\) and~\(0.126\) in accessibility.
For the six-model response group, the accessibility reduction is at most~\(0.033\).
Matched-input response decreases for every model, ranging from less than~\(0.001\) for TimesFM 2.5~(\texttt{F2}) to~\(0.209\) for Toto-Open-Base 1.0~(\texttt{TO}).
The three largest response reductions occur for Toto-Open-Base 1.0~(\texttt{TO}), Timer-XL~(\texttt{TX}), and Time-MoE~(\texttt{ME}), at~\(0.209\), \(0.154\), and~\(0.104\), respectively.
These reductions exceed their corresponding accessibility losses of~\(0.025\), \(0.036\), and~\(0.033\), showing that the added tone affects forecast response more strongly than parameter accessibility in these models.

\textbf{TSFM Processing Makes the Joint Parameters Linearly Accessible.}
\autoref{tab:representation_diag_multi} shows that the raw and matched-input controls remain near zero for all seven parameters, with absolute values reaching at most~\(0.02\), while TSFM accessibility ranges from approximately~\(0.61\) to~\(1.00\).
For low-tone frequency~(\texttt{TL}), high-tone frequency~(\texttt{TH}), and drive frequency~(\texttt{WC}), the conditional reference is~\(1.00\), yet the corresponding input controls remain near zero.
The selected hidden representations therefore expose these parameters substantially more linearly than the tested raw-input summaries across models and jointly varying settings.
This comparison does not establish that the parameter information is absent from the input or uniquely computed by the network.

Together, these results show that joint identification preserves strong accessibility for the frequency parameters while maintaining parameter-specific differences within each dynamical law.
The matched single-tone control further shows that adding a second frequency component has a modest effect on accessibility but can produce a larger reduction in matched-input forecast response.

\begin{figure}[t]
\centering
\includegraphics[width=\linewidth]{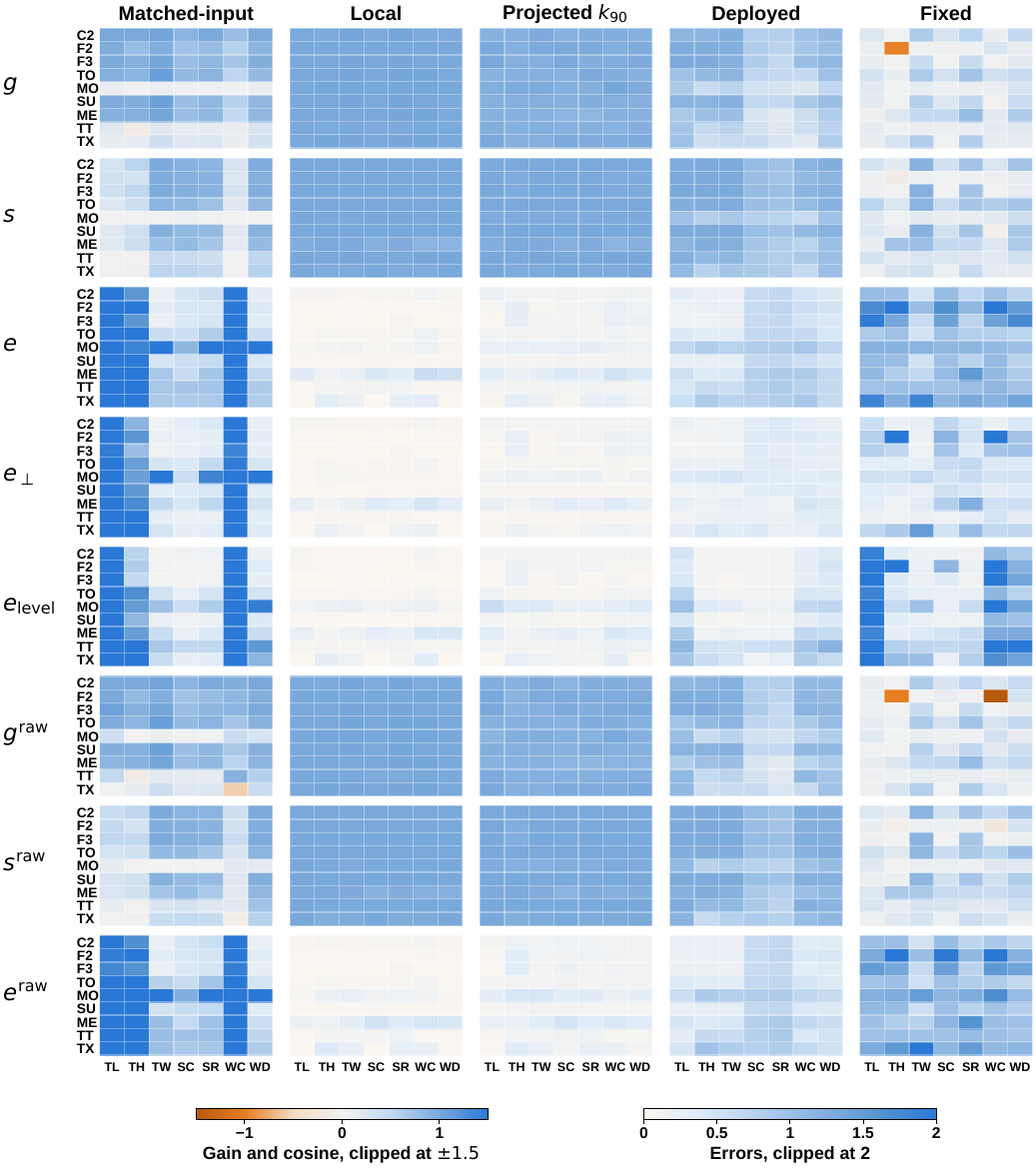}
\caption{
Centered and raw response-fidelity metrics across the multi-parameter panel.
Columns compare matched-input, local, projected, deployed, and fixed-direction responses.
Centered metrics capture response alignment and trajectory error, while raw metrics use uncentered forecast changes.
}
\label{fig:revision_rf_multi}
\end{figure}

\begin{figure}[t]
\centering
\begin{minipage}[t]{0.495\linewidth}
\centering
\includegraphics[width=\linewidth]{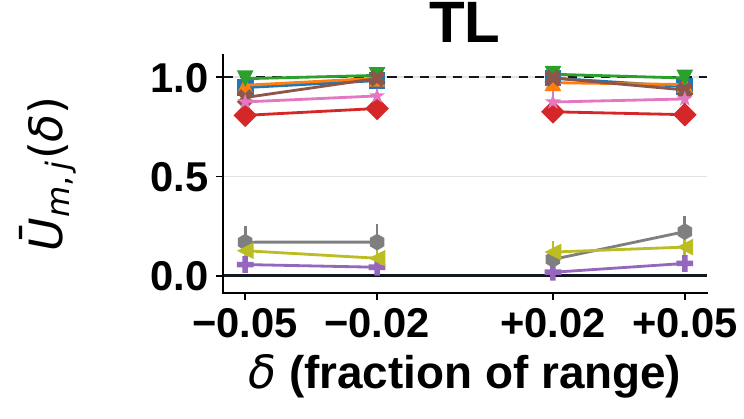}
\end{minipage}
\hfill
\begin{minipage}[t]{0.495\linewidth}
\centering
\includegraphics[width=\linewidth]{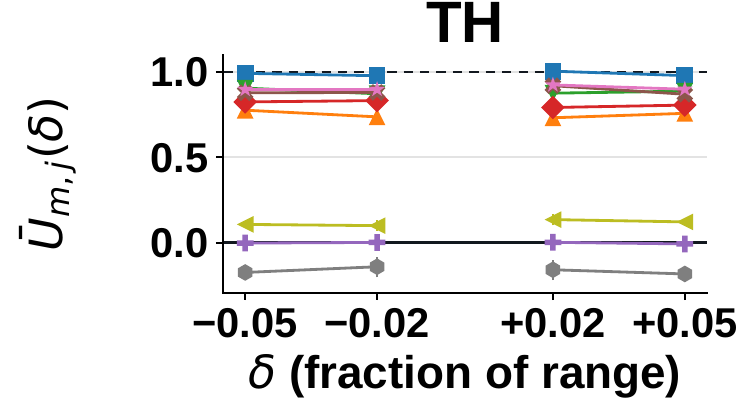}
\end{minipage}
\vspace{4pt}
\begin{minipage}[t]{0.495\linewidth}
\centering
\includegraphics[width=\linewidth]{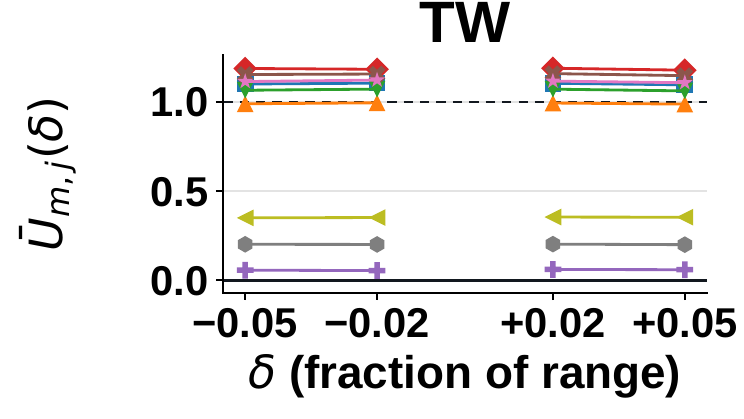}
\end{minipage}
\hfill
\begin{minipage}[t]{0.495\linewidth}
\centering
\includegraphics[width=\linewidth]{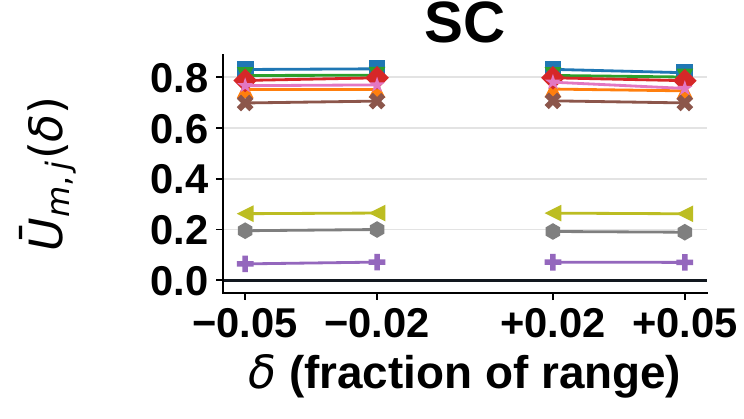}
\end{minipage}
\vspace{4pt}
\begin{minipage}[t]{0.495\linewidth}
\centering
\includegraphics[width=\linewidth]{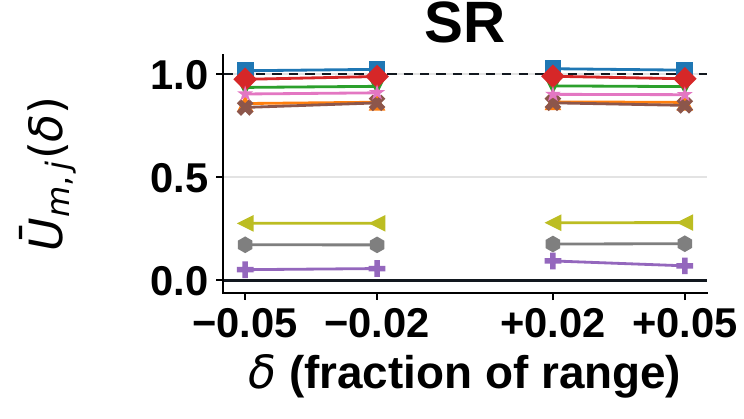}
\end{minipage}
\hfill
\begin{minipage}[t]{0.495\linewidth}
\centering
\includegraphics[width=\linewidth]{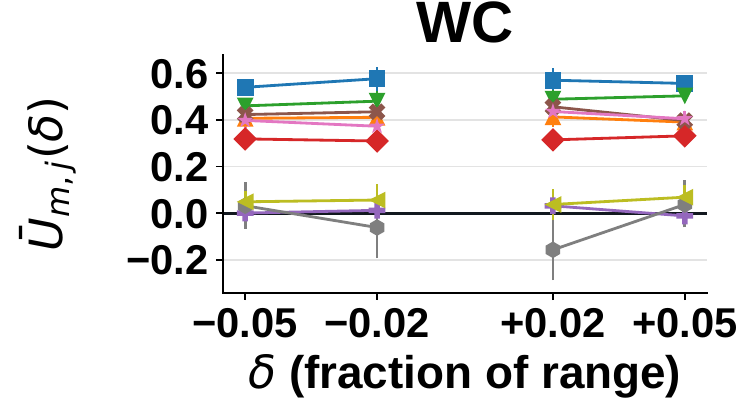}
\end{minipage}
\vspace{4pt}
\begin{minipage}[t]{0.495\linewidth}
\centering
\includegraphics[width=\linewidth]{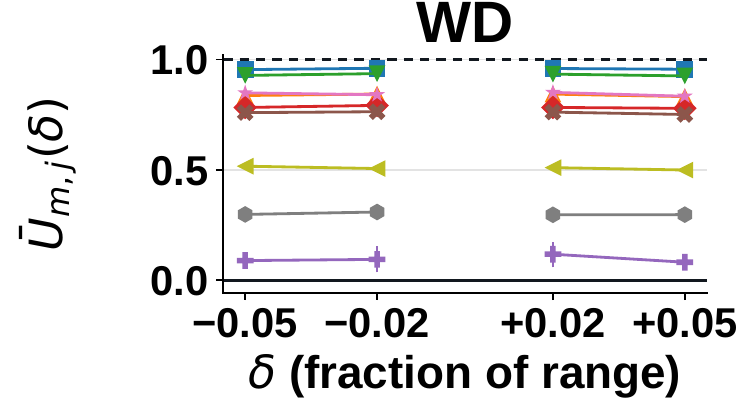}
\end{minipage}
\hfill
\begin{minipage}[t]{0.495\linewidth}
\centering
\includegraphics[width=\linewidth]{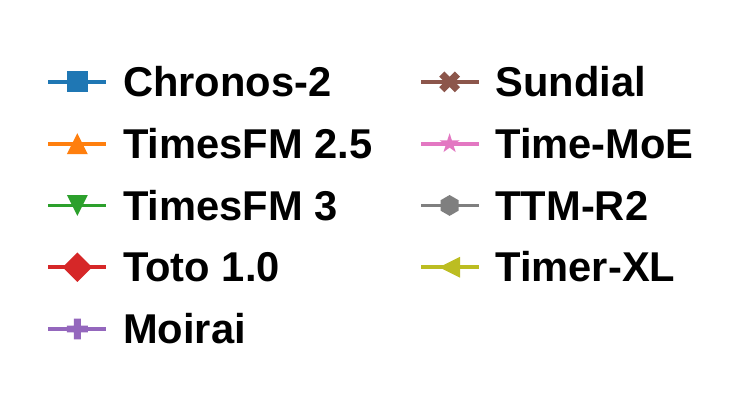}
\end{minipage}
\caption{
Conditional-reference-normalized multi-parameter response across signed intervention magnitudes.
The dashed unit-reference line is shown where it lies within the panel range.
}
\label{fig:law_U_dose_multi}
\end{figure}

\subsubsection{Forecast Response}
\label{app:multi_resp}

This analysis examines matched-input response for the seven jointly identified parameters across full-vector fidelity, intervention magnitudes, and response diagnostics.
\autoref{fig:revision_rf_multi} compares centered and raw full-vector fidelity across response conditions.
\autoref{fig:law_U_dose_multi} then shows conditional-reference-normalized response across signed intervention magnitudes, while~\autoref{tab:response_diag_multi} reports scale-aware and unpaired diagnostics.
Together, these analyses provide the detailed response results underlying~\S\ref{sec:eval_multi}.

\textbf{Full-Vector Fidelity Separates the Response Conditions.}
\autoref{fig:revision_rf_multi} compares centered and raw fidelity for matched-input, local, projected, deployed, and fixed-direction responses across the seven jointly identified parameters.
The local and projected interventions remain closest to the reference across the fidelity metrics, showing that the selected hidden states can support the reference response and that much of this response is retained after projection into the shared subspace.
Median pooled centered errors are~\(0.93\), \(0.14\), and~\(0.61\) for matched-input, projected, and deployed responses, respectively.
The deployed responses therefore recover a substantial part of the reference response but retain larger gain, directional, and trajectory errors than the reference-derived projected interventions.
Matched-input responses show the largest variation across parameters and models, while the fixed-direction control generally provides weaker fidelity than the input-specific deployed directions.
Differences between centered and raw metrics further indicate that agreement in overall forecast level can coexist with disagreement in the dynamical shape of the response.
Together, the figure separates response capacity in the hidden state from the fidelity achieved by natural input changes and predicted intervention coordinates across jointly varying multi-parameter dynamical systems.

\textbf{Weak Response Persists Across Tested Magnitudes.}
As shown in~\autoref{fig:law_U_dose_multi}, the median spread across four intervention magnitudes is at most~\(0.044\) for every parameter, while the largest spread among the six responsive models is~\(0.098\) for low-tone frequency~(\texttt{TL}).
The largest overall spreads occur for TTM-R2~(\texttt{TT}) on drive frequency~(\texttt{WC}) at~\(0.194\) and low-tone frequency~(\texttt{TL}) at~\(0.140\), where responses remain near zero.
The difference between positive and negative interventions of equal magnitude is at most~\(0.095\).
For drive frequency~(\texttt{WC}), the six responsive models remain between~\(0.31\) and~\(0.58\) at every magnitude, with a within-model spread of at most~\(0.063\).
The weak drive-frequency response therefore persists across the evaluated intervention magnitudes rather than arising from a particular magnitude.

\textbf{Drive-Frequency Response Is Weaker Than Unpaired Waveform Tracking.}
\autoref{tab:response_diag_multi} shows that the six responsive models track the unpaired conditional forecast substantially better than they respond to a controlled drive-frequency change.
For drive frequency~(\texttt{WC}), the unpaired coefficient ranges from~\(0.59\) to~\(0.88\), with shape~\(R^2\) between~\(0.92\) and~\(0.96\).
The two tone frequencies show comparable unpaired tracking, with coefficients between~\(0.85\) and~\(0.98\) and~\(R^2\) between~\(0.92\) and~\(0.97\).
By contrast, matched-input drive-frequency response remains between~\(0.32\) and~\(0.56\).
These models therefore track much of the unpaired waveform shape while responding more weakly to a controlled change in its frequency across the evaluated multi-parameter settings.

\textbf{Some Response Components Are Associated With Window Scale.}
The scale-aware diagnostics in~\autoref{tab:response_diag_multi} separate the reference-aligned component from a component associated with window scale.
For the six responsive models, adding the scale term reduces eigenvalue-modulus~(\texttt{SR}) responses from~\(0.85\)--\(1.02\) to~\(0.61\)--\(0.75\), and diffusion-time~(\texttt{WD}) responses from~\(0.76\)--\(0.96\) to~\(0.52\)--\(0.76\).
For amplitude ratio~(\texttt{TW}), the maximum decreases from~\(1.18\) to~\(1.09\), while the frequency parameters change less.
The responses remain positive, indicating that the scale term explains part, but not all, of the reference-aligned component.
This regression comparison is diagnostic and does not isolate a causal normalization pathway.
The four cells whose signs change under the refit all have primary responses near zero, with~\(\left|\bar U_{m,j}\right|\leq0.12\).

\textbf{Heavy-Tailed Cells Do Not Change the Conclusions.}
A small number of model--parameter cells have heavy-tailed per-pair response coefficients.
In five of the~63 cells,~\(\bar U_{m,j}\) differs by more than~\(0.2\) from the corresponding ratio formed using per-pair medians.
Four involve weakly responding models: TTM-R2~(\texttt{TT}) on high-tone frequency~(\texttt{TH}), drive frequency~(\texttt{WC}), and low-tone frequency~(\texttt{TL}), and Timer-XL~(\texttt{TX}) on diffusion time~(\texttt{WD}).
Under median aggregation, the three TTM-R2 responses remain between~\(0.35\) and~\(0.38\), still well below the conditional-reference level.
The fifth cell is TimesFM 2.5~(\texttt{F2}) on high-tone frequency~(\texttt{TH}), where~\(\bar U_{m,j}=0.75\), the median-based value is~\(1.01\), and the~\(10\%\)-trimmed value is~\(0.81\).
The heavy tail therefore understates rather than overstates the response in this case without altering the overall response pattern.

Together, these results show that weak multi-parameter responses persist across intervention magnitudes and alternative response diagnostics.
Full-vector fidelity further separates matched-input behavior from the stronger target recovery obtained through hidden-state interventions.

\begin{figure}[t]
\centering

\begin{minipage}[t]{0.45\linewidth}
\centering
\includegraphics[width=\linewidth]{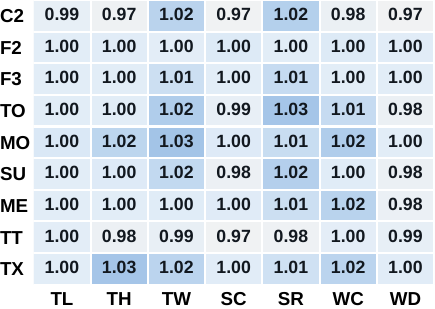}
\\[-0.2em]
{\small (a) Selected-layer local response}
\end{minipage}
\hfill
\begin{minipage}[t]{0.45\linewidth}
\centering
\includegraphics[width=\linewidth]{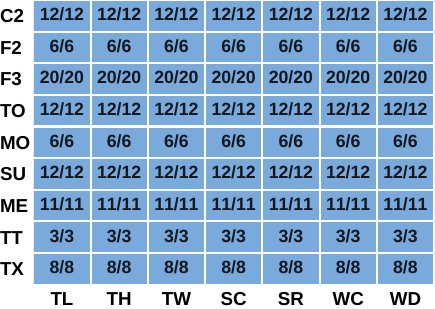}
\\[-0.2em]
{\small (b) Validation-admissible layer fraction}
\end{minipage}

\caption{
Local validity of Jacobian-derived hidden-state interventions across seven jointly identified parameters.
Panel~(a) reports test~\(U_{\mathrm{local}}\left(\ell\right)\) at the validation-selected layer, while panel~(b) reports the fraction of measured layers satisfying the local-validity criterion.
}
\vspace{-0.1in}
\label{fig:local_validity_multi}
\end{figure}

\begin{table}[t]
\centering
\caption{
Scale-aware and unpaired response diagnostics for seven jointly identified parameters.
Panel~A summarizes results across nine~TSFMs, with brackets indicating model minima and maxima.
Panel~B reports weak, reversed, or amplified response cells.
}
\label{tab:response_diag_multi}
\footnotesize
\setlength{\tabcolsep}{2.4pt}
\renewcommand{\arraystretch}{1.08}

\resizebox{\linewidth}{!}{%
\begin{tabular}{@{}llrrrrrrr@{}}
\toprule

\multicolumn{9}{@{}l}{\textbf{PANEL A: PARAMETER-LEVEL SUMMARY}} \\
\midrule

&
&
\multicolumn{5}{c}{Matched-Input Diagnostics}
&
\multicolumn{2}{c}{Unpaired Check}
\\
\cmidrule(lr){3-7}
\cmidrule(l){8-9}

Parameter
& Model
& \(\bar U_{m,j}\)
& \(U^{(3)}\)
& \(\gamma\)
& \(\bar\rho\)
& \(U^{\star(3)}/U^{\star}\)
& Coefficient
& \(R^{2}\)
\\
\midrule

Low-tone frequency (\texttt{TL})
& Nine models
& \([0.04, 1.00]\)
& \([-0.04, 1.01]\)
& \(-0.011\)
& \(0.50\)
& \(1.00\)
& \([0.03, 0.98]\)
& \(0.94\)
\\

High-tone frequency (\texttt{TH})
& Nine models
& \([-0.16, 0.99]\)
& \([-0.52, 1.01]\)
& \(-0.008\)
& \(0.25\)
& \(1.00\)
& \([0.03, 0.98]\)
& \(0.94\)
\\

Amplitude ratio (\texttt{TW})
& Nine models
& \([0.06, 1.18]\)
& \([0.04, 1.09]\)
& \(-0.001\)
& \(0.91\)
& \(0.97\)
& \([0.03, 0.98]\)
& \(0.94\)
\\

Eigenvalue angle (\texttt{SC})
& Nine models
& \([0.07, 0.83]\)
& \([0.05, 0.79]\)
& \(-0.016\)
& \(0.43\)
& \(1.03\)
& \([0.09, 0.81]\)
& \(0.78\)
\\

Eigenvalue modulus~(\texttt{SR})
& Nine models
& \([0.07, 1.02]\)
& \([0.06, 0.75]\)
& \(-0.001\)
& \(0.66\)
& \(0.89\)
& \([0.09, 0.78]\)
& \(0.76\)
\\

Drive frequency (\texttt{WC})
& Nine models
& \([-0.04, 0.56]\)
& \([-0.14, 0.49]\)
& \(-0.089\)
& \(0.43\)
& \(1.00\)
& \([-0.07, 0.88]\)
& \(0.92\)
\\

Diffusion time (\texttt{WD})
& Nine models
& \([0.10, 0.96]\)
& \([0.05, 0.76]\)
& \(-0.001\)
& \(0.76\)
& \(0.99\)
& \([-0.08, 0.88]\)
& \(0.92\)
\\

\midrule
\multicolumn{9}{@{}l}{\textbf{PANEL B: WEAK, REVERSED, OR AMPLIFIED RESPONSES}} \\
\midrule

High-tone frequency (\texttt{TH})
& Moirai (\texttt{MO})
& \(-0.00\)
& \(0.00\)
& \(-0.140\)
& \(0.25\)
& --
& \(0.03\)
& \(0.17\)
\\

High-tone frequency (\texttt{TH})
& TTM-R2 (\texttt{TT})
& \(-0.16\)
& \(-0.52\)
& \(-0.137\)
& \(0.64\)
& --
& \(0.16\)
& \(0.63\)
\\

Drive frequency (\texttt{WC})
& TTM-R2 (\texttt{TT})
& \(-0.04\)
& \(-0.13\)
& \(-0.099\)
& \(0.59\)
& --
& \(-0.07\)
& \(0.69\)
\\

\bottomrule
\end{tabular}%
}
\end{table}

\begin{figure}[t]
\centering

\begin{minipage}[t]{0.325\linewidth}
\centering
\includegraphics[width=\linewidth]{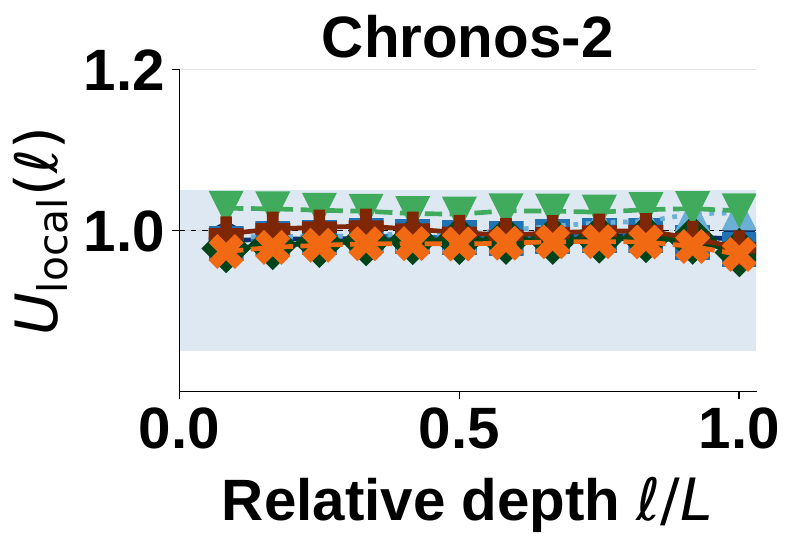}
\end{minipage}
\hfill
\begin{minipage}[t]{0.325\linewidth}
\centering
\includegraphics[width=\linewidth]{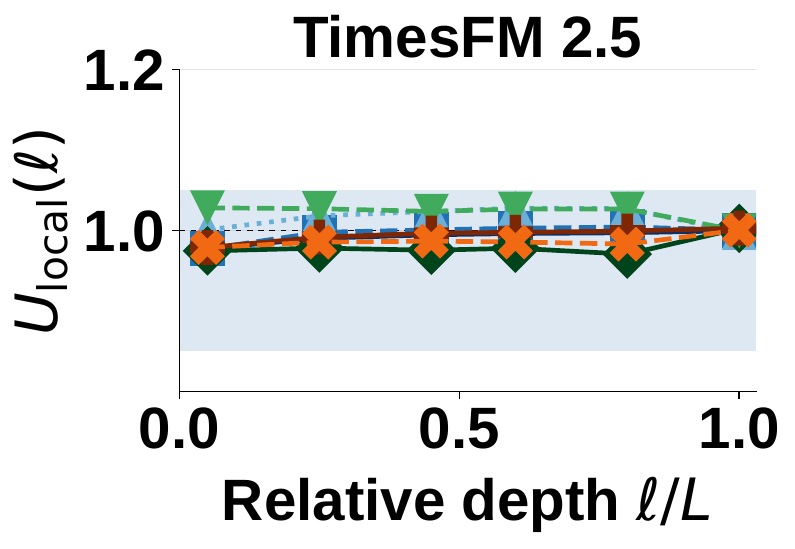}
\end{minipage}
\hfill
\begin{minipage}[t]{0.325\linewidth}
\centering
\includegraphics[width=\linewidth]{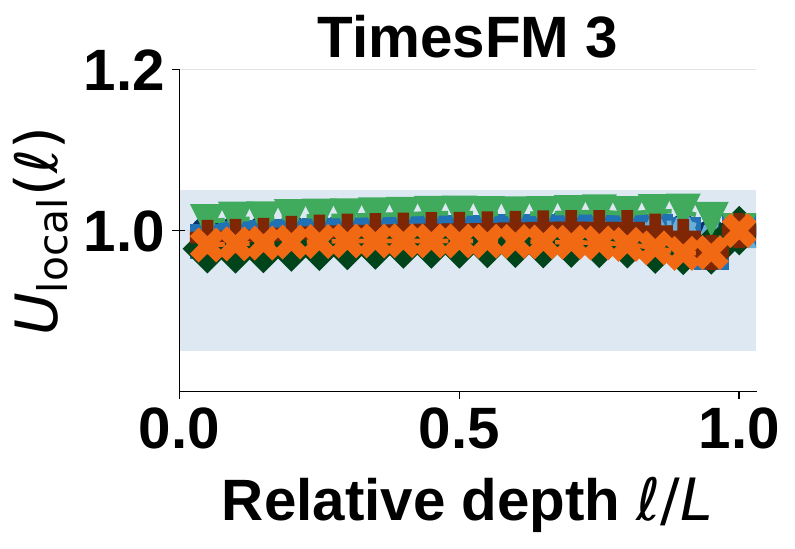}
\end{minipage}

\vspace{2pt}

\begin{minipage}[t]{0.325\linewidth}
\centering
\includegraphics[width=\linewidth]{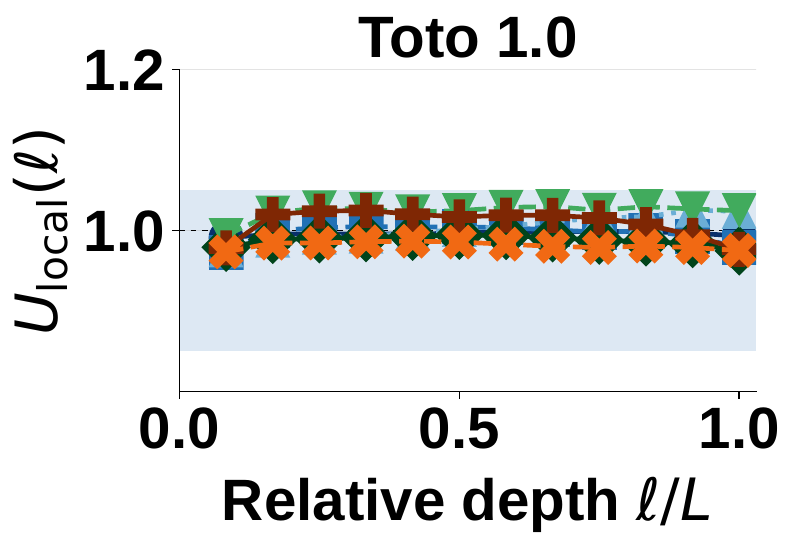}
\end{minipage}
\hfill
\begin{minipage}[t]{0.325\linewidth}
\centering
\includegraphics[width=\linewidth]{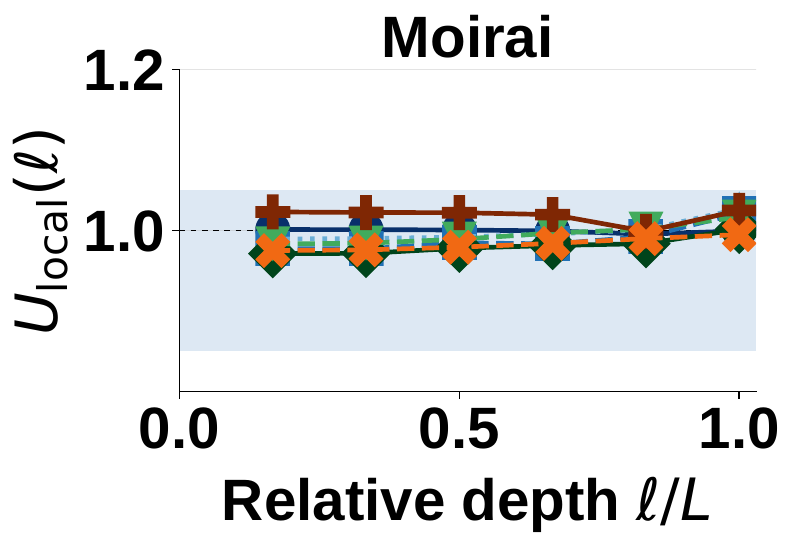}
\end{minipage}
\hfill
\begin{minipage}[t]{0.325\linewidth}
\centering
\includegraphics[width=\linewidth]{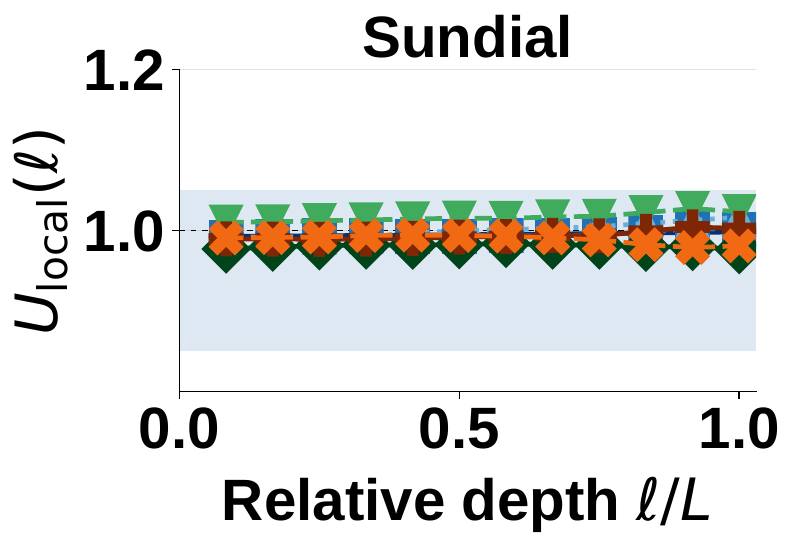}
\end{minipage}

\vspace{2pt}

\begin{minipage}[t]{0.325\linewidth}
\centering
\includegraphics[width=\linewidth]{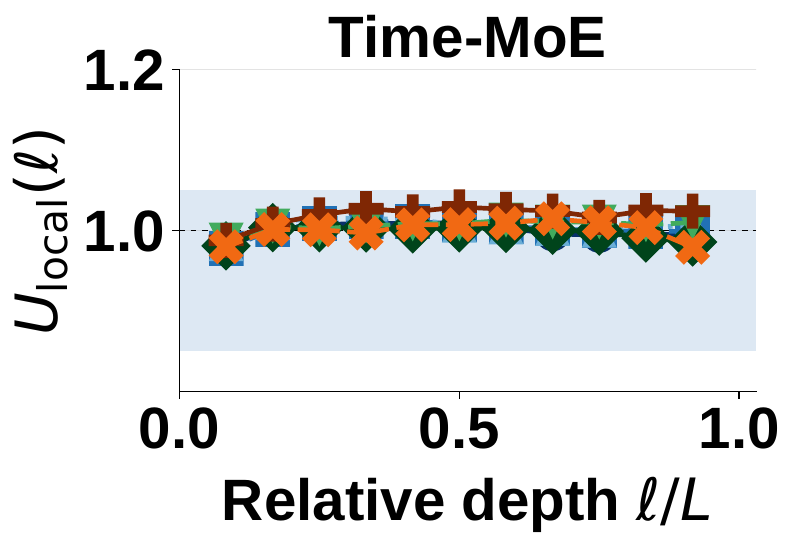}
\end{minipage}
\hfill
\begin{minipage}[t]{0.325\linewidth}
\centering
\includegraphics[width=\linewidth]{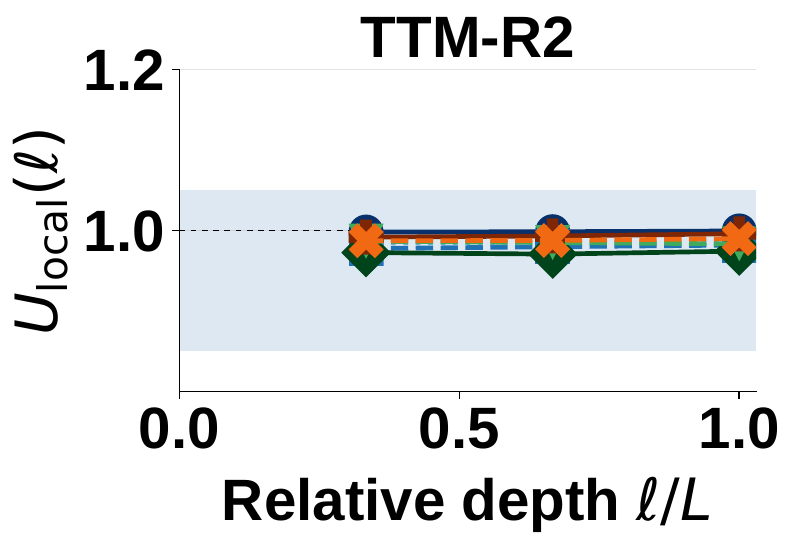}
\end{minipage}
\hfill
\begin{minipage}[t]{0.325\linewidth}
\centering
\includegraphics[width=\linewidth]{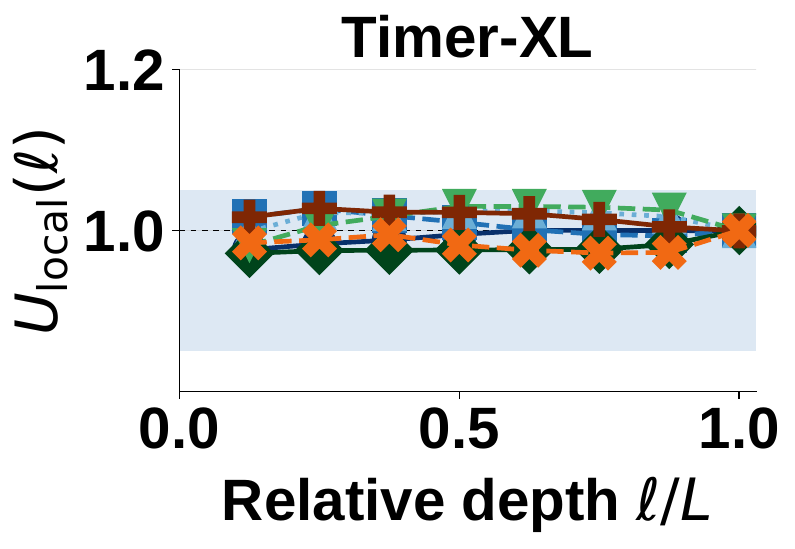}
\end{minipage}

\vspace{2pt}

\includegraphics[width=\linewidth]{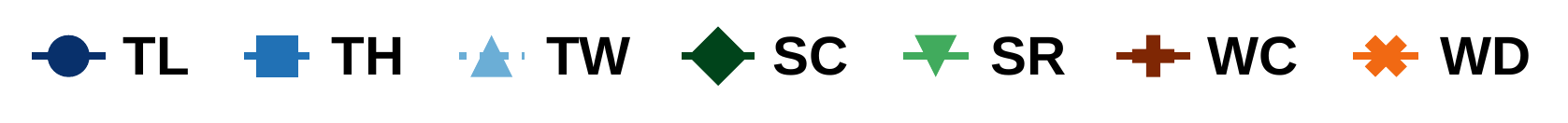}

\caption{
Validation local response across relative depth for seven jointly identified parameters and nine TSFMs.
The profiles show the nonlinear effectiveness of Jacobian-derived hidden-state interventions across measured layers.
}
\label{fig:local_layers_multi}
\end{figure}

\subsubsection{Causal Geometry}
\label{app:multi_geom}

\begin{figure}[t]
\centering

\begin{minipage}[t]{0.325\linewidth}
\centering
\includegraphics[width=\linewidth]{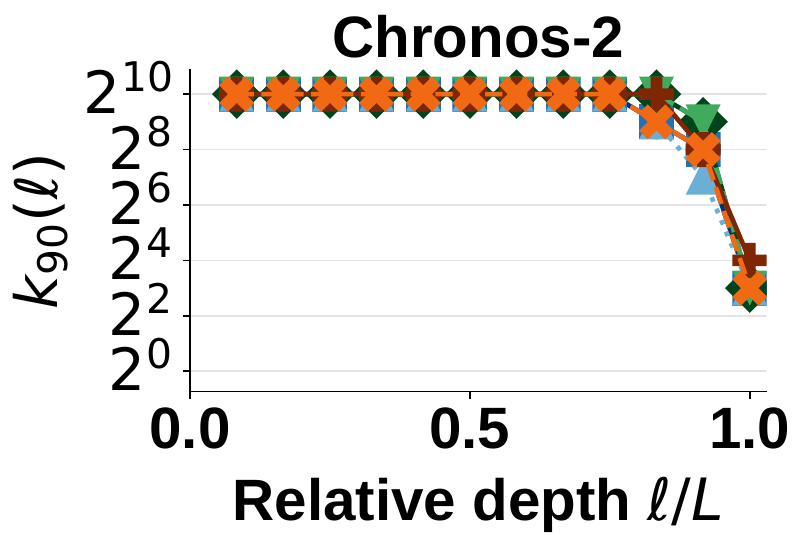}
\end{minipage}
\hfill
\begin{minipage}[t]{0.325\linewidth}
\centering
\includegraphics[width=\linewidth]{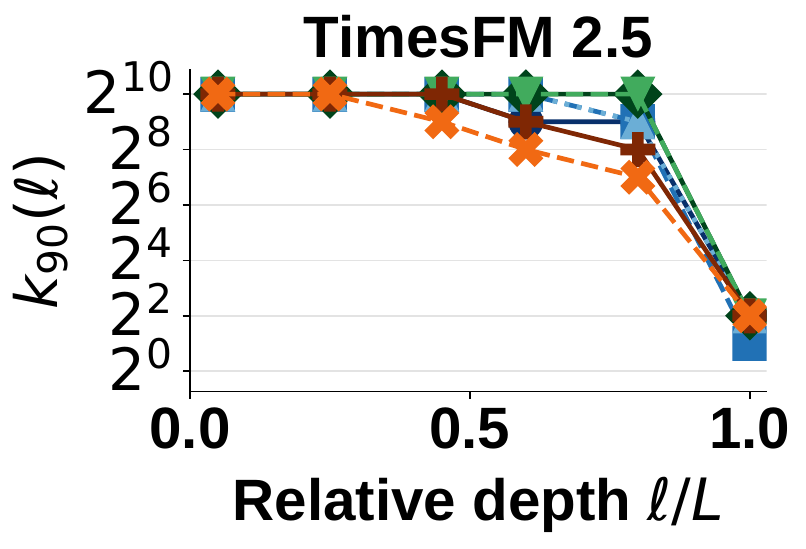}
\end{minipage}
\hfill
\begin{minipage}[t]{0.325\linewidth}
\centering
\includegraphics[width=\linewidth]{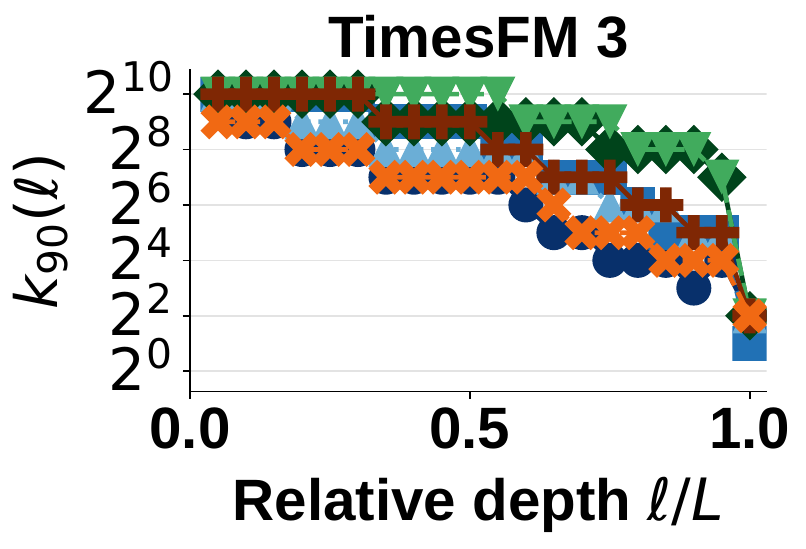}
\end{minipage}

\vspace{2pt}

\begin{minipage}[t]{0.325\linewidth}
\centering
\includegraphics[width=\linewidth]{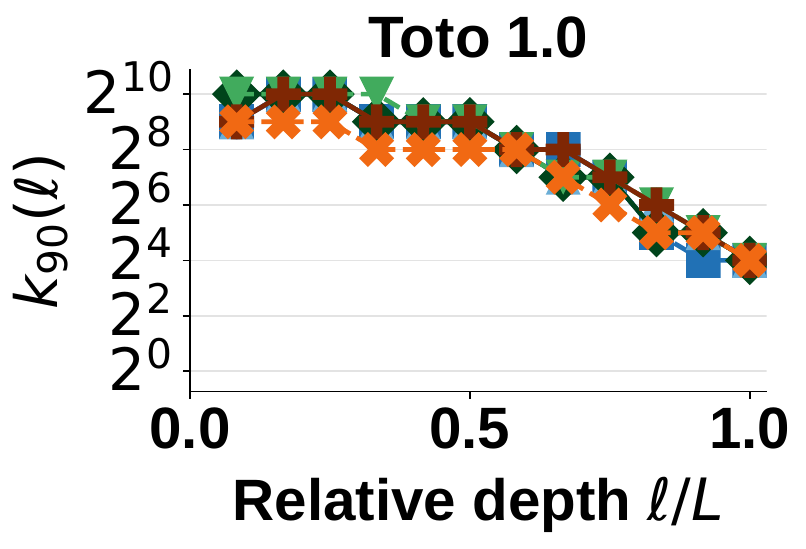}
\end{minipage}
\hfill
\begin{minipage}[t]{0.325\linewidth}
\centering
\includegraphics[width=\linewidth]{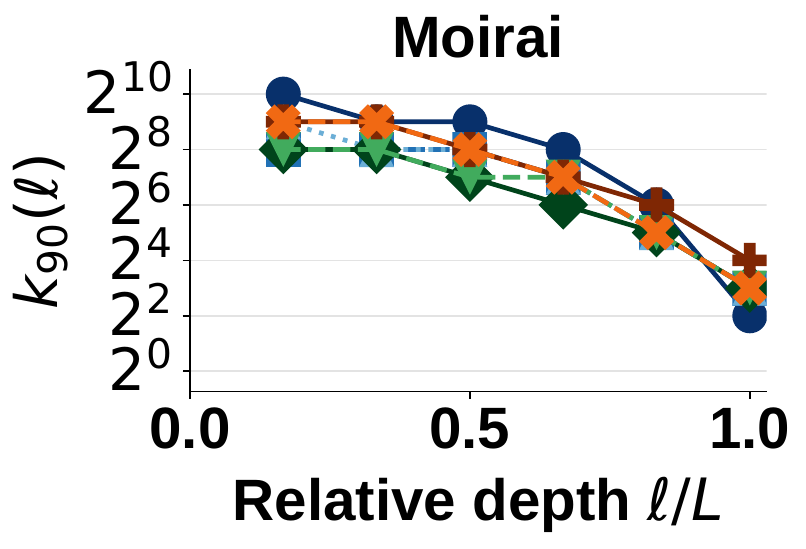}
\end{minipage}
\hfill
\begin{minipage}[t]{0.325\linewidth}
\centering
\includegraphics[width=\linewidth]{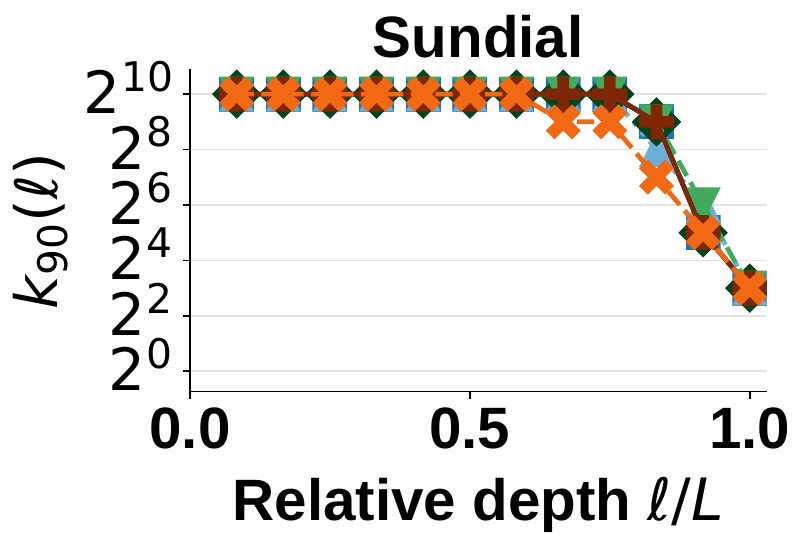}
\end{minipage}

\vspace{2pt}

\begin{minipage}[t]{0.325\linewidth}
\centering
\includegraphics[width=\linewidth]{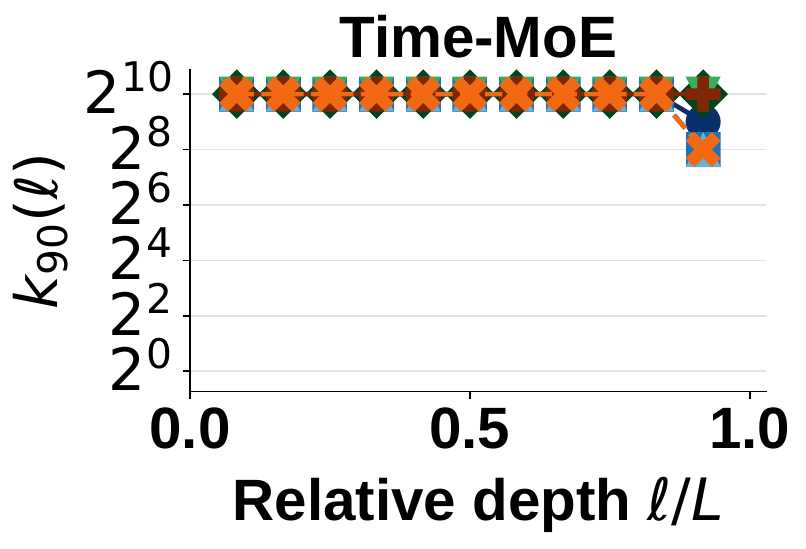}
\end{minipage}
\hfill
\begin{minipage}[t]{0.325\linewidth}
\centering
\includegraphics[width=\linewidth]{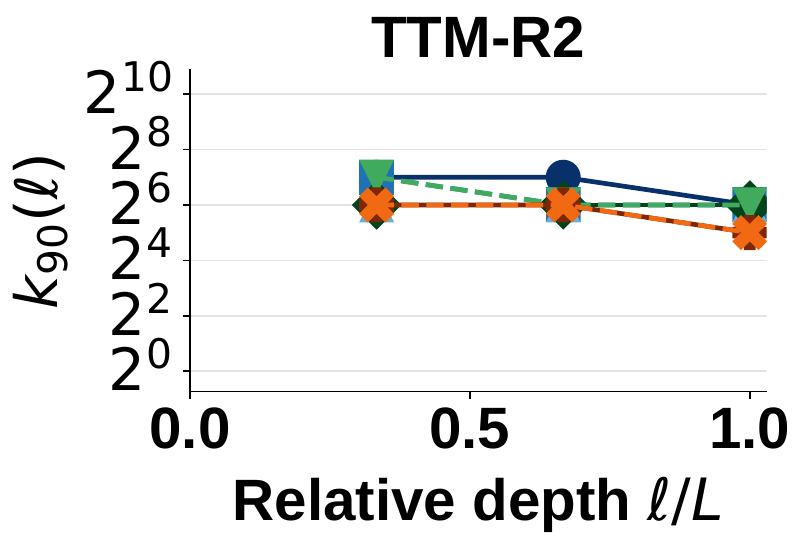}
\end{minipage}
\hfill
\begin{minipage}[t]{0.325\linewidth}
\centering
\includegraphics[width=\linewidth]{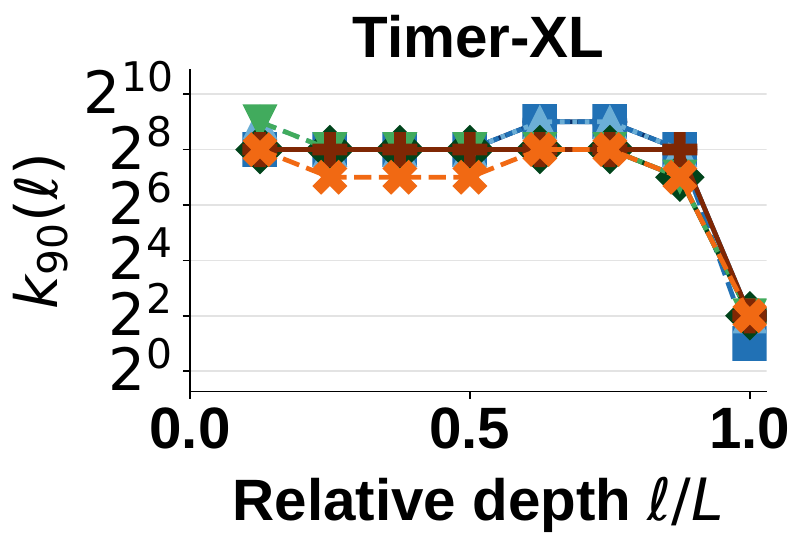}
\end{minipage}

\vspace{2pt}

\includegraphics[width=\linewidth]{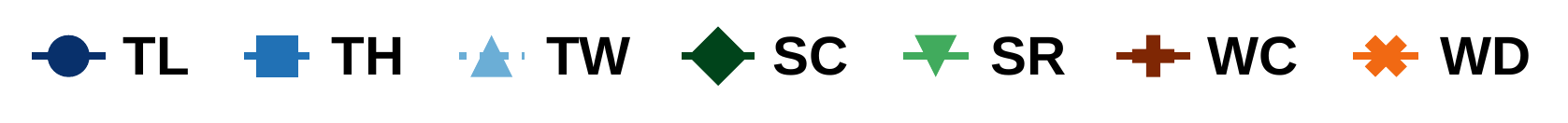}

\caption{
Shared forecast-control rank across relative depth for seven jointly identified parameters and nine TSFMs.
Values at the top grid point~\(2^{10}\) include unreached response-retention thresholds and therefore represent lower bounds where censored.
}
\label{fig:jspace_compact_multi}
\end{figure}

\begin{figure}[t]
\centering

\begin{minipage}[t]{0.325\linewidth}
\centering
\includegraphics[width=\linewidth]{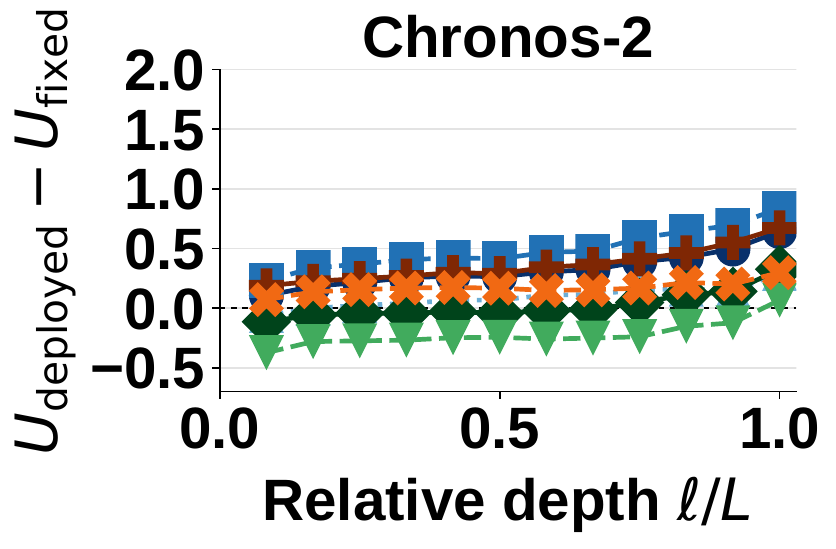}
\end{minipage}
\hfill
\begin{minipage}[t]{0.325\linewidth}
\centering
\includegraphics[width=\linewidth]{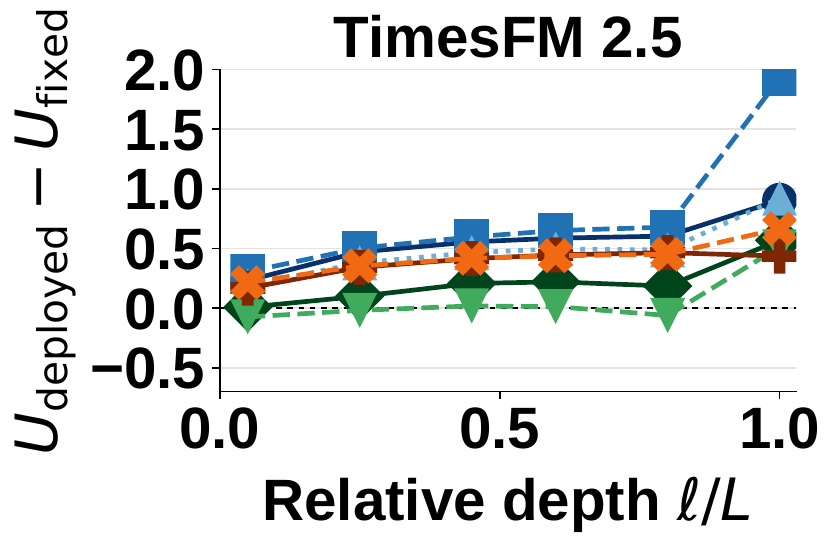}
\end{minipage}
\hfill
\begin{minipage}[t]{0.325\linewidth}
\centering
\includegraphics[width=\linewidth]{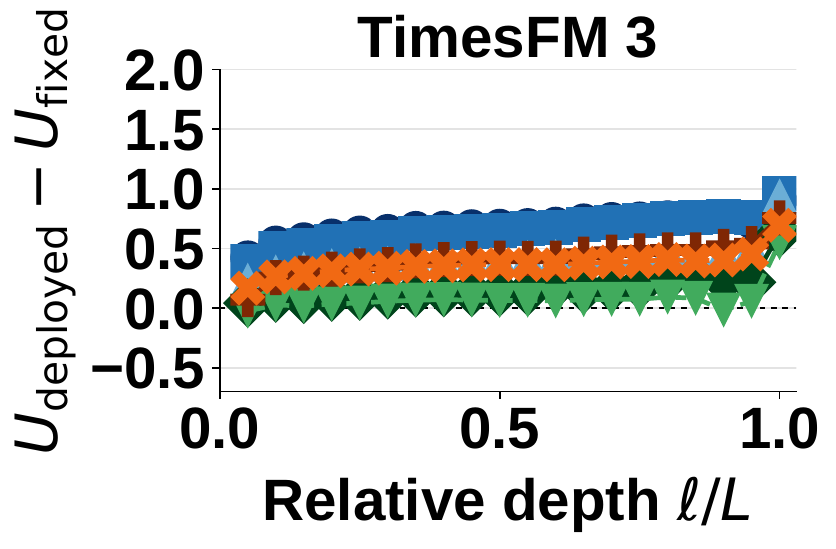}
\end{minipage}

\vspace{2pt}

\begin{minipage}[t]{0.325\linewidth}
\centering
\includegraphics[width=\linewidth]{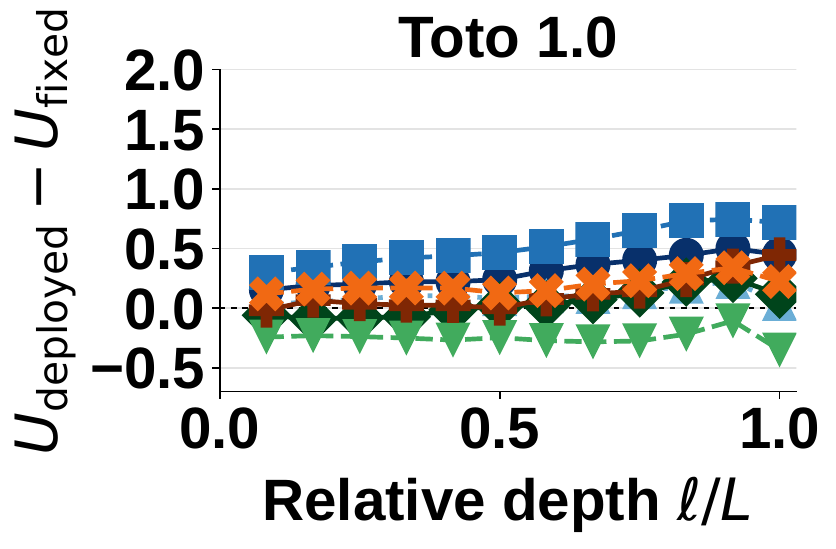}
\end{minipage}
\hfill
\begin{minipage}[t]{0.325\linewidth}
\centering
\includegraphics[width=\linewidth]{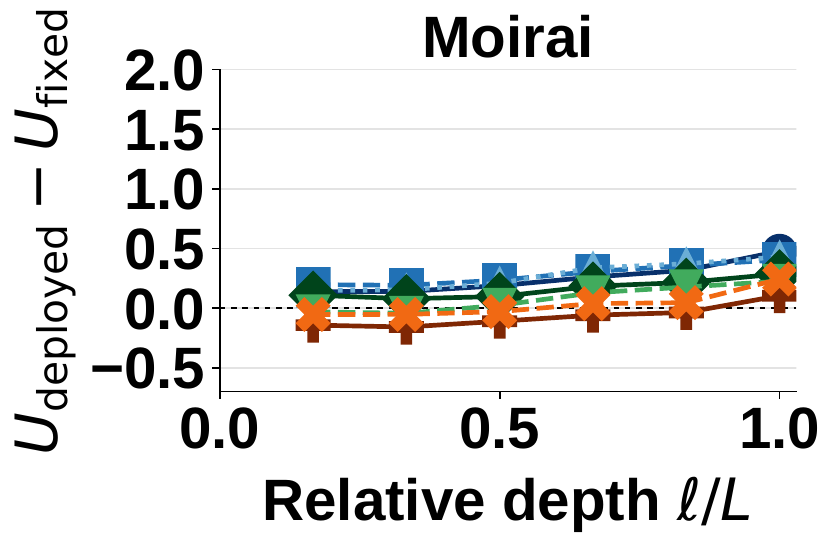}
\end{minipage}
\hfill
\begin{minipage}[t]{0.325\linewidth}
\centering
\includegraphics[width=\linewidth]{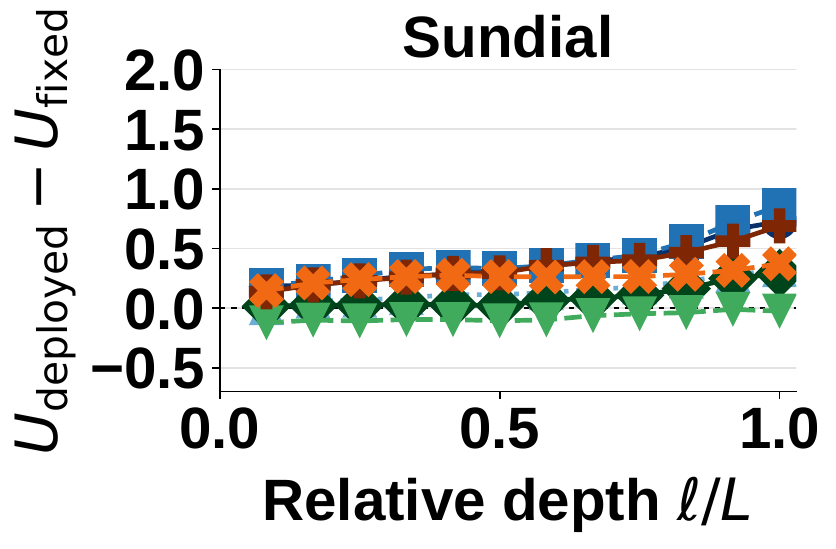}
\end{minipage}

\vspace{2pt}

\begin{minipage}[t]{0.325\linewidth}
\centering
\includegraphics[width=\linewidth]{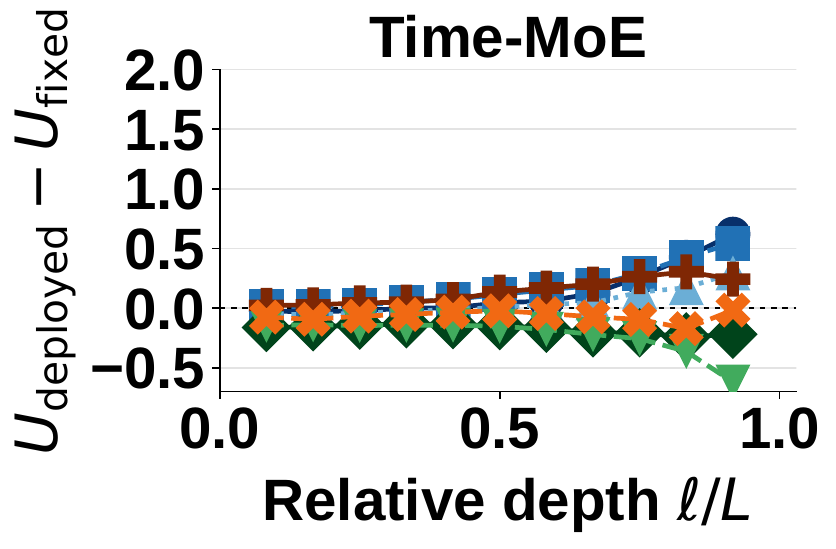}
\end{minipage}
\hfill
\begin{minipage}[t]{0.325\linewidth}
\centering
\includegraphics[width=\linewidth]{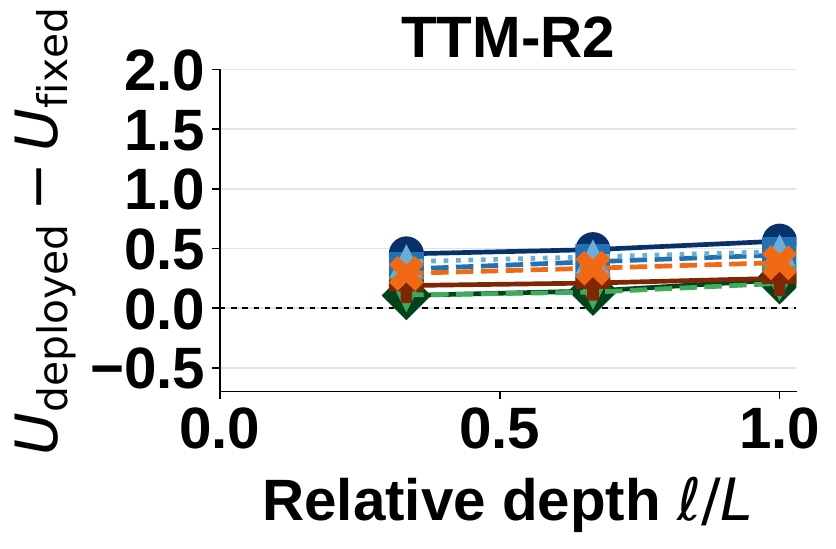}
\end{minipage}
\hfill
\begin{minipage}[t]{0.325\linewidth}
\centering
\includegraphics[width=\linewidth]{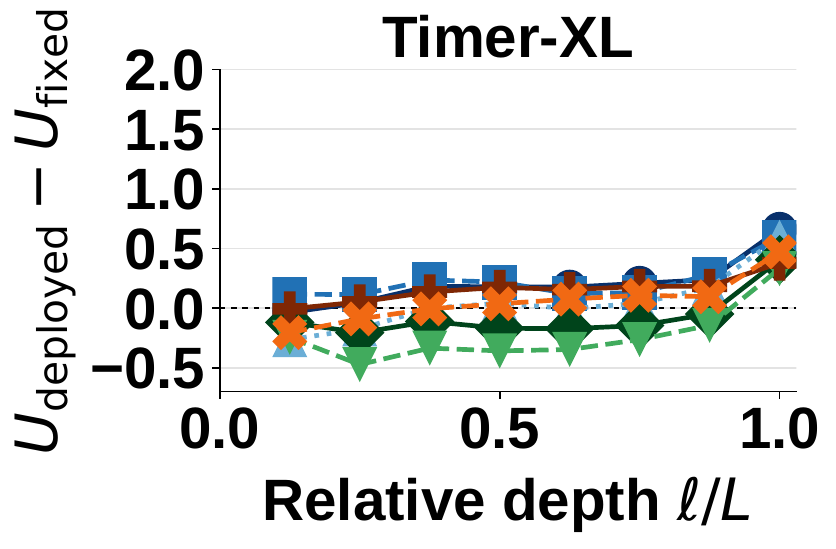}
\end{minipage}

\vspace{2pt}

\includegraphics[width=\linewidth]{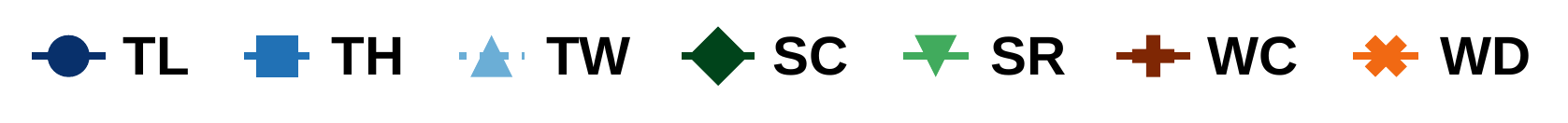}

\caption{
Deployed-over-fixed response gain across validation-admissible layers for seven jointly identified parameters and nine TSFMs.
Positive values indicate an advantage of the input-specific deployed direction over the fixed-direction control.
}
\label{fig:jspace_gain_multi}
\vspace{-0.15in}
\end{figure}

This analysis examines whether the jointly identified parameters admit effective and structured hidden-state interventions across model depth.
\autoref{fig:local_validity_multi} summarizes selected-layer local validity and the fraction of validation-admissible layers, while~\autoref{fig:local_layers_multi} shows the complete layer-wise local-response profiles.
\autoref{fig:jspace_compact_multi} then characterizes how the required shared causal-subspace rank changes with depth.
Finally, \autoref{fig:jspace_gain_multi} evaluates whether input-specific deployed directions improve over the fixed-direction control across admissible layers.
The calibrated intervention settings are summarized in~\autoref{tab:geometry_settings}, and these analyses provide the detailed causal-geometry results underlying~\S\ref{sec:eval_multi}.

\textbf{Local Causal Directions Remain Effective Across Depth.}
\autoref{fig:local_validity_multi} shows that the validation local response lies between~\(0.970\) and~\(1.030\) at all~630 measured layers, so every measured layer satisfies the local-validity criterion.
At the validation-selected layers, the corresponding test responses also remain between~\(0.970\) and~\(1.030\).
Across all measured test layers, the broader range is~\(0.942\) to~\(1.035\), indicating only limited deviation from the reference response outside the selected configurations.
The calibrated intervention step converges at every layer, as reported in~\autoref{tab:geometry_settings}.
These results show that the Jacobian-derived directions remain effective after nonlinear evaluation rather than succeeding only under the first-order approximation.

\textbf{Layer-Wise Profiles Show Broad Local Validity Rather Than Isolated Valid Layers.}
\autoref{fig:local_layers_multi} provides the full depth profiles behind the aggregate validity results.
The local response stays near the unit target across nearly the entire depth of each model and parameter, rather than peaking only at a small number of layers.
This stability indicates that hidden-state directions that produce the reference response are available across a broad portion of the representation hierarchy.
The selected layer therefore identifies a configuration for the subsequent shared-subspace and deployed analyses rather than creating the local response through a narrow layer-selection effect.

\textbf{The Shared Causal Subspace Contracts with Depth.}
\autoref{fig:jspace_compact_multi} shows lower recorded rank-grid outcomes at the last measured layer in~58 of~63 model--parameter cells.
The median displayed rank or lower bound decreases from~\(1{,}024\) at the first measured layer to~\(8\) at the last measured layer.
This contraction indicates that later representations often require fewer shared directions to retain the local reference-aligned response.
Values at the top grid point remain censored when the~\(90\%\) response-retention threshold is not reached, so unchanged values of~\(1{,}024\) do not imply unchanged true subspace dimension.
Time-MoE~(\texttt{ME}) is the clearest exception, with final-layer displayed ranks or lower bounds between~\(256\) and~\(1{,}024\), while TTM-R2~(\texttt{TT}) provides only three measured layers.
The dominant pattern nevertheless indicates increasing concentration of shared causal-subspace structure with depth across the multi-parameter panel.

\textbf{Input-Specific Coordinates Usually Improve on a Fixed Direction.}
\autoref{fig:jspace_gain_multi} compares the deployed coordinate-readout intervention with the fixed-direction control across validation-admissible layers.
The median deployed-minus-fixed response difference is~\(+0.19\), and the deployed direction remains better at the final admissible layer in~58 of~63 model--parameter cells.
For TTM-R2~(\texttt{TT}), the deployed direction exceeds the fixed direction at every admissible layer for all seven parameters, with a minimum advantage of~\(+0.11\), despite its weak matched-input responses.
This separation indicates that the unperturbed hidden state contains information useful for selecting input-specific coordinates even when the original input intervention does not produce the desired forecast response.
The advantage is not uniform across all models and depths.
Time-MoE~(\texttt{ME}) and Timer-XL~(\texttt{TX}) have only two and one parameters, respectively, with positive deployed-over-fixed gain at every admissible layer, with minimum differences of~\(-0.62\) and~\(-0.47\).

Together, the four analyses separate three properties of the multi-parameter causal geometry.
Local directions establish that the selected hidden states can support the reference response, shared-subspace results show that these directions often become more compact with depth, and deployed-over-fixed gains show that input-specific coordinates can often be inferred from the unperturbed representation.
The jointly identified parameters therefore retain broadly valid hidden-state response pathways whose shared structure becomes more concentrated with depth, while coordinate prediction remains model- and parameter-dependent across the evaluated multi-parameter settings.

\begin{figure}[t]
\centering

\includegraphics[width=0.9\linewidth]{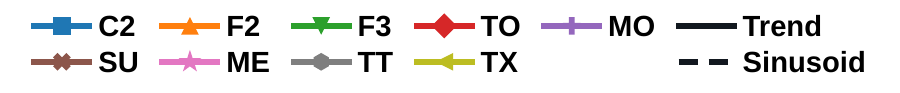}

\begin{minipage}[t]{0.32\linewidth}
\centering
\includegraphics[width=\linewidth]{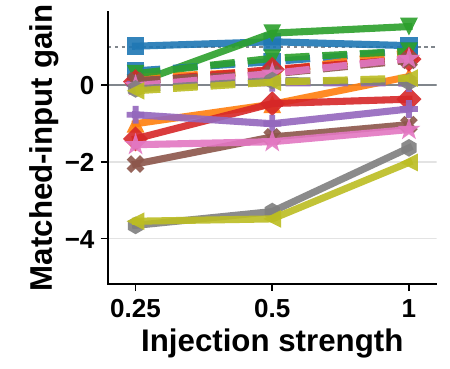}
\\[-0.2em]
{\small (a) ETTh1}
\end{minipage}
\hfill
\begin{minipage}[t]{0.32\linewidth}
\centering
\includegraphics[width=\linewidth]{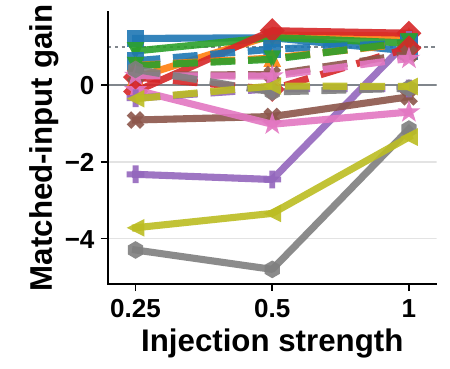}
\\[-0.2em]
{\small (b) Electricity}
\end{minipage}
\hfill
\begin{minipage}[t]{0.32\linewidth}
\centering
\includegraphics[width=\linewidth]{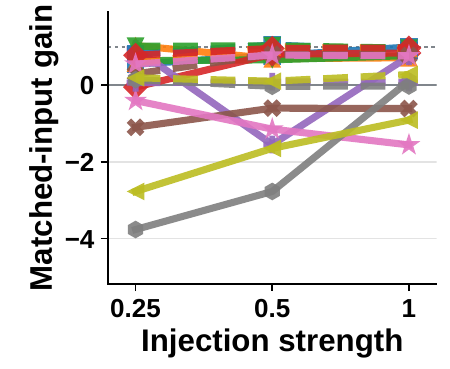}
\\[-0.2em]
{\small (c) Weather}
\end{minipage}

\caption{
Matched-input responses to controlled linear trend and sinusoid injections on real-series backgrounds.
Injection strength is component root-mean-square~(RMS) amplitude relative to context standard deviation for all three evaluated real datasets.
}
\label{fig:app_real_all}
\end{figure}

\begin{figure}[t]
    \centering
    \includegraphics[width=\linewidth]{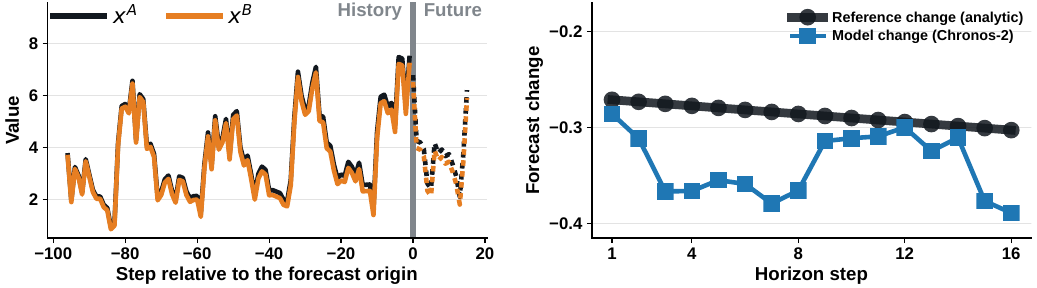}
    \caption{
    Illustrative ETTh1 linear trend injection and Chronos-2~(\texttt{C2}) forecast-change traces at injection strength~\(0.5\) for one representative matched input pair.
    }
    \label{fig:revision_real_example}
    \vspace{-0.1in}
\end{figure}

\subsection{Additional Results on Real-Series Backgrounds}
\label{app:eval_real}

This subsection extends the controlled real-series evaluation in~\S\ref{sec:eval_real} from ETTh1 to Electricity and Weather.
All datasets use the same linear trend and sinusoid injections.
Injection strength is the component root-mean-square~(RMS) amplitude relative to the context standard deviation, evaluated at~\(0.25\), \(0.5\), and~\(1.0\).
Solid lines denote linear trend and dashed lines denote sinusoid.
\autoref{fig:app_real_all} shows the complete cross-dataset results across all settings.

\textbf{ETTh1.}
Panel~(a) shows substantial model variation under the same injected dynamics.
Several models retain negative linear trend responses across all three injection strengths, while sinusoid responses are generally closer to the expected positive direction.
At strength~\(1.0\), the pooled linear trend gain remains~\(-0.62\), whereas the sinusoid gain reaches~\(0.67\).
Increasing the injected amplitude therefore improves the response for many models but does not eliminate the trend-response mismatch.

\autoref{fig:revision_real_example} shows one representative ETTh1 matched pair at injection strength~\(0.5\).
The left panel shows that the two inputs share the same ETTh1 background and differ only in the controlled linear trend component.
The right panel compares the resulting Chronos-2~(\texttt{C2}) forecast change with the analytic reference response across the forecast horizon.
The model follows the same response direction but remains more negative than the analytic reference across much of the horizon.
This example illustrates that a precisely controlled input change need not produce the corresponding analytic continuation uniformly across forecast steps.

\textbf{Electricity.}
Panel~(b) shows a stronger dependence on injection strength.
Several models have strongly negative linear trend responses at strengths~\(0.25\) and~\(0.5\), but move substantially toward the injected continuation at strength~\(1.0\).
The pooled linear trend gain reaches~\(0.89\) at strength~\(1.0\), although some models remain below the reference response.
Sinusoid responses are generally positive and vary less strongly with injection strength.

\textbf{Weather.}
Panel~(c) shows a similar but model-dependent transition.
The pooled linear trend gain reaches~\(0.71\) at strength~\(1.0\), while sinusoid gain reaches~\(0.77\).
Several models nevertheless retain weak or negative linear trend responses, particularly at smaller injection strengths.
Increasing component amplitude can therefore strengthen the expected response without producing uniform behavior across models.

\textbf{Cross-Dataset Comparison.}
Across the three backgrounds, the same controlled dynamical change produces substantially different forecast responses across models and datasets.
Stronger injections often move the response toward the injected continuation, but both its magnitude and direction remain dependent on the underlying background.
The ETTh1 example in~\autoref{fig:revision_real_example} further shows that this mismatch can vary across the forecast horizon within a single matched pair.
Together, these results extend the main-text finding by showing that high representation accessibility can coexist with weak, reversed, or horizon-dependent forecast responses when controlled dynamics are embedded in real-series backgrounds across the evaluated real-series settings.

\subsection{Cross-Measure and Mechanistic Diagnostics}
\label{app:cross_measure}

The main evaluation shows that representation accessibility, forecast response, and deployed response can diverge for the same Dynamical Parameter.
This subsection tests the robustness of this separation and examines mechanisms underlying model-specific trend responses.
We first compare the three measurements across the model panel and test whether weak response can be explained by ordinary forecasting competence.
We then examine reversed linear trend~(\texttt{TR}) responses using law-identification controls, TimesFM 3~(\texttt{F3}) detrending, continuation behavior, a shape-and-scale decomposition, and normalization replay across several targeted mechanistic diagnostic settings.

\begin{table}[t]
\centering
\caption{
Thresholded agreement and cross-model correlations among representation accessibility, matched-input response, and deployed response.
Counts use original-bank cells with~\(D_{m,j}\geq0.95\), so the trend comparison is exploratory rather than common-regime.
}
\label{tab:cross_measure}
\footnotesize
\setlength{\tabcolsep}{3.0pt}
\renewcommand{\arraystretch}{1.08}

\resizebox{\linewidth}{!}{%
\begin{tabular}{@{}lcccccc@{}}
\toprule
&
\multicolumn{4}{c}{Cells With \(D_{m,j}\geq0.95\)}
&
\multicolumn{2}{c}{Spearman~\(\rho\)}
\\
\cmidrule(lr){2-5}
\cmidrule(l){6-7}

Law
& All
& \(\bar U_{m,j}<0\)
& \(0\leq\bar U_{m,j}<0.5\)
& \(\bar U_{m,j}>1.5\)
& \(D_{m,j},\bar U_{m,j}\)
& \(U_{\mathrm{deployed}},\bar U_{m,j}\)
\\
\midrule

Linear trend (\texttt{TR})
& 9 & 3 & 2 & 0 & 0.62 & 0.78 \\

Oscillation (\texttt{OS})
& 0 & -- & -- & -- & 0.92 & 0.78 \\

Mean reversion (\texttt{MR})
& 8 & 0 & 1 & 0 & 0.92 & \(-0.03\) \\

Saturating growth (\texttt{SG})
& 9 & 2 & 2 & 3 & 0.18 & 0.45 \\

Damped oscillation (\texttt{DO})
& 8 & 0 & 2 & 0 & 0.38 & \(-0.20\) \\

Seasonal autoregression (\texttt{SA})
& 0 & -- & -- & -- & 0.82 & 0.32 \\

Transient pulse (\texttt{TP})
& 9 & 0 & 8 & 0 & 0.78 & \(-0.03\) \\

\midrule
\textbf{All}
& \textbf{43}
& \textbf{5}
& \textbf{15}
& \textbf{3}
& \textbf{0.33}
& \textbf{0.37}
\\

\bottomrule
\end{tabular}%
}
\end{table}

\subsubsection{Cross-Measure Robustness}
\label{app:cross_measure_controls}

\begin{figure}[t]
\centering
\includegraphics[width=\linewidth]{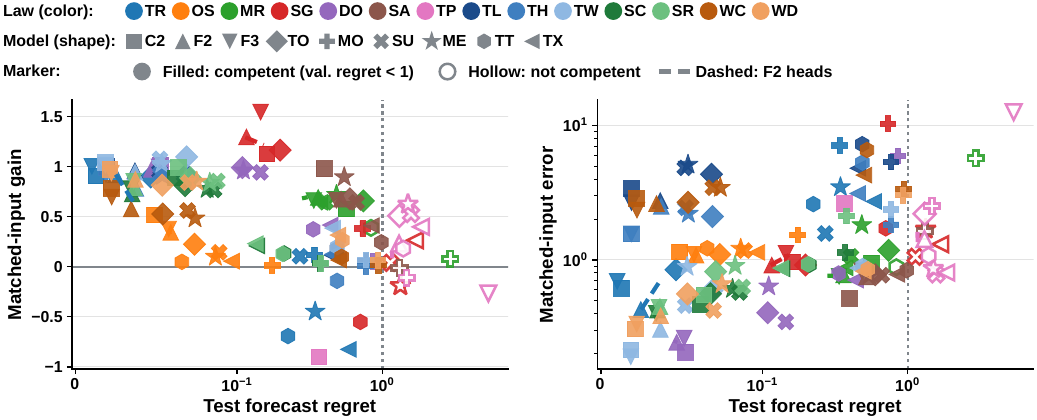}
\caption{
Forecast regret versus matched-input response fidelity across the single-parameter panel.
The figure highlights validation-selected competent forecasters and compares the two TimesFM 2.5~(\texttt{F2}) forecast heads under the same response evaluation setting.
}
\label{fig:revision_quality}
\end{figure}

This analysis tests whether the separation among representation accessibility, matched-input forecast response, and deployed aligned gain persists across the single-parameter panel and among models with stronger ordinary forecasting performance.
\autoref{tab:cross_measure} summarizes thresholded disagreements and cross-model correlations among the three measurements.
\autoref{tab:revision_competence} and~\autoref{fig:revision_quality} then examine whether weak matched-input response can be explained by ordinary forecast competence.
Because the cross-measure comparison uses the original accessibility datasets, the common-regime results in~\S\ref{sec:eval_response} remain the primary comparison between accessibility and forecast response.

\textbf{The Three Measurements Remain Distinct Across the Panel.}
As shown in~\autoref{tab:cross_measure}, among the~43 model--parameter cells with accessibility~\(D_{m,j}\geq0.95\), five have negative matched-input response, 15 have response between~\(0\) and~\(0.5\), and three exceed~\(1.5\).
These disagreements vary substantially across dynamical laws.
Accessibility and matched-input response are rank-correlated across models for oscillation~(\texttt{OS}), mean reversion~(\texttt{MR}), and seasonal autoregression~(\texttt{SA}), with correlations between~\(0.82\) and~\(0.92\), while the correlation falls to~\(0.38\) for damped oscillation~(\texttt{DO}).
Deployed and matched-input responses also need not agree, indicating that natural forecast expression and response capacity under predicted hidden-state intervention remain distinct measurements across the evaluated model--parameter cells and across different dynamical response regimes.

\textbf{Forecast Competence Does Not Eliminate Response Mismatch.}
We next test whether weak matched-input response simply reflects poor ordinary forecasting.
For each law, normalized forecast regret is defined relative to the conditional reference and fixed baseline as
\[
\frac{
L\left(F_{\theta},y\right)-L\left(F_m^{\star},y\right)
}{
L\left(F_{\mathrm{base}},y\right)-L\left(F_m^{\star},y\right)+\epsilon
},
\]
where~\(L\) is horizon mean-squared error, \(F_m^{\star}\) is the conditional reference, and \(F_{\mathrm{base}}\) is a fixed law-specific baseline.
A separate validation dataset of~\(4{,}096\) histories per law determines the forecast-competence subsets before accessibility and response are evaluated on their respective test datasets.

As shown in~\autoref{fig:revision_quality}, each point represents one model--parameter cell, with color indicating the dynamical law and marker shape indicating the TSFM.
The horizontal axis reports test forecast regret, where smaller values indicate forecasts closer to the conditional reference relative to the fixed baseline.
Filled markers denote cells selected as forecast competent using validation regret below one, while hollow markers denote the remaining cells.
Because competence is selected on the validation dataset, the filled and hollow markers need not fall on opposite sides of the vertical test-regret line.
The highlighted TimesFM 2.5~(\texttt{F2}) markers compare its two forecast heads under the same response evaluation for a controlled within-model head comparison.

The left panel plots matched-input aligned gain~\(g\) against test forecast regret.
A gain near~\(1\) indicates a response with the expected reference-aligned magnitude, a value near~\(0\) indicates little aligned response, and a negative value indicates an opposing response.
Although lower-regret cells tend to have larger gain, cells with similar forecast regret span a wide range of response gains.
Several validation-selected competent cells still have weak or negative gain, showing that accurate ordinary forecasts do not guarantee a reference-aligned response to the parameter intervention.

The right panel plots centered response error~\(e\), where smaller values indicate closer agreement with the full reference trajectory.
Lower-regret cells generally have smaller response error, but substantial variation remains even among the competent cells.
Several filled markers have centered error near or above~\(1\), indicating that their forecast-change trajectories can remain far from the reference despite competent ordinary forecasting.
The overlap of competent and non-competent cells in both panels therefore shows that forecast quality alone does not determine matched-input response fidelity.

Consistent with the scatter plots, test regret and matched-input gain have Spearman correlation~\(\rho=-0.41\), with a~\(95\%\) interval of~\(\left[-0.60,-0.20\right]\).
Among the~49 cells with validation regret below one, five still have negative matched-input gain, while median accessibility remains~\(0.976\).
As summarized in~\autoref{tab:revision_competence}, their median matched-input gain and centered error are~\(0.64\) and~\(0.92\), respectively.
The stricter regret-below-\(0.5\) and within-law best-tercile subsets increase median gain only to~\(0.65\) and~\(0.69\), while centered errors remain~\(0.92\) and~\(0.83\).
Ordinary forecasting competence therefore reduces, but does not remove, the separation between parameter accessibility and response fidelity across the evaluated single-parameter model panel.

\begin{table}[t]
\centering
\caption{
Single-parameter cells stratified by forecast competence.
\(D_{m,j}\) denotes original-bank representation accessibility, while \(g\) and \(e\) are pooled matched-input response metrics.
Subset thresholds use validation regret, with metrics evaluated on the corresponding test banks.
}
\label{tab:revision_competence}
\footnotesize
\setlength{\tabcolsep}{5pt}
\renewcommand{\arraystretch}{1.08}

\begin{tabular}{@{}lrrrr@{}}
\toprule
Subset
& Cells
& Median \(D_{m,j}\)
& Median \(g\)
& Median \(e\)
\\
\midrule
All
& 63
& 0.985
& 0.42
& 1.06
\\

Regret \(<1\)
& 49
& 0.976
& 0.64
& 0.92
\\

Regret \(<0.5\)
& 36
& 0.990
& 0.65
& 0.92
\\

Best tercile
& 21
& 0.989
& 0.69
& 0.83
\\

\bottomrule
\end{tabular}
\end{table}

\subsubsection{Trend-Response Mechanisms}
\label{app:trend_mechanisms}

The reversed linear trend~(\texttt{TR}) responses provide a concrete setting for examining why accessible parameter information need not produce the expected forecast change.
We first test whether the reversing models confuse trend with other dynamical laws using~\autoref{tab:lawid_confusion}.
We then examine the TimesFM 3~(\texttt{F3}) detrending pathway in~\autoref{tab:tfm3_gate} and compare model continuation behavior in~\autoref{tab:continuation}.
Next, \autoref{tab:ddecomp} separates the standardized-shape and window-scale contributions to the trend response.
Finally, \autoref{tab:norm_replay} tests whether the standard normalization pipeline conceals or alters the observed response.
Together, these diagnostics examine the mechanisms underlying the trend-response patterns reported in~\S\ref{sec:eval_response}.

\textbf{(I)~Reversed Trend Response Is Not Explained by Law Confusion.}
\autoref{tab:lawid_confusion} uses the trend row of the thirteen-law identification experiment in~\S\ref{sec:eval_law} to test whether reversed trend responses coincide with confusion between linear trend~(\texttt{TR}) and related dynamics.
No model assigns more than~\(0.7\%\) of true trend sequences to mean reversion~(\texttt{MR}), and the three models with negative trend response assign none.
Time-MoE~(\texttt{ME}), TTM-R2~(\texttt{TT}), and Timer-XL~(\texttt{TX}) classify~\(94.0\%\), \(87.0\%\), and~\(96.3\%\) of trend sequences correctly, respectively.
For TTM-R2~(\texttt{TT}), the remaining errors fall primarily on random walk with drift~(\texttt{RW}) at~\(4.9\%\) and the remaining laws at~\(8.1\%\), rather than on mean reversion.
The validation results further show that trend identity becomes substantially more accessible with depth for several models.
For example, Time-MoE~(\texttt{ME}) increases from~\(50.4\%\) trend accuracy at the first layer to~\(94.6\%\) at the selected layer, while TimesFM 3~(\texttt{F3}) increases from~\(62.8\%\) to~\(94.8\%\) and Moirai~(\texttt{MO}) from~\(59.1\%\) to~\(92.2\%\).
Thus, models with weak or reversed trend responses can still develop hidden representations that distinguish the trend law accurately across multiple depths and model architectures.
The response reversal therefore does not appear to arise from a general failure to identify linear-trend dynamics in the inspected representations.

\begin{table}[t]
\centering
\caption{
Trend-law classification and validation accuracy across nine TSFMs.
Classification columns report the percentage of true linear-trend~(\texttt{TR}) sequences assigned to each law group.
}
\label{tab:lawid_confusion}
\footnotesize
\setlength{\tabcolsep}{3.0pt}
\renewcommand{\arraystretch}{1.08}

\resizebox{\linewidth}{!}{%
\begin{tabular}{@{}lrrrrrrc@{}}
\toprule
&
\multicolumn{4}{c}{True Trend Classified As~(\%)}
&
\multicolumn{2}{c}{Validation Trend Accuracy~(\%)}
&
\\
\cmidrule(lr){2-5}
\cmidrule(lr){6-7}

Model
& Trend
& Mean Rev.
& Rand. Walk
& Other
& Layer 1
& Selected
& Selected Layer
\\
\midrule

Chronos-2 (\texttt{C2})
& 97.2 & 0.0 & 1.4 & 1.4
& 93.7 & 97.1 & 10/12
\\

TimesFM 2.5 (\texttt{F2})
& 97.6 & 0.0 & 0.9 & 1.5
& 92.8 & 97.2 & 10/20
\\

TimesFM 3 (\texttt{F3})
& 94.6 & 0.7 & 0.7 & 4.0
& 62.8 & 94.8 & 16/20
\\

Toto-Open-Base 1.0 (\texttt{TO})
& 96.1 & 0.0 & 2.0 & 1.9
& 82.8 & 96.2 & 6/12
\\

Moirai (\texttt{MO})
& 92.9 & 0.0 & 4.4 & 2.7
& 59.1 & 92.2 & 5/6
\\

Sundial (\texttt{SU})
& 96.2 & 0.0 & 1.9 & 1.9
& 82.3 & 97.5 & 12/12
\\

Time-MoE (\texttt{ME})
& 94.0 & 0.0 & 3.5 & 2.6
& 50.4 & 94.6 & 12/12
\\

TTM-R2 (\texttt{TT})
& 87.0 & 0.0 & 4.9 & 8.1
& 82.0 & 87.5 & 3/3
\\

Timer-XL (\texttt{TX})
& 96.3 & 0.0 & 2.3 & 1.3
& 84.9 & 95.0 & 5/8
\\

\bottomrule
\end{tabular}%
}
\end{table}

\textbf{(II)~TimesFM 3 Uses a Detrending Pathway That Bypasses the Inspected State.}
TimesFM 3~(\texttt{F3}) provides the opposite pattern from the reversing models.
On the wider trend-response histories, its slope accessibility falls to~\(D_{m,S}=0.372\), while its conditional-reference-normalized response reaches~\(\bar U_{m,S}=1.01\).
\autoref{tab:tfm3_gate} shows that the detrending gate fires on~\(48.0\%\) of the~\(32{,}768\) response histories, with a sharp dependence on absolute slope.
The gate remains inactive for~\(\left|S\right|<0.2\), fires on only~\(17.6\%\) of histories in~\(\left[0.200,0.234\right)\), rises to~\(92.5\%\) in~\(\left[0.234,0.300\right)\), and reaches~\(100\%\) for~\(\left|S\right|\geq0.300\).
This transition explains why the wider slope distribution used for forecast-response evaluation exposes a different representation regime from the narrow-slope accessibility setting.
When the gate fires, an ordinary least-squares trend is removed before the transformer, while the fitted trend is extrapolated and added back outside the inspected hidden-state pathway.

The gate-stratified accessibility results further support this mechanism.
For histories on which the gate fires, linear slope accessibility remains near zero across layers, ranging from~\(-0.018\) to~\(0.001\).
For histories on which the gate remains inactive, accessibility instead ranges from~\(0.990\) to~\(0.993\).
These two regimes together produce the substantially lower aggregate accessibility of~\(0.372\) on the response-history distribution.
Disabling the detrending step restores accessibility to~\(0.998\)--\(0.999\) across layers, including~\(0.999\) on histories that would otherwise activate the gate.
The detrending pathway therefore accounts for the measured loss of slope accessibility in the inspected transformer states while preserving a separate route through which the fitted trend contributes to the forecast.

\begin{table}[t]
\centering
\caption{
TimesFM 3~(\texttt{F3}) detrending-gate firing rate across~\(32{,}768\) trend histories grouped by absolute slope.
The gate remains inactive at small slopes and becomes nearly universal once the slope exceeds the analytic crossover region across the evaluated response-history distribution.
}
\label{tab:tfm3_gate}
\footnotesize
\setlength{\tabcolsep}{6pt}
\renewcommand{\arraystretch}{1.08}

\begin{tabular}{@{}lrr@{}}
\toprule
Absolute-Slope Bin~\(\left|S\right|\)
& Histories
& Gate Fires
\\
\midrule

All
& 32{,}768
& 48.0\%
\\

\(\left[0.000,0.100\right)\)
& 7{,}324
& 0.0\%
\\

\(\left[0.100,0.200\right)\)
& 7{,}302
& 0.0\%
\\

\(\left[0.200,0.234\right)\)
& 2{,}480
& 17.6\%
\\

\(\left[0.234,0.300\right)\)
& 4{,}798
& 92.5\%
\\

\(\left[0.300,0.450\right)\)
& 10{,}864
& 100.0\%
\\

\bottomrule
\end{tabular}
\end{table}

\textbf{(III)~Continuation Behavior Tracks the Direction of Trend Response.}
The trend-response experiment measures how the forecast changes under a finite slope intervention.
A complementary diagnostic measures the model's continuation behavior without intervention.
For an unintervened window, we regress the tilt of the standardized forecast on the standardized input slope and denote the resulting continuation gain by~\(k\).
A value of~\(k=1\) indicates a unit average continuation component, \(k=0\) indicates no such component, and~\(k<0\) indicates an opposing component.

As shown in~\autoref{tab:continuation}, continuation behavior differs strongly across models.
Chronos-2~(\texttt{C2}), TimesFM 2.5~(\texttt{F2}), TimesFM 3~(\texttt{F3}), and Toto-Open-Base 1.0~(\texttt{TO}) have positive continuation gains between~\(0.79\) and~\(0.94\).
Time-MoE~(\texttt{ME}), TTM-R2~(\texttt{TT}), and Timer-XL~(\texttt{TX}) instead have gains of~\(-0.82\), \(-0.86\), and~\(-1.36\), respectively.
Across the nine models, continuation gain and conditional-reference-normalized trend response are rank-correlated at~\(0.97\), showing that the intervention response closely follows the models' underlying continuation behavior across the evaluated trend-response model panel.
The slope-bin results further show that this behavior varies with the baseline slope, particularly for the models with opposing continuation.

TimesFM 2.5~(\texttt{F2}) further isolates the role of the output head.
Its pinball-trained median head has continuation gain~\(0.81\), while its MSE-trained mean head has gain~\(0.54\), despite sharing the same trunk and input representation.
Their matched-input trend gains are~\(0.92\) and~\(0.65\), respectively.
The common trunk therefore does not determine a unique continuation policy, since the two forecast heads produce different continuation and intervention responses.

\begin{table}[t]
\centering
\caption{
Continuation gain across nine TSFMs, reported overall and by absolute-slope bin.
The TimesFM 2.5~(\texttt{F2}) rows additionally compare its median and mean forecast heads.
}
\label{tab:continuation}
\footnotesize
\setlength{\tabcolsep}{3.2pt}
\renewcommand{\arraystretch}{1.08}

\begin{tabular}{@{}lrrrrrr@{}}
\toprule
&
&
\multicolumn{3}{c}{Continuation Gain \(k\) by \(\left|S\right|\)}
&
&
\\
\cmidrule(lr){3-5}

Model
& Overall \(k\)
& \(\left[0,0.05\right)\)
& \(\left[0.05,0.2\right)\)
& \(\left[0.2,0.45\right)\)
& \(k^{\star}\)
& \(\bar U_{\mathrm{TR}}\)
\\
\midrule

Chronos-2 (\texttt{C2})
& \(+0.87\)
& \(+0.54\)
& \(+0.84\)
& \(+0.89\)
& \(+1.00\)
& \(+0.91\)
\\

TimesFM 2.5 (\texttt{F2})
& \(+0.81\)
& \(+0.50\)
& \(+0.69\)
& \(+0.86\)
& \(+1.00\)
& \(+0.92\)
\\
\quad Median head (pinball)
& \(+0.81\)
& --
& --
& --
& --
& --
\\
\quad Mean head (MSE)
& \(+0.54\)
& --
& --
& --
& --
& --
\\

TimesFM 3 (\texttt{F3})
& \(+0.94\)
& \(+0.32\)
& \(+0.80\)
& \(+0.99\)
& \(+1.00\)
& \(+1.01\)
\\

Toto-Open-Base 1.0 (\texttt{TO})
& \(+0.79\)
& \(+0.14\)
& \(+0.70\)
& \(+0.84\)
& \(+1.00\)
& \(+0.90\)
\\

Moirai (\texttt{MO})
& \(-0.20\)
& \(-0.61\)
& \(-0.50\)
& \(-0.09\)
& \(+1.00\)
& \(+0.12\)
\\

Sundial (\texttt{SU})
& \(-0.18\)
& \(-1.05\)
& \(-0.56\)
& \(-0.02\)
& \(+1.00\)
& \(+0.11\)
\\

Time-MoE (\texttt{ME})
& \(-0.82\)
& \(-1.89\)
& \(-1.53\)
& \(-0.54\)
& \(+1.00\)
& \(-0.45\)
\\

TTM-R2 (\texttt{TT})
& \(-0.86\)
& \(-2.89\)
& \(-1.10\)
& \(-0.75\)
& \(+1.00\)
& \(-0.69\)
\\

Timer-XL (\texttt{TX})
& \(-1.36\)
& \(-1.00\)
& \(-1.76\)
& \(-1.21\)
& \(+1.00\)
& \(-0.82\)
\\

\bottomrule
\end{tabular}%
\end{table}

\textbf{(IV)~Shape and Scale Carry Complementary Parts of the Trend Response.}
For models using whole-window standardization, the end-to-end trend response contains both a change in the standardized forecast shape and a contribution introduced when the forecast is returned to the original window scale.
Writing~\(u_A\) and~\(u'_B\) for the two standardized inputs, \(g\) for the map from standardized input to standardized forecast, and~\(\sigma_A\) and~\(\sigma_B\) for the window standard deviations, the raw forecast change can be written as the sum of two complementary components:
\[
\begin{aligned}
\sigma_B g\left(u'_B\right)-\sigma_A g\left(u_A\right)
&=
\underbrace{
\sigma_A
\left[
g\left(u'_B\right)-g\left(u_A\right)
\right]
}_{\text{shape}}
\\
&\quad+
\underbrace{
\left(\sigma_B-\sigma_A\right)
g\left(u'_B\right)
}_{\text{scale}}.
\end{aligned}
\]
The first term captures the network response to the standardized-shape change across the forecast horizon.
The second captures the contribution of changing the output scale while holding the arm-B standardized forecast fixed.
This end-to-end decomposition is distinct from the deployed coordinate-readout intervention defined in~\S\ref{sec:causal_geometry}.

As shown in~\autoref{tab:ddecomp}, the relative contributions of the two channels vary substantially across models and slope regimes.
Chronos-2~(\texttt{C2}) remains positive in both channels, with overall shape and scale contributions of~\(+0.39\) and~\(+0.53\), producing a live response of~\(+0.92\) across the evaluated trend-response slope regimes.
The three reversing models instead have negative overall scale contributions, ranging from~\(-0.22\) for Time-MoE~(\texttt{ME}) to~\(-0.77\) for Timer-XL~(\texttt{TX}).

In the lowest absolute-slope bin, the standardized-shape contribution is itself negative for all three reversing models.
It reaches~\(-1.60\) for Time-MoE~(\texttt{ME}), \(-2.93\) for TTM-R2~(\texttt{TT}), and~\(-0.77\) for Timer-XL~(\texttt{TX}).
The corresponding scale contributions are smaller in magnitude, showing that the strong reversal in this regime arises primarily from the standardized forecast response rather than from output rescaling alone across these low-slope intervention settings.

The balance changes as the baseline slope increases.
In the highest slope bin, Time-MoE~(\texttt{ME}) and Timer-XL~(\texttt{TX}) have positive shape contributions of~\(+0.42\) and~\(+0.93\), while negative scale contributions of~\(-0.33\) and~\(-1.03\) reduce the corresponding live responses to~\(+0.09\) and~\(-0.10\).
TTM-R2~(\texttt{TT}) shows the same competition in the middle bin, where a positive shape contribution of~\(+0.57\) is offset by a scale contribution of~\(-0.61\), leaving a live response near zero at~\(-0.04\).
The decomposition therefore shows that trend-response reversal can arise through different balances of standardized forecast behavior and output scaling, with those balances changing across models and slope regimes throughout the evaluated trend-response conditions.

\begin{table}[t]
\centering
\caption{
Trend response decomposed into standardized-shape and window-scale contributions across slope bins.
Rows report live response together with the corresponding shape and scale components for each evaluated model and slope regime.
}
\label{tab:ddecomp}
\small
\setlength{\tabcolsep}{4pt}
\renewcommand{\arraystretch}{1.08}

\begin{tabular}{@{}llrrrr@{}}
\toprule
Model
& Absolute-Slope Bin~\(\left|S\right|\)
& Pairs
& Live
& Shape
& Scale
\\
\midrule

Chronos-2 (\texttt{C2})
& \(\left[0,0.05\right)\)
& 122
& \(+0.86\)
& \(+0.76\)
& \(+0.09\)
\\

&
\(\left[0.05,0.2\right)\)
& 357
& \(+0.96\)
& \(+0.58\)
& \(+0.38\)
\\

&
\(\left[0.2,0.45\right)\)
& 545
& \(+0.91\)
& \(+0.18\)
& \(+0.73\)
\\

\quad All
&
& 1{,}024
& \(+0.92\)
& \(+0.39\)
& \(+0.53\)
\\

\addlinespace[0.3em]

Time-MoE (\texttt{ME})
& \(\left[0,0.05\right)\)
& 122
& \(-1.33\)
& \(-1.60\)
& \(+0.27\)
\\

&
\(\left[0.05,0.2\right)\)
& 357
& \(-0.98\)
& \(-0.74\)
& \(-0.23\)
\\

&
\(\left[0.2,0.45\right)\)
& 545
& \(+0.09\)
& \(+0.42\)
& \(-0.33\)
\\

\quad All
&
& 1{,}024
& \(-0.45\)
& \(-0.23\)
& \(-0.22\)
\\

\addlinespace[0.3em]

TTM-R2 (\texttt{TT})
& \(\left[0,0.05\right)\)
& 237
& \(-3.22\)
& \(-2.93\)
& \(-0.29\)
\\

&
\(\left[0.05,0.2\right)\)
& 709
& \(-0.04\)
& \(+0.57\)
& \(-0.61\)
\\

&
\(\left[0.2,0.45\right)\)
& 1{,}102
& \(-0.56\)
& \(-0.10\)
& \(-0.46\)
\\

\quad All
&
& 2{,}048
& \(-0.69\)
& \(-0.20\)
& \(-0.49\)
\\

\addlinespace[0.3em]

Timer-XL (\texttt{TX})
& \(\left[0,0.05\right)\)
& 237
& \(-0.91\)
& \(-0.77\)
& \(-0.13\)
\\

&
\(\left[0.05,0.2\right)\)
& 709
& \(-1.92\)
& \(-1.35\)
& \(-0.57\)
\\

&
\(\left[0.2,0.45\right)\)
& 1{,}102
& \(-0.10\)
& \(+0.93\)
& \(-1.03\)
\\

\quad All
&
& 2{,}048
& \(-0.83\)
& \(-0.06\)
& \(-0.77\)
\\

\bottomrule
\end{tabular}%
\end{table}

\textbf{(V)~Normalization Replay Does Not Reveal a Hidden Positive Trend Response.}
We finally test whether whole-window normalization conceals the expected linear-trend~(\texttt{TR}) response.
The replay covers Chronos-2~(\texttt{C2}), Moirai~(\texttt{MO}), Sundial~(\texttt{SU}), Time-MoE~(\texttt{ME}), TTM-R2~(\texttt{TT}), and Timer-XL~(\texttt{TX}), whose evaluated interfaces standardize the context using its mean and standard deviation.
For the frozen-statistics replay, arm~B is forecast using the normalization statistics of arm~A.
An amplitude-only control separately measures the response induced by presenting the arm-A shape at the arm-B amplitude under the same frozen statistics.

As shown in~\autoref{tab:norm_replay}, freezing the normalization statistics does not recover a positive trend response for the weak or reversing models.
Moirai~(\texttt{MO}) changes from~\(+0.12\) to~\(-0.46\), Sundial~(\texttt{SU}) from~\(+0.11\) to~\(-1.69\), Time-MoE~(\texttt{ME}) from~\(-0.45\) to~\(-1.25\), TTM-R2~(\texttt{TT}) from~\(-0.69\) to~\(-2.70\), and Timer-XL~(\texttt{TX}) from~\(-0.82\) to~\(-2.40\).
Chronos-2~(\texttt{C2}) changes only modestly from~\(+0.91\) to~\(+0.81\).
Thus, holding the normalization statistics fixed generally strengthens rather than removes the weak or reversed behavior.

The amplitude-only control closely accounts for these changes.
Its responses are~\(-0.56\), \(-1.76\), \(-0.80\), \(-2.02\), and~\(-1.58\) for the five weak or reversing models, compared with only~\(-0.10\) for Chronos-2~(\texttt{C2}).
After subtracting the live and amplitude-only contributions from the frozen-statistics response, the residual remains within~\(0.03\) of zero for every model.
The large changes under frozen statistics are therefore closely associated with the accompanying amplitude perturbation rather than with recovery of an expected slope-response pathway.
The normalization replay consequently provides no evidence that the standard normalization pipeline is concealing a positive trend response in these models across the evaluated normalization replay settings.

\begin{table}[t]
\centering
\caption{
Trend response under the standard normalization pipeline, frozen statistics, and an amplitude-only control.
The residual reports the difference between the frozen-statistics response and the sum of the live and amplitude-only responses across all evaluated normalization replay models.
}
\label{tab:norm_replay}
\footnotesize
\setlength{\tabcolsep}{4.0pt}
\renewcommand{\arraystretch}{1.08}

\begin{tabular}{@{}lrrrrr@{}}
\toprule
Model
& Live Pairs
& Live
& Frozen
& Amplitude
& Residual
\\
\midrule

Chronos-2 (\texttt{C2})
& 32{,}768
& \(+0.91\)
& \(+0.81\)
& \(-0.10\)
& \(-0.00\)
\\

Moirai (\texttt{MO})
& 131{,}072
& \(+0.12\)
& \(-0.46\)
& \(-0.56\)
& \(-0.02\)
\\

Sundial (\texttt{SU})
& 32{,}768
& \(+0.11\)
& \(-1.69\)
& \(-1.76\)
& \(-0.03\)
\\

Time-MoE (\texttt{ME})
& 32{,}768
& \(-0.45\)
& \(-1.25\)
& \(-0.80\)
& \(+0.01\)
\\

TTM-R2 (\texttt{TT})
& 32{,}768
& \(-0.69\)
& \(-2.70\)
& \(-2.02\)
& \(+0.00\)
\\

Timer-XL (\texttt{TX})
& 32{,}768
& \(-0.82\)
& \(-2.40\)
& \(-1.58\)
& \(+0.00\)
\\

\bottomrule
\end{tabular}
\end{table}

\textbf{In Summary.}
Together, these diagnostics rule out several simple explanations for the observed accessibility--response separation.
Weak response persists among models with competitive ordinary forecasting performance, and reversed trend response is not explained by confusion between linear trend and mean reversion.
The trend-specific analyses instead show that preprocessing pathways, continuation behavior, standardized forecast shape, and output scaling can each alter how a slope change reaches the final forecast across different models and response regimes.
These mechanisms help explain why parameter information that is accessible in a hidden representation need not be expressed through the corresponding end-to-end forecast response.

\subsection{Conditional-Reference Sensitivity}
\label{app:revision_bayes}

The main evaluation conditions parameter and response references on the generating law and documented nuisance variables.
To test sensitivity to this information, the analysis compares three settings: known law with documented nuisances, known law with nuisances marginalized under the generator priors, and an eleven-law mixture with the law posterior inferred from history.
The primary references are~\(\mathbb{E}\left[z\mid x,m,\eta\right]\) for the parameter and~\(F_m^\star\left(x;\eta\right)\) for the response.
All settings use identical histories, signed doses, future targets, and scoring rules.

The comparison covers seasonal autoregression~(\texttt{SA}), transient pulse~(\texttt{TP}), damped oscillation~(\texttt{DO}), drive frequency~(\texttt{WC}), and diffusion time~(\texttt{WD}).
Parameter correlations use~\(8{,}192\) histories, known-law response references use~\(4{,}096\) pairs per signed dose, and law-unknown references use~\(2{,}048\) pairs.
On a balanced dataset of~\(2{,}200\) histories, the exact and calibrated discriminative law posteriors achieve accuracies of~\(0.964\) and~\(0.955\), with expected calibration errors of~\(0.011\) and~\(0.006\).
Numerical refinement changes every retained metric by at most~\(0.0064\).

\begin{table}[t]
\centering
\small
\caption{
Sensitivity of parameter correlation and pair-weighted response-reference gain to nuisance conditioning.
Response columns report reference values rather than TSFM scores.
}
\label{tab:revision_bayes}
\setlength{\tabcolsep}{5pt}
\renewcommand{\arraystretch}{1.08}
\begin{tabular}{@{}lrrrr@{}}
\toprule
Parameter
& \(r^\star_{\mathrm{cond}}\)
& \(r^\star_{\mathrm{hist}}\)
& \(U^\star_{\mathrm{cond}}\)
& \(U^\star_{\mathrm{hist}}\) \\
\midrule
Seasonal autoregression~(\texttt{SA}) & 0.975 & 0.956 & 0.848 & 0.883 \\
Transient pulse~(\texttt{TP})         & 0.997 & 0.989 & 0.983 & 0.768 \\
Damped oscillation~(\texttt{DO})      & 0.986 & 0.986 & 0.957 & 0.958 \\
Drive frequency~(\texttt{WC})         & 1.000 & 1.000 & 1.000 & 0.917 \\
Diffusion time~(\texttt{WD})          & 0.991 & 0.942 & 0.960 & 0.971 \\
\bottomrule
\end{tabular}
\end{table}

\begin{figure}[t]
    \centering
    \includegraphics[width=\linewidth]{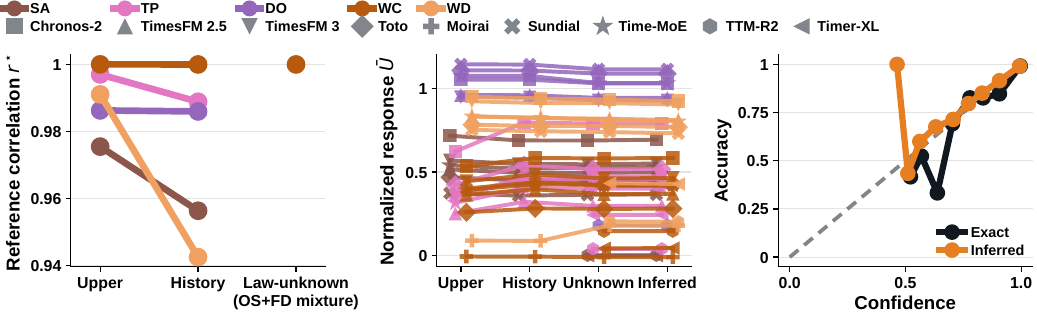}
    \caption{
    Reference and normalized-score changes when nuisance variables or the generating law are not supplied.
    Left: conditional and history-only parameter correlations.
    Middle: model responses normalized by each reference.
    Right: calibration of exact and classifier-based law posteriors.
    }
    \label{fig:revision_bayes}
\end{figure}

\autoref{tab:revision_bayes} and~\autoref{fig:revision_bayes} show that parameter references are generally more stable than response normalization when conditioning information is reduced.

\textbf{Parameter References Remain Stable.}
The left panel of~\autoref{fig:revision_bayes} shows that history-only parameter correlations remain high for all five evaluated parameters.
Damped oscillation~(\texttt{DO}) and drive frequency~(\texttt{WC}) are unchanged at~\(0.986\) and~\(1.000\), while seasonal autoregression~(\texttt{SA}) and transient pulse~(\texttt{TP}) decrease only from~\(0.975\) to~\(0.956\) and from~\(0.997\) to~\(0.989\), respectively.
Diffusion time~(\texttt{WD}) shows the largest reduction, from~\(0.991\) to~\(0.942\), but remains strongly correlated with the conditional reference.
Thus, removing nuisance information changes parameter recovery only modestly across the evaluated laws.

\textbf{Response References Are More Sensitive.}
The middle panel of~\autoref{fig:revision_bayes} shows larger and more heterogeneous changes when model responses are renormalized using reduced-information references.
Consistent with~\autoref{tab:revision_bayes}, seasonal autoregression~(\texttt{SA}), damped oscillation~(\texttt{DO}), and diffusion time~(\texttt{WD}) change only modestly between conditional and history-only response references.
Transient pulse~(\texttt{TP}) decreases from~\(0.983\) to~\(0.768\), while drive frequency~(\texttt{WC}) decreases from~\(1.000\) to~\(0.917\).
For transient pulse~(\texttt{TP}), pooled gain instead increases from~\(0.325\) to~\(1.746\) because pooling places greater weight on higher-energy pairs.
The model-level points in the middle panel further show that these reference changes do not shift all TSFMs uniformly, indicating that the choice of response reference can alter both normalization magnitude and cross-model comparisons.

\textbf{Law Uncertainty Does Not Dominate the Result.}
The right panel of~\autoref{fig:revision_bayes} compares calibration for the exact and inferred law posteriors.
Both remain close to the diagonal over most confidence levels, consistent with accuracies of~\(0.964\) and~\(0.955\) and expected calibration errors of~\(0.011\) and~\(0.006\).
Using the eleven-law mixture, the~\(0.05\) agreement criterion is satisfied for seasonal autoregression~(\texttt{SA}), transient pulse~(\texttt{TP}), drive frequency~(\texttt{WC}), and diffusion time~(\texttt{WD}).
Damped oscillation~(\texttt{DO}) is the only exception, with a pooled difference of~\(0.058\).
These results indicate that the main sensitivity arises more from how the response reference changes under reduced conditioning than from inaccurate inference of the generating law across the evaluated reference settings.
Overall, parameter-reference correlation remains comparatively stable, whereas response normalization is more dependent on the information supplied to the reference.

\subsection{Intervention Side Effects and Output Semantics}
\label{app:revision_sideeffects}

The causal-geometry evaluation in~\S\ref{sec:eval_geometry} measures whether hidden-state edits recover the reference forecast response.
To test whether this recovery introduces unintended output changes, we evaluate target fidelity, non-target forecast properties, and distributional effects across~\(64\) combinations of model, parameter, and forecast horizon.
Target fidelity is measured by centered error against the desired parameter response.
The non-target score is the largest standardized deviation in forecast variance, first-difference energy, or lag-one autocorrelation from the ideal edited forecast.
For forced diffusion, edits targeting drive frequency~(\texttt{WC}) are also projected onto the sibling diffusion-time~(\texttt{WD}) response to measure cross-coordinate effects.

Chronos-2~(\texttt{C2}) and TimesFM 3~(\texttt{F3}) additionally provide quantile forecasts.
Their per-unit-step quantile changes are linearly extrapolated before comparing relative continuous ranked probability score~(CRPS), \(50\%\)-interval coverage, and quantile crossing with the corresponding arm-B forecast.
These distributional diagnostics therefore apply only to the two quantile models.

\begin{table}[t]
\centering
\small
\caption{
Median target and non-target errors across~\(64\) horizon runs.
Quantile metrics use linearly extrapolated changes for Chronos-2~(\texttt{C2}) and TimesFM 3~(\texttt{F3}), with coverage change measured for the~\(50\%\) interval across the two evaluated quantile models.
}
\label{tab:revision_sideeffects}
\setlength{\tabcolsep}{5pt}
\renewcommand{\arraystretch}{1.08}
\begin{tabular}{@{}lrrrr@{}}
\toprule
Condition
& Target \(e\)
& Non-target
& Relative \(\Delta\)CRPS
& \(\Delta\)Coverage \\
\midrule
Matched input      & 1.10 & 0.325 & --     & --     \\
Deployed           & 0.64 & 0.177 & 0.000  & -0.001 \\
Projected          & 0.21 & 0.030 & -0.002 & -0.001 \\
Fixed              & 1.12 & 0.155 & 0.013  & -0.005 \\
Equal-norm random  & 1.00 & 0.423 & 0.025  & -0.015 \\
\bottomrule
\end{tabular}
\end{table}

\begin{figure}[t]
    \centering
    \includegraphics[width=\linewidth]{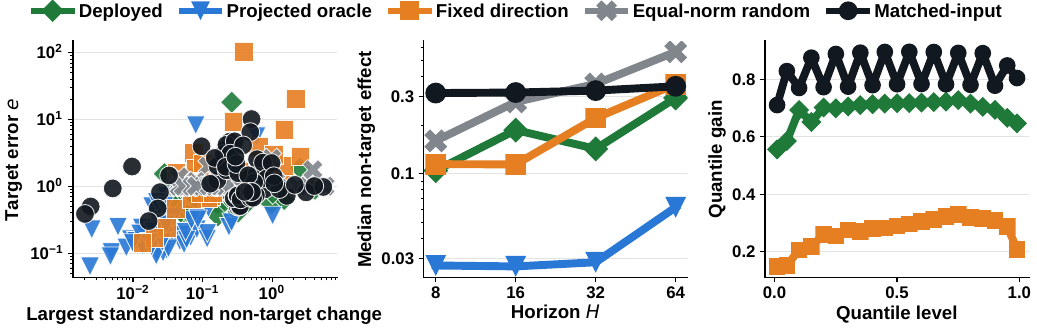}
    \caption{
    Target fidelity, non-target deviations, and quantile-level response changes for hidden-state edits.
    Quantile diagnostics apply only to Chronos-2~(\texttt{C2}) and TimesFM 3~(\texttt{F3}).
    }
    \label{fig:revision_sideeffects}
\end{figure}

\textbf{Target Fidelity and Non-Target Effects.}
\autoref{tab:revision_sideeffects} shows that projected edits achieve the lowest median target error and non-target deviation, at~\(0.21\) and~\(0.030\), respectively.
Deployed edits reduce target error from~\(1.00\) for equal-norm random edits to~\(0.64\), while reducing non-target deviation from~\(0.423\) to~\(0.177\).
Fixed edits have a slightly smaller non-target deviation of~\(0.155\), but their target error rises to~\(1.12\), indicating that a small side effect alone does not imply successful target control.
The left panel of~\autoref{fig:revision_sideeffects} shows the same tradeoff across individual runs.
Projected edits concentrate toward low target error and small non-target change, while deployed edits generally lie between the projected condition and the fixed or random controls.
Thus, improved target alignment is not explained simply by applying an edit of comparable magnitude.

\textbf{Side Effects Vary With Forecast Horizon.}
The middle panel of~\autoref{fig:revision_sideeffects} shows how non-target effects change with forecast horizon.
Projected edits remain lowest across horizons, whereas the deviations of deployed, fixed, and equal-norm random edits generally increase as the horizon grows.
The deployed condition nevertheless remains below the equal-norm random control across the evaluated horizons.
This pattern suggests that selective response control becomes harder to preserve over longer forecasts, even when the target direction is recovered more accurately.

\textbf{Distributional Outputs Remain Largely Stable.}
For Chronos-2~(\texttt{C2}) and TimesFM 3~(\texttt{F3}), projected and deployed edits produce median relative CRPS changes near zero and \(50\%\)-interval coverage changes of only~\(-0.001\), as shown in~\autoref{tab:revision_sideeffects}.
The right panel of~\autoref{fig:revision_sideeffects} provides a finer view across quantile levels.
Deployed edits preserve substantially more of the quantile-level response than the fixed-direction control, although they do not fully reproduce the matched-input response across the quantile range.
Thus, small aggregate CRPS and coverage changes can coexist with structured differences in how the intervention response is distributed across quantiles.

\textbf{Cross-Coordinate Effects.}
For forced diffusion, the deployed drive-frequency~(\texttt{WC}) edit has gain~\(-0.48\) along the diffusion-time~(\texttt{WD}) response, close to the projected value of~\(-0.46\).
Because the two response directions overlap, this quantity does not isolate parameter leakage.
Together, these analyses show that hidden-state edits improve target alignment while retaining measurable non-target, horizon-dependent, and distributional effects.

\subsection{Computational Cost and Operational Boundary}
\label{app:revision_cost}

The hidden-state intervention pipeline incurs different costs during direction discovery and test-time deployment.
To quantify this operational boundary, \(128\) linear trend histories are timed at each selected layer on one NVIDIA H100.
Discovery measures per-input Jacobian construction and inversion together with one-time basis construction using~\(16{,}384\) discovery displacements.
Deployment includes readout evaluation, coordinate lifting, and the edited forward pass, without constructing a new Jacobian.

\textbf{Results.}
As shown in~\autoref{tab:revision_cost}, inversion is negligible relative to the other components at~\(H=16\), requiring only~\(0.02\)--\(0.08\,\mathrm{ms}\) across the four evaluated models.
Jacobian construction ranges from~\(0.002\) to~\(0.036\,\mathrm{s}\), while one-time basis construction ranges from~\(4.6\) to~\(11.4\,\mathrm{s}\).
Deployment is substantially cheaper than basis construction, ranging from~\(0.2\) to~\(22.2\,\mathrm{ms}\) per input.

Across~\(H=8\) to~\(64\), Jacobian time increases from~\(0.001\) to~\(0.008\,\mathrm{s}\) for Chronos-2~(\texttt{C2}), \(0.004\) to~\(0.021\,\mathrm{s}\) for TimesFM 3~(\texttt{F3}), \(0.011\) to~\(0.062\,\mathrm{s}\) for Time-MoE~(\texttt{ME}), and \(0.001\) to~\(0.005\,\mathrm{s}\) for TTM-R2~(\texttt{TT}).
Time-MoE~(\texttt{ME}) also has the largest measured memory footprint, ranging from~\(17.6\) to~\(24.8\,\mathrm{GB}\).
Thus, the principal computational burden lies in discovery and validation, while the learned readout can be applied at test time without repeating the per-input Jacobian analysis.

\begin{table}[t]
\centering
\small
\caption{
Computational cost at~\(H=16\).
Jacobian construction and inversion are measured per pair, deployment per input, and basis construction once.
Shared-device contention and backend differences preclude a general speed comparison across the four evaluated TSFM architectures.
}
\label{tab:revision_cost}
\setlength{\tabcolsep}{5pt}
\renewcommand{\arraystretch}{1.08}
\begin{tabular}{@{}lrrrrr@{}}
\toprule
Model
& Jacobian (s)
& Inversion (ms)
& Basis (s)
& Peak (GB)
& Deployment (ms) \\
\midrule
Chronos-2~(\texttt{C2}) & 0.003 & 0.04 & 6.2  & 11.0 & 6.3  \\
TimesFM 3~(\texttt{F3}) & 0.007 & 0.02 & 4.6  & 14.6 & 22.2 \\
Time-MoE~(\texttt{ME})  & 0.036 & 0.08 & 11.4 & 21.1 & 9.2  \\
TTM-R2~(\texttt{TT})    & 0.002 & 0.03 & 5.2  & 10.1 & 0.2  \\
\bottomrule
\end{tabular}
\end{table}

\subsection{Limitations and Future Work}
\label{app:limitations}

The controlled laws isolate designated Dynamical Parameters but do not cover broader multivariate, nonstationary, or exogenous dynamics.
Linear probes measure linear representation accessibility and may miss information available only through nonlinear readouts.
The conditional reference receives law and nuisance information unavailable to the TSFM, so it serves as an informed response reference rather than a model-equivalent predictor.
Hidden-state interventions measure response capacity at calibrated steps and depend on the selected state, family of reference responses, and forecast horizon.
Controlled injections on real-series backgrounds preserve realistic temporal structure but do not identify the intrinsic dynamics of the underlying series.
Future work could extend the framework to richer dynamical systems, nonlinear accessibility measurements, and training objectives that better align accessible parameter information with forecast response.

\end{document}